%% file: main.tex
\documentclass{amsart}
\usepackage[foot]{amsaddr}

\newif\ifarxiv
\arxivtrue

\usepackage[T1]{fontenc}
\usepackage[scaled]{helvet}

\usepackage{placeins}

\usepackage{comment}  
\usepackage{graphicx} 
\usepackage{hyperref} 
\usepackage{tikz}     
\usepackage{pgfplots} 
\usepackage{siunitx}  
\usepackage{pdflscape}
\usepackage{subcaption}
\DeclareCaptionLabelFormat{bold}{\textbf{#2}}
\usepackage[font=small, margin= 0cm]{caption}

\DeclareSIUnit{\mmHg}{mmHg}

\usepackage{enumerate}
\usepackage{paralist}
\usepackage[margin=3.5cm]{geometry}

\usepackage{booktabs}
\usepackage{longtable}
\usepackage{pifont}
\usepackage{makecell}

\usetikzlibrary{arrows}
\usetikzlibrary{arrows.meta}
\usetikzlibrary{calc}
\usetikzlibrary{external}
\usetikzlibrary{patterns}
\usetikzlibrary{positioning}
\usetikzlibrary{scopes}

\pgfplotsset{
  compat = 1.16,
}

\usepgfplotslibrary{statistics}

\usepackage[numbers, compress, sort]{natbib}

\ifarxiv
\else
  \usepackage{lineno}
  \linenumbers
\fi

\usepackage[utf8]{inputenc}

\title{Predicting sepsis in multi-site, multi-national intensive care cohorts using deep learning} 

\author{Michael Moor$^{1,2,\ast}$, Nicolas Bennet$^{3,\ast}$, Drago Plecko$^{3,\ast}$, Max Horn$^{1,2\ast}$, \\ Bastian Rieck$^{1,2}$, Nicolai Meinshausen$^{3}$, Peter Bühlmann$^{3}$ and \\ Karsten Borgwardt$^{1,2}$}

\address{\footnotesize$^1$: Department of Biosystems Science and Engineering, ETH Zurich, 4058 Basel, Switzerland}

\address{\footnotesize$^2$: SIB Swiss Institute of Bioinformatics, Switzerland.}
\address{\footnotesize$^3$: Seminar for Statistics, Department of Mathematics, ETH Zurich, Switzerland.}
\address{\footnotesize$^{\ast}$: These authors contributed equally.}

\usepackage{fancyhdr} 

\fancyheadoffset{0cm}

\fancypagestyle{plain}{%
  \fancyhf{}%
  \fancyhead[R]{\thepage}%
}

\begin{document}

\maketitle
\thispagestyle{empty}

\begin{abstract}
Despite decades of clinical research, sepsis remains a global public health crisis with high mortality, and morbidity.
Currently, when sepsis is detected and the underlying pathogen is identified, organ damage may have already progressed to irreversible stages. Effective sepsis management is therefore highly time-sensitive. 
By systematically analysing trends in the plethora of clinical data available in the intensive care unit (ICU), an early prediction of sepsis could lead to earlier pathogen identification, resistance testing, and effective antibiotic and supportive treatment, and thereby become a life-saving measure. Here, we developed and validated a machine learning (ML) system for the prediction of sepsis in the ICU. Our analysis represents the largest multi-national, multi-centre in-ICU study for sepsis prediction using ML to date. Our dataset contains $156,309$ unique ICU admissions, which represent a refined and harmonised subset of five large ICU databases originating from three countries. Using the international consensus definition Sepsis-3, we derived hourly-resolved sepsis label annotations, amounting to $26,734$ ($17.1\%$) septic stays. We compared our approach, a deep self-attention model, to several clinical baselines as well as ML baselines and performed an extensive internal and external validation within and across databases. On average, our model was able to predict sepsis with an AUROC of $ 0.847 \pm 0.050$ (internal out-of sample validation) and $0.761 \pm 0.052$ (external validation). For a harmonised prevalence of $17\%$, at $80\%$ recall our model detects septic patients with $39\%$ precision \SI{3.7}{\hour} in advance.
\end{abstract}

\noindent
Sepsis remains a major public health crisis associated with high mortality, morbidity, and related health costs~\cite{dellinger2013, kaukonen2014, arise2007, hotchkiss2016}. 
Following the most recent definition (Sepsis-3)~\cite{singer2016third, seymour2016assessment}, sepsis was defined as a life-threatening organ dysfunction caused by a dysregulated host response to infection. From sepsis onset, each hour of delayed effective antimicrobial treatment increases mortality~\citep{ferrer2014empiric, seymour2017time, pruinelli2018delay}. However, identifying bacterial species in the blood can take up to 48 hours after blood sampling~\cite{osthoff2017impact}. Meanwhile, an abundance of clinical and laboratory data is being routinely collected, the richest set of which is accumulated in the intensive care unit~(ICU). But as it becomes harder for ICU clinicians to manually process increasing quantities of patient information, large parts thereof are merely stored in patient data management systems~(PDMS) and remain unused while the attending physician is challenged to identify the relevant subset of information to focus on and act upon~\citep{pickering2013data}. This situation is exacerbated by sepsis being a complex and heterogeneous syndrome, which may still be reversible in its early stages, yet difficult to identify, while its more advanced stages become easier to recognise, but also more challenging to successfully treat~\citep{levy2018surviving, rhodes2017surviving}. 

Even though singular biomarkers for sepsis have previously failed to demonstrate a sufficient specificity \citep{bozza2007cytokine, tsalik2012discriminative}, several candidates such as PCT, presepsin, CD64, suPAR, sTREM-1 and MiR-223-3p have shown some promise and need to be further investigated \citep{larsen2017novel, zhang2019circulating}. In the meantime, there is no clinical gold standard for the accurate and early identification of sepsis. Having entered the digital era, it is the hope that this notoriously hard task, i.e., to recognise sepsis early, can be solved by fully leveraging the data-rich environment of the ICU using state-of-the art machine learning (ML) techniques.

Recent years have brought forth several studies trying to address the sepsis early prediction problem via ML (see \citet{fleuren2020machine} and \citet{moor2020early} for an overview). However, according to these systematic reviews, more than half of the assessed studies using ML for predicting in-ICU sepsis are based on the same single centre, namely MIMIC~\citep{johnson2016mimic},
an established open-access critical care database. By contrast, the majority of remaining studies are based on restricted-access datasets~\citep{fleuren2020machine}. As a consequence, the field currently lacks accessible, annotated, and multi-centre EHR data for developing and validating sepsis early warning systems. The literature so far has therefore struggled to assess the generalisability of sepsis prediction models and their \emph{transferability} across hospitals or even across countries.

Sepsis prediction in the ICU has typically been considered a \emph{static} prediction task, which attempts to answer the question: ``given an observed onset of sepsis, how early in advance could we have predicted it?''~\citep{moor2020early}. For this, the time window preceding sepsis onset in cases is commonly compared to similarly-timed windows in control patients \citep{moor2020early}. However, since the models were typically trained offline on full patient stays (or up to sepsis onset), it is not clear how well such models generalise to the clinical reality of real-time continuous monitoring applications. Therefore, to minimise data distribution shift between model development and application, in this study we considered the prediction of sepsis as a \emph{real-time} prediction task. Hence, our early warning system was optimised for the continuous monitoring of patients in order to raise an early alarm for sepsis. 

Compared to more conventional clinical prediction targets such as mortality or length of stay, sepsis is a hard-to-define clinical entity of syndromic nature~\citep{singer2016third}. The exact onset of sepsis is generally unknown and in practice needs to be approximated in order to annotate patient data. To this end, a miscellaneous set of labelling strategies have been adopted ranging from consensus definitions (Sepsis-2~\citep{levy20032001}, Sepsis-3~\citep{singer2016third}) to more ad-hoc definitions combining ICD billing codes with clinical and laboratory signs of infection and inflammation~\citep{moor2020early}. Combined with the unavailability of annotated ICU data, these circumstances perpetuate an increasingly fragmented literature in which it becomes ever more challenging to pool or compare quantitative results regarding the early predictability of sepsis. The striking lack of accessible and annotated data has further manifested in a recent study demonstrating that a widely adopted proprietary sepsis prediction model performed surprisingly poorly in an external validation~\citep{Wong2021}.

The goal of this study was to address these challenges by unifying ICU data from three nations and five databases to build an open-access platform for developing and externally validating sepsis prediction approaches. After harmonising, cleaning, and filtering these data, we implemented sepsis annotations based on Sepsis-3~\citep{singer2016third} and developed sepsis early warning systems by leveraging state-of-the-art ML algorithms. We further devised an evaluation strategy that accounts for the inherent trade-off between \emph{accurate} and \emph{early} alarms for sepsis while keeping false alarms (and therefore alarm fatigue) at bay. Finally, our unique disposition with harmonised ICU data from three countries enabled us to perform an extensive, unprecedented external validation to assess transferability of models between hospitals, countries, and even continents. 

\begin{figure}[tbp]
    \centering
    \subcaptionbox{\label{fig:cleaning}}{
        \includegraphics[width=1.\linewidth]{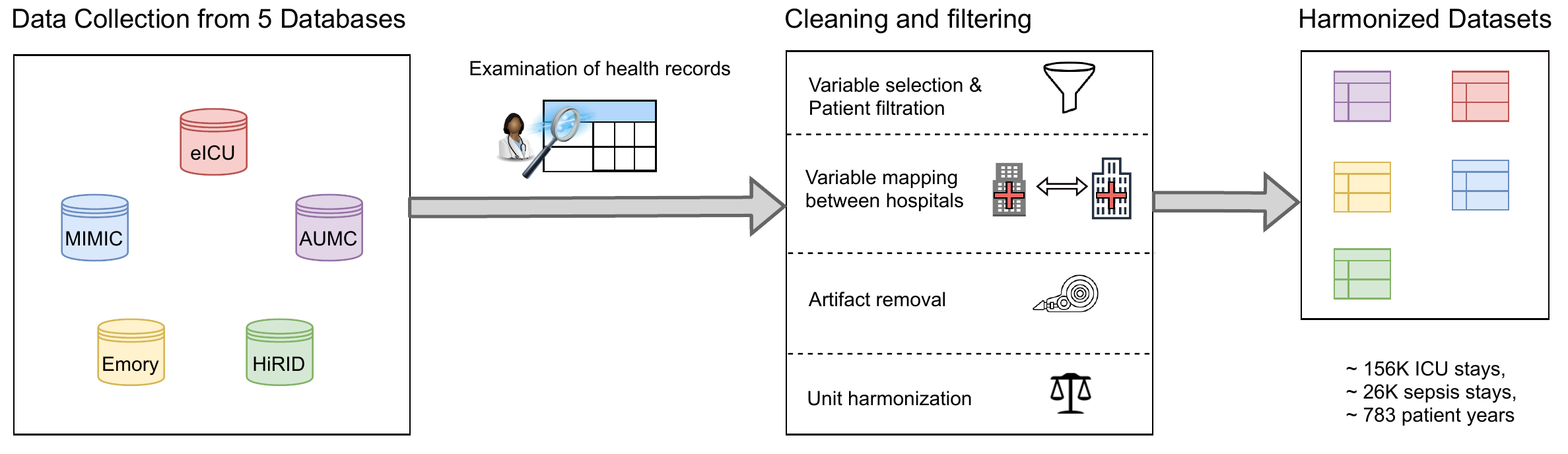}
    }
    \subcaptionbox{\label{fig:pipeline}}{
        \includegraphics[width=1.\linewidth]{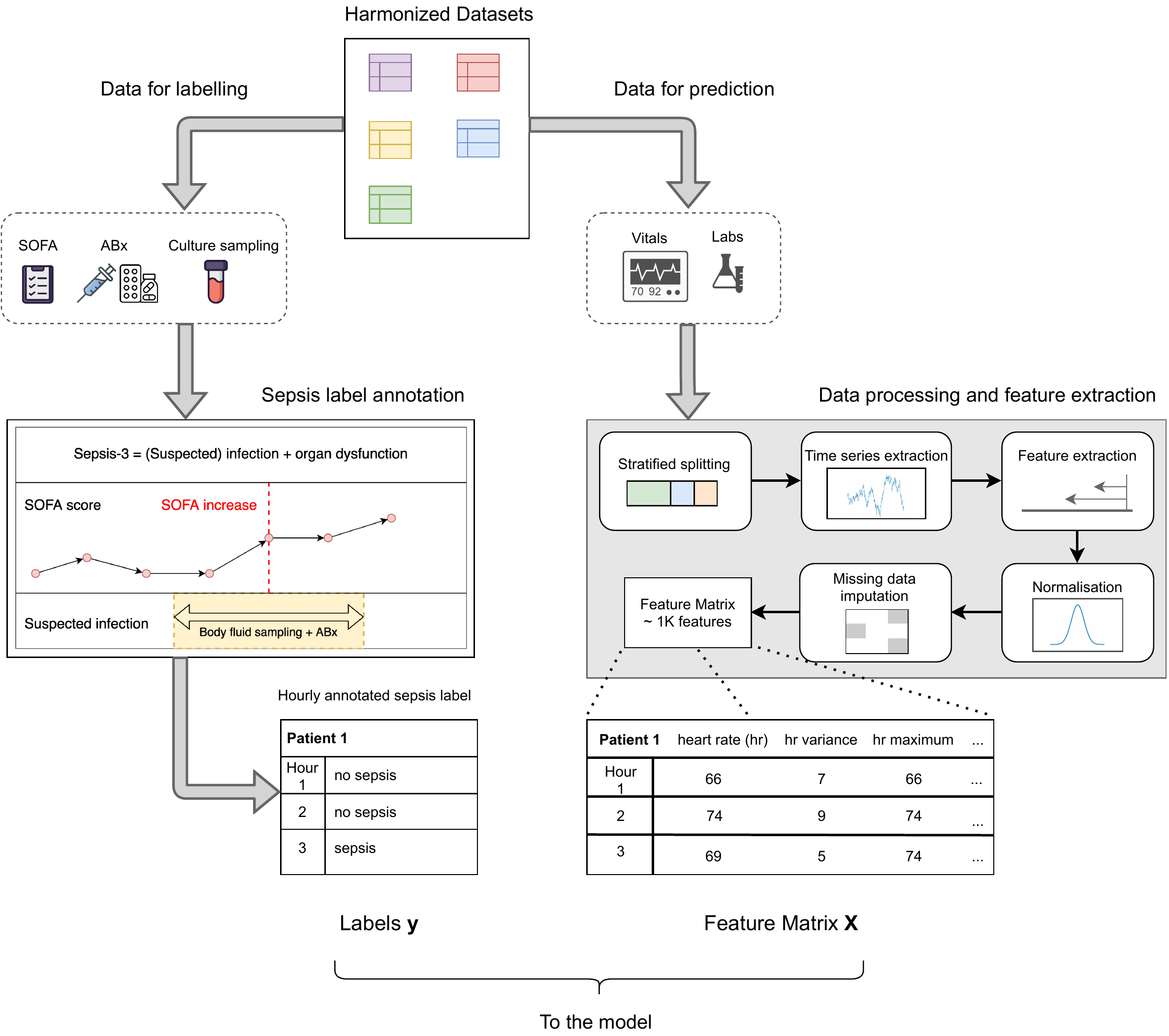}
    }
    \caption{Overview of the preprocessing pipeline.
    Panel~\subref{fig:cleaning}: Data from five ICU EHR data bases are collected, cleaned and harmonised.
    Panel~\subref{fig:pipeline}: Harmonised data are extracted for sepsis label annotation (left) as well as feature extraction (right) resulting in labels and features that are used for training the model.
    }
    \label{fig:preprocessin}
\end{figure}

\begin{figure}
    \centering
    \subcaptionbox{\label{fig:venn}}{
        \includegraphics[height=4.5cm]{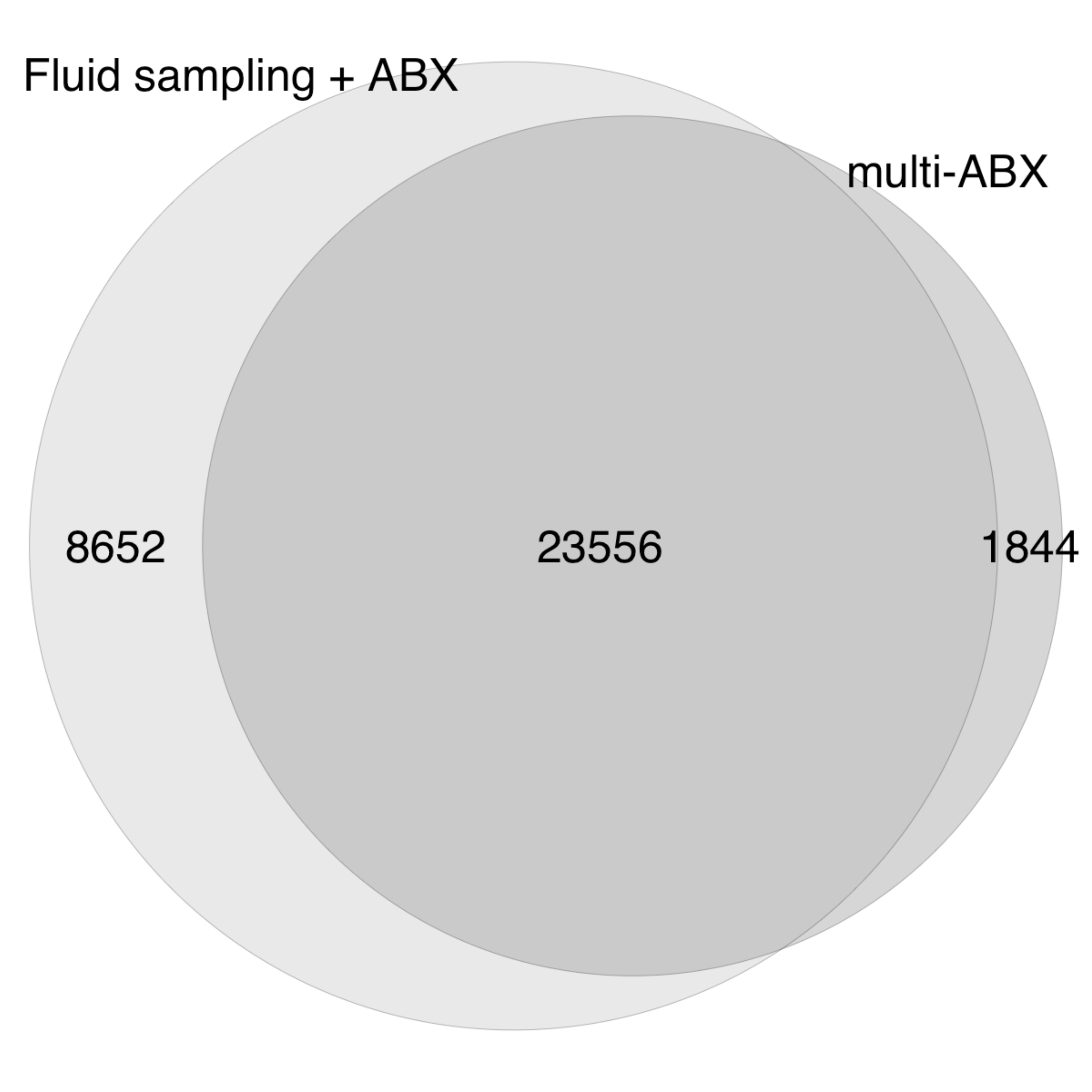}
    }
    \subcaptionbox{\label{fig:s3_def}}{
        \includegraphics[height=3.5cm]{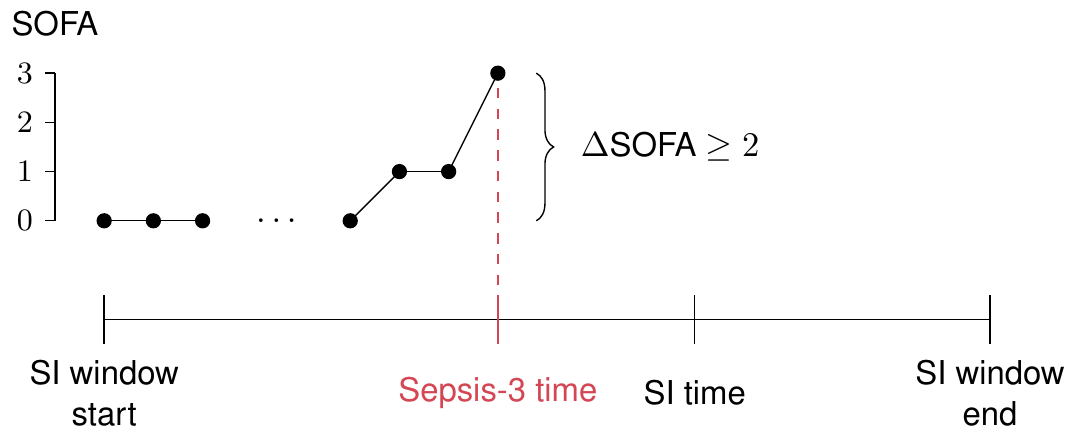}
    }\\
    \subcaptionbox{\label{fig:units}}{
    \includegraphics[width=0.95\linewidth]{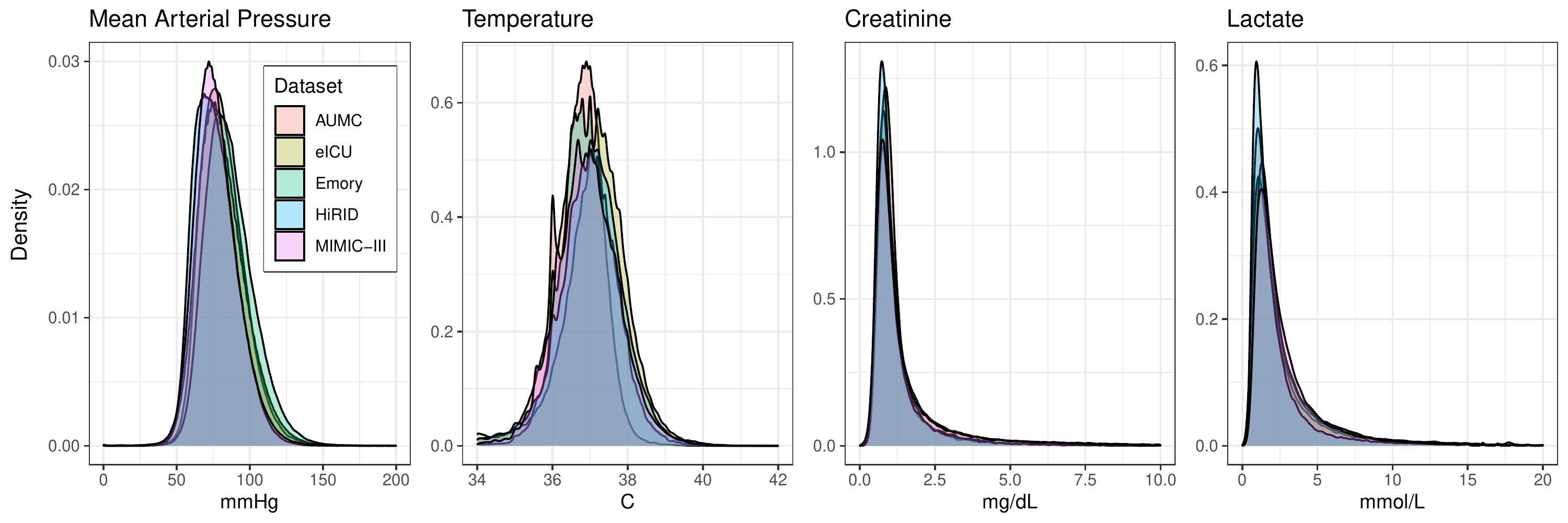}
    }\\
    \subcaptionbox{\label{fig:task}}{
        \includegraphics[width=0.9\linewidth]{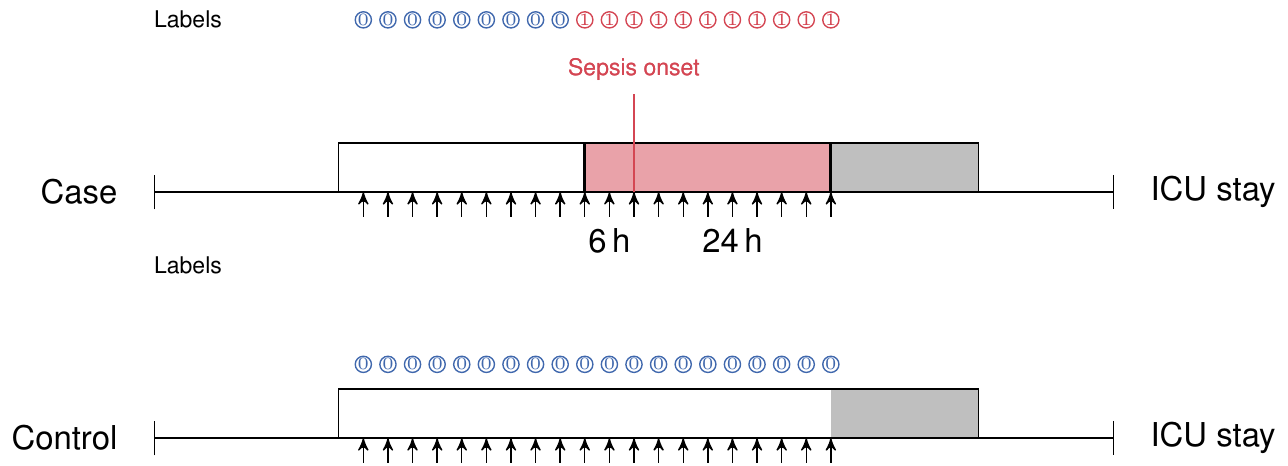}
    }
    \caption{Data harmonisation, annotation and task formulation.
    \subref{fig:venn}: Comparison of the original suspected infection definition (antibiotics + fluid sampling) against the alternative suspected infection definition based on multiple antibiotics (which was used when no culture sampling was available) displayed for MIMIC-III, a dataset that allows for both implementations. The two variations showed a Jaccard similarity of 0.69. For further visualisations, refer to \autoref{fig:siaumcnonsurg} in the Supplementary.  
    \subref{fig:s3_def}: Visual representation of how the onset time of sepsis was defined following Sepsis-3~\citep{singer2016third}. The label is constructed by evaluating the SOFA score on hourly basis and determining suspicion of infection (SI) windows based on body fluid sampling and antibiotics administration~\citep{singer2016third, seymour2016assessment}. The criterion is fulfilled if (and at the time of) an acute increase in SOFA of at least 2 points coincides with an SI window. 
    \subref{fig:units}: Monitoring of distributions of measurements across all five datasets. These checks were applied to ensure that units were harmonised.
    \subref{fig:task}: Prediction task for a given case and control patient. Data is available for the whole ICU stay. We set the label of the case to be $1$ starting from \SI{6}{\hour} \emph{before} sepsis onset~(red line) until
    \SI{24}{\hour} \emph{after} sepsis onset. During training, this allows for penalising late false negative predictions. If more than
    \SI{24}{\hour} after onset have passed, any further available data is not considered~(grey region).}
    \label{fig:vars_and_task}
\end{figure}

\section*{Results}

\subsection*{Creating a multi-centre ICU cohort for sepsis}
We present the largest multi-national in-ICU analysis for the early prediction of sepsis using machine learning to date. After cleaning, filtering and processing, our fully-interoperable cohort comprises $783$ ICU admission years and a total of $156,309$ ICU stays of which $26,734$ ($17.1\%$) develop sepsis. For this, we collected electronic health record (EHR) data from five databases representing three nations: HiRID~\citep{hyland2020early} from Switzerland, AUMC~\citep{thoral2021sharing} from the Netherlands, as well as Emory~\citep{reyna2019early}, MIMIC-III~\citep{johnson2016mimic}, and eICU~\citep{pollard2018eicu} from the US. The Emory dataset was the only dataset that was externally preprocessed and annotated as part of the ``PhysioNet Computing in Cardiology Challenge 2019''. Next, to prevent redundancy, we only reuse data originating from Emory Hospital, and discard the MIMIC-III part of the challenge data~\citep{reyna2019early}. For the input data of our model, in order to harmonise these distinct data sources, we extracted vital, laboratory, and static patient variables (see \autoref{tab:variables}) that were \begin{inparaenum}[i)] \item plausibly relevant for sepsis, while \item consistently measured and \item not suspected to lead to spurious models that merely wait for the clinician to treat sepsis. \end{inparaenum} To account for the last point, we excluded therapeutic variables such as antibiotics, intravenous fluids or vasopressors from the set of input data used for prediction.  
To ensure interoperability, we resampled all datasets to an hourly resolution, reporting the median value per hour and patient for each variable. \autoref{tab:variables} lists all input variables and indicates their availability per dataset. Whereas four out of five datasets largely overlap on a set $59$ variables~(we will henceforth refer to them as ``core'' datasets), the Emory dataset~(which was made available only in a preprocessed and annotated stage) exhibits a smaller overlap ($35$ out of $59$ variables) to our harmonised variable set. Therefore, and since it is the only dataset with predefined sepsis annotations, we report analyses about this distinct dataset separately~(see Section~\ref{sec:additional_results} in the Supplementary Materials). 

\subsection*{Developing an early warning score for sepsis}

We continuously monitored $59$ vital and laboratory parameters in hourly intervals (together with $4$ static variables) in order to raise an alarm when sepsis is about to occur. We incentivised our model to recognise that a sepsis will start within the next \SI{6}{\hour}.
To improve upon clinical baselines, we investigated two families of classifiers: deep learning approaches and non-deep ML approaches. As for deep models, we considered a self-attention model~(attn)~\citep{vaswani2017attention} as well as a recurrent neural network employing Gated Recurrent Units~(gru)~\citep{cho2014properties}, both of which are intrinsically capable of leveraging sequential data. Next, we included LightGBM~(lgbm)~\citep{ke2017lightgbm} and a LASSO-regularised Logistic regression~(lr)~\citep{Tibshirani96}, which were given access to a total of 1,269 features that were extracted in order to make temporal dynamics governing the data accessible to these methods. As for clinical baselines, we investigated how well sepsis could be predicted with NEWS~\citep{jones2012newsdig}, MEWS~\citep{subbe2001validation}, SIRS~\citep{bone1992definitions}, SOFA~\citep{vincent1996sofa}, and qSOFA~\citep{singer2016third}. 

The employed Sepsis-3 definition subsumes the SOFA score, which captures additional treatment information (e.g., antibiotics and vasopressors) that our ML models were \emph{not} intended to rely upon. Still, since the clinical baseline scores encode valuable domain knowledge, during feature engineering, we constructed partial scores by only including lab and vitals that belong to our list of readily measured, non-therapeutic $63$ input variables.

\subsection*{An encounter-focused evaluation strategy}
We formulated the problem of detecting sepsis as an online prediction
task, where a model is continuously provided with new data in order to
make the next, hourly prediction for a given patient. This setup was
chosen in order to simulate a realistic deployment setting already
during the (retrospective) model development phase. By contrast, in
order to devise a clinically sensible model evaluation strategy, we
report the following performance metrics on the \emph{encounter} level:
\begin{inparaenum}[i)] \item area under the ROC curve (AUROC) and \item
precision and earliness for a fixed recall of $80\%$.
\end{inparaenum}
In both cases, we considered the set of all observed unnormalised
prediction scores (logits) and removed the top and bottom $0.5$
percentiles, i.e., we kept the $99$ innermost percentiles~(for outlier
robustness and sake of better visualisability). Next, we partitioned the
remaining range of scores into $100$ evenly-spaced thresholds. For each
threshold, we swept through the prediction scores of an ICU stay and
triggered an alarm the first~(and only the first) time the prediction
score surpassed the current threshold. The set of raised alarms was then
used to fill the cells of a confusion matrix, where a sepsis case with
an alarm was considered true positive, a control stay without an alarm
a true negative, and so on. ROC curves were then computed from these
confusion matrices over all thresholds. For the second measure, we
determined the threshold with $80\%$ recall and report the precision~(or
positive predictive value) as obtained from the confusion matrix
corresponding to this threshold, as well as alarm \emph{earliness},
computed as the median number of hours the alarm preceded sepsis onset. In
case there was no threshold with exactly $80\%$ recall, we linearly
interpolated all measures based on the two closest thresholds. This
evaluation strategy is conservative in that no repeated alarms~(that
would improve recall at the cost of alarm fatigue) are permissible. While
this makes recognising sepsis cases more challenging, it also guarantees
that at most only a single false alarm can be raised in a control stay.

\subsection*{Internally validating sepsis warning systems in 5 databases from 3 countries}%

\begin{figure}
    \centering
    \subcaptionbox{\label{fig:in_aumc_subset}}{
        \includegraphics[width=0.73\linewidth]{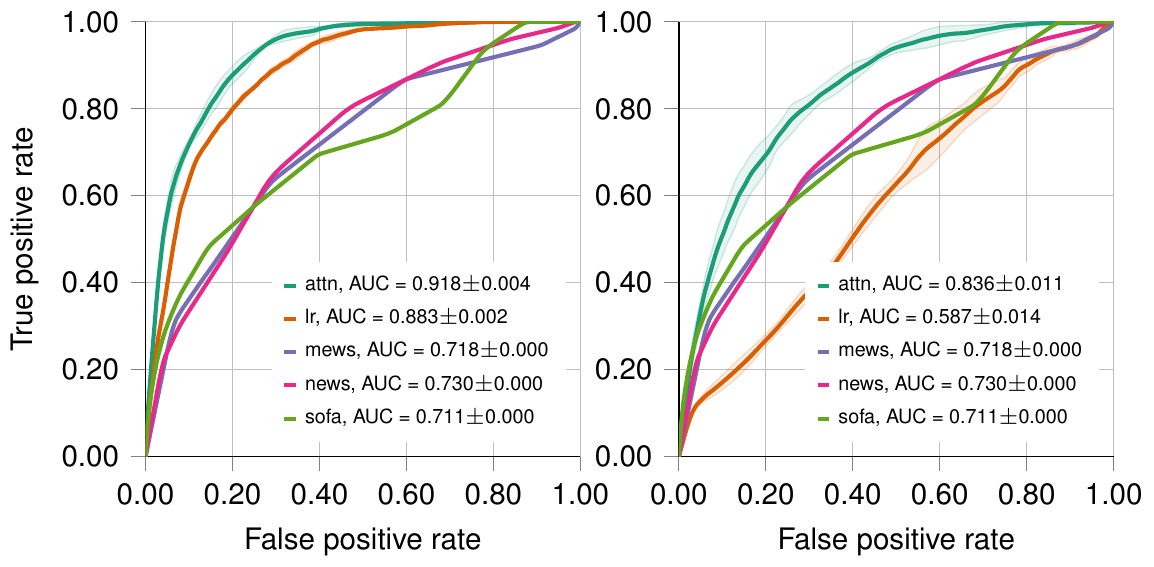} 
    }\\%
    \subcaptionbox{\label{fig:in_aumc_scatter_subset}}{
        \includegraphics[width=0.72\linewidth]{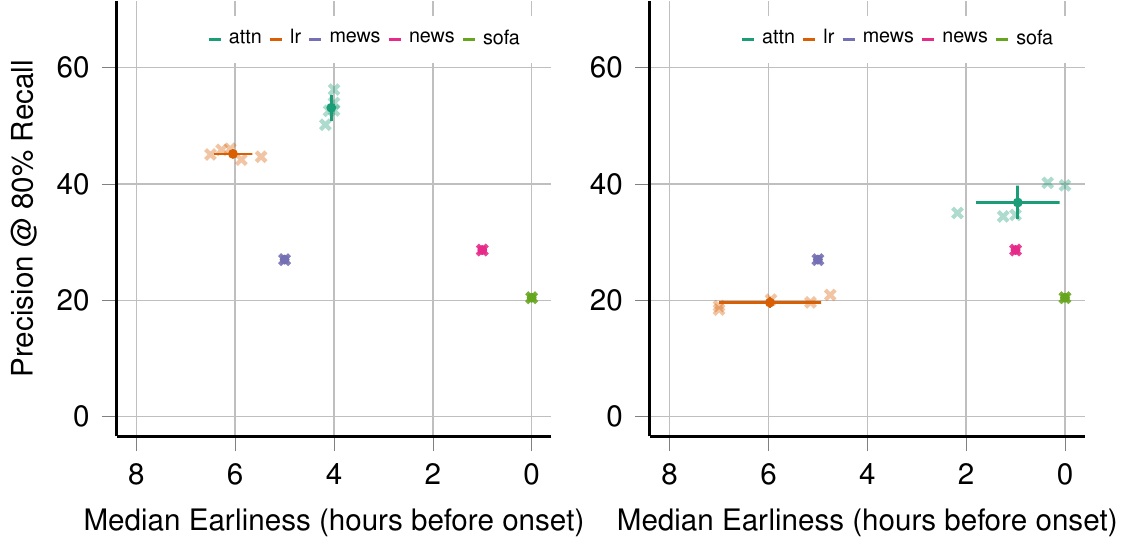} 
    }\\%
    \subcaptionbox{\label{fig:roc_heatmap}}{
        \includegraphics[width=0.6\textwidth]{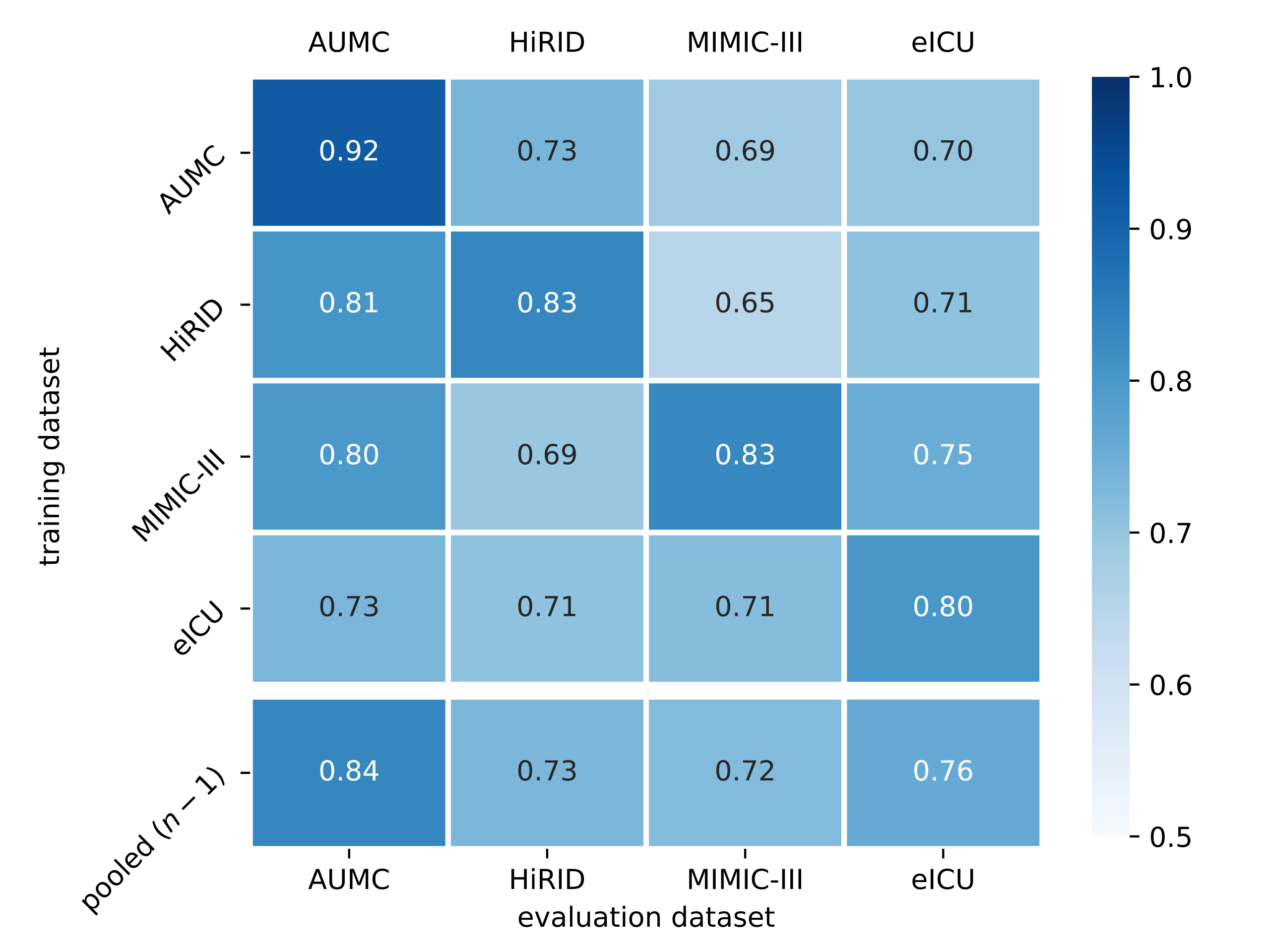}
    }%
    \caption{\scriptsize Predictive performance plots.
\subref{fig:in_aumc_subset}-\subref{fig:in_aumc_scatter_subset}: Internal (left) and external (right) validation of early sepsis prediction illustrated for the AUMC dataset. Our deep learning approach (attn) is visualised together with a subset of the comparison methods including clinical baselines (SOFA, MEWS and NEWS), as well as the logistic regression model (lr) (for all methods and datasets refer to Extended Data Figure~\ref{fig:aumc_roc}-\ref{fig:hirid_roc}).
    \subref{fig:in_aumc_subset}: ROC Curves are computed on an encounter level following our threshold-based evaluation strategy. The error bars indicate standard deviation over $5$ repetition of train-validation splitting.
    \subref{fig:in_aumc_scatter_subset}: We display the trade-off between prediction accuracy and earliness. For this, recall is fixed at $80\%$ and precision is scattered against the median number of hours that the alarm precedes sepsis onset.
    \subref{fig:roc_heatmap}: External validation of our deep self-attention model. The encounter-level AUROC (mean over five repeated training splits) is displayed. The columns indicate the evaluation dataset, in the rows the training dataset is listed. In the last row, for each evaluation dataset, the performance of an ensemble of all other datasets is shown (where the respective maximum prediction score was used). We observe that the pooling approach improves the generalisability to new datasets by outperforming or being on par with the best training dataset, which is not known a priori. To maximise comparability with the internal validation, the metrics are computed on the test split of each dataset.} 
    \label{fig:internal_roc}
\end{figure}

In a first step, we trained all models on all datasets separately and performed an internal validation by reporting out-of-sample performance (i.e., on a held-out test split) on the training dataset for each method and dataset. The results are shown in \autoref{fig:internal_roc}. The full set of methods and datasets is displayed in Extended Data Figures~\ref{fig:aumc_roc}--\ref{fig:hirid_roc}. Additionally, for the smaller variable set of the Emory database, the corresponding results are shown in Supplementary Figures~\ref{fig:emory_roc} and \ref{fig:heat_extended}. 

Our deep learning model (attn) achieved an average test AUROC of $0.847 \pm 0.050$ when internally validating on the four core datasets that were harmonised to the consensus list of $63$ variables~(see \autoref{tab:variables}). For a harmonised prevalence of $17\%$, at
$80\%$ recall, on average our model detects septic patients with \SI{39.3 \pm 1.2}{\percent} precision \SI{3.71 \pm 0.41}{\hour} in advance.
In the left column of Figures~\ref{fig:in_aumc_subset} and \ref{fig:in_aumc_scatter_subset}, these results are illustrated for the AUMC dataset, our deep attention model (attn), and a selection of clinical baselines (SOFA, NEWS, and MEWS). Notably, SOFA is a strong baseline for predicting the Sepsis-3 label, as it is a prominent component of the label while it also includes further information about treatment (antibiotics and vasopressor administrations), which our model (restricted to labs and vitals) was not given access to. Further clinical and ML baselines are shown in the Extended Data Figures~\ref{fig:aumc_roc}--\ref{fig:hirid_roc}. 

\subsection*{Can sepsis prediction models generalise to different centres? An external validation.}

In a second analysis, we performed an external validation by applying previously-trained models to independent testing databases. As a first step, we assessed the transfer from one database to another one using the harmonised variable set. For our deep learning model, this is displayed in \autoref{fig:roc_heatmap}, whereas all other models are depicted in Supplementary Figures~\ref{fig:ex_aumc}--\ref{fig:ex_scatter_physionet2019}. \autoref{fig:roc_heatmap} depicts a heatmap of AUROC values, with rows corresponding to the training database and columns corresponding to the testing database. The last row of this heatmap displays a maximum pooling strategy, where for a given database, all the prediction scores of the same model class trained on all of the remaining databases seperately are aggregated such that at each point in time, the maximal prediction score is used for prediction. 
Using this pooling strategy in our external validation (refer to the right columns in Figures~\ref{fig:in_aumc_subset} and \ref{fig:in_aumc_scatter_subset} as well as the Extended Data Figures~\ref{fig:aumc_roc}--\ref{fig:hirid_roc}), we achieve an average AUROC of $0.761 \pm 0.052$. When fixing the prediction threshold at $80\%$ recall this resulted in \SI{29.3 \pm 1.4}{\percent} precision \SI{1.75 \pm 0.94}{\hour} in advance.

Overall, \autoref{fig:roc_heatmap} demonstrates that applying the pooling strategy for a given testing database achieves better or equivalent performance as compared to the best-performing model trained on a single external database that could only be determined ex post hoc. In Extended Data Figure~\ref{fig:cohort_ablation}, we find that the higher performance in AUMC stems from the foremost surgical cohort, however we could not generalise this finding to other cohorts. In an auxillary analysis (Extended Data Figure~\ref{fig:pooled_datasets}), we retrained our deep learning model by pooling all datasets except for the respective evaluation dataset and found no improvement over our max pooling strategy which ensembles predictions without the need for costly retraining. 

\begin{figure}
    \centering
    \includegraphics[width=0.8\textwidth]{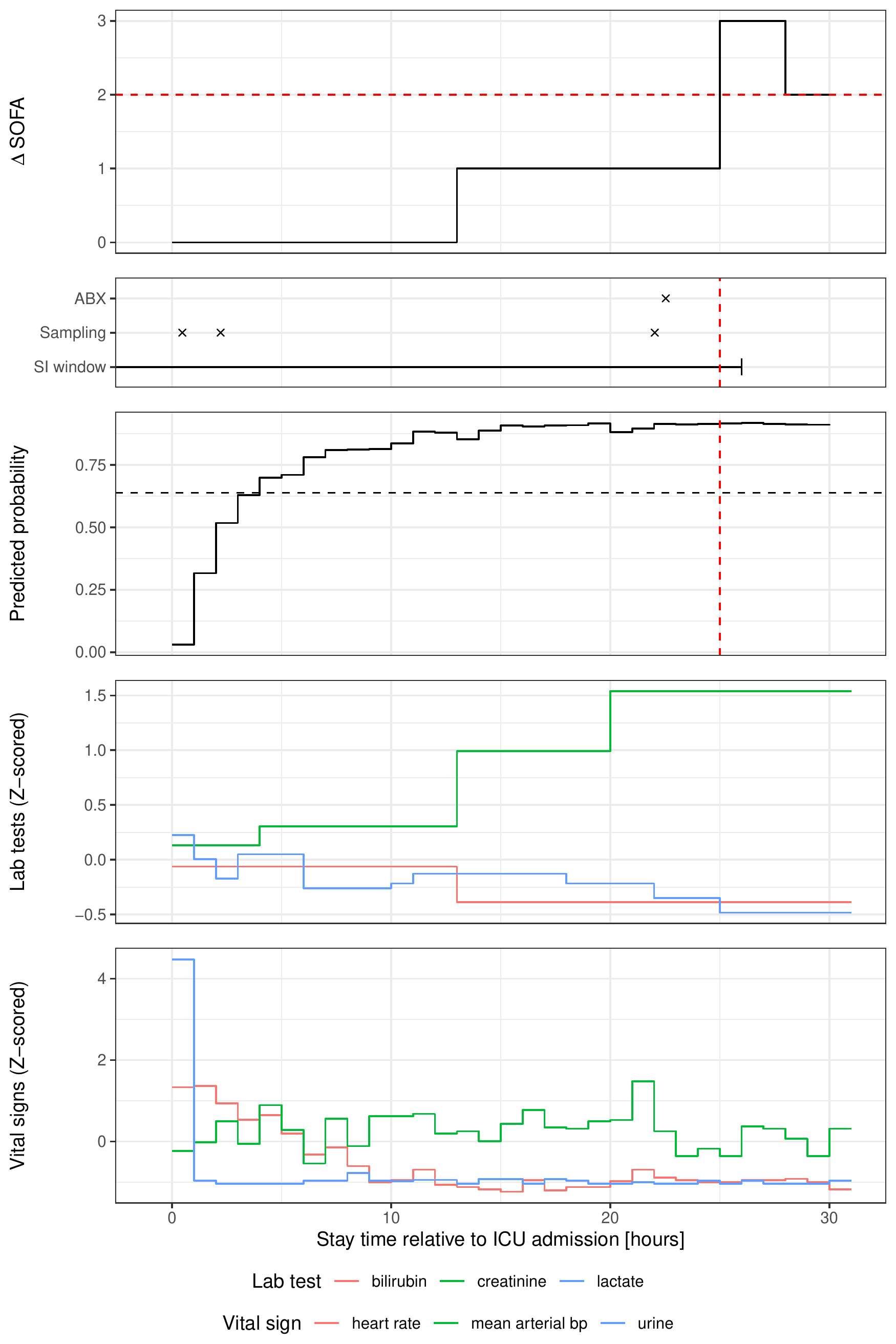}
    \caption{ Illustration of the sepsis warning score.
    Using a label derived from Sepsis-3 onset times, machine learning methods were trained to predict sepsis onset using a wide array of lab test measurements alongside vital signs. Here, a sepsis warning score based on our attention model (attn) is illustrated for one sample encounter (of an unseen testing database) together with a subset of vitals and labs that were used for prediction. In the top two rows, the sepsis label is shown decomposed into its components, the suspected infection (SI) window (consisting of antibiotics (ABX) administration coinciding with body fluid sampling), and an acute increase in SOFA ($\Delta\mathrm{SOFA}$) of two or more points. Red lines indicate the point at which $\Delta\mathrm{SOFA} \leq 2$. A decision threshold based on $80\%$ recall is indicated by the black horizontal dashed line. The displayed model was trained on eICU and here applied to AUMC.
} 
    \label{fig:example}
\end{figure}

\subsection*{Inspection of model features}

To interpret the predictions of our attention model, we calculated
Shapley values for each dataset~\citep{Lundberg17}. 
This enables us to explain and interpret the predictions of a model in
a unified manner, which even accounts for the directionality of an
effect. 
In the following analysis, we focused on (patho-)physiological signals by restricting ourselves to ``raw'' measurements, i.e., we
discarded count variables, missingness indicator variables, and derived
features.

\begin{figure}[tbp]
  \centering
  \newsavebox{\imga}
  \newsavebox{\imgb}
  \newsavebox{\imgc}
  \sbox{\imga}{%
    \subcaptionbox{\label{fig:shap_all}}{%
      \resizebox{0.35\linewidth}{!}{%
        \input{figures/shapley/shapley_16h_raw_bar.pgf}
      }
      \vspace{-0.75cm}
    }%
  }%
  \sbox{\imgb}{%
    \subcaptionbox{\label{fig:shap_beeplot}}{%
      \resizebox{0.60\linewidth}{!}{%
        \graphicspath{{figures/shapley/}}
        \input{figures/shapley/shapley_vx8vbt08_16h_raw_EICU_dot.pgf}
      }
    }
  }%
  \sbox{\imgc}{%
    \subcaptionbox{\label{fig:shap_scatter}}{%
      \resizebox{0.35\linewidth}{!}{%
        \graphicspath{{figures/shapley/}}
        \input{figures/shapley/shapley_vx8vbt08_16h_raw_EICU_scatter_map.pgf}
      }
    }%
  }%
   
  \begin{minipage}[t][\ht\imgb]{0.35\linewidth}
    \usebox{\imga}
    \usebox{\imgc}
  \end{minipage}%
  \begin{minipage}[b][\ht\imgb]{0.60\linewidth}
    \usebox{\imgb}
  \end{minipage}%
  \caption{%
  Feature inspection via Shapley analysis.
\protect\subref{fig:shap_all}: Mean absolute Shapley values averaged over all datasets (error bar indicates standard deviation over datasets). The top $20$ variables are displayed. Large values indicate large contributions to the model's prediction~(i.e., the prediction
of sepsis).
\protect\subref{fig:shap_beeplot}: Shapley value distributions illustrated for eICU. Features are sorted according to their overall contribution to the model prediction on this dataset. Positive Shapley values are associated with positive predictions of the model and vice versa.
\protect\subref{fig:shap_scatter}: Shapley values for
mean arterial pressure measurements exemplified on the eICU dataset.
  }
  \label{fig:Shapley}
\end{figure}
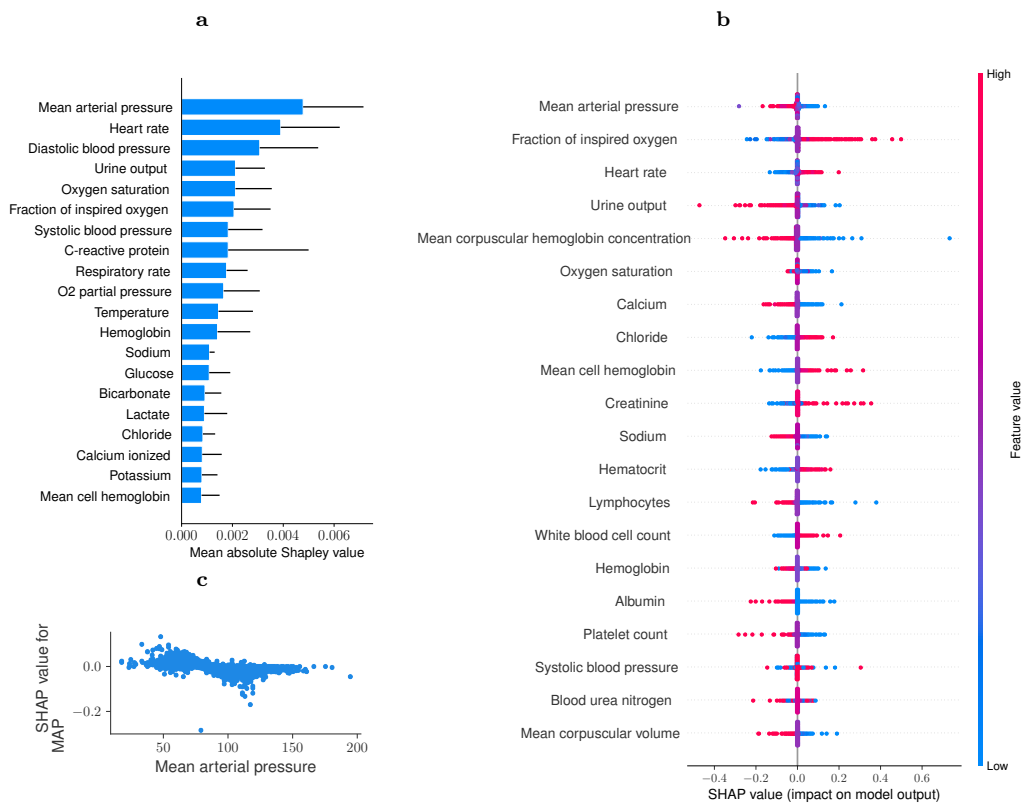

\autoref{fig:Shapley} depicts Shapley values at different granularities:
first, \autoref{fig:shap_all} shows Shapley values of the top $20$
raw variables pooled across all datasets. 
Mean arterial pressure, followed by heart rate exhibits the largest
overall contribution to predictions of our model, which suggests that the model has learned to attend to variables relevant to the assessment of hemodynamic stability. 

On a more detailed level, \autoref{fig:shap_beeplot} depicts a beeswarm plot, i.e., distributions of Shapley values for
a single dataset~(eICU), making it possible to understand the effects of increasing or decreasing an individual feature. Specifically, for each encounter we focus on how the model utilised such changes in the period of up to \SI{16}{\hour} preceding the time step at which the prediction score was maximal.
For instance, high values in mean arterial pressure resulted in
a lower prediction score, while high heart rate values led to a higher prediction score, thus encouraging a positive prediction, i.e., an alarm for sepsis.
This relationship is further illustrated in \autoref{fig:shap_scatter}.
Here, all raw mean arterial pressure measurements are shown with respect to their
Shapley value. We observe that low values in this variable~(below \SI{60}{\mmHg}) are associated with high Shapley values, meaning they are associated with positive predictions of the model. This is in line with the definition of~(septic) shock and clinical knowledge, which associates low MAP values with adverse outcomes~\citep{Leon15}. 
When comparing the beeswarm plots across datasets (see Extended Figure~\ref{fig:extended_bee}), we observe that depending on the dataset, the top ranking effects are more (e.g., eICU) or less (e.g., AUMC) aligned with clinical assumptions about sepsis. Please refer to Supplemental Figure~\ref{fig:Shapley full} for more visualisations on other datasets and a depiction of all variable types. We also assessed the performances of singular feature types (Extended Data Figure~\ref{fig:feature_ablation}) confirming that lab tests carry relevant sampling information whereas this was less the case for vital signs.

\section*{Discussion}

In this study, we presented the first sepsis prediction effort on a multi-centre,
multi-national ICU dataset. To this end, we harmonised five
ICU databases from three countries, provided hourly sepsis label
annotations, and developed an early warning system based on a deep
learning model, which we subsequently validated~(both internally and
externally) and compared to other machine learning techniques as well as
clinical baselines.
Using our rich and comprehensive ICU dataset, we
demonstrated that state-of-the-art deep learning models can
automatically learn and leverage salient features in clinical and
laboratory measurements in order to predict sepsis several hours in
advance. Moreover, when performing careful and extensive hand-crafted feature engineering of the raw time series
data, more conventional models based on statistical learning managed to
at least approach the performance of the deep learning model.
Given certain non-reducible heterogeneities inherent
to the underlying data sources~(such as the availability of specific
variables, their measurement frequency, or varying treatment policies), we observed moderate performance when training a model on a single dataset and testing it on another one. However, by pooling predictions from
models trained on different sites, which served to attenuate effects as mentioned above to some extent, we measured an average AUROC of $0.761 \pm 0.052$,
which suggests the general transferability of the model to
hitherto-unseen hospitals. While in our external
validations we observed only a minor reduction in precision, we found that
alarm earliness suffers when applying a model to a previously-unseen
data distribution. When considering deployment of our early warning system in a new hospital site, an on-site fine-tuning of pretrained models may be necessary, in particular to account for a new (and possibly unknown) prevalence of sepsis in the target hospital centre.

Comparing our model explanations with the distribution of variables most frequently used in sepsis prediction, we find that the fraction of inspired oxygen, a parameter that indicates the percentage amount of oxygen a patient requires, may need to be further investigated as a hitherto-neglected pulmonary parameter for predicting sepsis~\citep{fleuren2020machine}
While initial retrospective sepsis prediction analyses have shown promising results~\citep{futoma2017learning, moor2019early}, the recent literature has been diverging: on the one hand, proprietary interests have enabled rapid model deployment~\citep{burdick2020effect, shimabukuro2017effect}, while on the other hand, studies have criticised the deployment of insufficiently validated models, and questioned the clinical relevance of studies that predicted post-hoc billing codes~\citep{moor2020early, Wong2021}. 
As outlined in a current survey~\citep{moor2020early}, the
external validation of sepsis prediction models is typically absent
from previous studies. This is a consequence of the difficulty in
getting access to high-quality annotated datasets for development as well as for validation.
With a recent study revealing the poor practical performance of
a widely-deployed sepsis prediction model by employing external
validation data~\citep{Wong2021}, our publicly-accessible multi-centre
ICU cohort promises to play a critical part in performing such an external
validation of sepsis prediction models before their clinical deployment.
Furthermore, we found that it is not necessary to physically pool  data for achieving high external validation performance, but that pooling models trained on different databases is sufficient. Having presented a framework for harmonising, annotating and modelling such data, we hope these findings and this platform invites researchers to further contribute interoperable models to further improve sepsis prediction in a distributed learning setting that allows for side stepping challenges with sharing sensitive hospital data.

Finally, our study faced certain limitations: despite the
large resulting sample size, many patients and even sites (in the case
of the eICU dataset) had to be excluded from all analyses due to their
insufficient data quality. Such exclusions may introduce selection
effects, which is also exacerbated by the fact that our cohort is mostly
Caucasian. Moreover, because of the differences in availability of culture sampling between the datasets, two slightly different definitions of the suspected infection cohort were adopted when implementing Sepsis-3. Last, even though we
simulated a real-time prediction scenario in this study, a prospective
evaluation is nevertheless necessary in order to assess the clinical utility of bed-side sepsis predictions. We consider our cohort and
analysis to pave the way for clinical validation
studies to deploy models that could be ex-ante externally validated using our cohort. In this context, we hypothesise that the use of domain adaptation
techniques~\citep{farahani2020brief} will further
increase model transferability between different sites, thus
potentially facilitating deployments in practice via pre-trained models.

\section*{Methods}

\subsection*{Study design}

This investigation was designed as a retrospective multi-centre cohort study. This involved the creation of a harmonised and annotated multi-centre ICU cohort as well as the development, internal and external validation of a sepsis early warning system. Data cleaning, harmonisation, sepsis endpoint implementation, model implementation, model training, and evaluation were performed at the Department of Biosystems Science and Engineering as well as the Department of Mathematics, both of which are a part of ETH Zurich, Switzerland.

\subsection*{Data sources}

The multi-centre cohort of septic and non-septic ICU patients was constructed via the harmonisation of the following five databases~(versions are provided wherever available):
\begin{inparaenum}[i)]
\item AUMC: 1.0.2, \item eICU: 2.0, \item MIMIC-III: 1.4, \item HiRID: 1.1.1, and \item Emory (as part of the Physionet 2019 Challenge dataset). Notice that hourly-resolved sepsis annotations were available only in the Emory cohort, whereas for all remaining datasets the Sepsis-3 definition was implemented~\citep{singer2016third}.
\end{inparaenum}

\subsection*{Exclusion criteria}

This section defines the exact exclusion criteria that were used for patient filtering. The steps are also visually shown in the flow diagram in \autoref{fig:studyflow}. Namely, the following filtering steps were applied:
\begin{enumerate}[a)]
    \item non-adult patients ($< 14$ years) were excluded,
    \item hospital centres in the eICU dataset that had low Sepsis-3 prevalence were removed ($< 15$ \%), as it is likely that such hospitals would contribute negative cases (controls), which might in fact correspond to Sepsis-3 cases, but are not labelled as such due to data missingness; the list of hospital IDs that were used can be found in the \texttt{config/cohorts.json} file in the code repository.
\end{enumerate}
Furthermore, we excluded patients that satisfied at least one of the conditions outlined below:
\begin{enumerate}[i)]
    \item an ICU length of stay shorter than 6 hours,
    \item recorded measurements at fewer than 4 different hourly time points,
    \item a missing data window longer than 12 hours,
    \item the onset of sepsis outside ICU stay,
    \item the onset of sepsis before 4 hours into ICU stay, or after 168 hours into ICU stay.
\end{enumerate}

\subsection*{Data extraction, cleaning, mapping and preprocessing}

Since the data originates from multiple hospitals~(in different countries and even on different continents), some additional checks were required in order to guarantee their harmonisation. Specifically, for the four databases that were preprocesssed~(MIMIC-III, eICU, HiRID, AUMC), unit synchronisation and filtering of values outside of clinically valid ranges (determined by an experienced ICU clinician) were applied automatically in the \texttt{ricu} package. Furthermore, we manually inspected whether the distributions of biomarkers were similar across all five datasets. We plotted the density of all biomarkers, stratified by dataset. An example of such a plot, for some vitals and labs, is given in \autoref{fig:units}.

\subsection*{Sepsis label annotation} \label{sec:sep3}

The Sepsis-3 criterion \citep{singer2016third} defined sepsis as co-occurrence of suspected infection and a SOFA increase of two or more points. We followed the approach of the original authors as closely as possible. Suspected infection was defined as co-occurrence of antibiotic treatment and body fluid sampling. More concretely, if antibiotic treatment occured first, it needed to be followed by fluid sampling within \SI{24}{\hour}. Alternatively, if fluid sampling occurred first, it needed to be followed by antibiotic treatment within \SI{72}{\hour} in order for it to be classified as suspected infection. The earlier of the two times is taken as the suspected infection~(SI) time. After this, the SI window is defined to begin \SI{48}{\hour} before the SI time and end \SI{24}{\hour} after the SI time. Within the SI window, a SOFA increase of two or more points is defined as the time of sepsis onset. \autoref{fig:s3_def} provides a visual representation.

\autoref{tab:sofa} lists all concepts that were used for constructing the Sepsis-3 label. For the SOFA components, there are two remarks we make about the score construction. For all patients who were sedated, we set the GCS score to 15. Furthermore, the urine output in the preceding 24 hours was not evaluated before 12 hours into ICU stay.

\paragraph{Suspected Infection.}

The eICU dataset reports only a small number of body fluid samplings, while the HiRID dataset reports no body fluid samplings at all. For this reason, the original definition of suspected infection is hard to implement on these datasets. Therefore, on the two datasets, we used an alternative definition of suspected infection, which was defined as a co-occurrence of multiple antibiotic administrations. We then tested whether the alternative definition coincided with the original definition on the MIMIC-III and AUMC datasets, where both antibiotic treatment and fluid sampling were reported. The comparison of the two definitions, in terms of Venn diagrams, is given in \autoref{fig:venn}, where we use data prior to filtering of early onsets.

What can be observed is that on the MIMIC-III dataset, the two definitions overlap to a large extent~(Jaccard similarity $0.69$). On the AUMC dataset, the multiple antibiotics definition was broader than the original definition, but it included almost all patients in the original definitions~(Jaccard similarity $0.42$). The reason for this was that the majority of admissions in the AUMC database were surgical, and such patients are frequently on antibiotics for prophylactic purposes. A comparison of the two definitions on the subset of non-surgical admissions in the AUMC database showed a very strong overlap~(Jaccard similarity $0.78$) and is shown in \autoref{fig:siaumcnonsurg}.

\subsection*{Machine learning methods}

In this study, we investigated a comprehensive selection of supervised ML approaches. This includes
\begin{inparaenum}[i)]
    \item deep self-attention models (attn)~\citep{vaswani2017attention}
    \item recurrent neural networks employing gated recurrent units (gru)~\citep{cho2014properties}
    \item LightGBM gradient boosting trees (lgbm)~\citep{ke2017lightgbm}, and
    \item LASSO-regularised~\citep{Tibshirani96} logistic regression (lr).
\end{inparaenum}

All models were tuned to minimise the binary cross-entropy loss of the prediction score with the binary sepsis label on the time-point level.
For the non-deep classifiers we ran a randomised hyperparameter search with $50$ iterations of a stratified $5$-fold cross-validation on the first training split. For the deep learning models depth, width, learning rate, batch size, and weight decay were selected based on a dedicated online validation split of the training data. The performance in terms of cross-entropy loss on this split was monitored during training and the model state where the performance was best was used for the final evaluation. The hyperparameter selection process was split into two stages: first, a search over a coarse grid (see \autoref{tab:deep_hparams} for the exact values) was performed by randomly selecting $25$ of the possible parameter combinations and evaluating the models on them. Afterwards a second finer grid around the previously selected parameters was constructed and further $25$ randomly selected parameter combinations were evaluated.  The model architecture (depth and width) was not changed in the second stage.  All models were trained with a positive weight dependent on the class imbalance.  Training was stopped after 100 epochs (iterations through the dataset) and the hyperparameter search was always performed on the training split of the first repetition of train-validation splitting. Finally, using the best hyperparameters, for each method $5$ repetition models were fitted on $5$ different repetitions of train-validation splitting of the development data. This strategy was applied to each database independently.

\subsection*{Feature Engineering}

For the included non-deep classifiers, which were not originally designed for sequential data, we extracted $1,269$ features based on multi-scale look-back statistics (mean, median, variance, minimum and maximum over the last 4, 8, and 16 hours), measurement counts, missingness indicators, the raw measurement values of the current time step, and hand-crafted features derived from domain knowledge (for instance shock index, oxygenation index, or available lab and vital components of baseline scores like SOFA, SIRS, or MEWS). For the deep models, which are capable of automatically learning feature representations from sequential data, we refrained from employing manual multi-scale look-back statistics, resulting in $190$ features including $59$ observed measurements, $59$ missingness indicators, $59$ measurement counts, $9$ derived domain-knowledge features (ratios and partial scores) and $4$ static variables representing demographic information. During preliminary experiments, we observed that including the static information had a slight detrimental effect on performance of the attention model; we therefore discarded such information for this model in our experiments.

\subsection*{Experimental setup}

All datasets were split into a development set ($90\%$) and a held-out test set ($10\%$). For $5$ independent repetitions, the development set was randomly divided into a training set and validation set such that they account for $80\%$ and $10\%$ of the total data. All splits were stratified to preserve the dataset-specific sepsis prevalence. We optimised all ML models on the first partition~(out of $5$ repetitions) of the development set, whereas the validation split was used for fine-tuning hyperparameters. Next, each model was refit on the respective train split of each of the five repetitions, which allowed to assess model robustness with regard to varying training data. Finally, all performance measures are reported on the respective held-out test split. To maximise comparability between internal and external evaluations, in both settings, identical test splits are evaluated.
For the sake of performance metrics being comparable across datasets, upon testing time we harmonised the prevalence of sepsis cases to the across-dataset average of $17\%$. This was achieved by randomly subsampling either controls for increasing the prevalence, or cases for decreasing the prevalence. To ensure that valuable case information is not lost during this procedure, we repeated all subsamplings $10$ times and confirmed that the coverage of sepsis cases is above $98.3\%$. Finally, all results are computed as averages over these $10$ subsamplings.

\subsection*{Feature inspection via Shapley values}
We measured the importance and relevance of individual features by calculating Shapley values~\citep{Lundberg17} via the integrated gradients method~\citep{sundararajan17a} for our attention-based deep learning model. This method ``explains'' a prediction of a trained model by assessing the contribution of each feature. All computations were performed on the dataset-specific held-out test split~(with the model having no access to such test data during training, as described in ``Experimental setup''). For this, we consistently inspected the model as fitted on the first repetition split of the development split. We decided to sample $500$ patient stays at random from each dataset and repeat this procedure five times, as the implementation of integrated gradients precludes calculations on the full dataset due to excessive memory requirements.
Since we are specifically interested in contributions ahead of a potential sepsis onset, we chose the time point with the maximum model prediction as an anchor for each patient stay. We then restricted the stay to the \SI{16}{\hour} preceding this time point~(making the resulting time windows at most \SI{16}{\hour} long) and calculated Shapley values for each remaining time point.
The advantage of this procedure is that it 
restricts our view to \emph{recent} time points, making the Shapley values more pertinent for a specific prediction, as opposed to incorporating a potentially large number of time points, which may in turn result in de-emphasising a potential signal.

\subsection*{Human subject data}

The data in MIMIC-III has been previously de-identified, and the institutional review boards (IRBs) of the Massachusetts Institute of Technology (No. 0403000206) and BIDMC (2001-P-001699/14) approved the use of the database for research. The eICU database is released under the Health Insurance Portability and Accountability Act (HIPAA) safe harbour provision. The re-identification risk was certified as meeting safe harbour standards by Privacert (Cambridge, MA) (HIPAA Certification no. 1031219-2). The IRB of the Canton of Bern approved the HiRID data collection and its anonymised use for research.
The AUMC dataset was certified according to NEN7510 (ISO 27001), which ensures that
strict information governance protocols are in place for the protection of source data. Furthermore, according to the original source~\citep{thoral2021sharing}, two independent teams auditing the project reported that the design, database management and governance were state-of-the-art, and therefore the data was considered as anonymous information in the context of the General Data Protection Regulation.

\subsection*{Data availability}

All the data used in this manuscript is publicly available. The MIMIC-III, eICU, HiRID, and the Emory part of the Physionet 2019 Computing in Cardiology Challenge data are available via \href{https://physionet.org/}{Physionet}~\citep{goldberger2000physiobank}. Data from the AUMC database is available via the website of \href{https://amsterdammedicaldatascience.nl/}{Amsterdam Medical Data Science}. The data loading was performed using the \texttt{ricu} R-package \citep{bennett2021ricu} (available from https://cran.r-project.org/package=ricu). All the code used for extracting, cleaning, filtering and modelling will be made available immediately upon publication, ensuring end-to-end reproducibility of all results presented. 

\subsection*{Acknowledgements}
This study was supported by the grant \#2017-110 of the Strategic Focal Area “Personalized Health and Related Technologies (PHRT)” of the ETH Domain for the SPHN/PHRT Driver Project “Personalized Swiss Sepsis Study”. 
The study was also funded by the Alfried Krupp Prize for Young University Teachers of the Alfried Krupp von Bohlen und Halbach-Stiftung (K.B.). We thank \url{https://diagrams.net/} for their open-access tool for creating figures of unrestricted usage.

\subsection*{Author contributions}
M.M., N.B., D.P and K.B. conceived the study. N.M., P.B. and K.B. supervised the study.
M.M., N.B., D.P., M.H., B.R., P.B., K.B. designed the experiments. N.B., D.P. performed the cleaning, harmonisation and label annotation. M.M., M.H. implemented the filtering and feature extraction. M.H., M.M implemented the deep learning models. M.M., N.B., D.P. implemented the non-deep ML models. N.B. implemented the clinical baselines. M.M. designed the encounter-focused evaluation. M.M., B.R. implemented encounter-focused evaluation plots. B.R., M.M. implemented and designed the performance plots. M.H., B.R. implemented the Shapley value calculation. B.R. designed, implemented, and performed the Shapley value analysis. M.M. ran the internal and external validation experiments for all methods. M.M. ran the hyperparameter search of the deep learning models and LightGBM. B.R. ran the hyperparameter search of Logistic regression. N.B. investigated different feature sets. M.M. implemented and ran the max pooling strategy. M.M. designed the pipeline overview figure. D.P., B.R. designed the data harmonisation figure. N.B. designed the risk score illustration and the study flow chart. D.P. devised the dataset table. P.B. and K.B. advised on algorithmic modelling, statistical interpretation and evaluation. 
All authors contributed to the interpretation of the findings and to the writing of the manuscript. 

\subsection*{Competing interests}

The authors declare no competing interests. The funders of this study had no role in study design, data
collection \& analysis, decision to publish, or preparation of the manuscript.

\bibliographystyle{abbrvnat}
\bibliography{ref}

\ifarxiv
  \clearpage
\else

\section{Figure Legends}
\subsection*{\autoref{fig:preprocessin} | Overview of the preprocessing pipeline.}
Panel~\subref{fig:cleaning}: Data from five ICU EHR data bases are collected, cleaned and harmonised.
Panel~\subref{fig:pipeline}: Harmonised data are extracted for sepsis label annotation (left) as well as feature extraction (right) resulting in labels and features that are used for training the model.

\subsection*{\autoref{fig:vars_and_task} | Data harmonisation, annotation and task formulation.}
\subref{fig:venn}: Comparison of the original suspected infection definition (antibiotics + fluid sampling) against the alternative suspected infection definition based on multiple antibiotics (which was used when no culture sampling was available) displayed for MIMIC-III, a dataset that allows for both implementations. The two variations showed a Jaccard similarity of 0.69. For further visualisations, refer to \autoref{fig:siaumcnonsurg} in the Supplementary.  
\subref{fig:s3_def}: Visual representation of how the onset time of sepsis was defined following Sepsis-3~\citep{singer2016third}. The label is constructed by evaluating the SOFA score on hourly basis and determining suspicion of infection (SI) windows based on body fluid sampling and antibiotics administration~\citep{singer2016third, seymour2016assessment}. The criterion is fulfilled if (and at the time of) an acute increase in SOFA of at least 2 points coincides with an SI window. 
\subref{fig:units}: Monitoring of distributions of measurements across all five datasets. These checks were applied to ensure that units were harmonised.
\subref{fig:task}: Prediction task for a given case and control patient. Data is available for the whole ICU stay. We set the label of the case to be $1$ starting from \SI{6}{\hour} \emph{before} sepsis onset~(red line) until
\SI{24}{\hour} \emph{after} sepsis onset. During training, this allows for penalising late false negative predictions. If more than
\SI{24}{\hour} after onset have passed, any further available data is not considered~(grey region).

\subsection*{\autoref{fig:internal_roc} | Predictive performance plots. }
\subref{fig:in_aumc_subset}-\subref{fig:in_aumc_scatter_subset}: Internal (left) and external (right) validation of early sepsis prediction illustrated for the AUMC dataset. Our deep learning approach (attn) is visualised together with a subset of the comparison methods including clinical baselines (SOFA, MEWS and NEWS), as well as the logistic regression model (lr) (for all methods and datasets refer to Extended Data Figure~\ref{fig:aumc_roc}-\ref{fig:hirid_roc}).
    \subref{fig:in_aumc_subset}: ROC Curves are computed on an encounter level following our threshold-based evaluation strategy. The error bars indicate standard deviation over $5$ repetition of train-validation splitting.
    \subref{fig:in_aumc_scatter_subset}: We display the trade-off between prediction accuracy and earliness. For this, recall is fixed at $80\%$ and precision is scattered against the median number of hours that the alarm precedes sepsis onset.
    \subref{fig:roc_heatmap}: External validation of our deep self-attention model. The encounter-level AUROC (mean over five repeated training splits) is displayed. The columns indicate the evaluation dataset, in the rows the training dataset is listed. In the last row, for each evaluation dataset, the performance of an ensemble of all other datasets is shown (where the respective maximum prediction score was used). We observe that the pooling approach improves the generalisability to new datasets by outperforming or being on par with the best training dataset, which is not known a priori. To maximise comparability with the internal validation, the metrics are computed on the test split of each dataset.

\subsection*{\autoref{fig:example} | Illustration of the sepsis warning score.}
    Using a label derived from Sepsis-3 onset times, machine learning methods were trained to predict sepsis onset using a wide array of lab test measurements alongside vital signs. Here, a sepsis warning score based on our attention model (attn) is illustrated for one sample encounter (of an unseen testing database) together with a subset of vitals and labs that were used for prediction. In the top two rows, the sepsis label is shown decomposed into its components, the suspected infection (SI) window (consisting of antibiotics (ABX) administration coinciding with body fluid sampling), and an acute increase in SOFA ($\Delta\mathrm{SOFA}$) of two or more points. Red lines indicate the point at which $\Delta\mathrm{SOFA} \leq 2$. A decision threshold based on $80\%$ recall is indicated by the black horizontal dashed line. The displayed model was trained on eICU and here applied to AUMC.

\subsection*{\autoref{fig:Shapley} | Feature inspection via Shapley analysis.}
\protect\subref{fig:shap_all}: Mean absolute Shapley values averaged over all datasets (error bar indicates standard deviation over datasets). The top $20$ variables are displayed. Large values indicate large contributions to the model's prediction~(i.e., the prediction
of sepsis).
\protect\subref{fig:shap_beeplot}: Shapley value distributions illustrated for eICU. Features are sorted according to their overall contribution to the model prediction on this dataset. Positive Shapley values are associated with positive predictions of the model and vice versa.
\protect\subref{fig:shap_scatter}: Shapley values for
mean arterial pressure measurements exemplified on the eICU dataset.

\subsection*{Extended Data Figure~\ref{fig:studyflow} | Study flow chart.}
Patient filtering steps applied to the data sets yielding the final study cohort of 156,309 patients. Parenthesized numbers refer to the individual data sets MIMIC-III (M), eICU (E), HiRID (H), AUMC (A) and Emory (P), respectively.

\subsection*{Extended Data Figure~\ref{fig:aumc_roc} | Predictive performance plots for the AUMC dataset and all methods.} 
    Internal (left) and external (right) validation of early sepsis prediction illustrated for the AUMC dataset. Our deep learning approach (attn) is visualised together with the comparison methods including clinical baselines (SOFA, MEWS, NEWS, qSOFA and SIRS), and ML methods: logistic regression model (lr), LightGBM (lgbm), and recurrent neural network employing Gated Recurrent Units~(gru).
    \subref{fig:in_aumc}: ROC Curves are computed on an encounter level following our threshold-based evaluation strategy. The error bars indicate standard deviation over $5$ repetition of train-validation splitting.
    \subref{fig:in_aumc_scatter}: We display the trade-off between prediction accuracy and earliness. For this, recall is fixed at $80\%$ and precision is plotted against the median number of hours that the alarm precedes sepsis onset.

\subsection*{Extended Data Figure~\ref{fig:mimic_roc} | Predictive performance plots for the MIMIC-III dataset and all methods.} 
    Internal (left) and external (right) validation of early sepsis prediction illustrated for the MIMIC-III dataset. Our deep learning approach (attn) is visualised together with the comparison methods including clinical baselines (SOFA, MEWS, NEWS, qSOFA and SIRS), and ML methods: logistic regression model (lr), LightGBM (lgbm), and recurrent neural network employing Gated Recurrent Units~(gru).
    \subref{fig:in_mimic}: ROC Curves are computed on an encounter level following our threshold-based evaluation strategy. The error bars indicate standard deviation over $5$ repetition of train-validation splitting.
    \subref{fig:in_mimic_scatter}: We display the trade-off between prediction accuracy and earliness. For this, recall is fixed at $80\%$ and precision is plotted against the median number of hours that the alarm precedes sepsis onset.
    
\subsection*{Extended Data Figure~\ref{fig:eicu_roc} | Predictive performance plots for the eICU dataset and all methods.} 
    Internal (left) and external (right) validation of early sepsis prediction illustrated for the eICU dataset. Our deep learning approach (attn) is visualised together with the comparison methods including clinical baselines (SOFA, MEWS, NEWS, qSOFA and SIRS), and ML methods: logistic regression model (lr), LightGBM (lgbm), and recurrent neural network employing Gated Recurrent Units~(gru).
    \subref{fig:in_eicu}: ROC Curves are computed on an encounter level following our threshold-based evaluation strategy. The error bars indicate standard deviation over $5$ repetition of train-validation splitting.
    \subref{fig:in_eicu_scatter}: We display the trade-off between prediction accuracy and earliness. For this, recall is fixed at $80\%$ and precision is plotted against the median number of hours that the alarm precedes sepsis onset.
    
\subsection*{Extended Data Figure~\ref{fig:hirid_roc} | Predictive performance plots for the HiRID dataset and all methods.} 
    Predictive performance plots for the HiRID dataset and all methods.
    Internal (left) and external (right) validation of early sepsis prediction illustrated for the HiRID dataset. Our deep learning approach (attn) is visualised together with the comparison methods including clinical baselines (SOFA, MEWS, NEWS, qSOFA and SIRS), and ML methods: logistic regression model (lr), LightGBM (lgbm), and recurrent neural network employing Gated Recurrent Units~(gru).
    \subref{fig:in_hirid}: ROC Curves are computed on an encounter level following our threshold-based evaluation strategy. The error bars indicate standard deviation over $5$ repetition of train-validation splitting.
    \subref{fig:in_hirid_scatter}: We display the trade-off between prediction accuracy and earliness. For this, recall is fixed at $80\%$ and precision is plotted against the median number of hours that the alarm precedes sepsis onset.
    
\subsection*{Extended Data Figure~\ref{fig:extended_bee} | Shapley value distribution plots for AUMC, eICU, HiRID, and MIMIC-III} 
The top $20$ raw measurements are being shown in the visualisation. While the precise ranking differs between databases, mean arterial pressure and heart rate are consistently part of the top $10$ features.

\subsection*{Extended Data Figure~\ref{fig:ablation} | Ablations of feature categories and cohorts.}
     \subref{fig:feature_ablation}: Having observed that count features appeared frequently among the top $20$ features explaining high prediction scores in \autoref{fig:Shapley full}, we investigated how a model performed when trained solely on counts or raw measurements (here internally validated on MIMIC-III). We further subdivided the features into lab tests and vital signs which confirmed our hypothesis that lab tests carry sampling information (as a clinician requested them), whereas this is less the case for vital signs. 
    \subref{fig:cohort_ablation}: We observed the best internal validation performance on AUMC, a dataset with a large proportion of surgical patients. Thus, we applied a model that was trained on AUMC to the medical and surgical cohorts both in-distribution (AUMC) and out-of-distribution (MIMIC-III) and observed that the surgical cohort within AUMC is indeed easier to classify, whereas this does not necessarily generalise to other datasets.

\subsection*{Extended Data Figure~\ref{fig:pooled_datasets} | Pooling datasets during training. }
An auxiliary analysis of our attention model by pooling datasets during \emph{training}, as opposed to pooling predictions of models trained on different datasets. For each displayed dataset the other datasets were pooled.

\section{Tables}
\fi

\begin{landscape}
    \begin{table}[p]
\begin{tabular}{lccccc}
  \hline
Variable & MIMIC-III & eICU & HiRID & AUMC & Emory \\ 
  \hline
Cohort size (n) & 36,591 & 56,765 & 27,278 & 15,844 & 19,831 \\ 
  Sepsis-3 prevalence (n (\%)) & 9,541 (26) & 4,708 (8) & 10,170 (37) & 1,275 (8) & 1,050 (5) \\ 
  Age, years (Median (IQR)) & 65 (52-77) & 65 (53-76) & 65 (55-75) & 65 (55-75) & 62 (50-72) \\ 
  Ethnicity (\%) \\
  \hspace{3mm} African American & 9 & 10 & - & - & - \\
  \hspace{3mm} Asian & 2 & 1 & - & - & - \\
  \hspace{3mm} Caucasian & 71 & 82 & - & - & - \\
  \hspace{3mm} Hispanic & 3 & 2 & - & - & - \\
  \hspace{3mm} Other & 15 & 5 & - & - & - \\
  In-hospital mortality (\%) & 8 & 7 & 5 & 5 & - \\ 
  ICU LOS, days (Median (IQR)) & 1.99 (1.15-3.63) & 1.71 (0.95-3.01) & 0.97 (0.8-1.95) & 0.97 (0.81-1.82) & - \\ 
  Hospital LOS, days (Median (IQR)) & 6.43 (3.82-11.14) & 5.53 (2.99-9.89) & - & - & - \\ 
  Gender, female (\%) & 44 & 45 & 37 & 35 & 46 \\ 
  Gender, male (\%) & 56 & 55 & 63 & 65 & 54 \\ 
  Ventilated patients (n (\%)) & 16,499 (45) & 24,534 (43) & 14,021 (51) & 10,469 (66) & - \\ 
  Patients on vasopressors (n (\%)) & 9,669 (26) & 6,769 (12) & 7,721 (28) & 7,980 (50) & - \\ 
  Patients on antibiotics (n (\%)) & 21,598 (59) & 21,847 (38) & 17,152 (63) & 11,165 (70) & - \\ 
  Patients with suspected infection (n (\%)) & 16,349 (45) & 9,739 (17) & 15,160 (56) & 1,639 (10) & - \\ 
  Initial SOFA (Median (IQR)) & 3 (1-4) & 3 (1-5) & 5 (3-8) & 6 (3-7) & - \\ 
  SOFA components (Median (IQR))\\
  \hspace{3mm} Respiratory & 1 (0-2) & 1 (0-2) & 3 (2-4) & 2 (1-3) & - \\ 
  \hspace{3mm} Coagulation & 0 (0-1) & 0 (0-1) & 0 (0-1) & 0 (0-1) & - \\ 
  \hspace{3mm} Hepatic & 0 (0-1) & 0 (0-0) & 0 (0-1) & 0 (0-0) & - \\ 
  \hspace{3mm} Cardiovascular & 1 (1-1) & 1 (0-1) & 1 (1-4) & 2 (1-4) & - \\ 
  \hspace{3mm} CNS & 0 (0-1) & 0 (0-2) & 0 (0-1) & 0 (0-1) & - \\ 
  \hspace{3mm} Renal & 0 (0-1) & 0 (0-1) & 0 (0-0) & 0 (0-1) & - \\ 
  Admission type (\%)\\
  \hspace{3mm} Surgical & 38 & 19 & - & 80 & - \\ 
  \hspace{3mm} Medical & 61 & 79 & - & 15 & - \\
  \hspace{3mm} Other & 1 & 3 & - & 5 & - \\ 
  Comorbidities ICD-9 (n (\%))\\
  \hspace{3mm} Chronic Renal Failure & 5,137 (14) & 3,950 (7) & - & - & - \\ 
  \hspace{3mm} Cancer & 4,687 (13) & 2,517 (4) & - & - & - \\ 
  \hspace{3mm} Chronic Pulmonary Disease & 7,994 (22) & 4,669 (8) & - & - & - \\ 
  \hspace{3mm} Liver Disease & 3,142 (9) & 1,507 (3) & - & - & - \\ 
  \hspace{3mm} Diabetes & 9,792 (27) & 1,919 (3) & - & - & - \\ 
   \hline
\end{tabular}
    \caption{Description of our multi-center ICU cohort. This table illustrates that the different cohorts vary in the degree to which additional cohort information and metadata is available. Whereas most of the data is available for our four core datasets that we processed, many statistics are missing for the Emory dataset which was externally pre-processed and published.}
    \label{tab:datasets}
    \end{table}
\end{landscape}

\setcounter{figure}{0}
\setcounter{table}{0}
\makeatletter
\def\fnum@figure{Extended Data Figure~\thefigure}
\makeatother

\section{Extended Data Figures}
\begin{figure}[h]
    \centering
    \includegraphics[width=1.\linewidth]{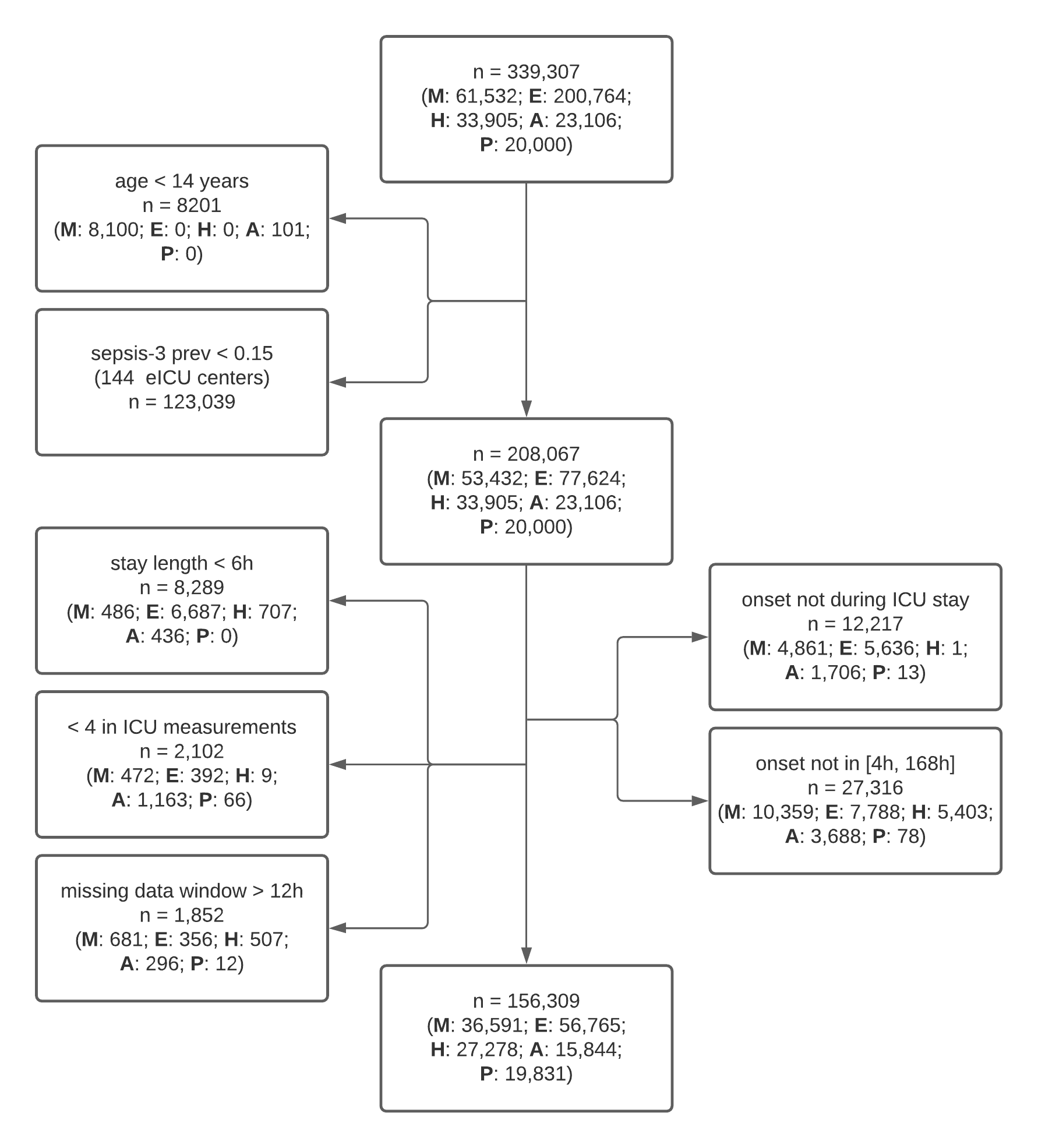}
    \caption{Patient filtering steps applied to the data sets yielding the final study cohort of 156,309 patients. Parenthesised numbers refer to the individual data sets MIMIC-III (M), eICU (E), HiRID (H), AUMC (A) and Emory (P), respectively.}
    \label{fig:studyflow}
\end{figure}

\begin{figure}
    \centering
    \subcaptionbox{\label{fig:in_aumc}}{
        \includegraphics[width=0.8\linewidth]{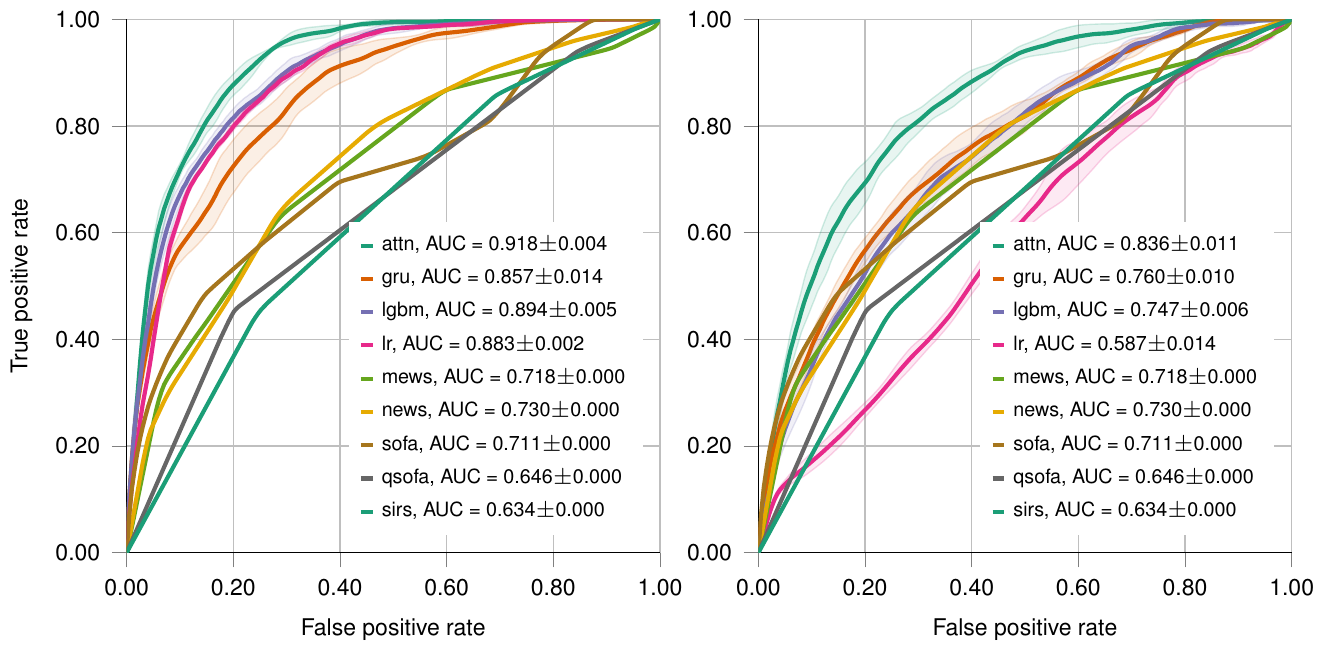}
    }\\%
    \subcaptionbox{\label{fig:in_aumc_scatter}}{
        \includegraphics[width=0.82\linewidth]{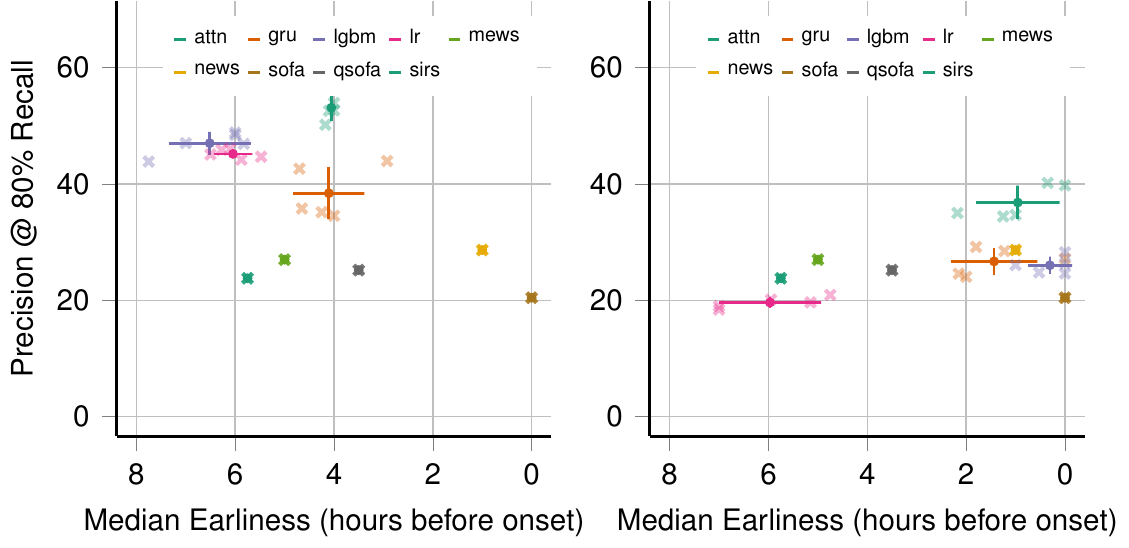}
    }\\%
    \caption{Predictive performance plots for the AUMC dataset and all methods. Internal (left) and external (right) validation of early sepsis prediction illustrated for the AUMC dataset. Our deep learning approach (attn) is visualised together with the comparison methods including clinical baselines (SOFA, MEWS, NEWS, qSOFA and SIRS), and ML methods: logistic regression model (lr), LightGBM (lgbm), and recurrent neural network employing Gated Recurrent Units~(gru).
    \subref{fig:in_aumc}: ROC Curves are computed on an encounter level following our threshold-based evaluation strategy. The error bars indicate standard deviation over $5$ repetition of train-validation splitting.
    \subref{fig:in_aumc_scatter}: We display the trade-off between prediction accuracy and earliness. For this, recall is fixed at $80\%$ and precision is plotted against the median number of hours that the alarm precedes sepsis onset.} 
    \label{fig:aumc_roc}
\end{figure}

\begin{figure}
    \centering
    \subcaptionbox{\label{fig:in_mimic}}{
        \includegraphics[width=0.8\linewidth]{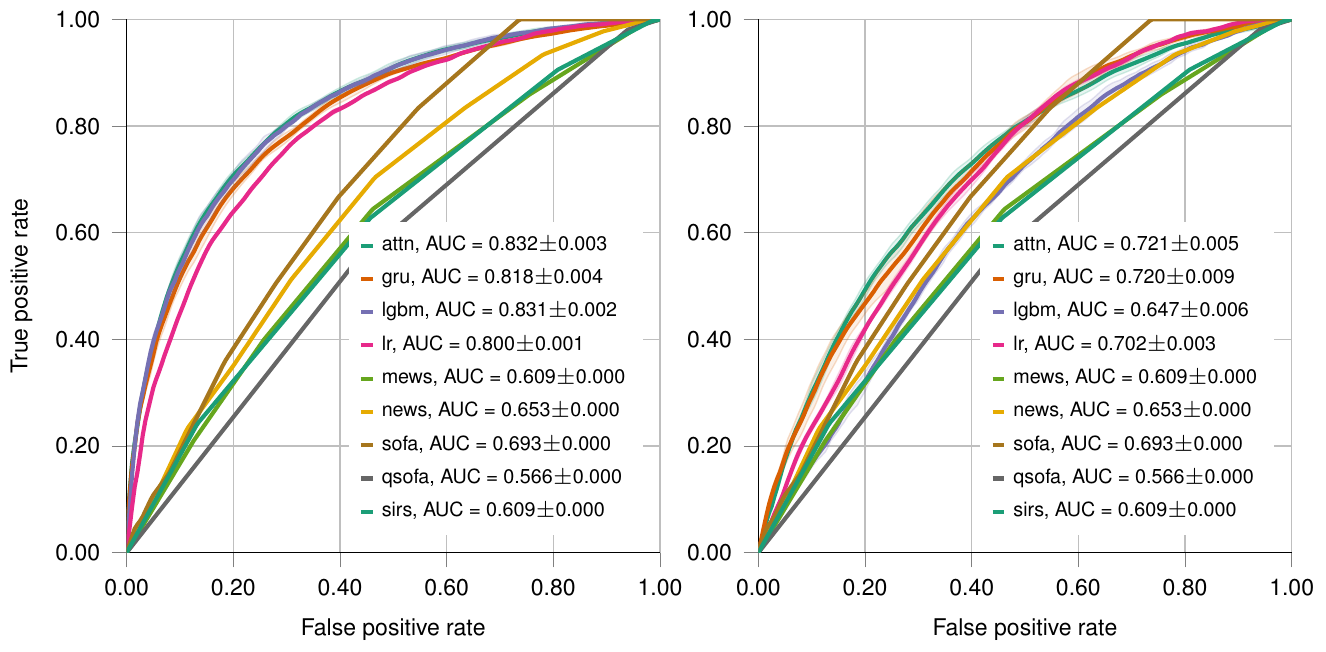}
    }\\%
    \subcaptionbox{\label{fig:in_mimic_scatter}}{
        \includegraphics[width=0.82\linewidth]{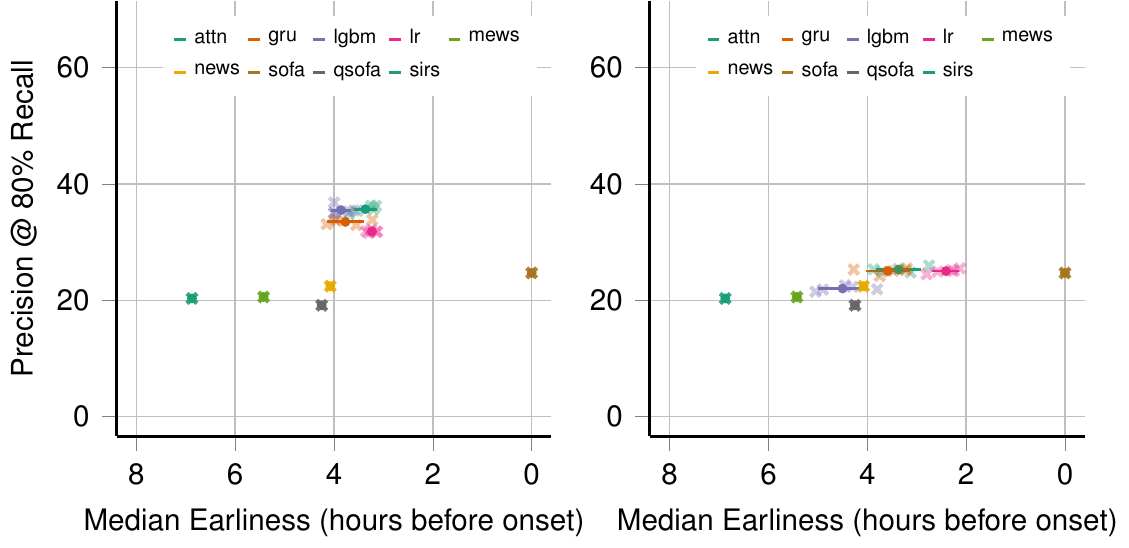}
    }\\%
    \caption{
    Predictive performance plots for the MIMIC-III dataset and all methods.
    Internal (left) and external (right) validation of early sepsis prediction illustrated for the MIMIC-III dataset. Our deep learning approach (attn) is visualised together with the comparison methods including clinical baselines (SOFA, MEWS, NEWS, qSOFA and SIRS), and ML methods: logistic regression model (lr), LightGBM (lgbm), and recurrent neural network employing Gated Recurrent Units~(gru).
    \subref{fig:in_mimic}: ROC Curves are computed on an encounter level following our threshold-based evaluation strategy. The error bars indicate standard deviation over $5$ repetition of train-validation splitting.
    \subref{fig:in_mimic_scatter}: We display the trade-off between prediction accuracy and earliness. For this, recall is fixed at $80\%$ and precision is plotted against the median number of hours that the alarm precedes sepsis onset.
    } 
    \label{fig:mimic_roc}
\end{figure} 

\begin{figure}
    \centering
    \subcaptionbox{\label{fig:in_eicu}}{
        \includegraphics[width=0.8\linewidth]{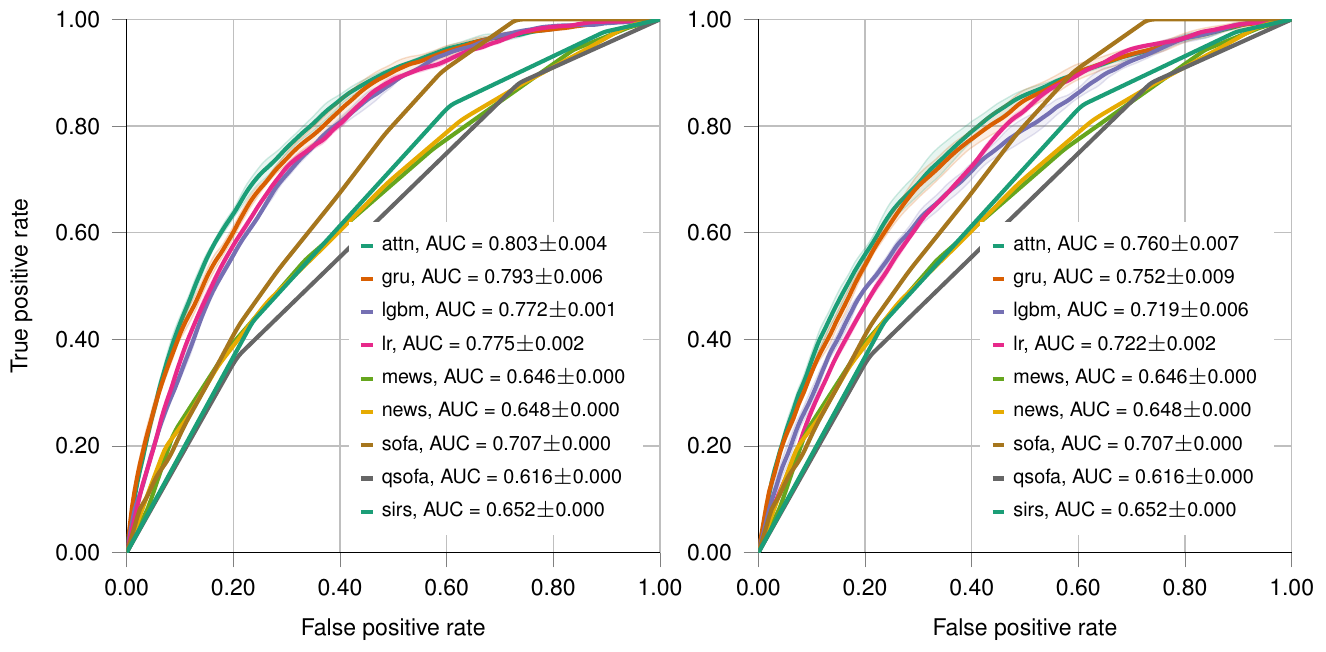}
    }\\%
    \subcaptionbox{\label{fig:in_eicu_scatter}}{
        \includegraphics[width=0.82\linewidth]{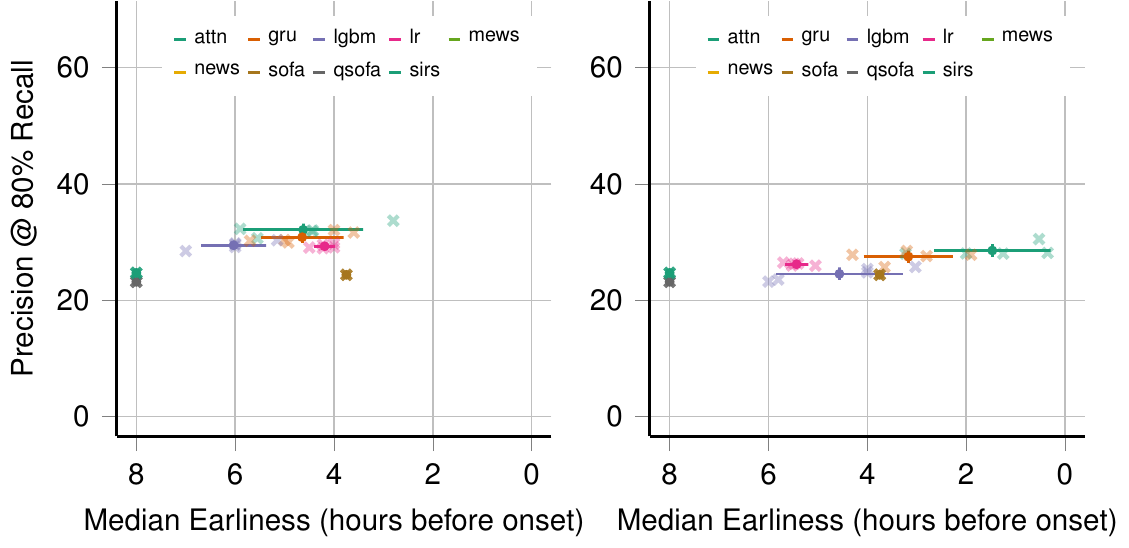}
    }\\%
    \caption{
      Predictive performance plots for the eICU dataset and all methods.
     Internal (left) and external (right) validation of early sepsis prediction illustrated for the eICU dataset. Our deep learning approach (attn) is visualised together with the comparison methods including clinical baselines (SOFA, MEWS, NEWS, qSOFA and SIRS), and ML methods: logistic regression model (lr), LightGBM (lgbm), and recurrent neural network employing Gated Recurrent Units~(gru).
    \subref{fig:in_eicu}: ROC Curves are computed on an encounter level following our threshold-based evaluation strategy. The error bars indicate standard deviation over $5$ repetition of train-validation splitting.
    \subref{fig:in_eicu_scatter}: We display the trade-off between prediction accuracy and earliness. For this, recall is fixed at $80\%$ and precision is plotted against the median number of hours that the alarm precedes sepsis onset.
    } 
    \label{fig:eicu_roc}
\end{figure} 

\begin{figure}
    \centering
    \subcaptionbox{\label{fig:in_hirid}}{
        \includegraphics[width=0.8\linewidth]{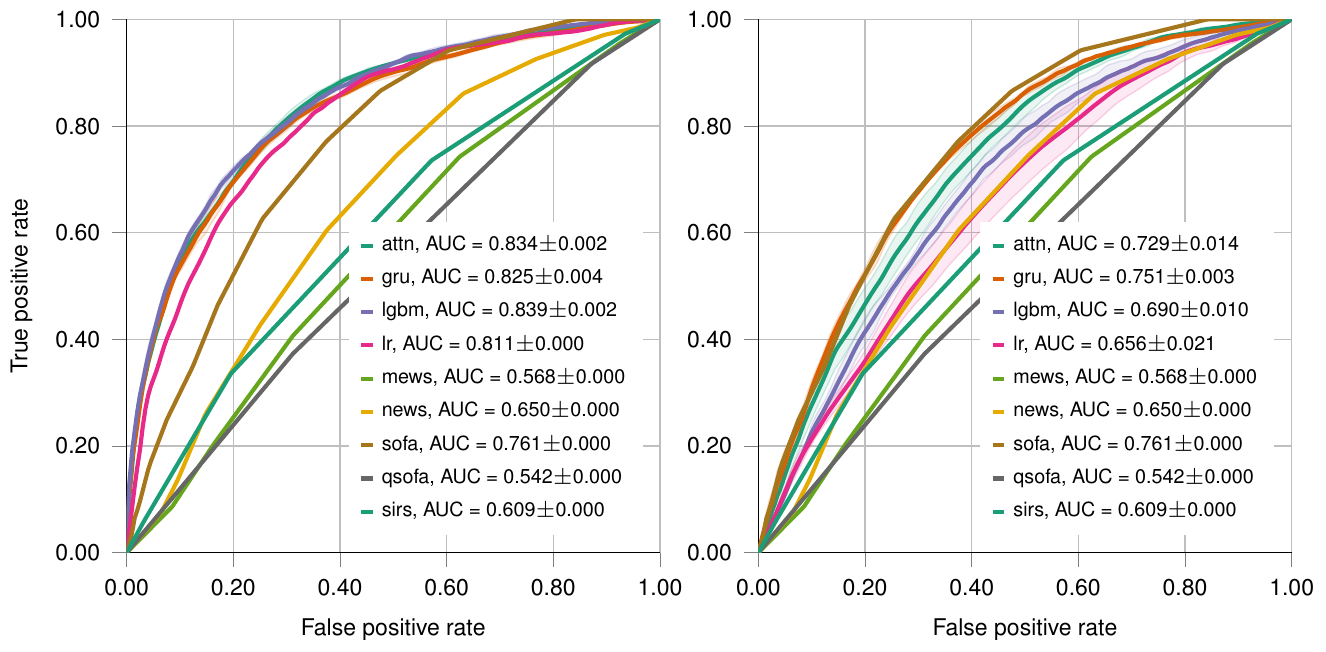}
    }\\%
    \subcaptionbox{\label{fig:in_hirid_scatter}}{
        \includegraphics[width=0.82\linewidth]{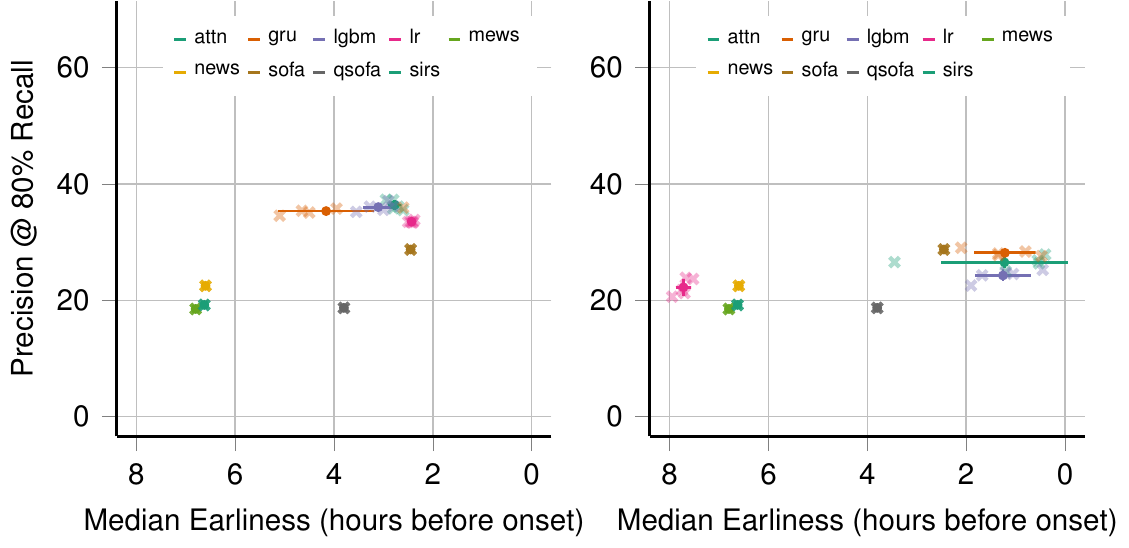}
    }\\%
    \caption{
     Predictive performance plots for the HiRID dataset and all methods.
    Internal (left) and external (right) validation of early sepsis prediction illustrated for the HiRID dataset. Our deep learning approach (attn) is visualised together with the comparison methods including clinical baselines (SOFA, MEWS, NEWS, qSOFA and SIRS), and ML methods: logistic regression model (lr), LightGBM (lgbm), and recurrent neural network employing Gated Recurrent Units~(gru).
    \subref{fig:in_hirid}: ROC Curves are computed on an encounter level following our threshold-based evaluation strategy. The error bars indicate standard deviation over $5$ repetition of train-validation splitting.
    \subref{fig:in_hirid_scatter}: We display the trade-off between prediction accuracy and earliness. For this, recall is fixed at $80\%$ and precision is plotted against the median number of hours that the alarm precedes sepsis onset.
    } 
    \label{fig:hirid_roc}
\end{figure} 

\begin{figure}
  \centering
  \subcaptionbox{AUMC\label{fig:extended_bee_AUMC}}{
    \resizebox{0.50\linewidth}{!}{%
      \graphicspath{{figures/shapley/}}
      \input{figures/shapley/shapley_j76ft4wm_16h_raw_AUMC_dot.pgf}
    }
  }%
  \subcaptionbox{eICU\label{fig:extended_bee_EICU}}{
    \resizebox{0.50\linewidth}{!}{%
      \graphicspath{{figures/shapley/}}
      \input{figures/shapley/shapley_vx8vbt08_16h_raw_EICU_dot.pgf}
    }
  }\\
  \subcaptionbox{HiRID~\label{fig:extended_bee_HIRID}}{
    \resizebox{0.50\linewidth}{!}{%
      \graphicspath{{figures/shapley/}}
      \input{figures/shapley/shapley_gjtf48im_16h_raw_Hirid_dot.pgf}
    }
  }%
  \subcaptionbox{MIMIC-III\label{fig:extended_bee_MIMIC}}{
    \resizebox{0.50\linewidth}{!}{%
      \graphicspath{{figures/shapley/}}
      \input{figures/shapley/shapley_pr9pa8oa_16h_raw_MIMIC_dot.pgf}
    }
  }%
  \caption{%
    Shapley value distribution plots for AUMC, eICU, HiRID, and MIMIC-III. The top $20$ raw measurements are being shown in the visualisation. While the precise ranking differs between databases, mean arterial pressure and heart rate are consistently part of the top $10$ features.
  }
  \label{fig:extended_bee}
\end{figure}
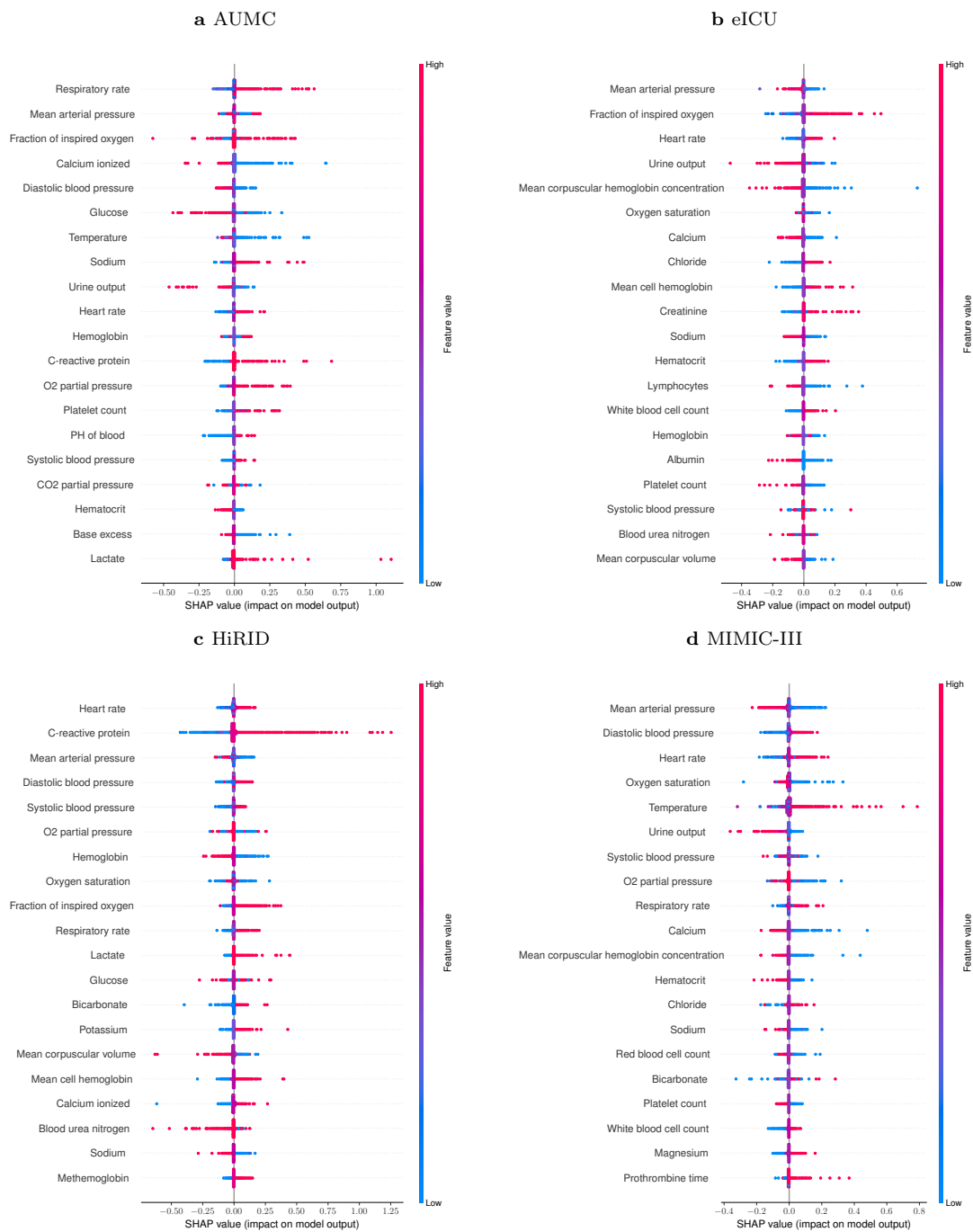 

\begin{figure}
    \centering
    \subcaptionbox{\label{fig:feature_ablation}}{
        \includegraphics[width=0.5\linewidth]{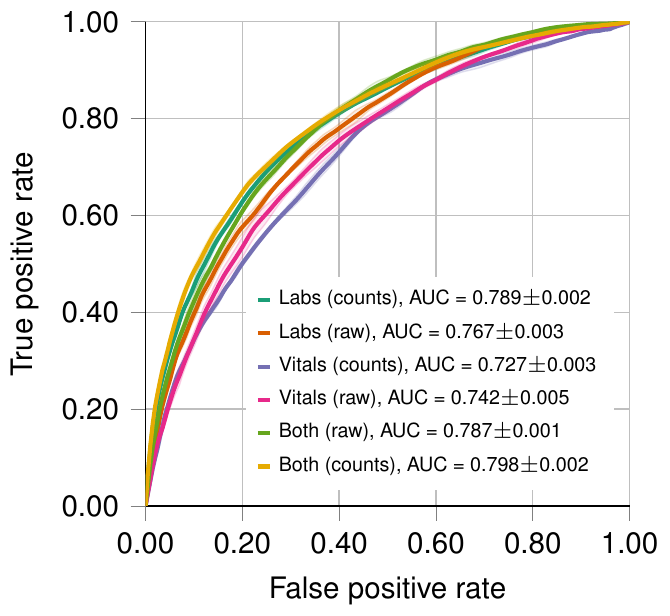}
    }%
    \subcaptionbox{\label{fig:cohort_ablation}}{
        \includegraphics[width=0.5\linewidth]{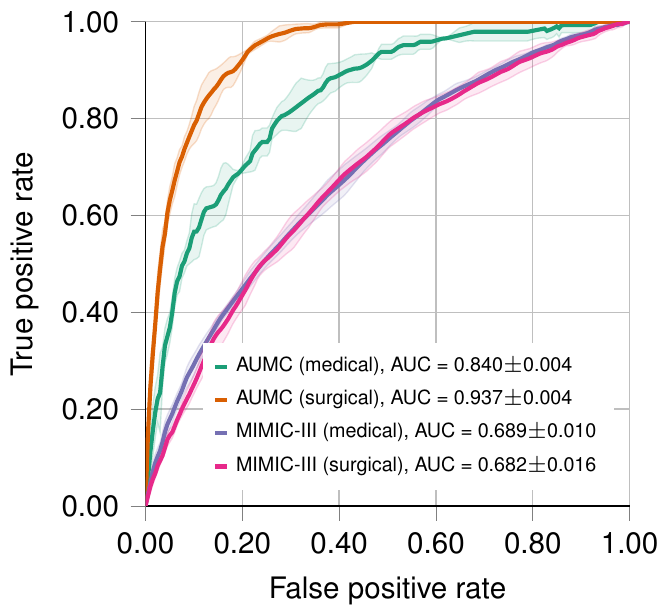}
    }
    \caption{
    Ablations of feature categories and cohorts. \subref{fig:feature_ablation}: Having observed that count features appeared frequently among the top $20$ features explaining high prediction scores in \autoref{fig:Shapley full}, we investigated how a model performed when trained solely on counts or raw measurements (here internally validated on MIMIC-III). We further subdivided the features into lab tests and vital signs which confirmed our hypothesis that lab tests carry sampling information (as a clinician requested them), whereas this is less the case for vital signs. 
    \subref{fig:cohort_ablation}: We observed the best internal validation performance on AUMC, a dataset with a large proportion of surgical patients. Thus, we applied a model that was trained on AUMC to the medical and surgical cohorts both in-distribution (AUMC) and out-of-distribution (MIMIC-III) and observed that the surgical cohort within AUMC is indeed easier to classify, whereas this does not necessarily generalise to other datasets.
    }
    \label{fig:ablation}
\end{figure}

\begin{figure}
    \centering
    \includegraphics[width=0.5\linewidth]{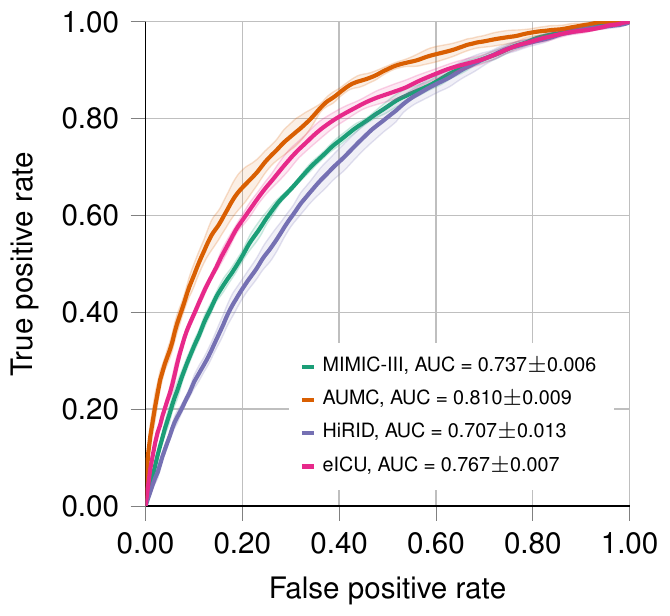}
    \caption{Pooling datasets during training. An auxiliary analysis of our attention model by pooling datasets during \emph{training}, as opposed to pooling predictions of models trained on different datasets. For each displayed dataset the other datasets were pooled.}
    \label{fig:pooled_datasets}
\end{figure}

\clearpage
\newpage
\appendix

\setcounter{figure}{0}
\setcounter{table}{0}

\makeatletter
\def\fnum@table{Supplementary Table~\thetable}
\def\fnum@figure{Supplementary Figure~\thefigure}
\makeatother

\section{Supplementary Materials}

\begin{figure}[bp]
    \centering
    \subcaptionbox{eICU}{
        \includegraphics[width=0.50\linewidth]{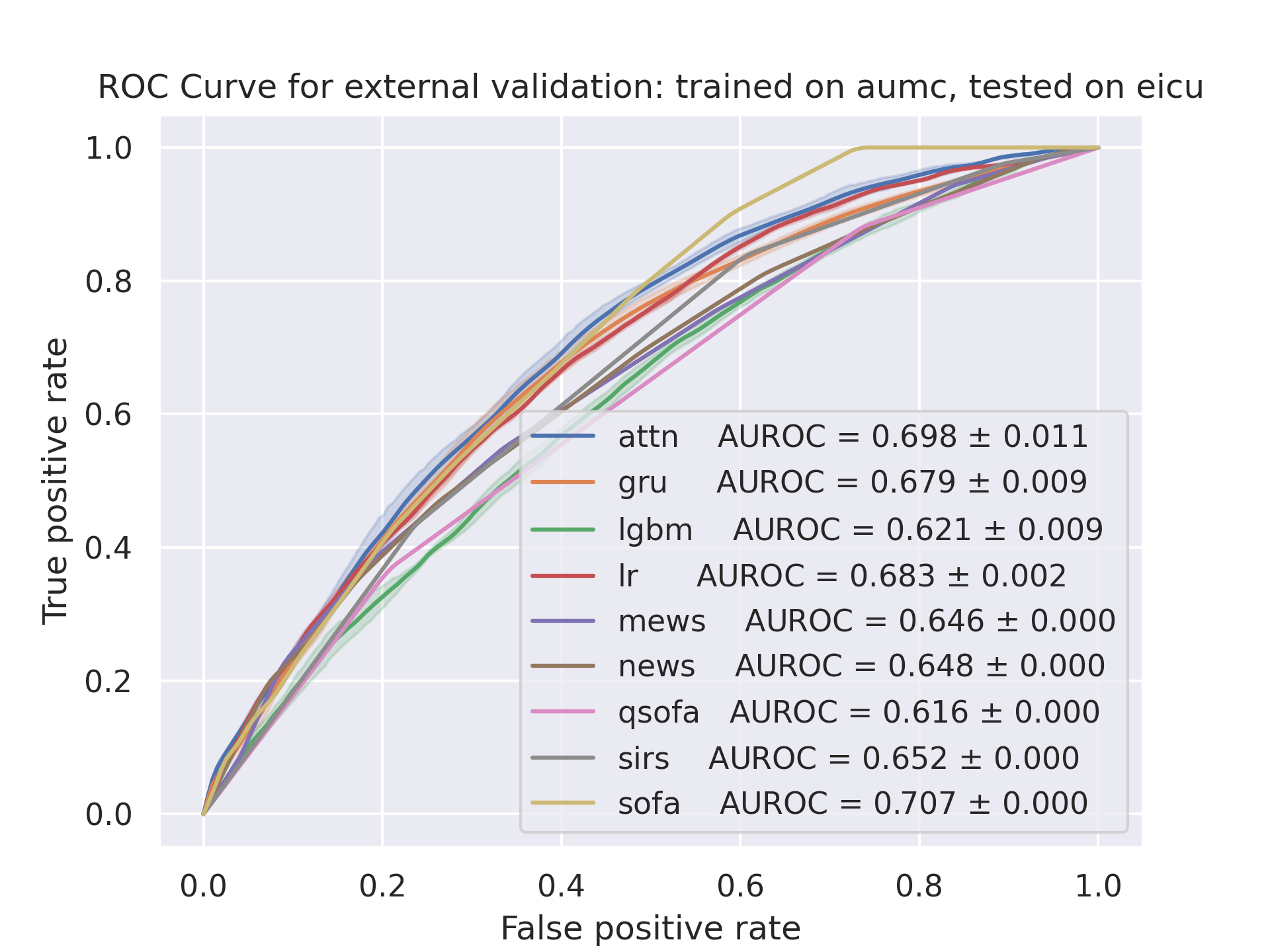}
    }%
    \subcaptionbox{HiRID}{
        \includegraphics[width=0.50\linewidth]{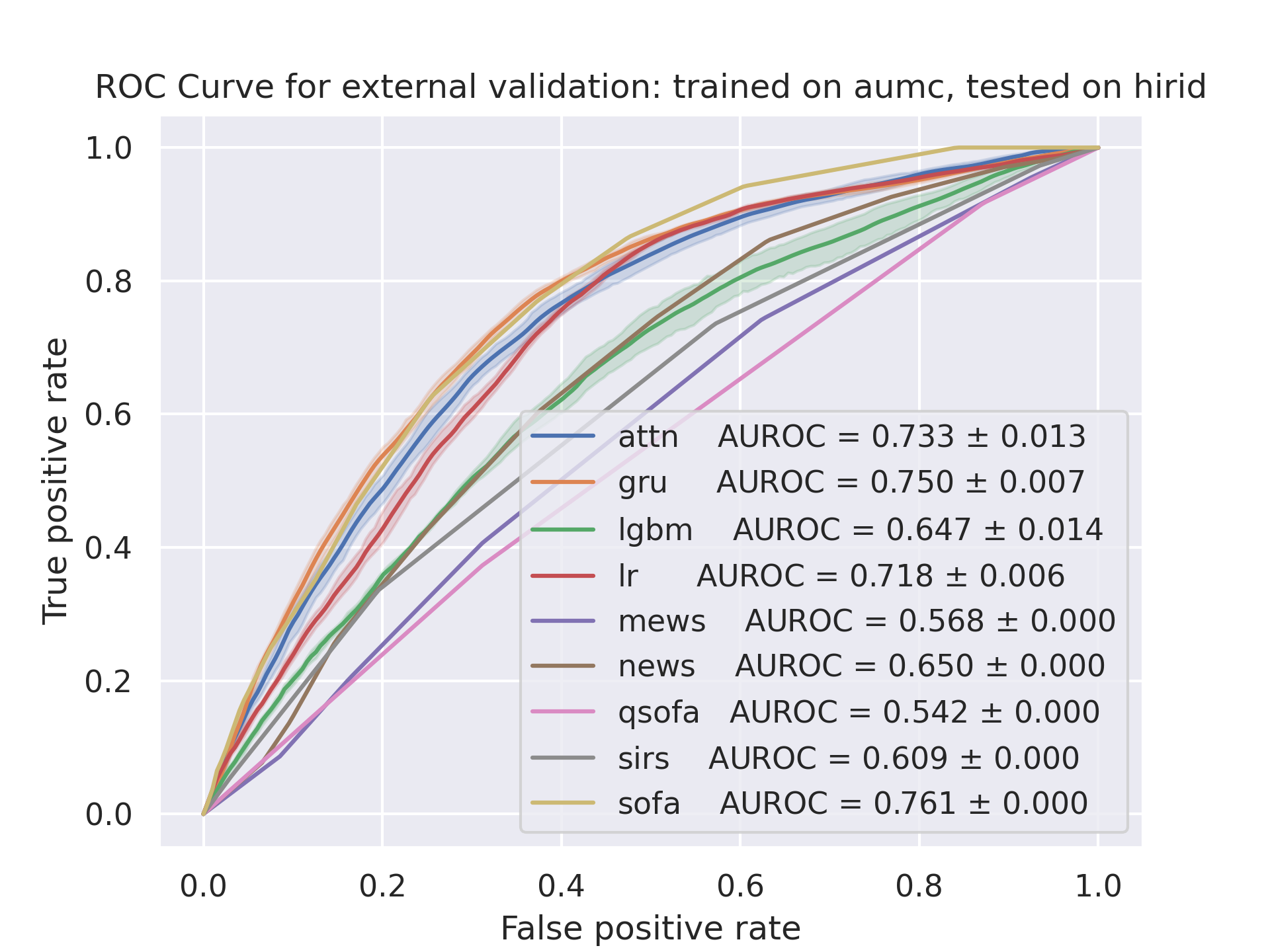}
    }\\%
    \subcaptionbox{MIMIC-III}{
        \includegraphics[width=0.50\linewidth]{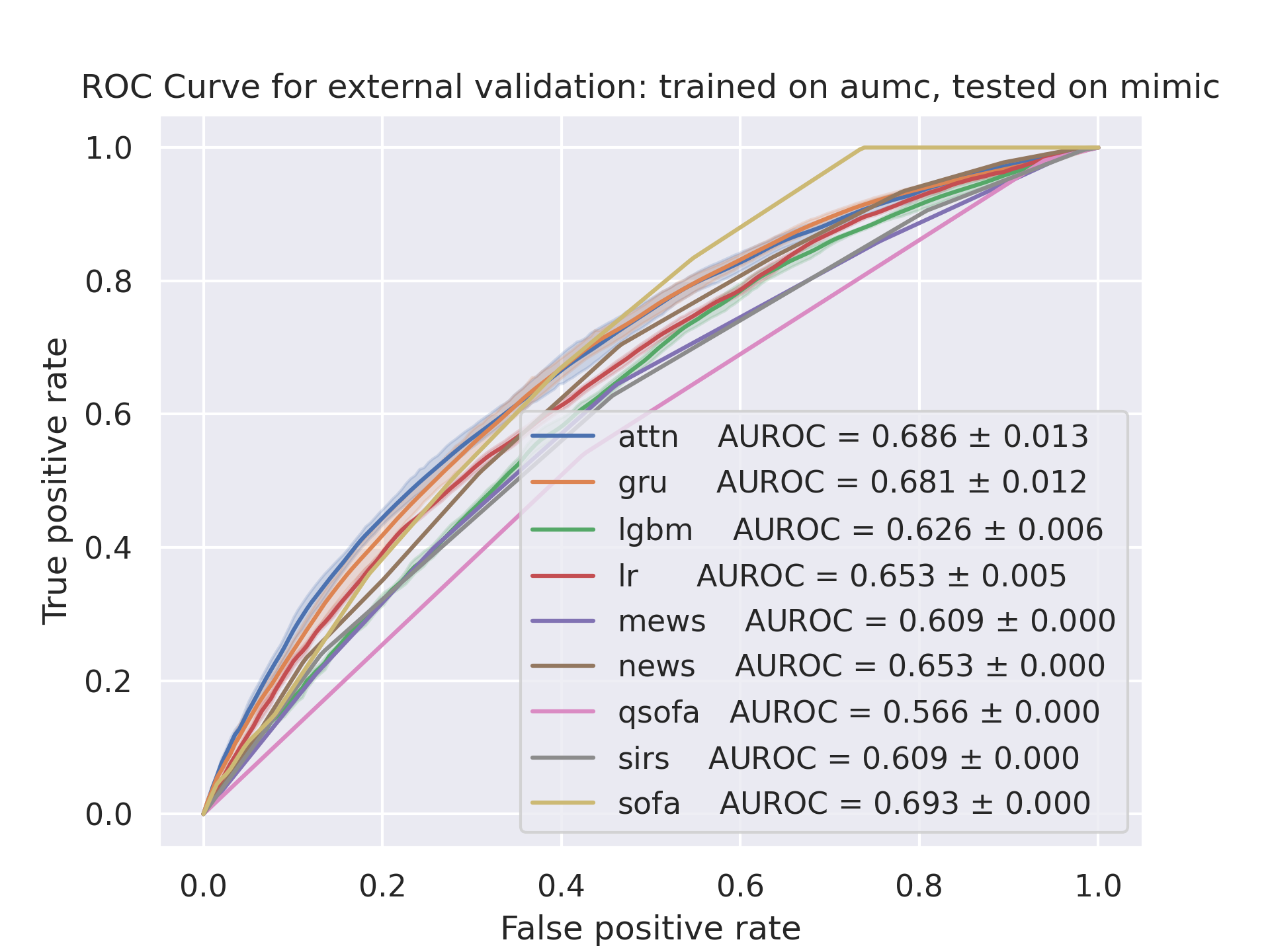}
    }%
    \subcaptionbox{Emory}{
        \includegraphics[width=0.50\linewidth]{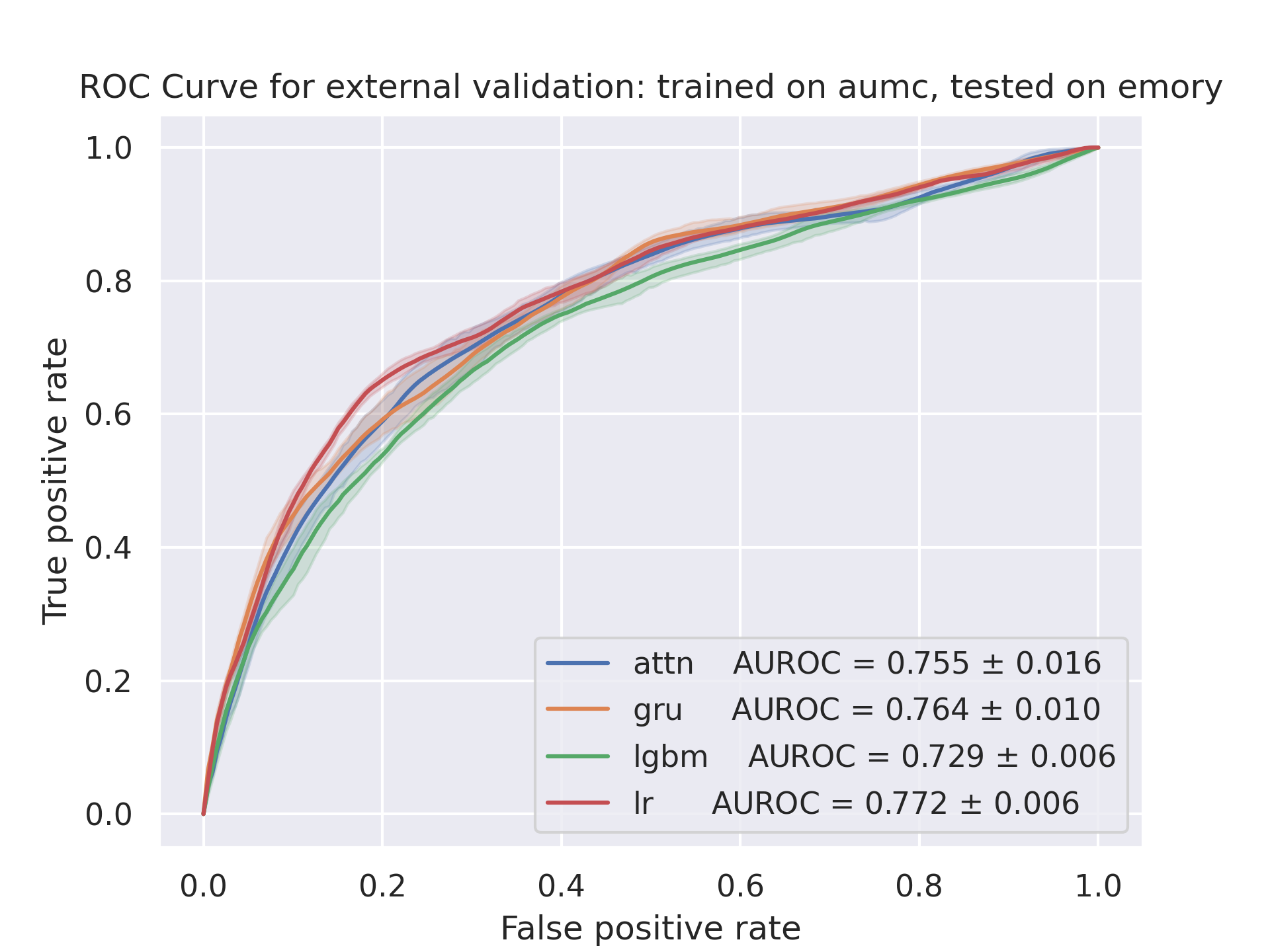}
    }
    \caption{
    ROC curves for the external validation scenarios. Each subplot depicts models trained on the AUMC database, and evaluated on one of the remaining databases. Each subfigure label indicates the respective evaluation database.
    }
    \label{fig:ex_aumc}
\end{figure}

\begin{figure}
    \centering
    \subcaptionbox{AUMC}{
        \includegraphics[width=0.50\linewidth]{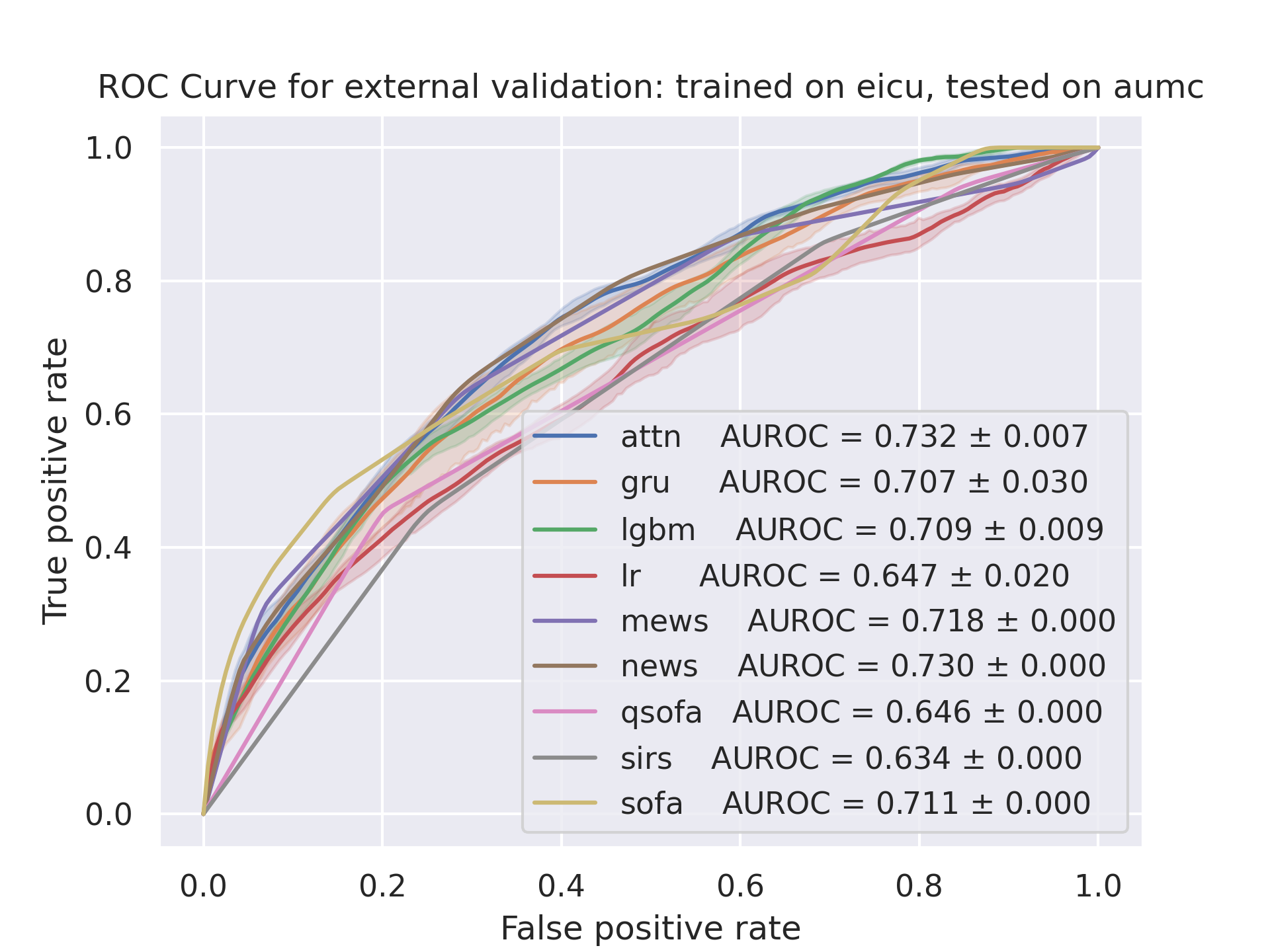}
    }%
    \subcaptionbox{HiRID}{
        \includegraphics[width=0.50\linewidth]{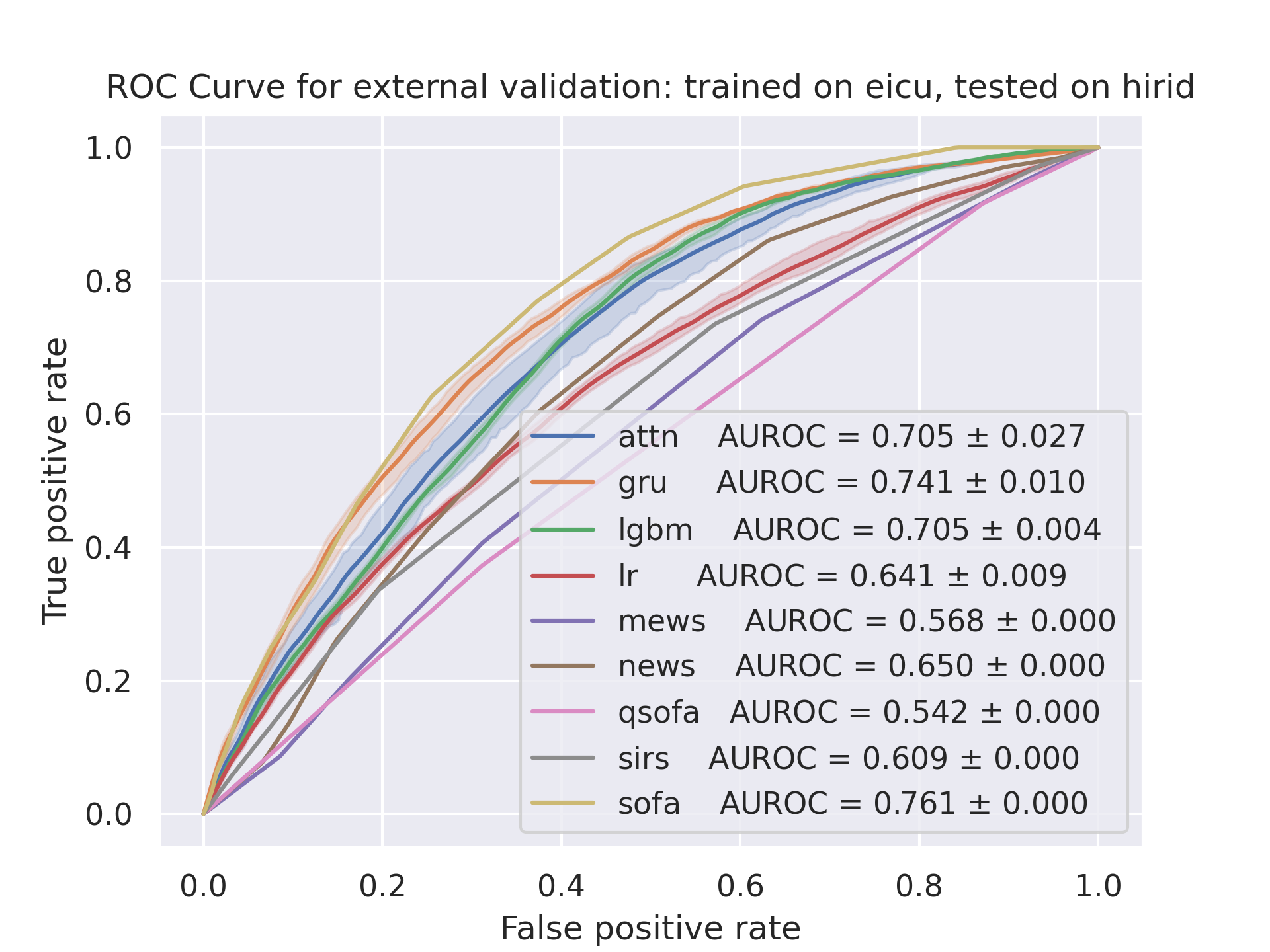}
    }\\%
    \subcaptionbox{MIMIC-III}{
        \includegraphics[width=0.50\linewidth]{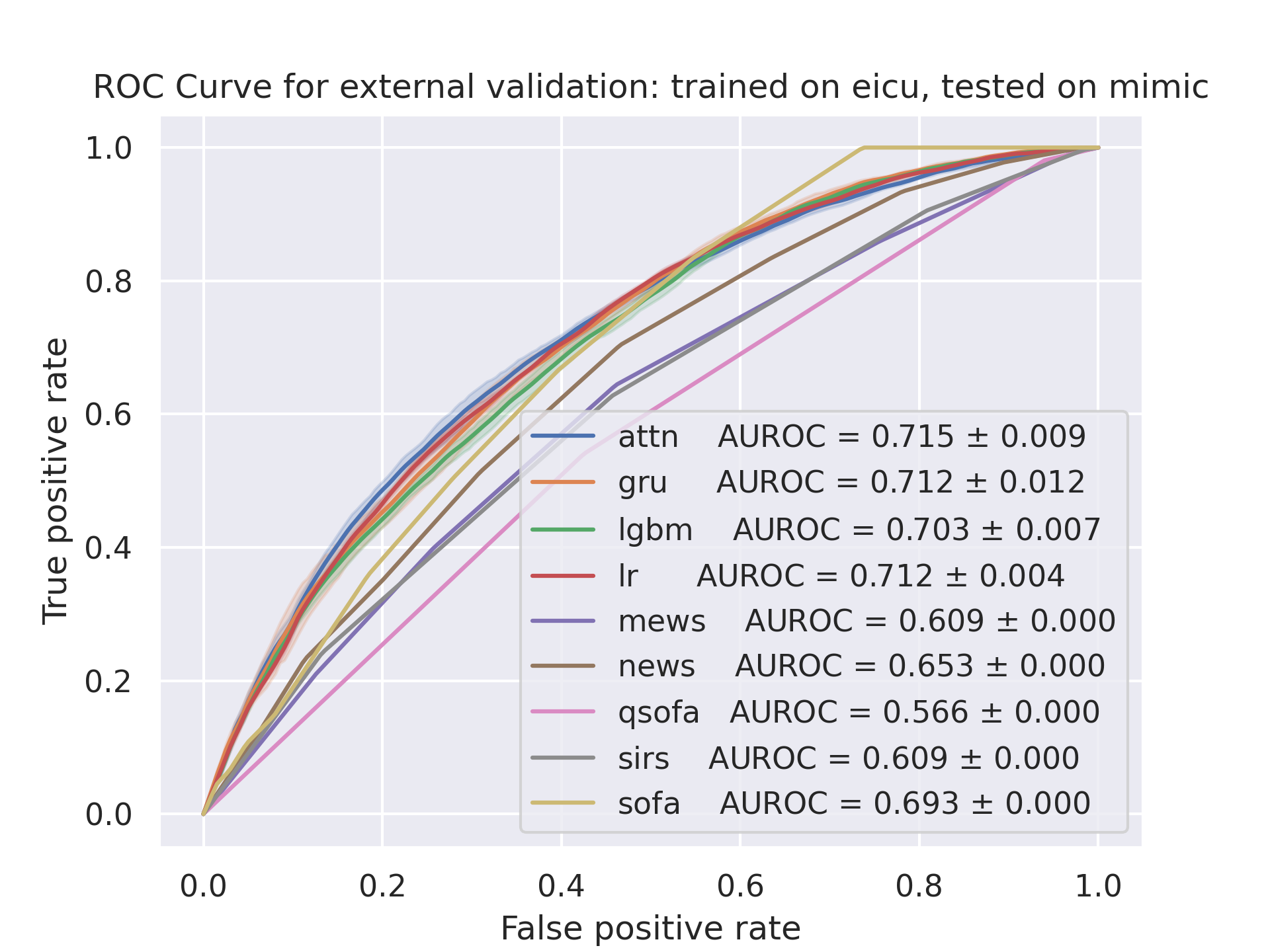}
    }%
    \subcaptionbox{Emory}{
        \includegraphics[width=0.50\linewidth]{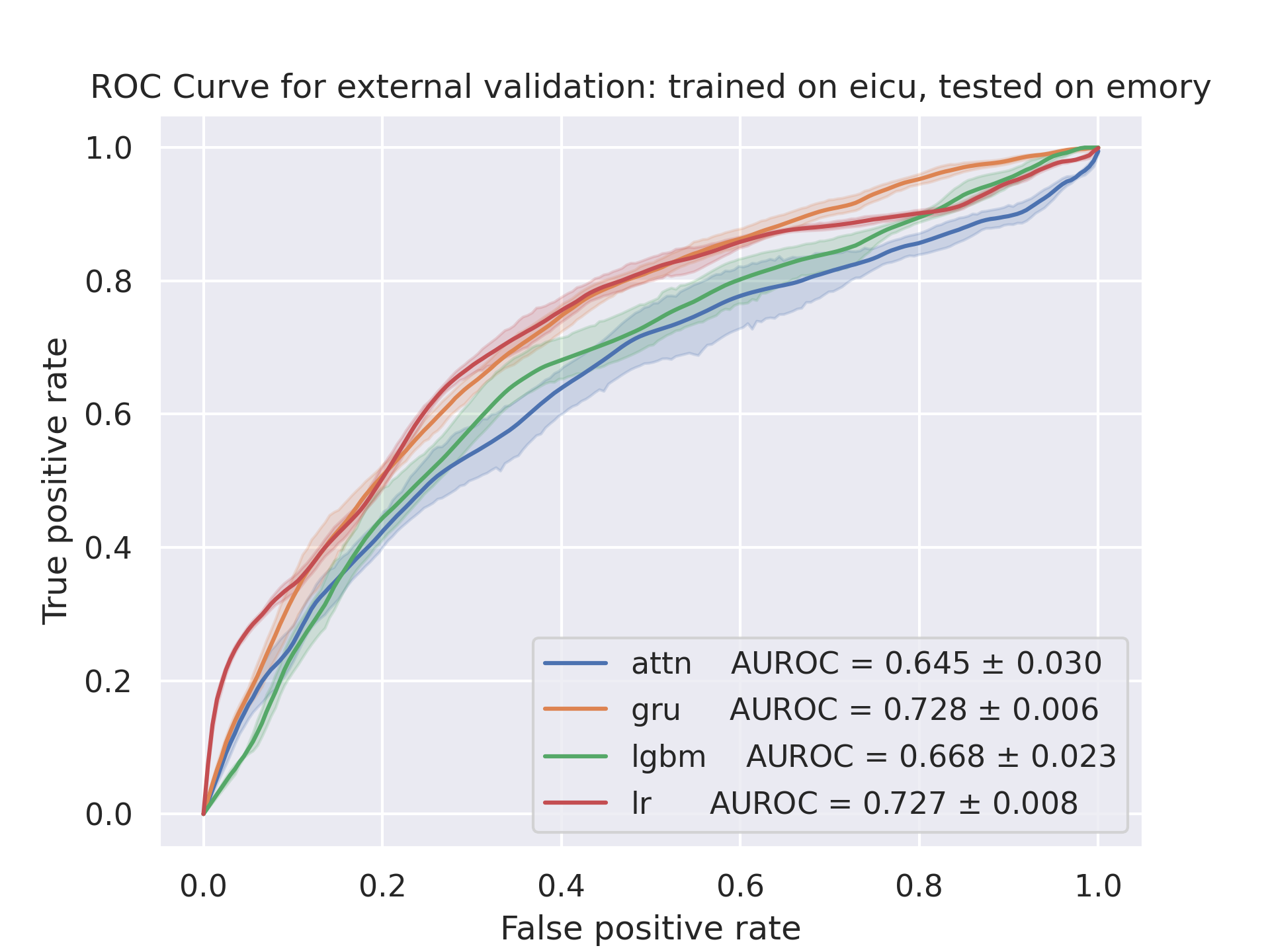}
    }
    \caption{
    ROC curves for the external validation scenarios. Each subplot depicts models trained on the eICU database, and evaluated on one of the remaining databases. Each subfigure label indicates the respective evaluation database. 
    }
\end{figure}

\begin{figure}
    \centering
    \subcaptionbox{AUMC}{
        \includegraphics[width=0.50\linewidth]{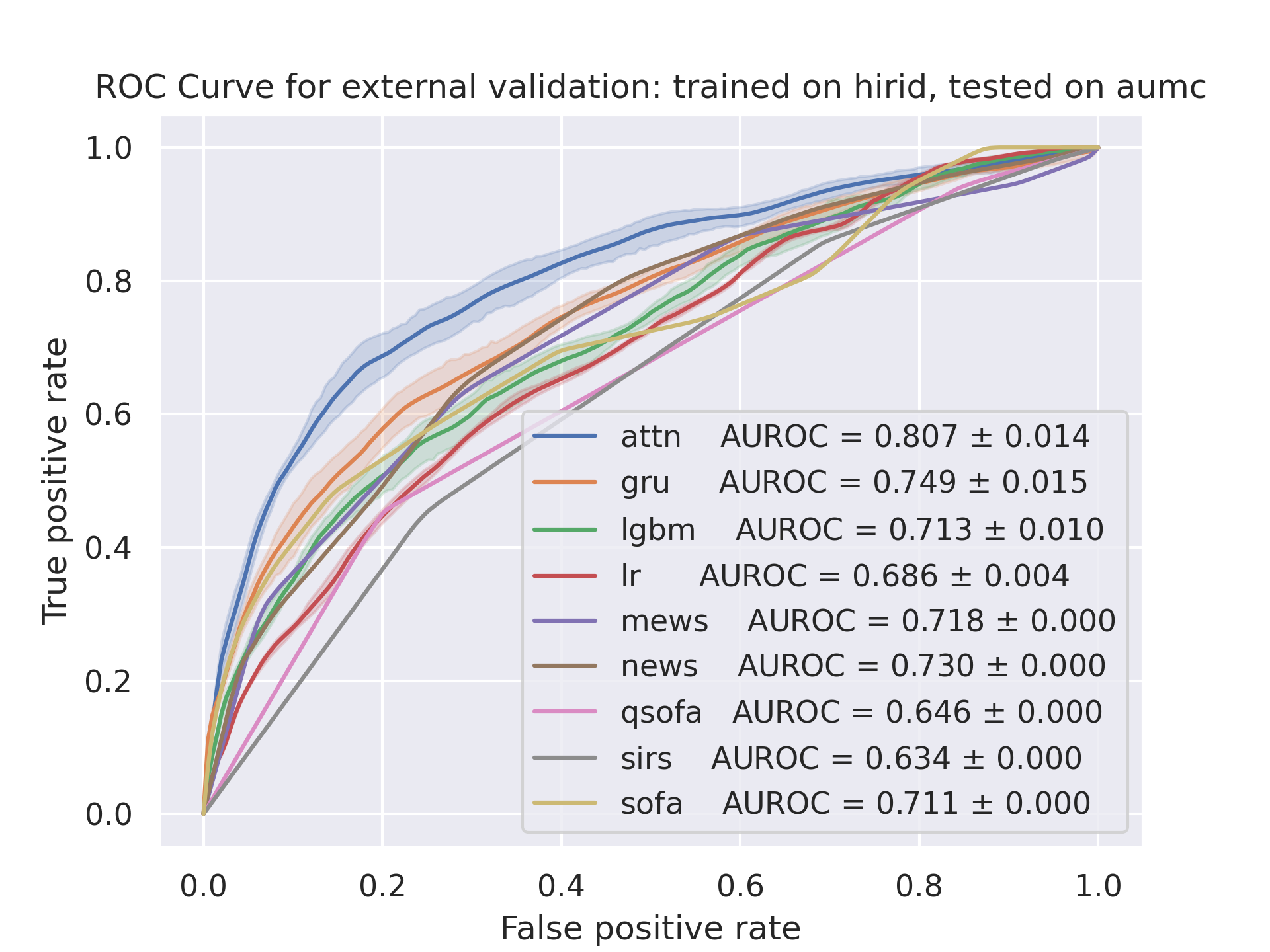}
    }%
    \subcaptionbox{eICU}{
        \includegraphics[width=0.50\linewidth]{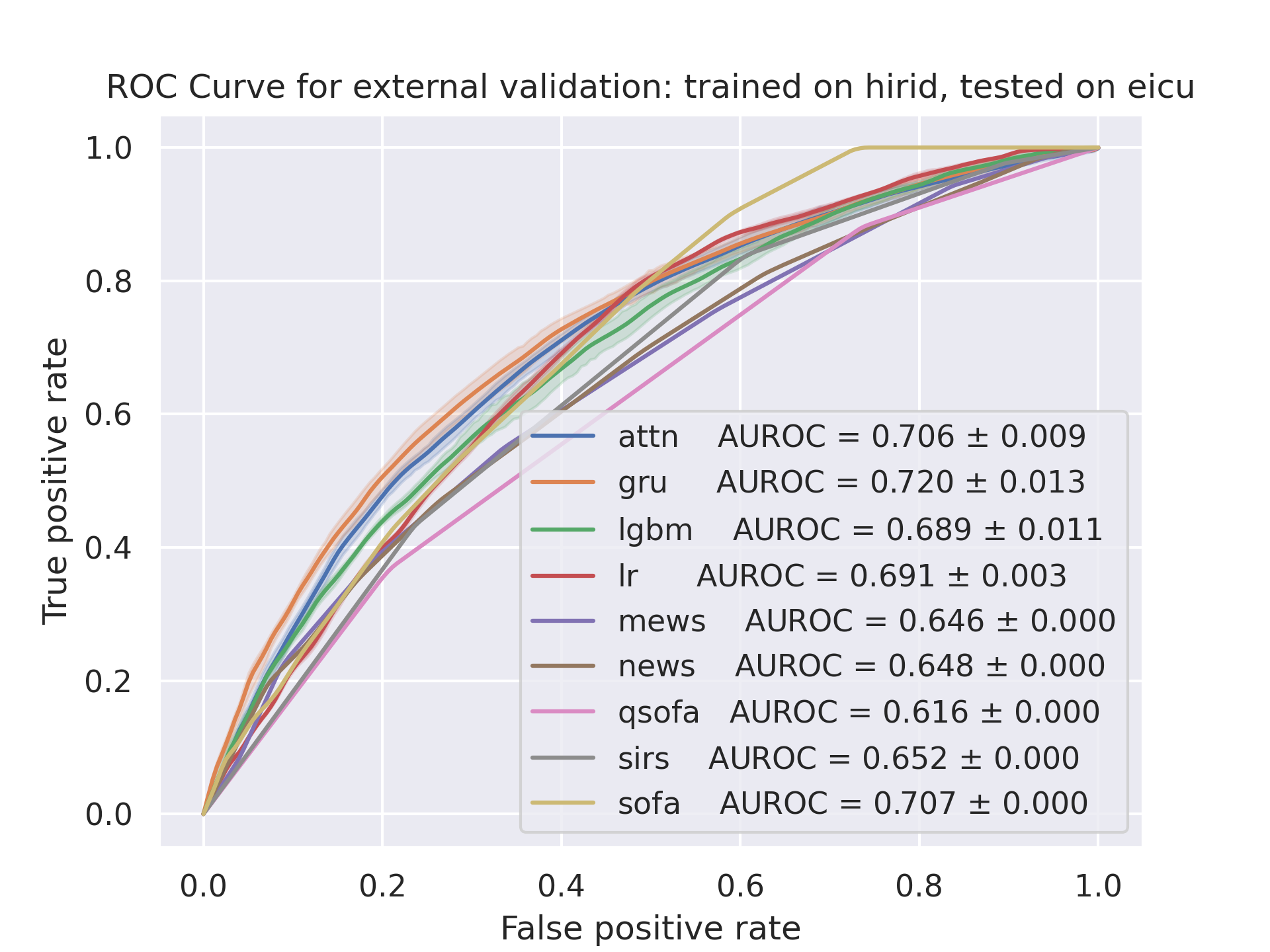}
    }\\%
    \subcaptionbox{MIMIC-III}{
        \includegraphics[width=0.50\linewidth]{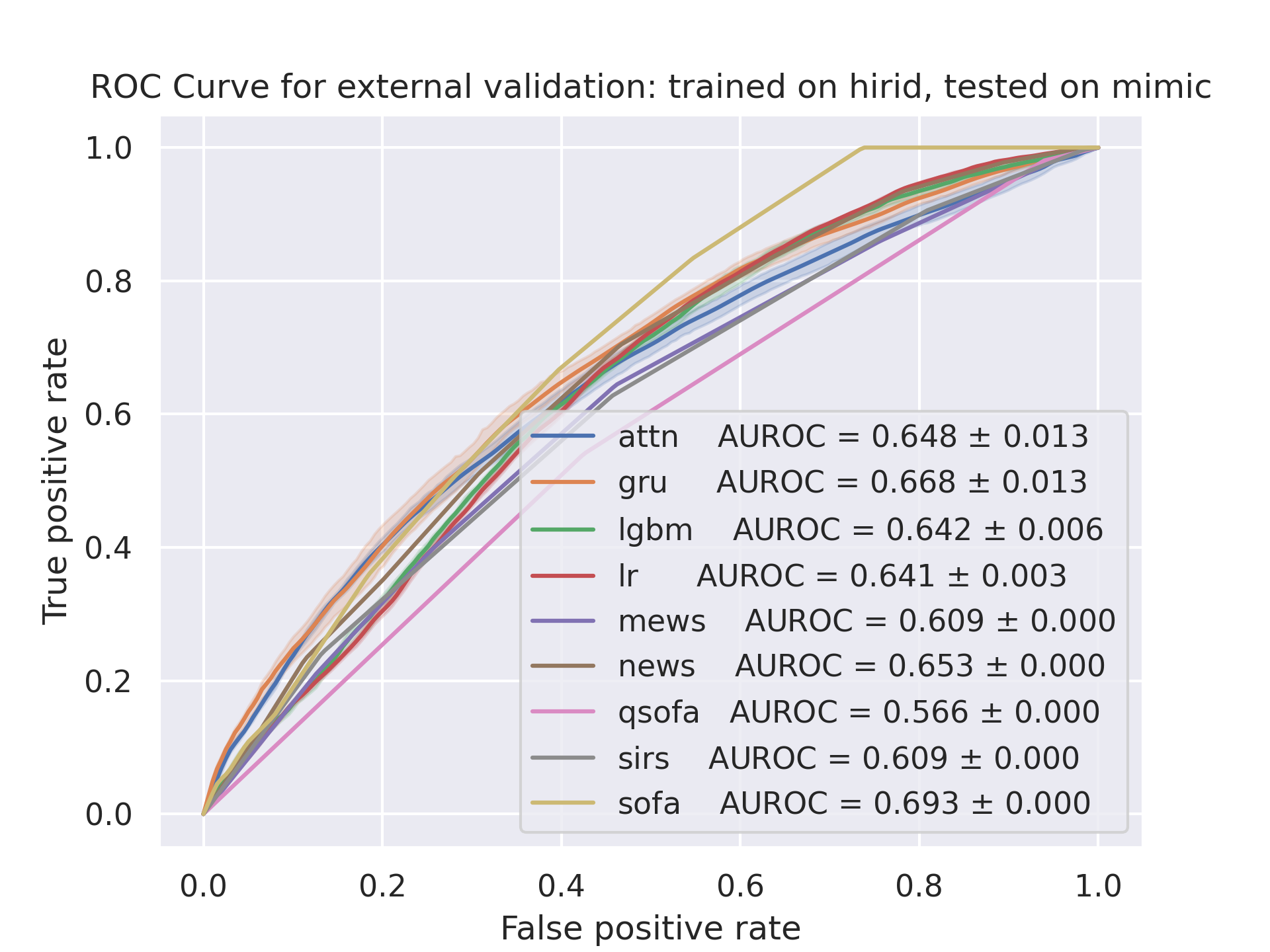}
    }%
    \subcaptionbox{Emory}{
        \includegraphics[width=0.50\linewidth]{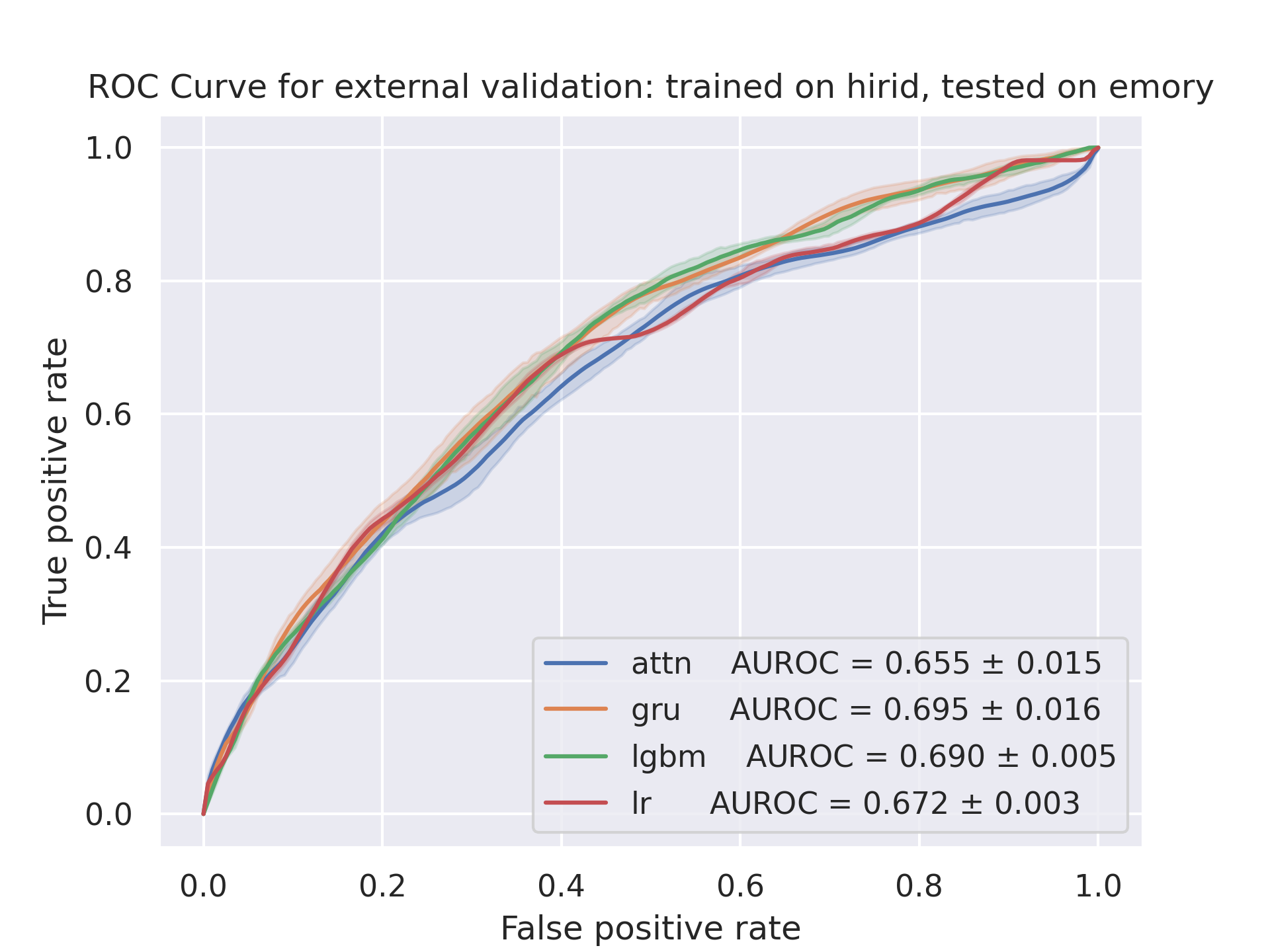}
    }
    \caption{%
      ROC curves for the external validation scenarios. Each subplot depicts models trained on the HiRID database, and evaluated on one of the remaining databases. Each subfigure label indicates the respective evaluation database.
    }
\end{figure}

\begin{figure}
    \centering
    \subcaptionbox{AUMC}{
        \includegraphics[width=0.50\linewidth]{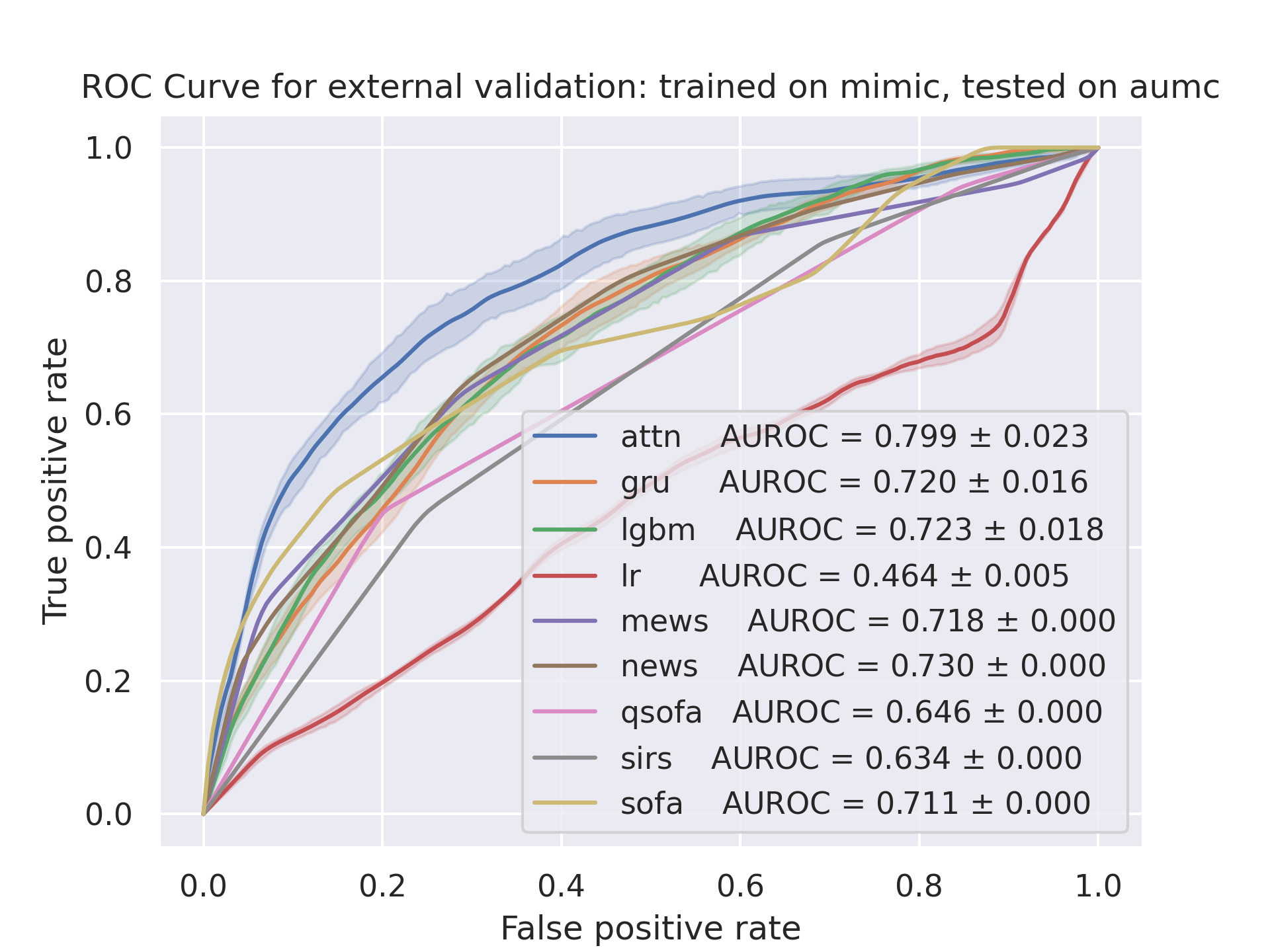}
    }%
    \subcaptionbox{eICU}{
        \includegraphics[width=0.50\linewidth]{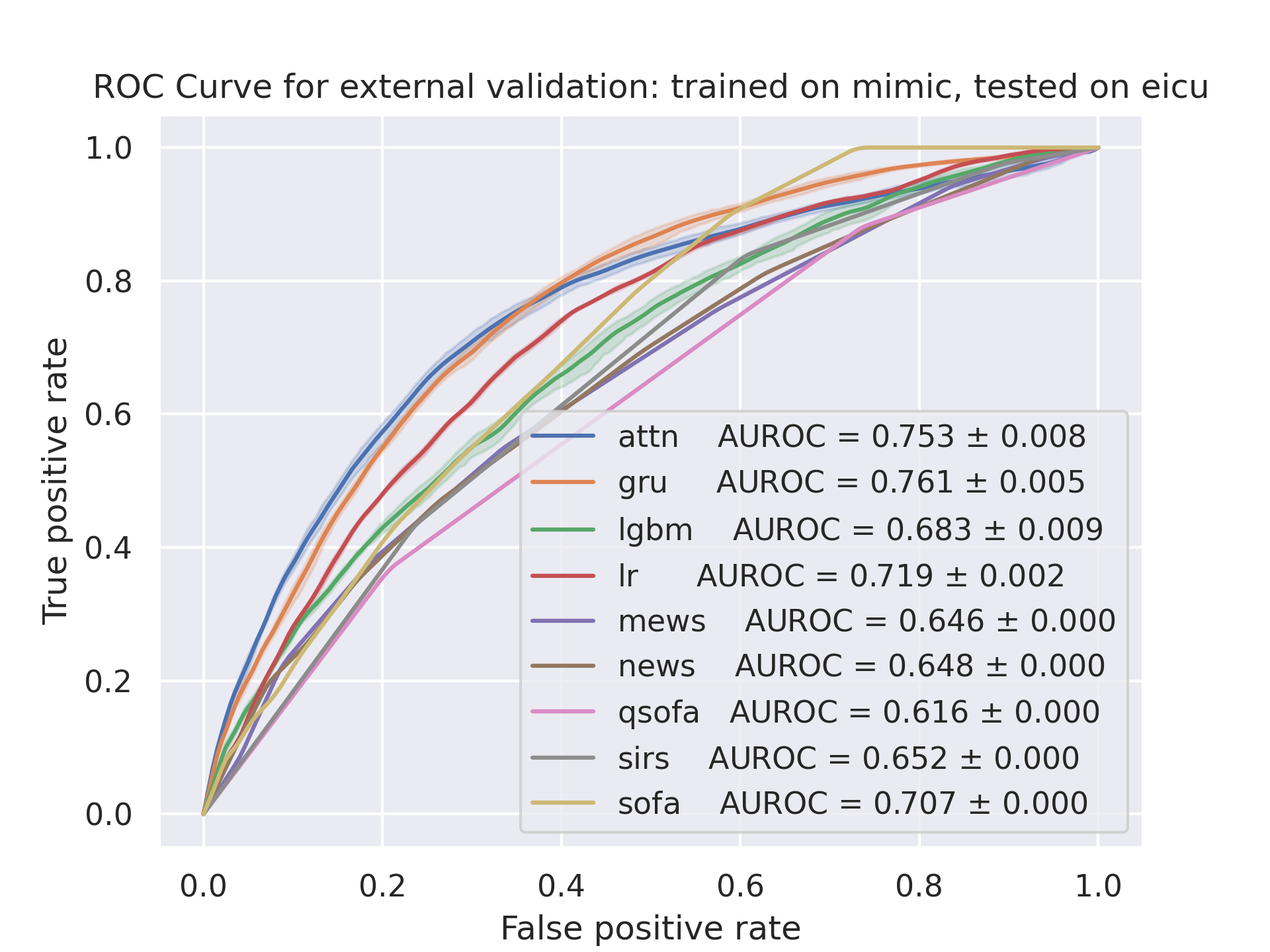}
    }\\%
    \subcaptionbox{HiRID}{
        \includegraphics[width=0.50\linewidth]{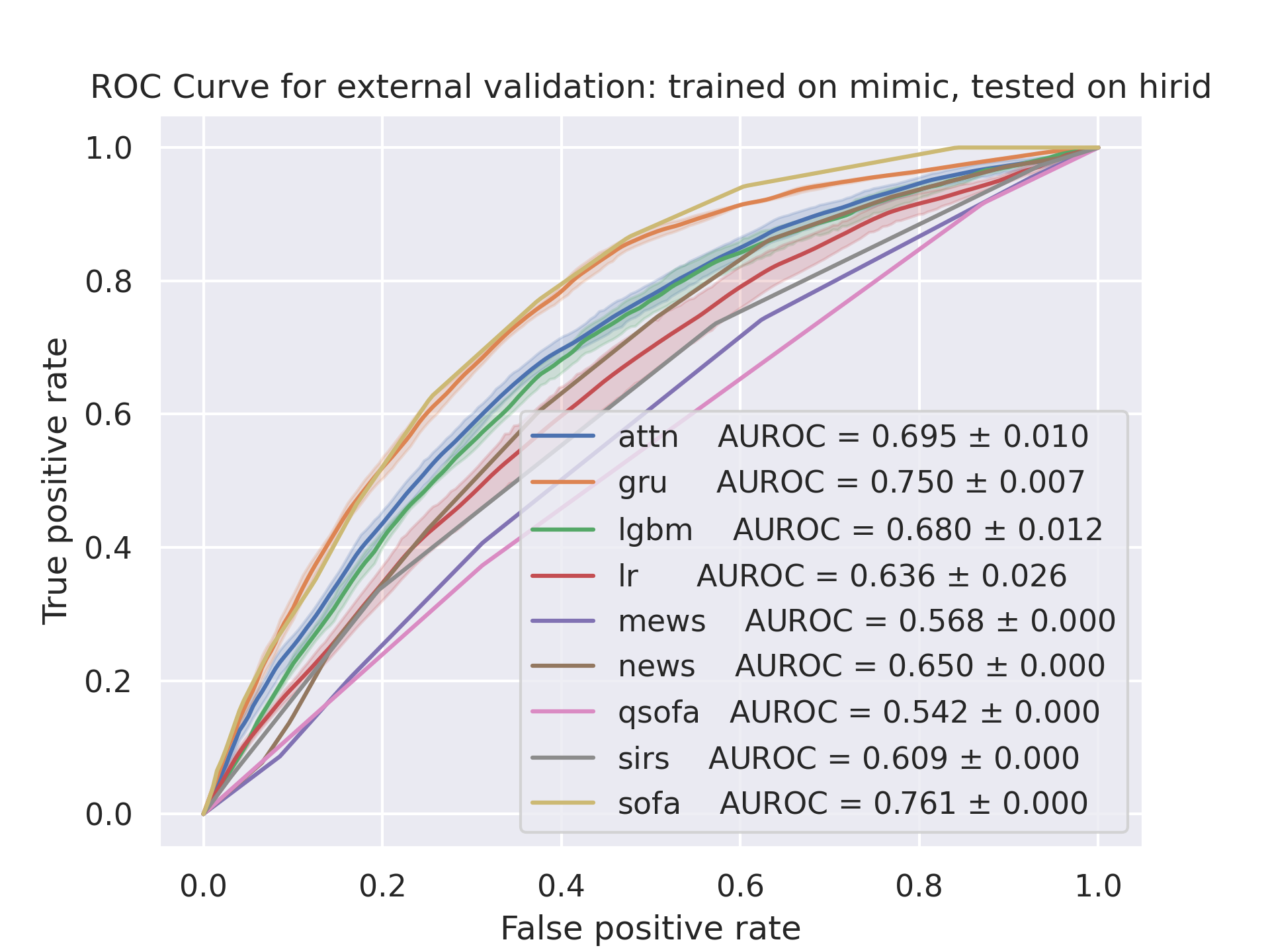}
    }%
    \subcaptionbox{Emory}{
        \includegraphics[width=0.50\linewidth]{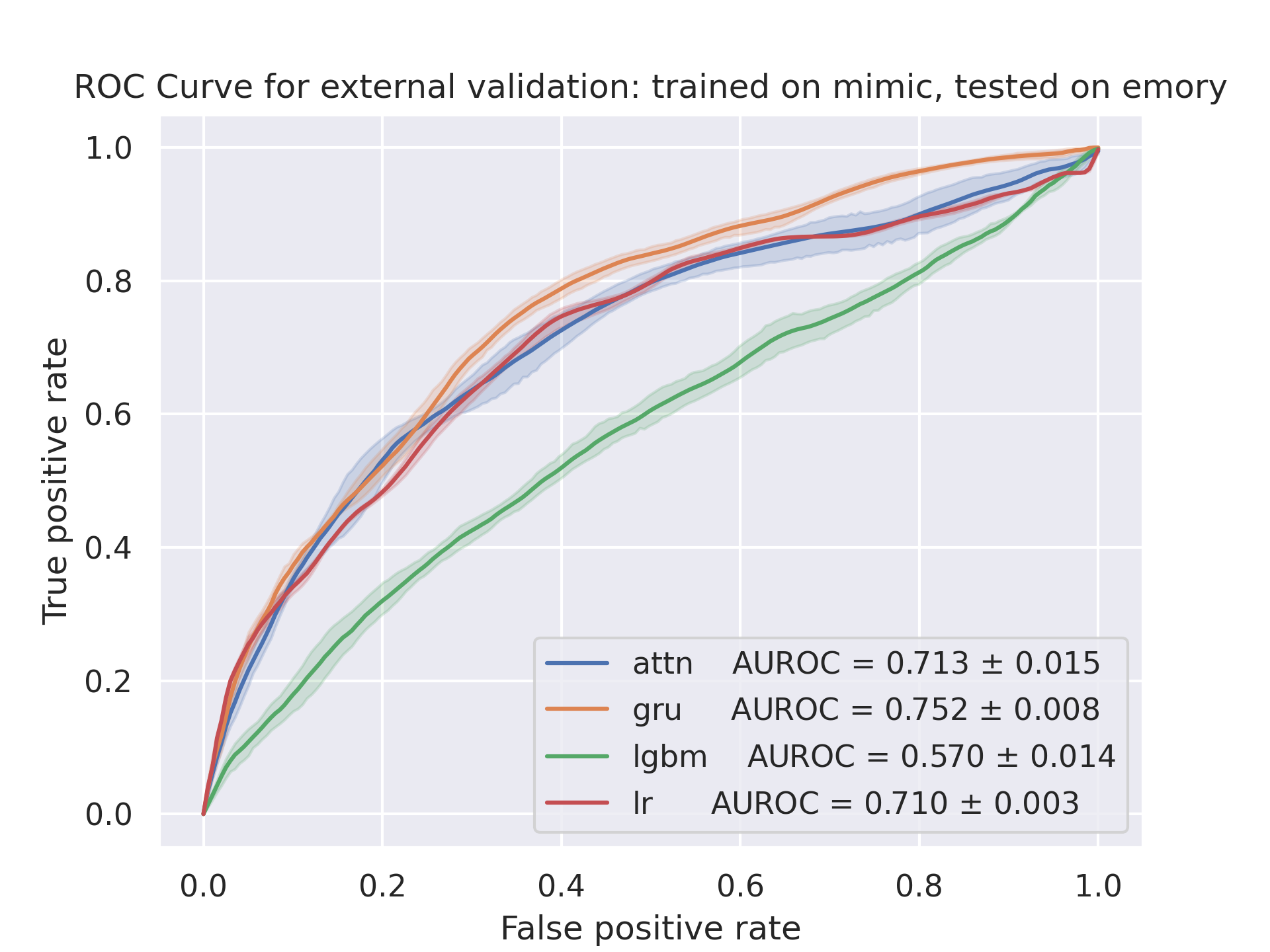}
    }
    \caption{
     ROC curves for the external validation scenarios. Each subplot depicts models trained on the \mbox{MIMIC-III} database, and evaluated on one of the remaining databases. Each subfigure label indicates the respective evaluation database. 
    }
\end{figure}

\begin{figure}
    \centering
    \subcaptionbox{AUMC}{
        \includegraphics[width=0.50\linewidth]{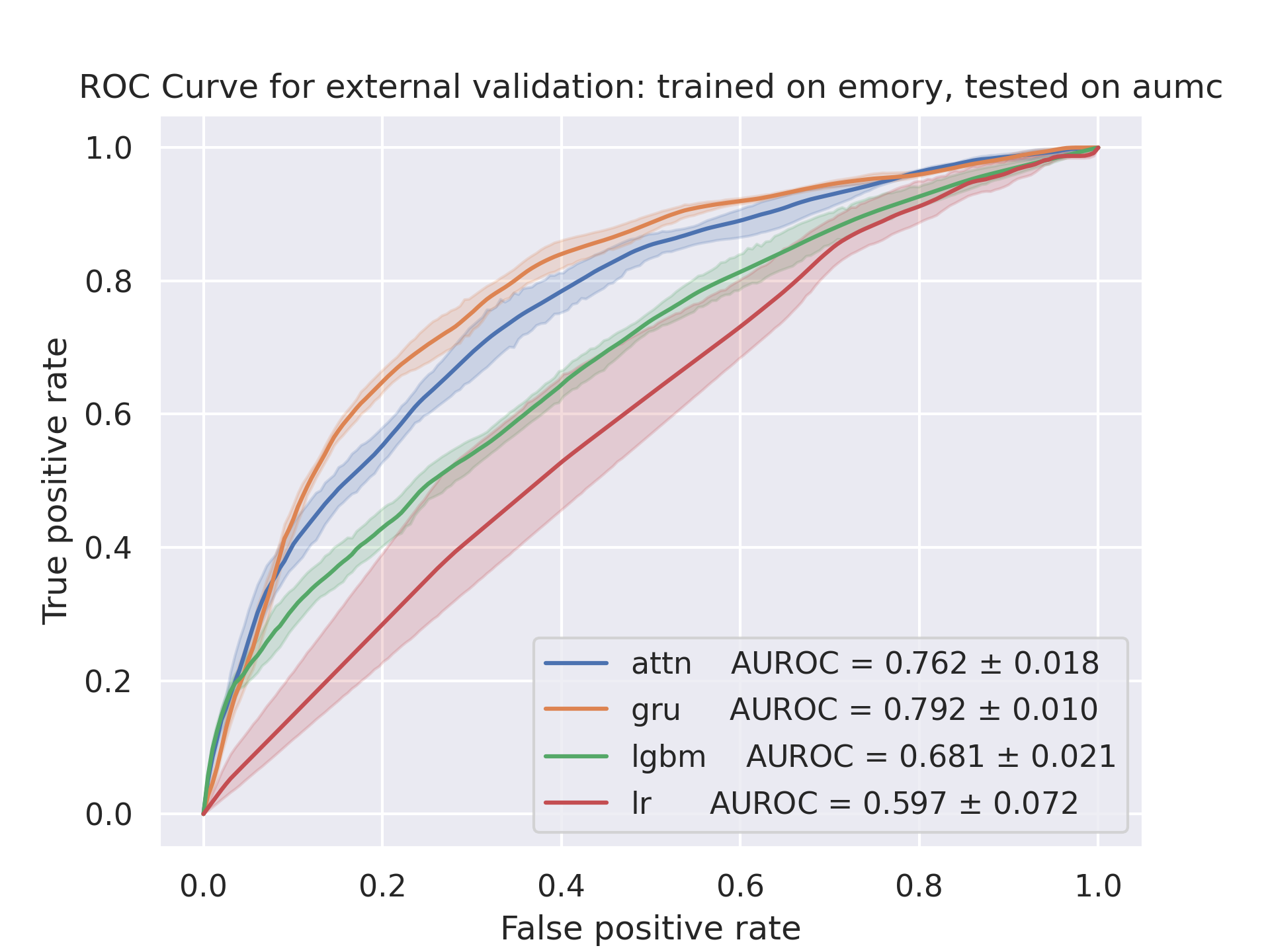}
    }%
    \subcaptionbox{eICU}{
        \includegraphics[width=0.50\linewidth]{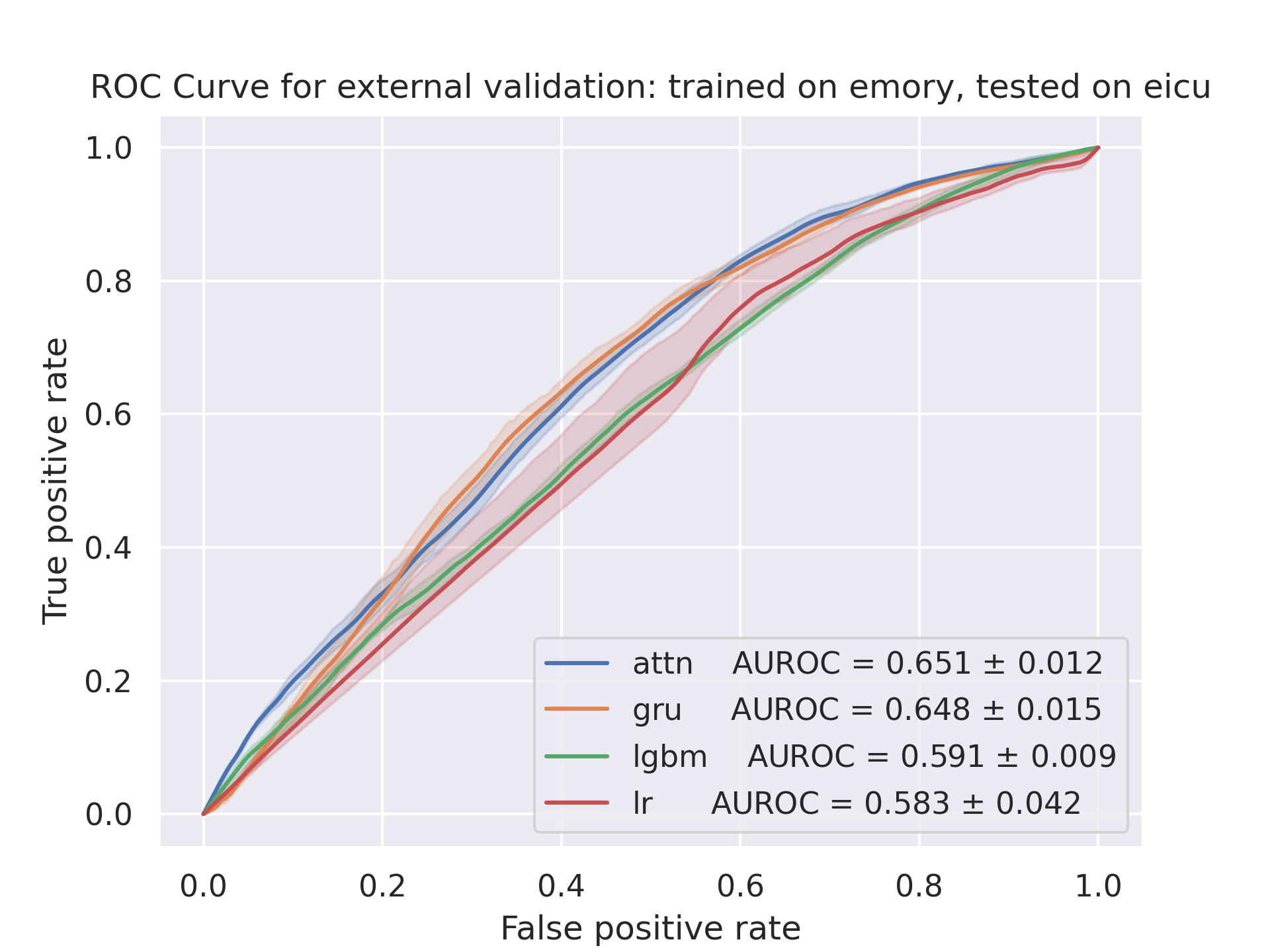}
    }\\%
    \subcaptionbox{HiRID}{
        \includegraphics[width=0.50\linewidth]{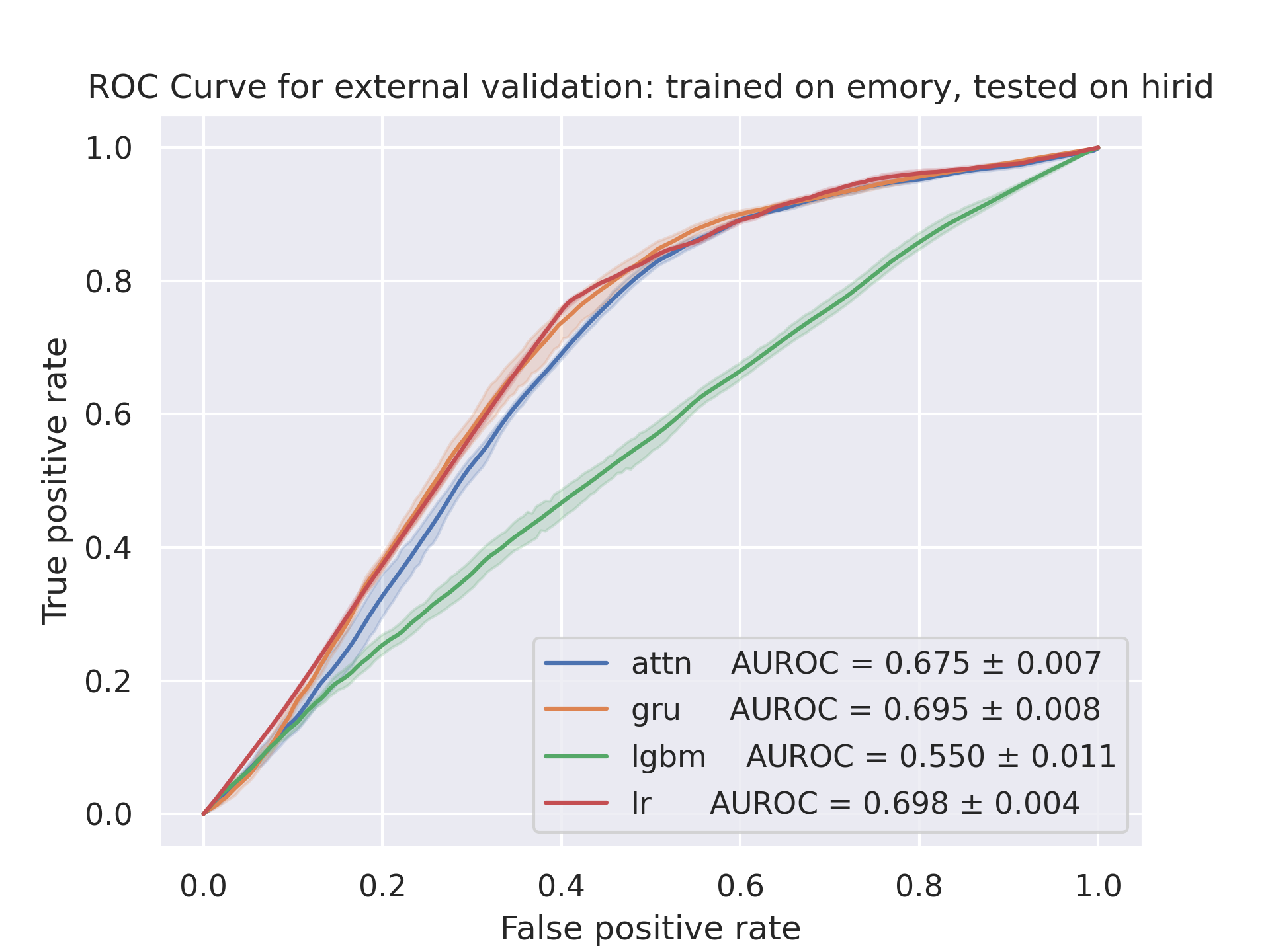}
    }%
    \subcaptionbox{MIMIC-III}{
        \includegraphics[width=0.50\linewidth]{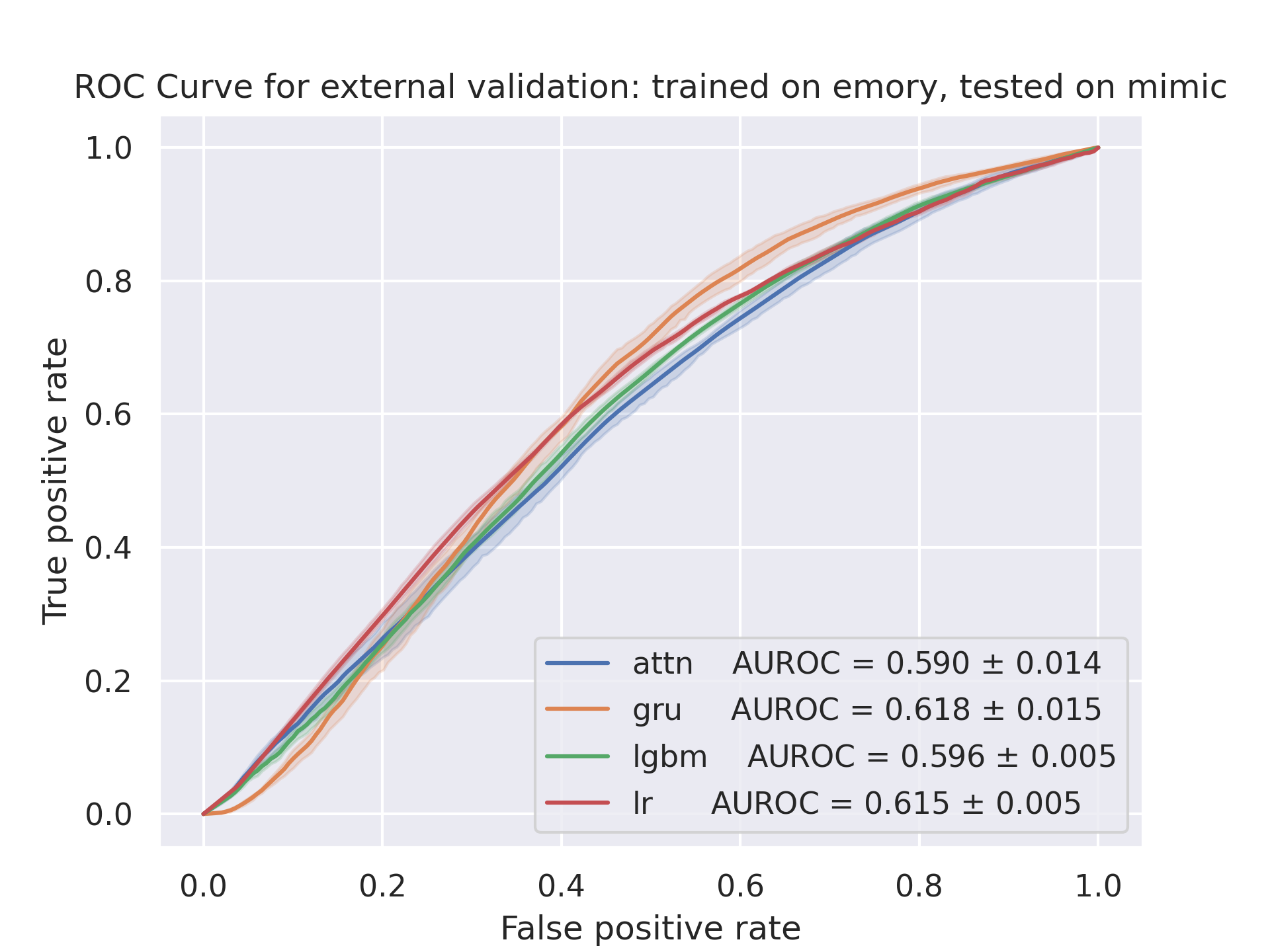}
    }
    \caption{
      ROC curves for the external validation scenarios. Each subplot depicts models trained on the Emory database, and evaluated on one of the remaining databases. Each subfigure label indicates the respective evaluation database.
    }
    \label{fig:ex_emory}
\end{figure}

\begin{figure}
    \centering
    \subcaptionbox{}{
      \includegraphics[width=0.45\linewidth]{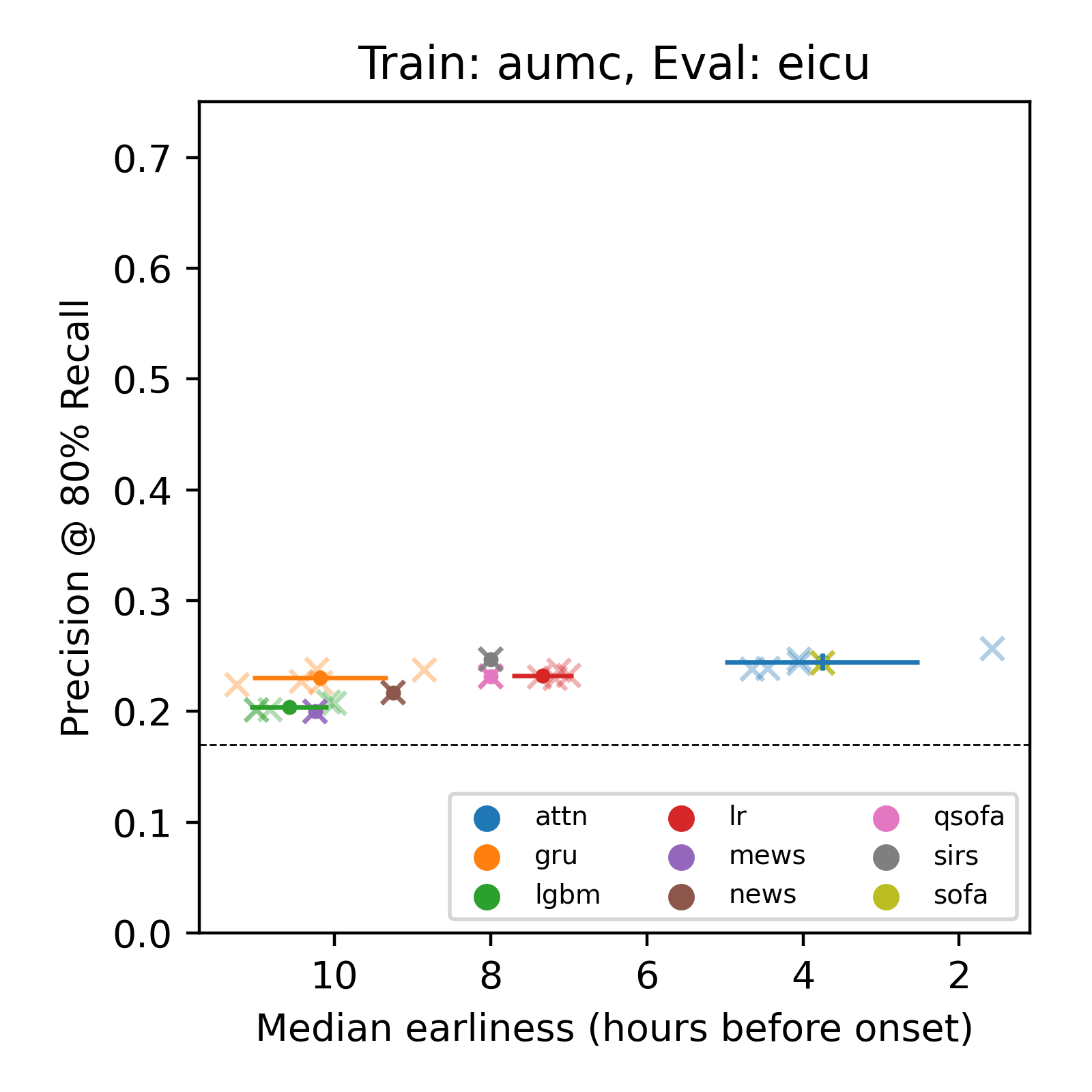}
    }
    \subcaptionbox{}{
        \includegraphics[width=0.45\linewidth]{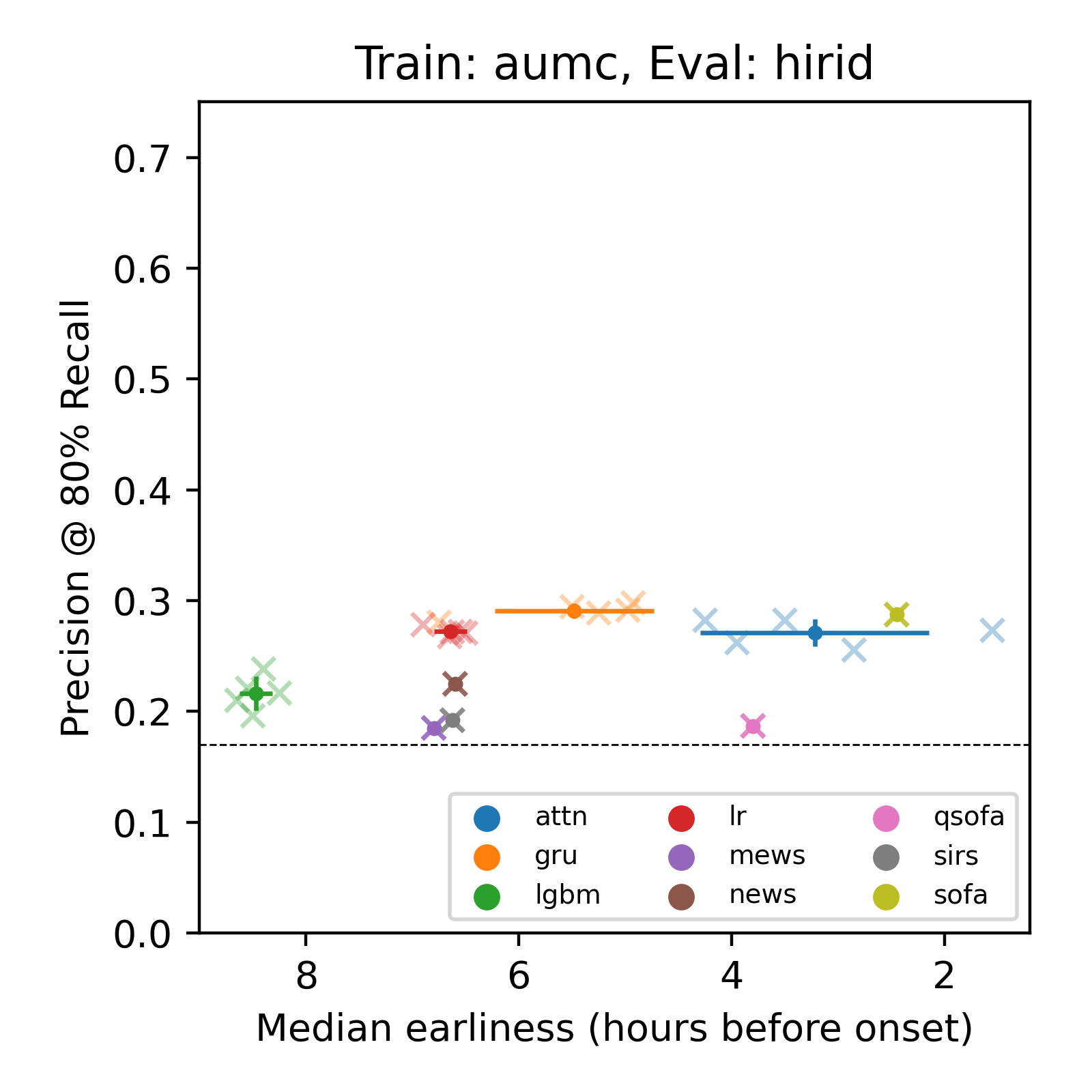}
    }\\
    \subcaptionbox{}{
        \includegraphics[width=0.45\linewidth]{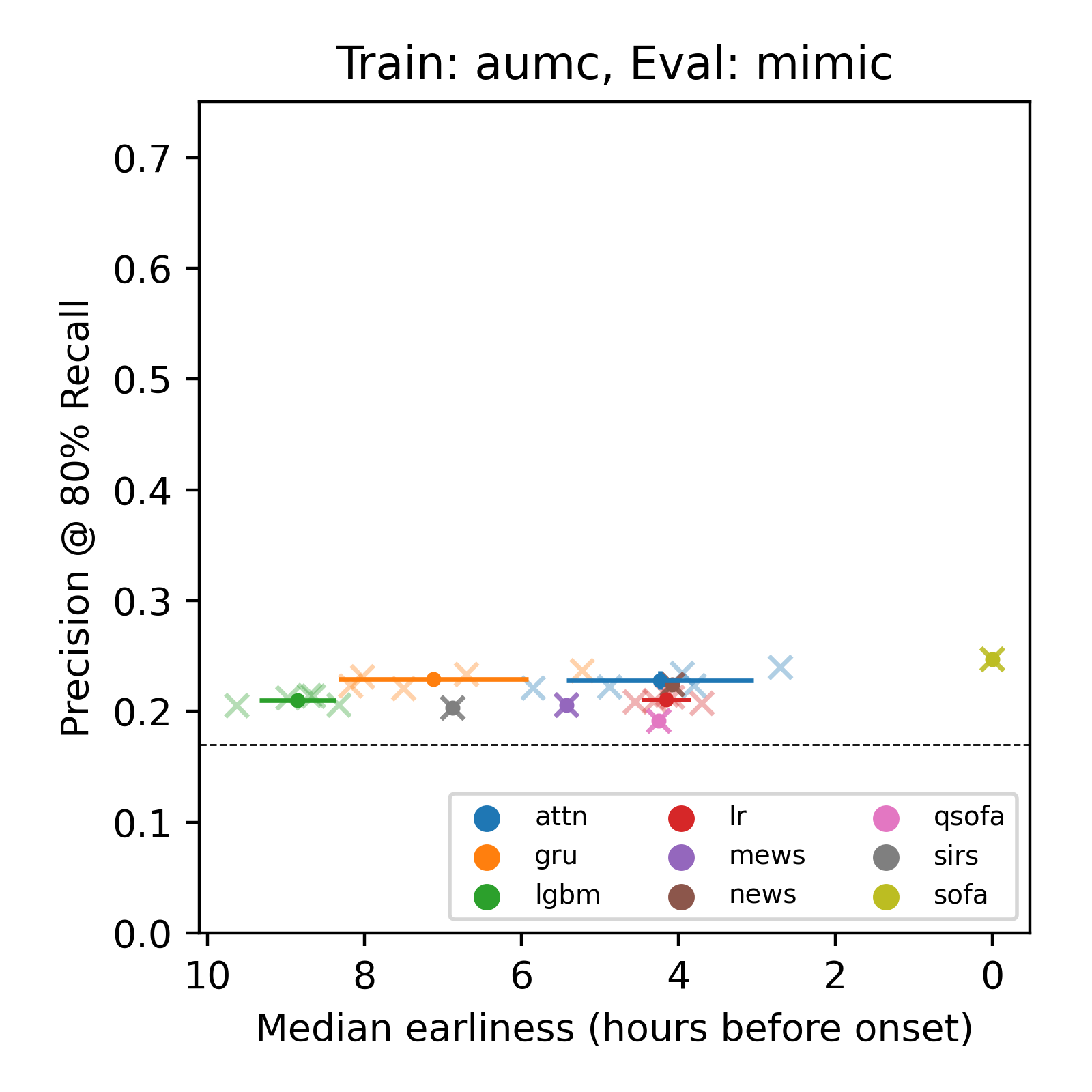}
    }
    \subcaptionbox{}{
        \includegraphics[width=0.45\linewidth]{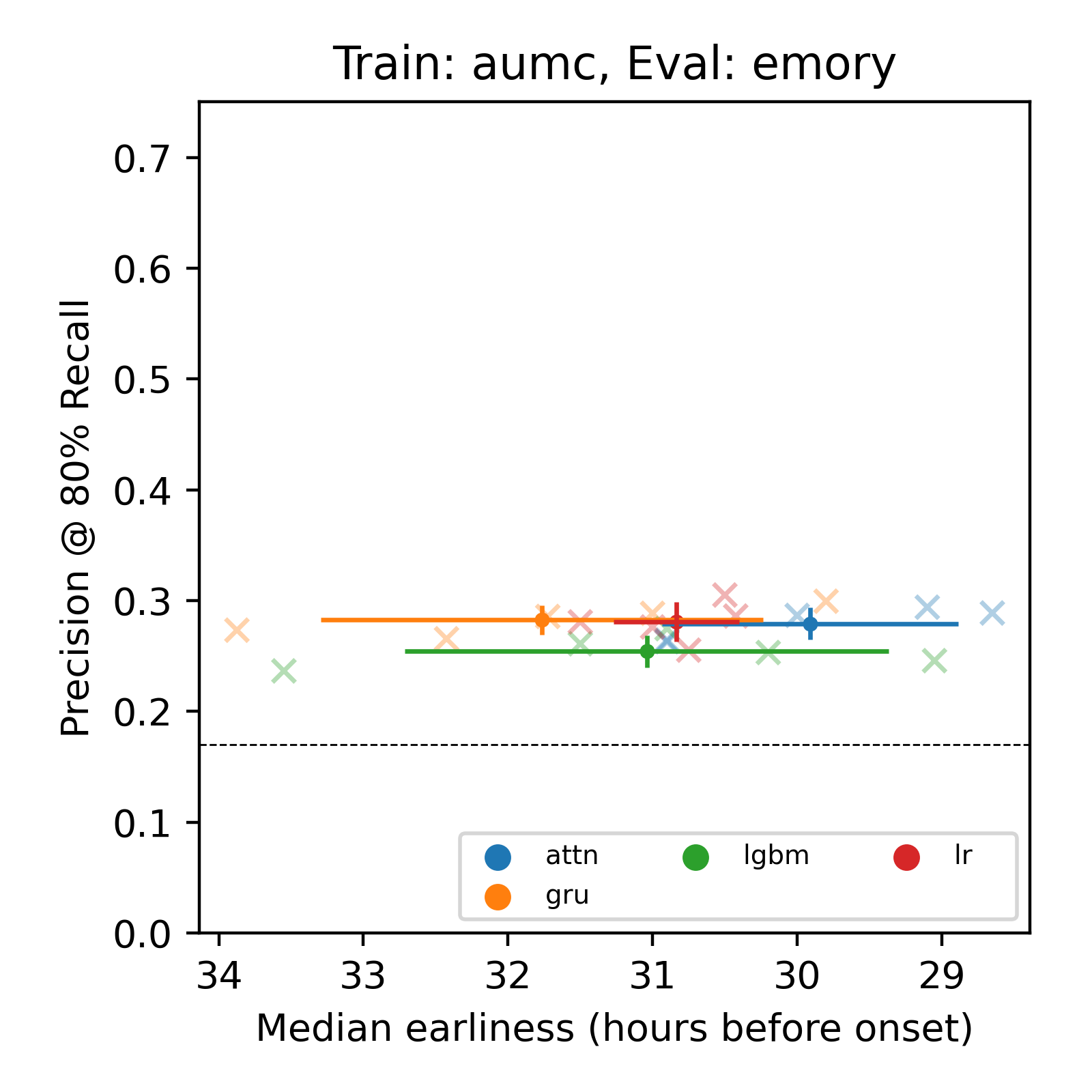}
    }
    \caption{Scatter plots for the external validation scenarios based on a single training and testing database. Each subplot depicts models trained on the AUMC database, and evaluated on one of the remaining databases. Each subfigure label indicates the respective evaluation database.}
    \label{fig:ex_scatter_aumc}
\end{figure}

\begin{figure}
    \centering
    \subcaptionbox{}{
      \includegraphics[width=0.45\linewidth]{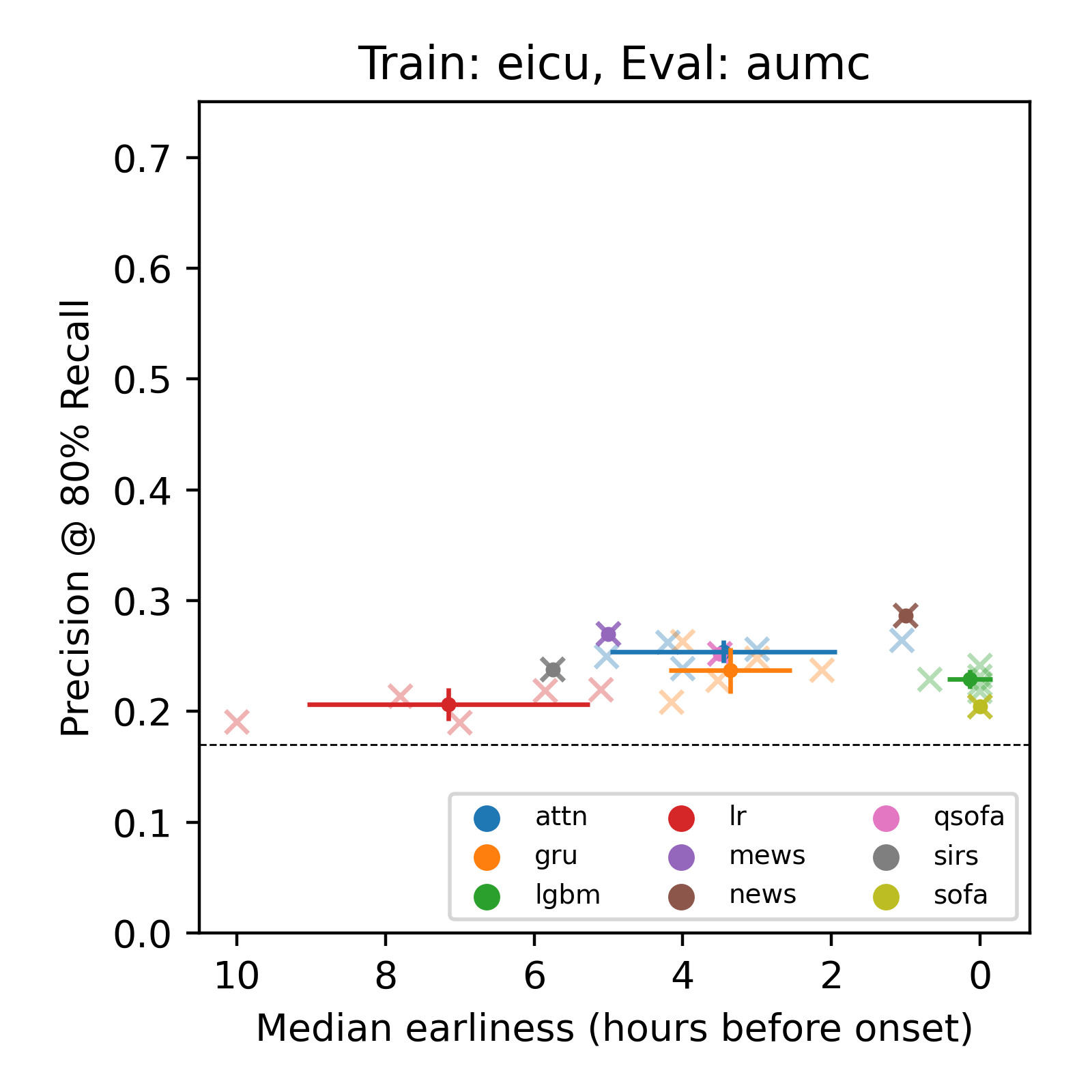}
    }
    \subcaptionbox{}{
        \includegraphics[width=0.45\linewidth]{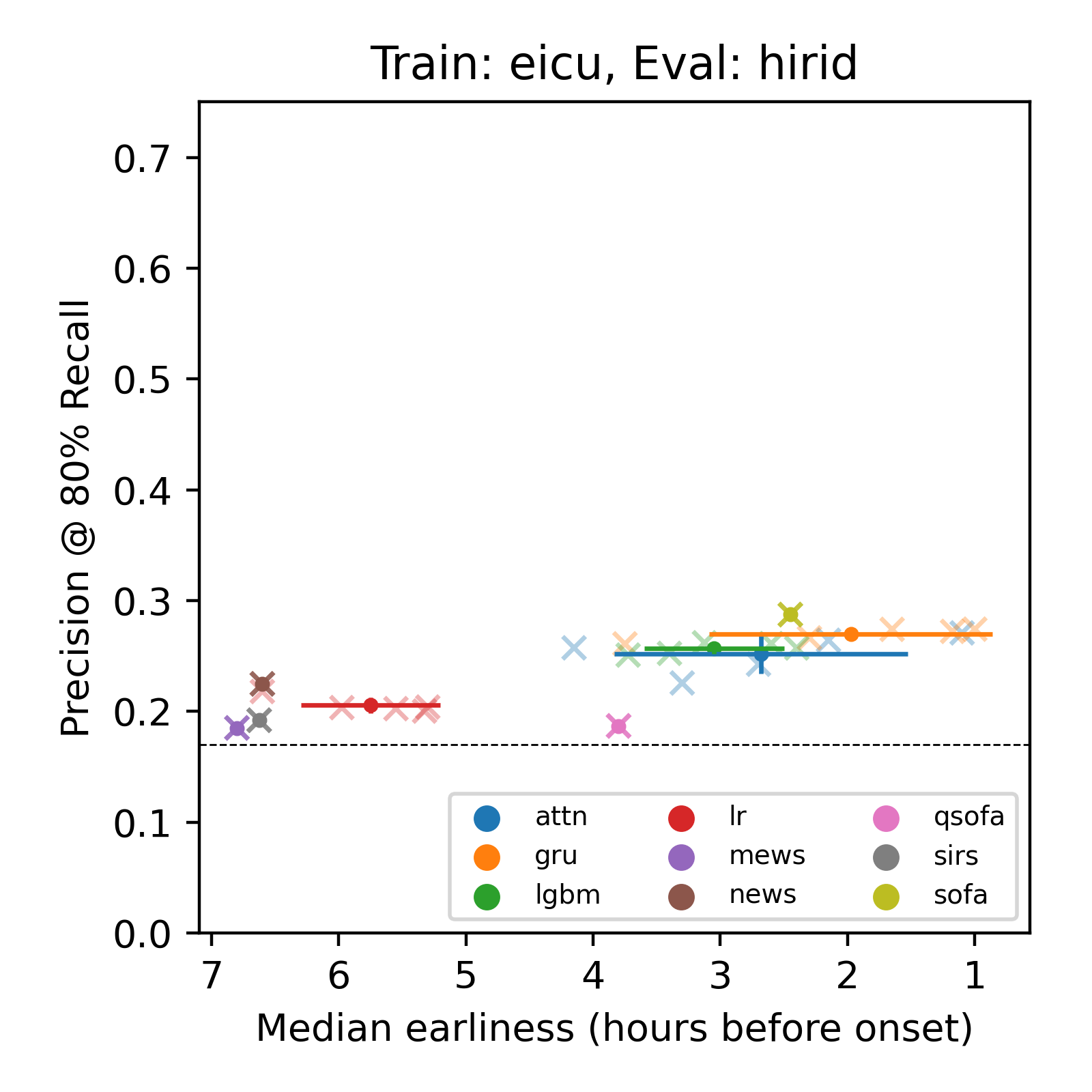}
    }\\
    \subcaptionbox{}{
        \includegraphics[width=0.45\linewidth]{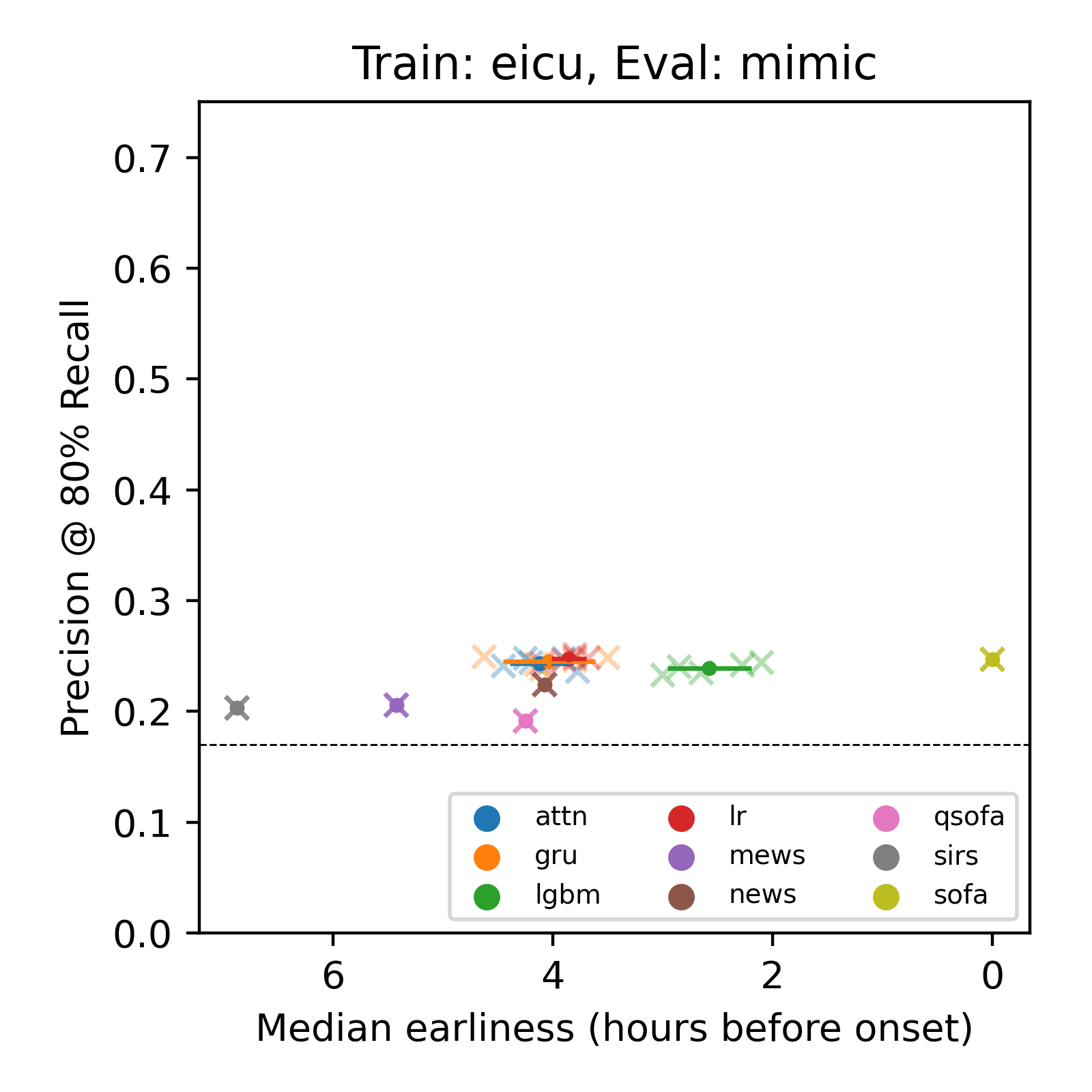}
    }
    \subcaptionbox{}{
        \includegraphics[width=0.45\linewidth]{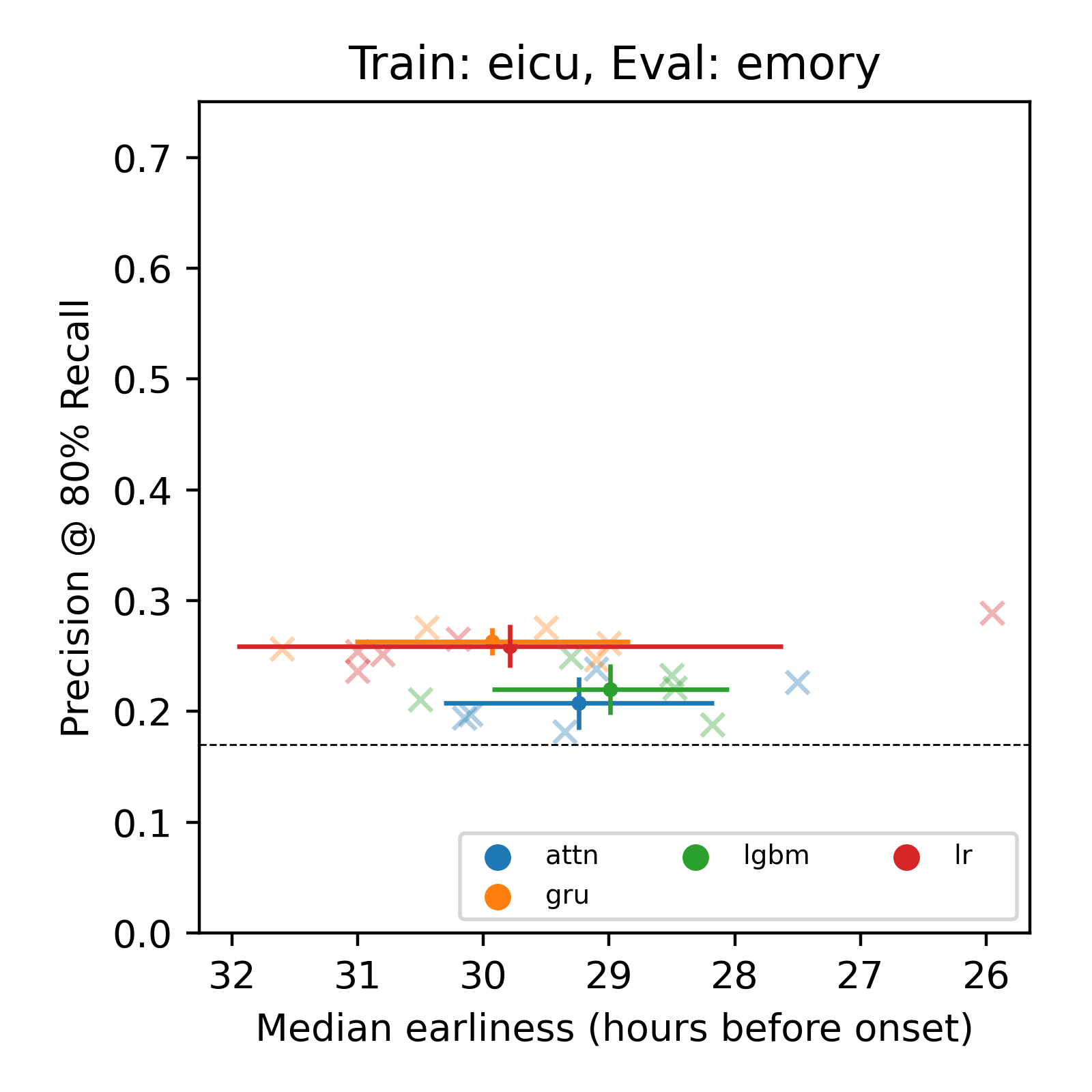}
    }
    \caption{Scatter plots for the external validation scenarios based on a single training and testing database. Each subplot depicts models trained on the eICU database, and evaluated on one of the remaining databases. Each subfigure label indicates the respective evaluation database.}
\end{figure}

\begin{figure}
    \centering
    \subcaptionbox{}{
      \includegraphics[width=0.45\linewidth]{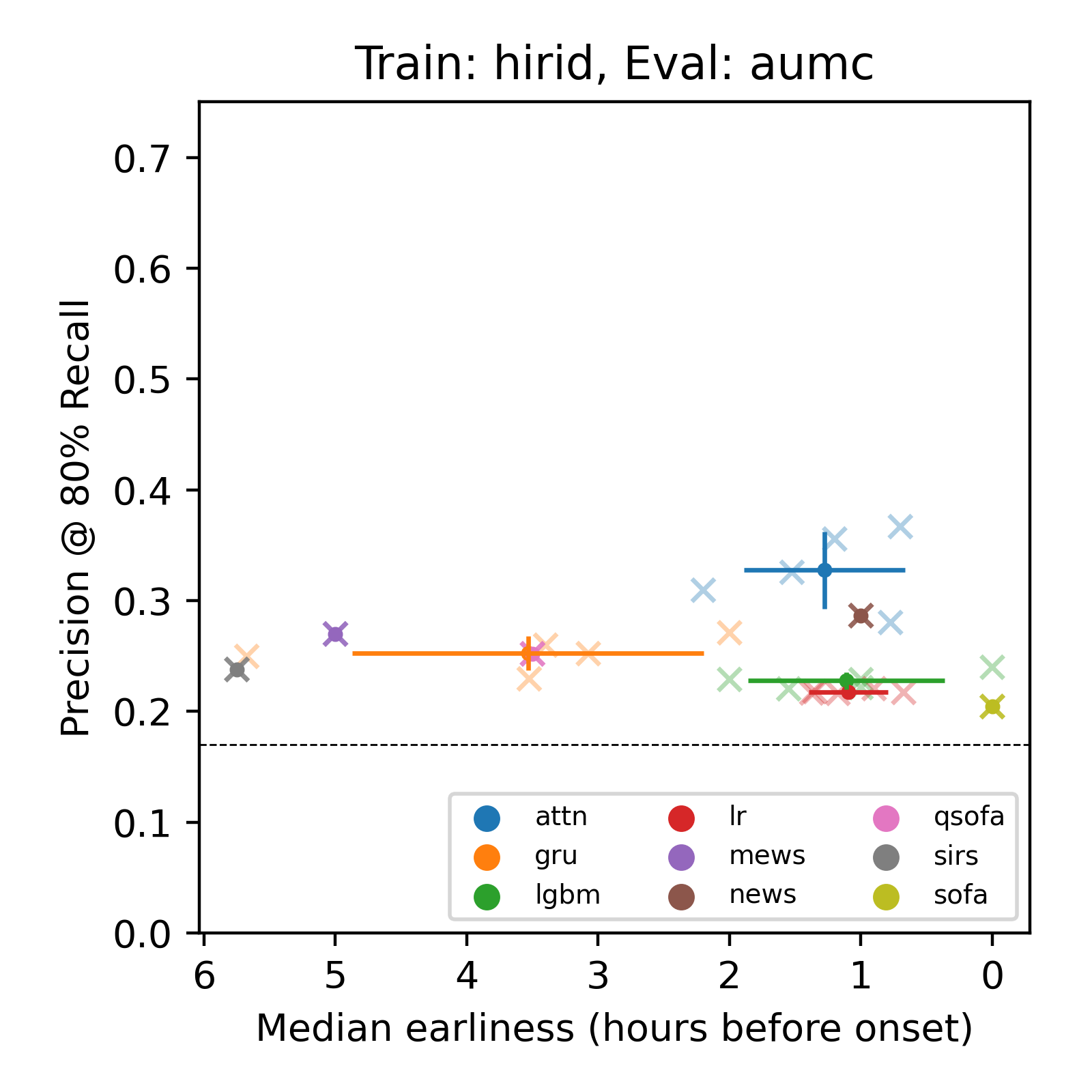}
    }
    \subcaptionbox{}{
        \includegraphics[width=0.45\linewidth]{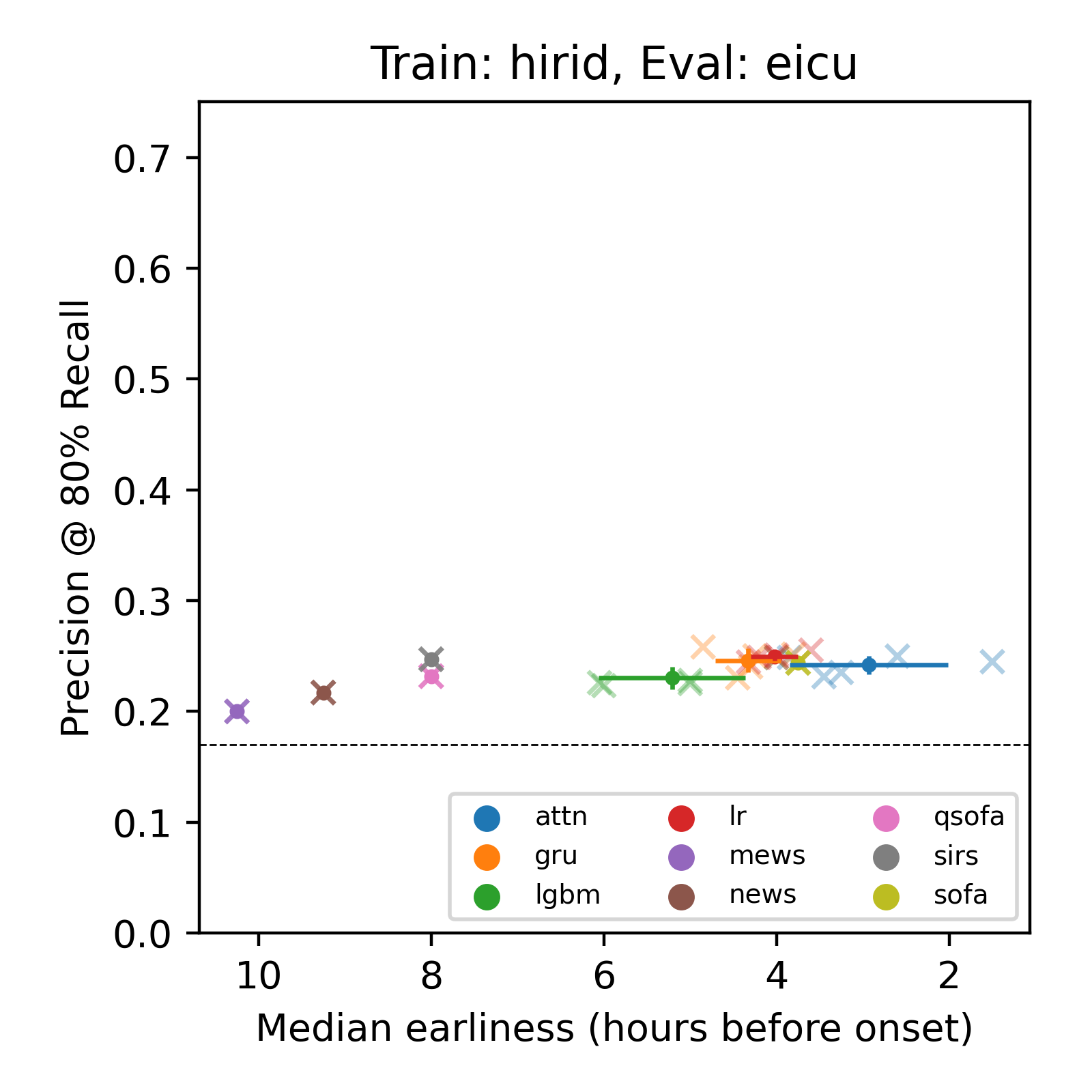}
    }\\
    \subcaptionbox{}{
        \includegraphics[width=0.45\linewidth]{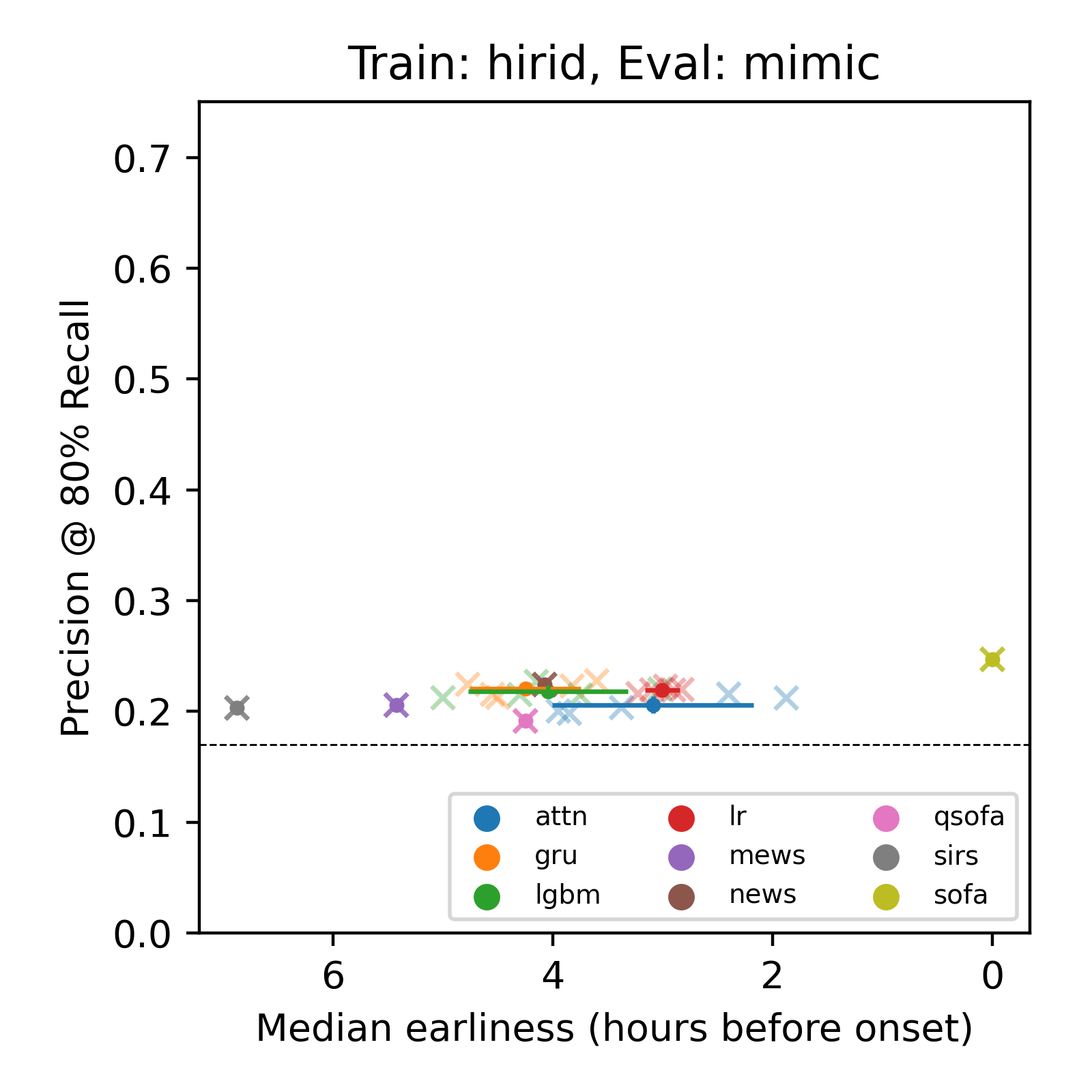}
    }
    \subcaptionbox{}{
        \includegraphics[width=0.45\linewidth]{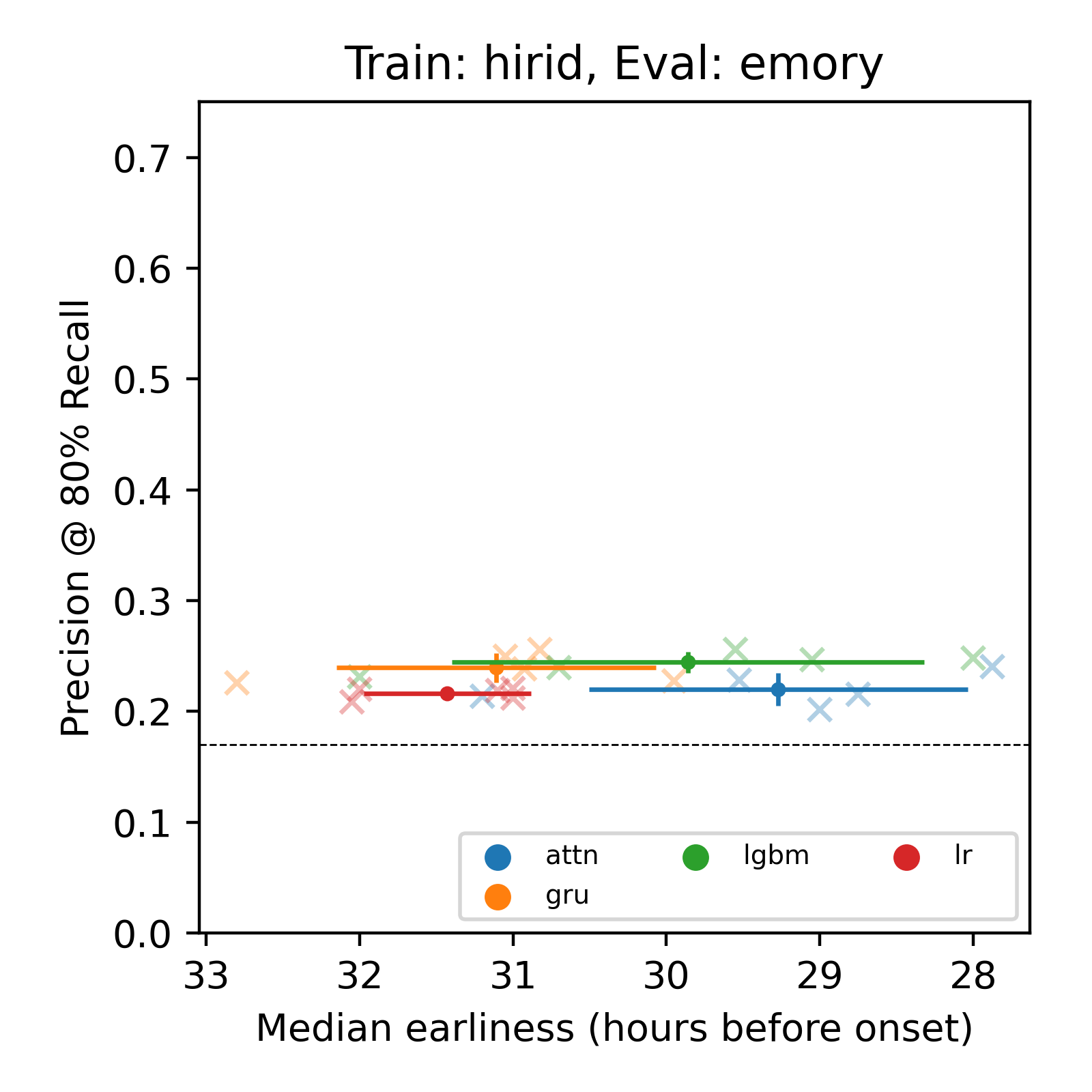}
    }
    \caption{Scatter plots for the external validation scenarios based on a single training and testing database. Each subplot depicts models trained on the HiRID database, and evaluated on one of the remaining databases. Each subfigure label indicates the respective evaluation database.}
\end{figure}

\begin{figure}
    \centering
    \subcaptionbox{}{
      \includegraphics[width=0.45\linewidth]{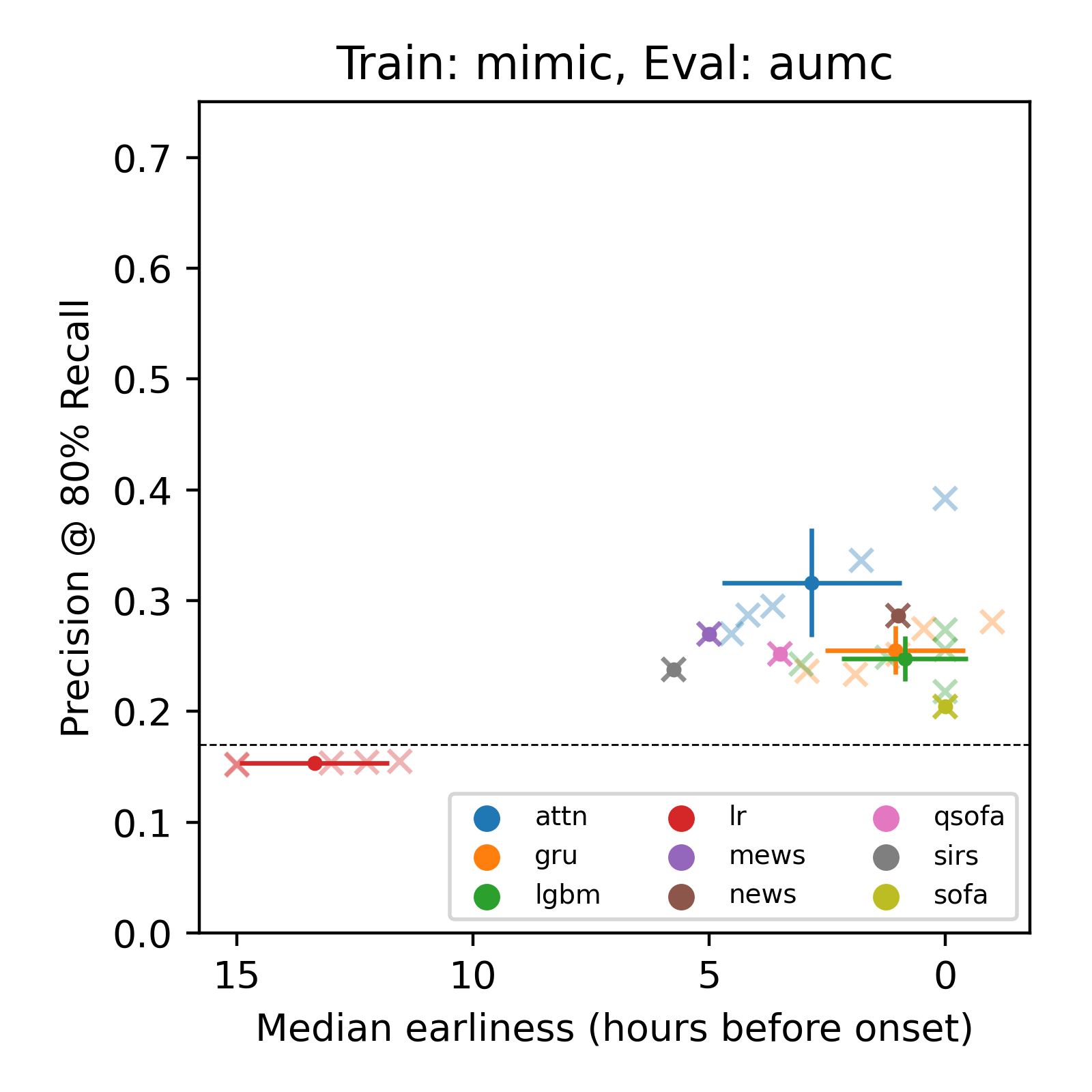}
    }
    \subcaptionbox{}{
        \includegraphics[width=0.45\linewidth]{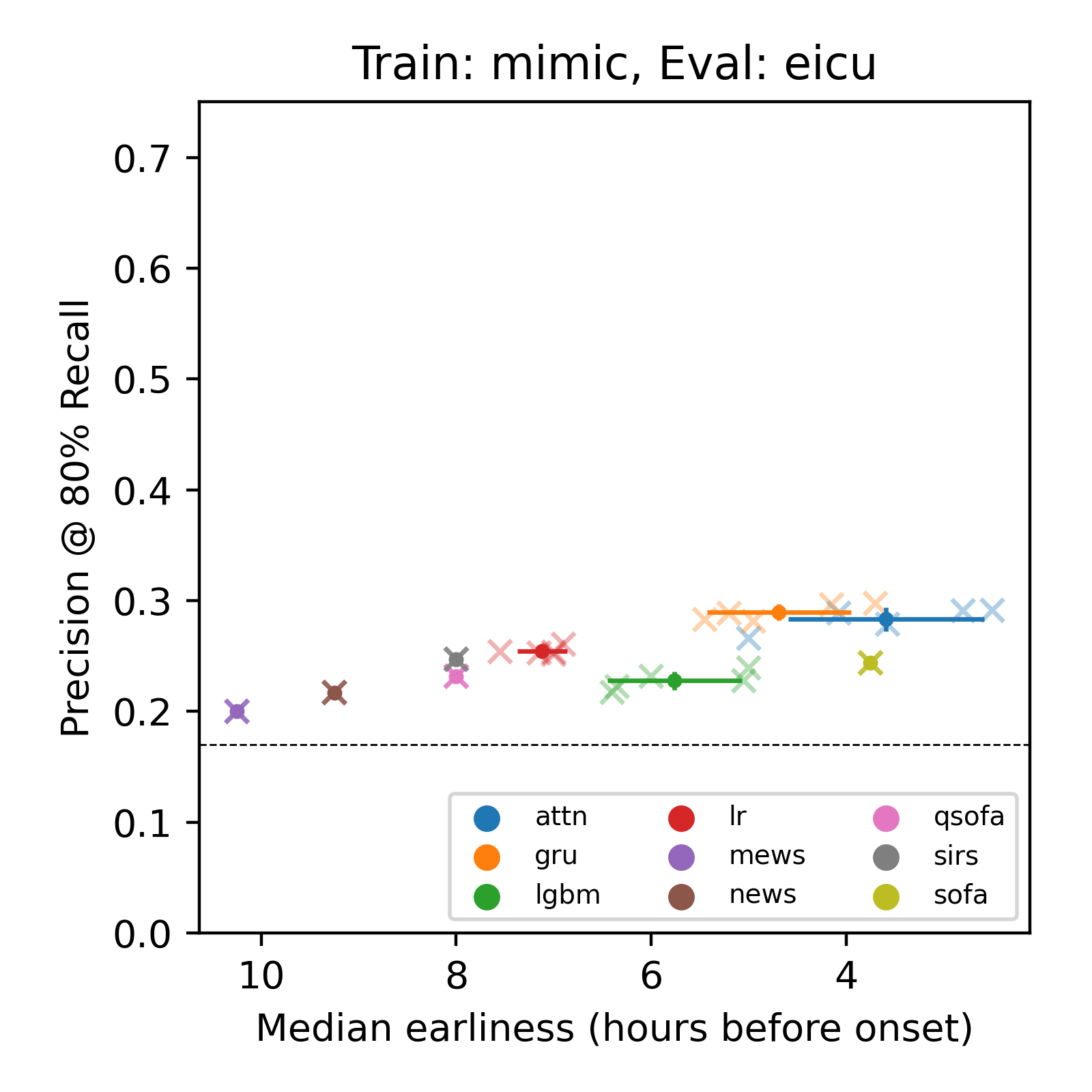}
    }\\
    \subcaptionbox{}{
        \includegraphics[width=0.45\linewidth]{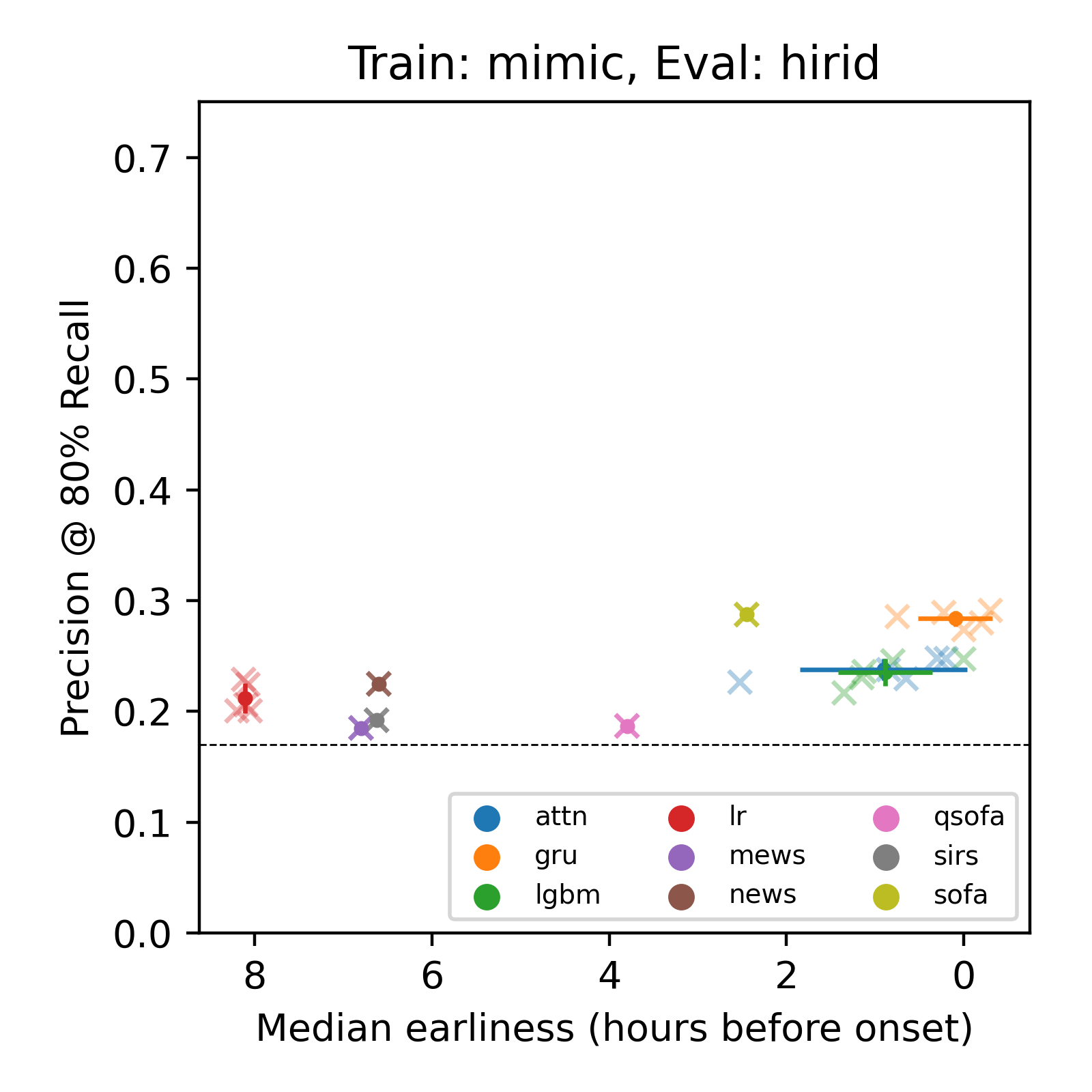}
    }
    \subcaptionbox{}{
        \includegraphics[width=0.45\linewidth]{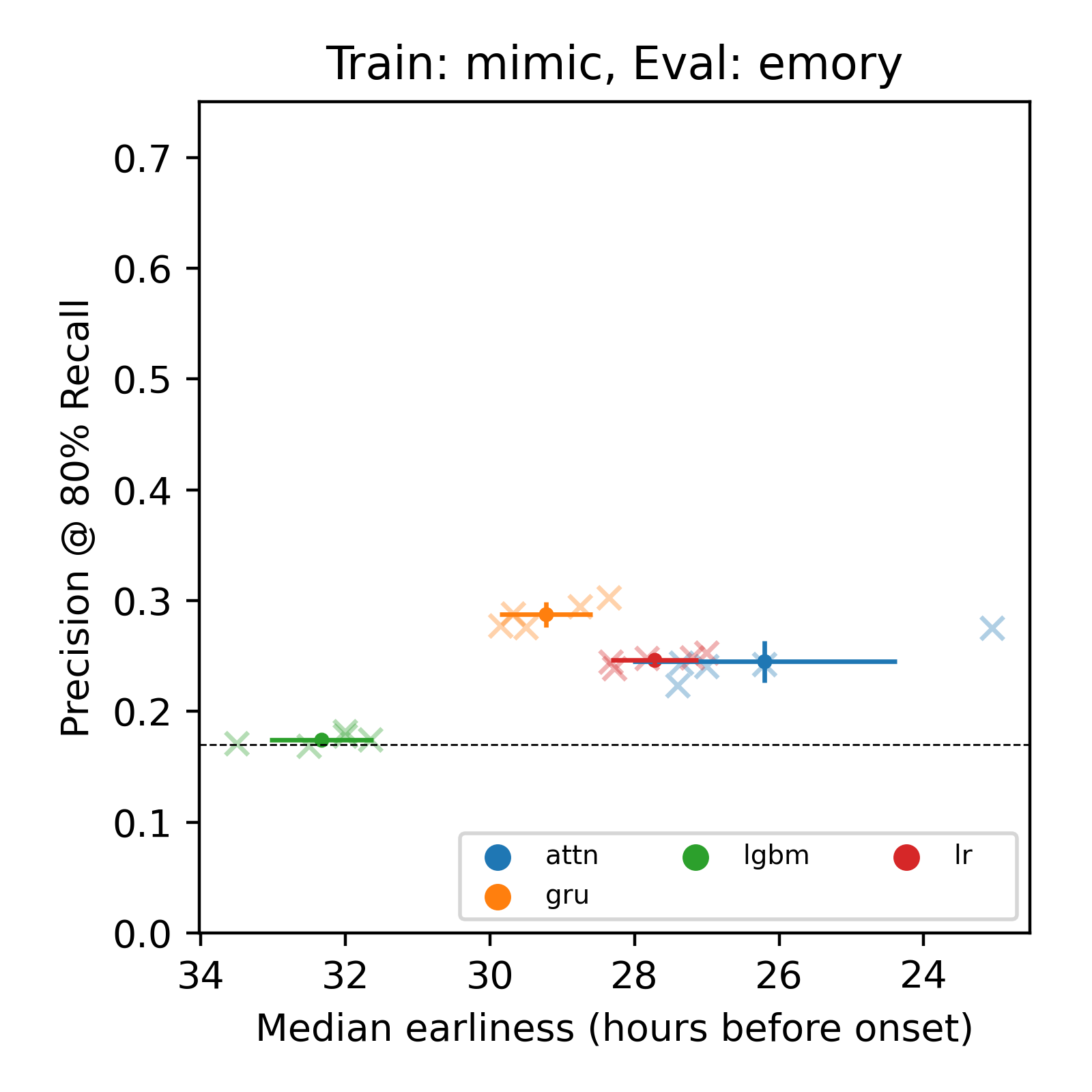}
    }
    \caption{Scatter plots for the external validation scenarios based on a single training and testing database. Each subplot depicts models trained on the MIMIC-III database, and evaluated on one of the remaining databases. Each subfigure label indicates the respective evaluation database.}
\end{figure}

\begin{figure}
    \centering
    \subcaptionbox{}{
      \includegraphics[width=0.45\linewidth]{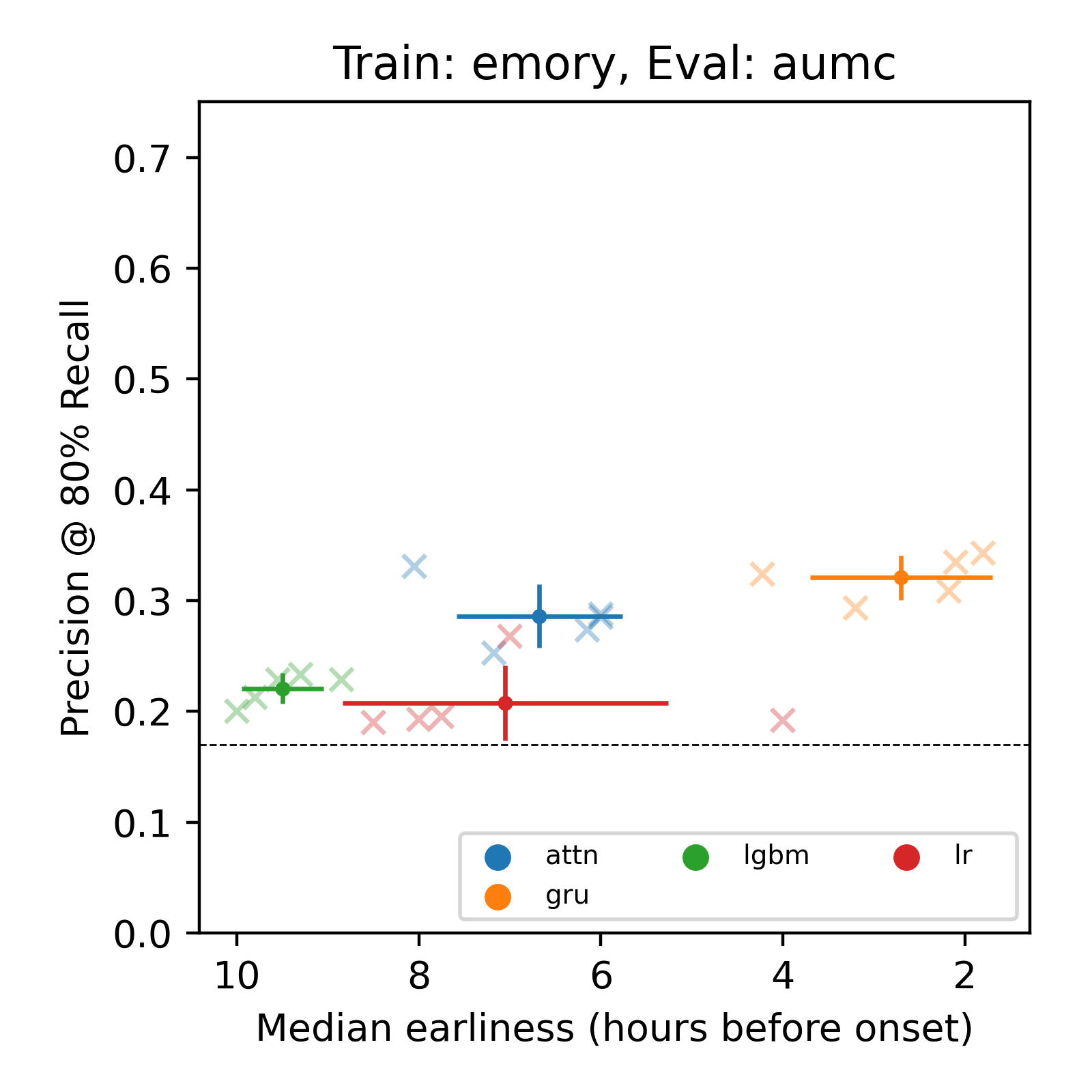}
    }
    \subcaptionbox{}{
        \includegraphics[width=0.45\linewidth]{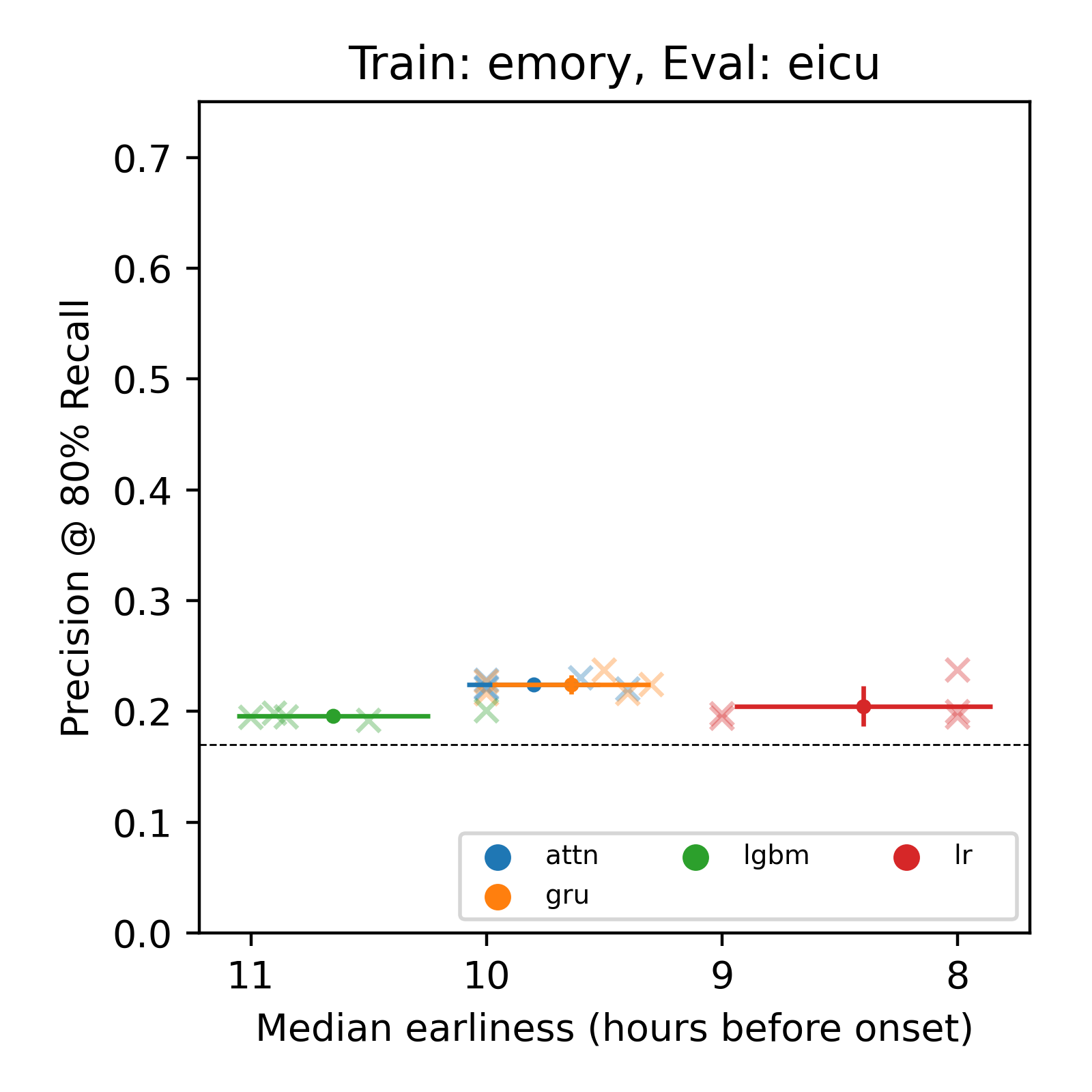}
    }\\
    \subcaptionbox{}{
        \includegraphics[width=0.45\linewidth]{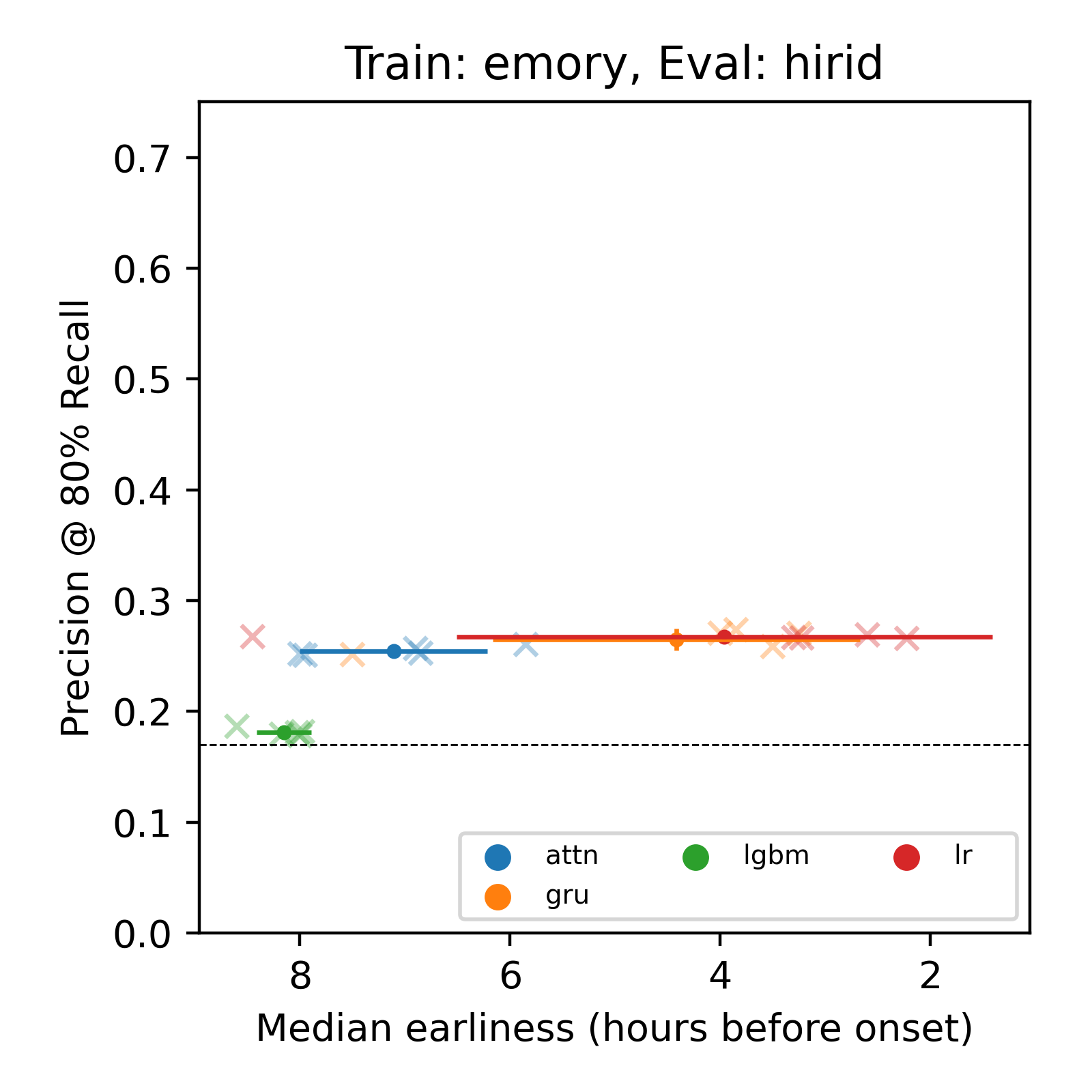}
    }
    \subcaptionbox{}{
        \includegraphics[width=0.45\linewidth]{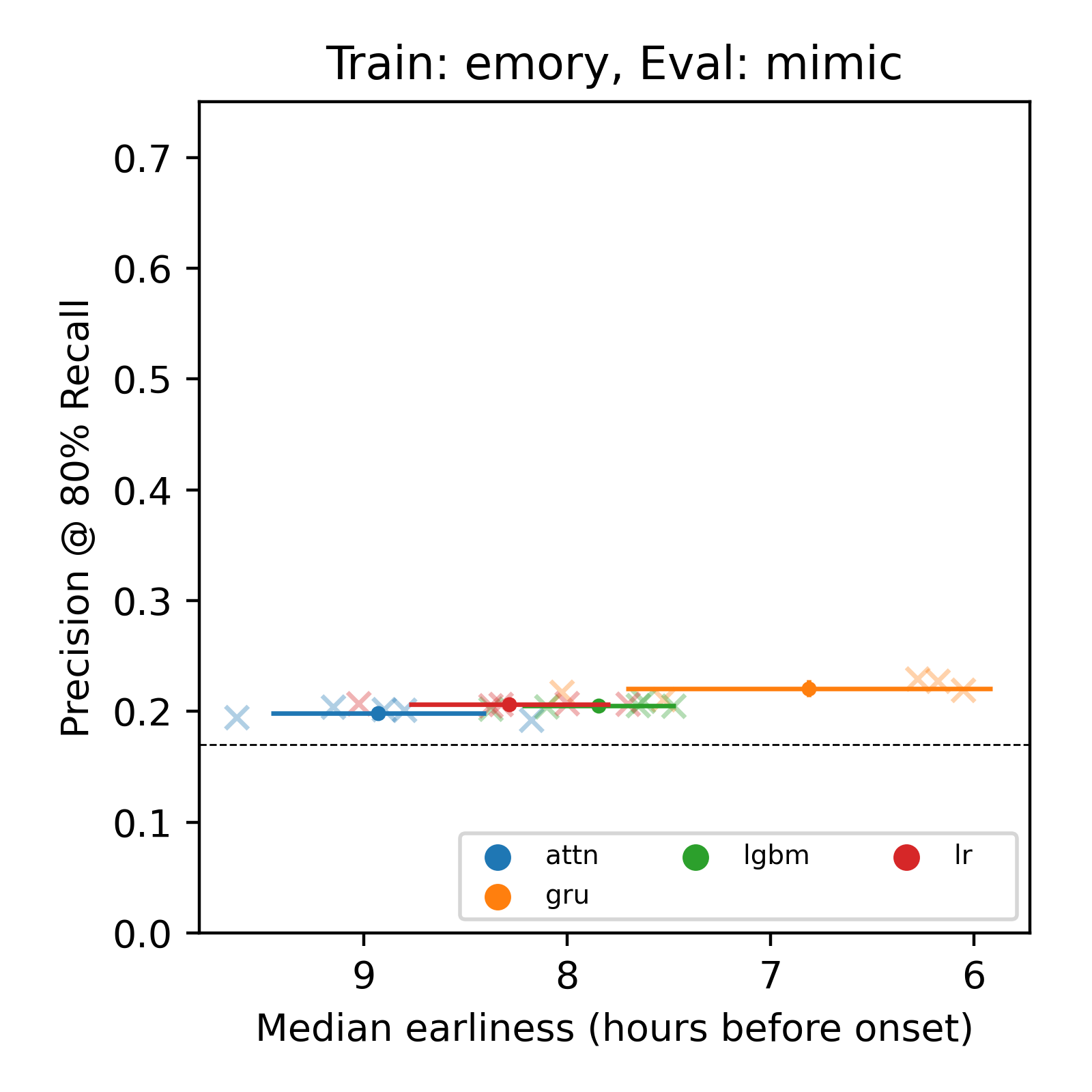}
    }
    \caption{Scatter plots for the external validation scenarios based on a single training and testing database. Each subplot depicts models trained on the Emory database, and evaluated on one of the remaining databases. Each subfigure label indicates the respective evaluation database.}
    \label{fig:ex_scatter_physionet2019}
\end{figure}

\begin{figure}
    \centering
    \subcaptionbox{\label{fig:in_emory}}{
        \includegraphics[width=0.55\linewidth]{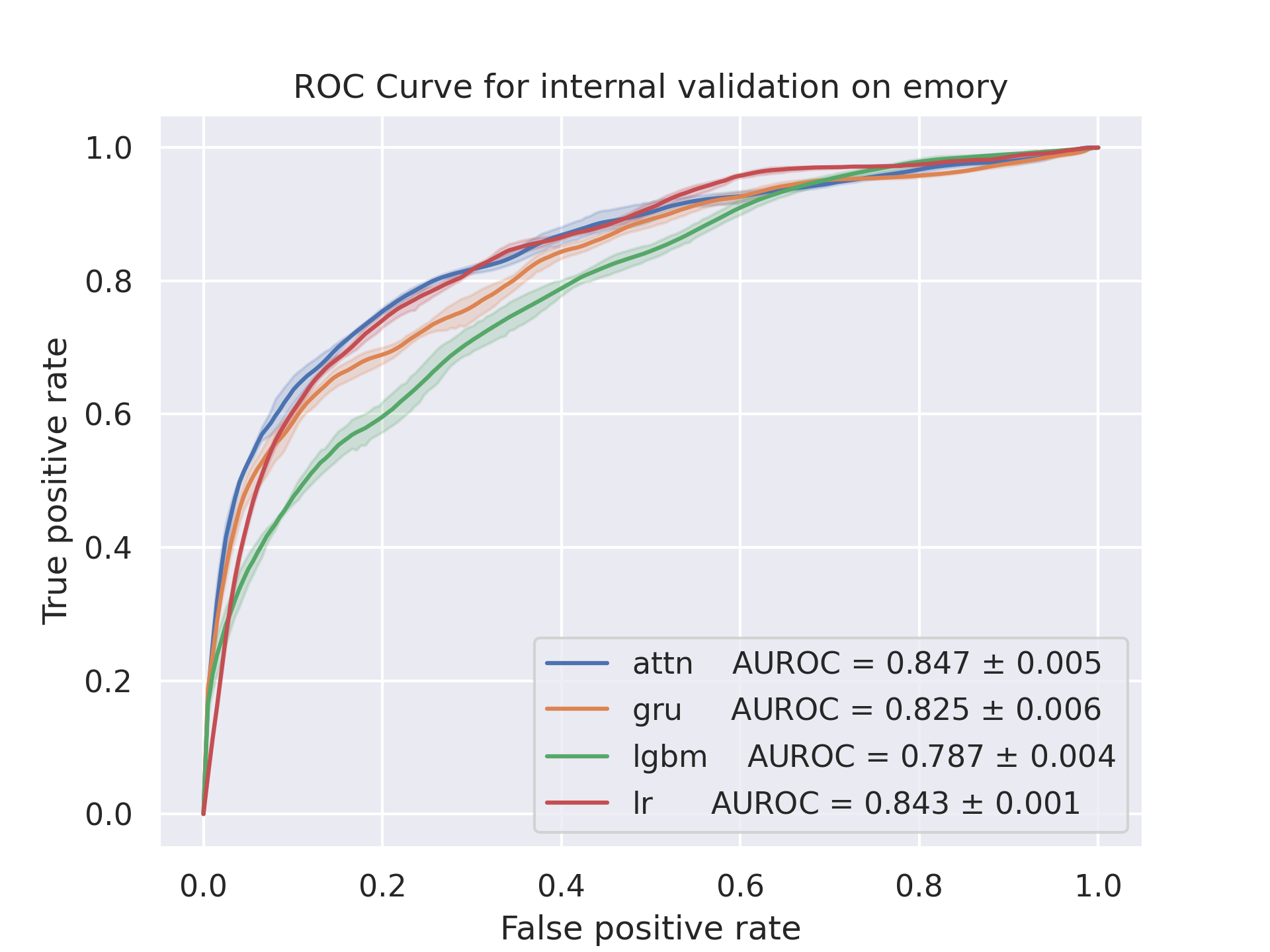}
    }%
    \subcaptionbox{\label{fig:in_emory_scatter}}{
        \includegraphics[width=0.41\linewidth]{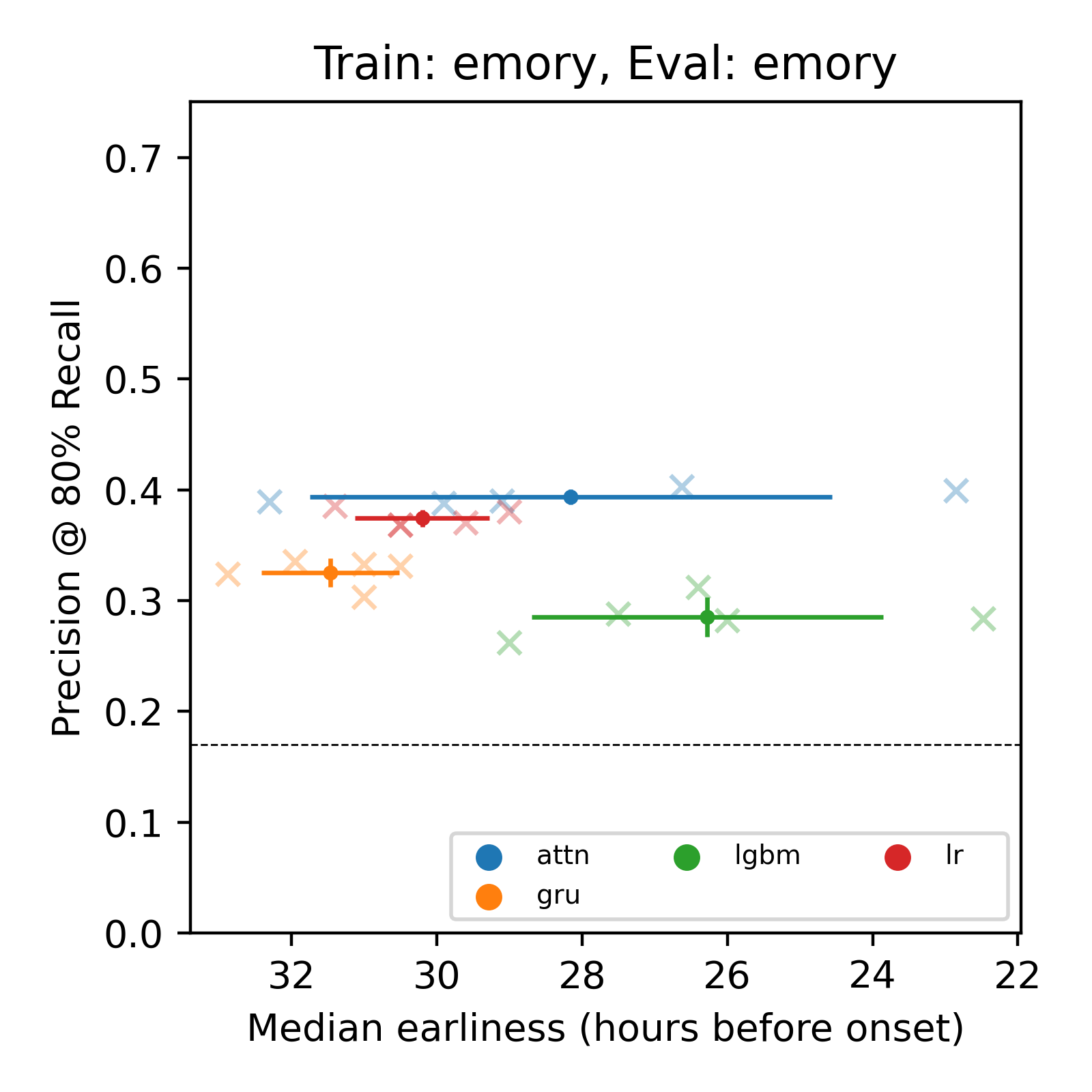}
    }%
    \caption{Internal validation results for the Emory dataset. \autoref{fig:in_emory} depicts the ROC Curves, whereas \autoref{fig:in_emory_scatter} depicts the earliness scatter plot.  This dataset was externally preprocessed and annotated and comprises a smaller variable set. Due to missingness of additional variables the clinical baseline scores were not extracted on this dataset.}
    \label{fig:emory_roc}
\end{figure}

\begin{figure}
    \centering
    \includegraphics[width=0.6\linewidth]{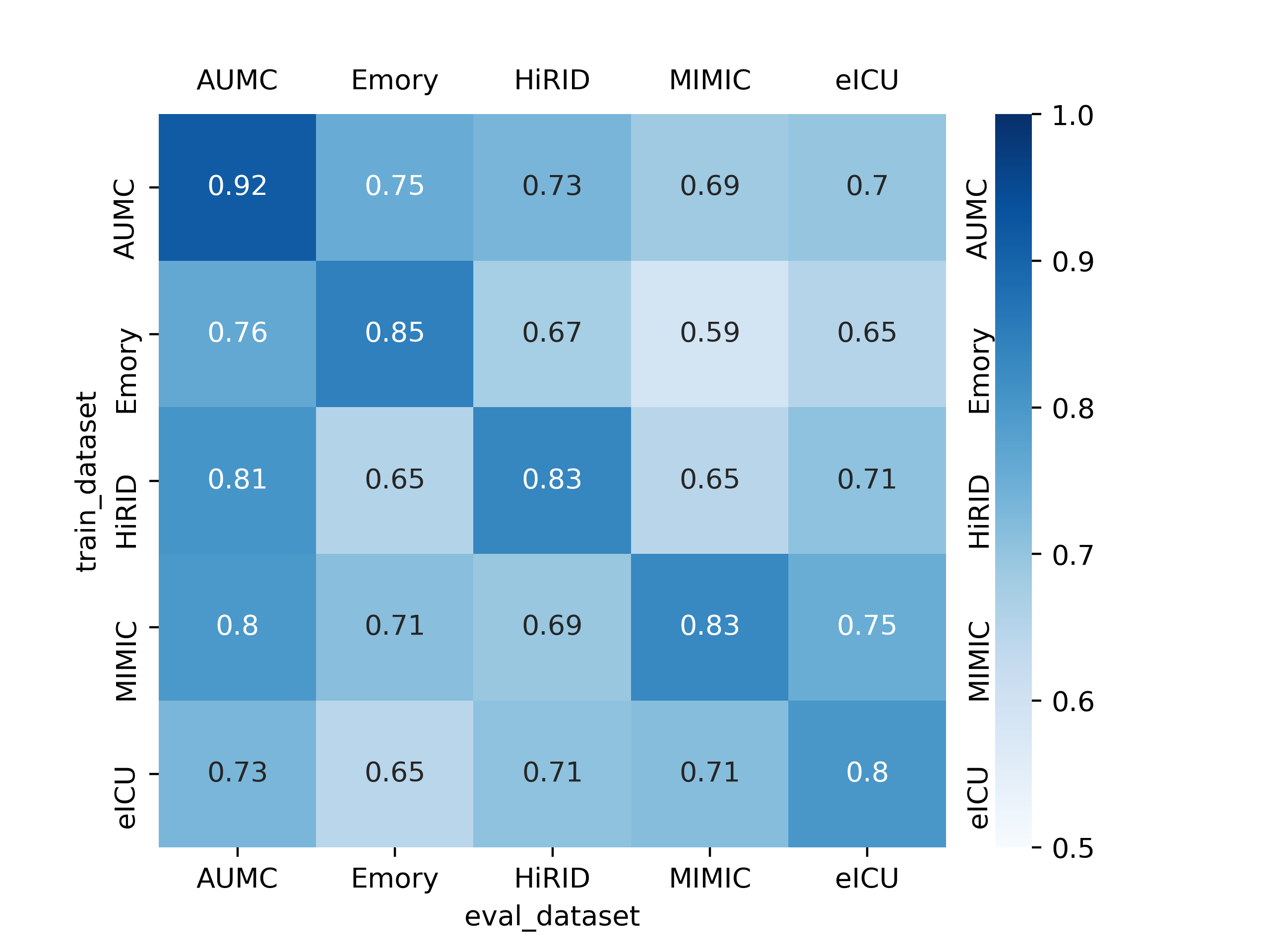}
    \caption{Extended AUROC results figure including the Emory dataset which was made available in a preprocessed and annoated stage and has a smaller variable set. In order to evaluate on this dataset models trained on other datasets were trained on the same smaller Emory variable set.}
    \label{fig:heat_extended}
\end{figure}

\begin{figure}
    \centering
    \subcaptionbox{\label{fig:feat-set_aumc}}{
        \includegraphics[width=0.9\linewidth]{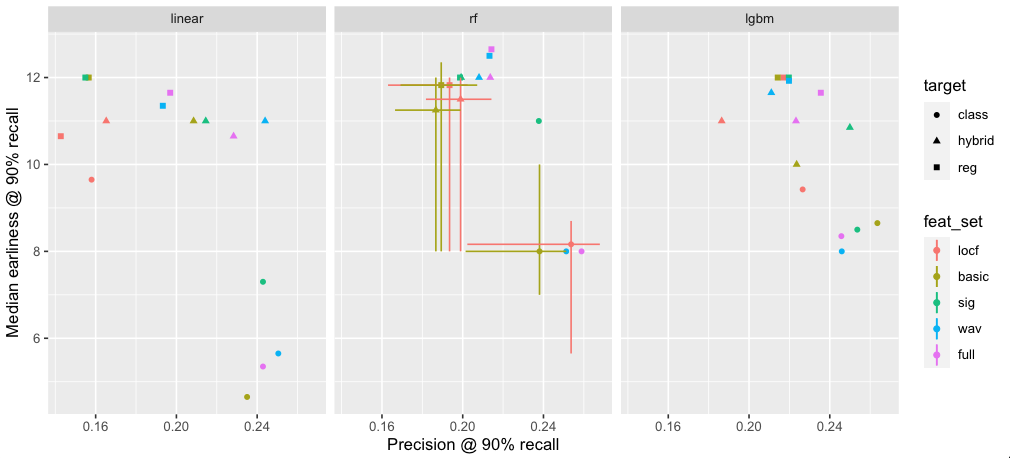}
    }\\
    \subcaptionbox{\label{fig:feat-set_mimic}}{
        \includegraphics[width=0.9\linewidth]{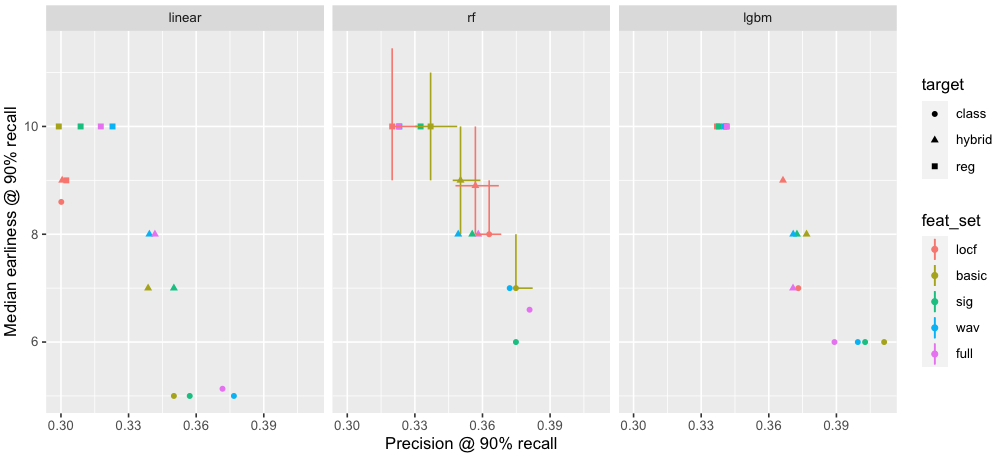}
    }
    \caption{Feature engineering exploration yields no significant increase in predictive performance when moving to larger feature sets which include wavelet scattering and/or signature transformed features. This was investigated using the AUMC (a) and MIMIC-III (b) data sets and linear logistic regression with LASSO penalty (linear), random forest (rf) and lightGBM models (lgbm). Investigated feature sets include last observation carried forward imputed values (locf); lookback, locf and derived features (basic); wavelet tranformed, alsongside basic features (wav); signature transformed, alongside basic features (sig); and a combination of all previously enumerated feature groups (full). Additionally, several objectives were looked at, including a classification target (class), where the label is false util 6h prior to sepsis onset and true afterwards; a hybrid target where the label switches back to false 6h after onset; and a regression target which was modelled after the PhysioNet challenge utility function \citep{reyna2019early}. Bars in random forest panels represent ranges resulting from different data splitting.}
    \label{fig:explore}
\end{figure}

\begin{figure}
    \centering
    \subcaptionbox{}{
        \includegraphics[width=0.30\linewidth]{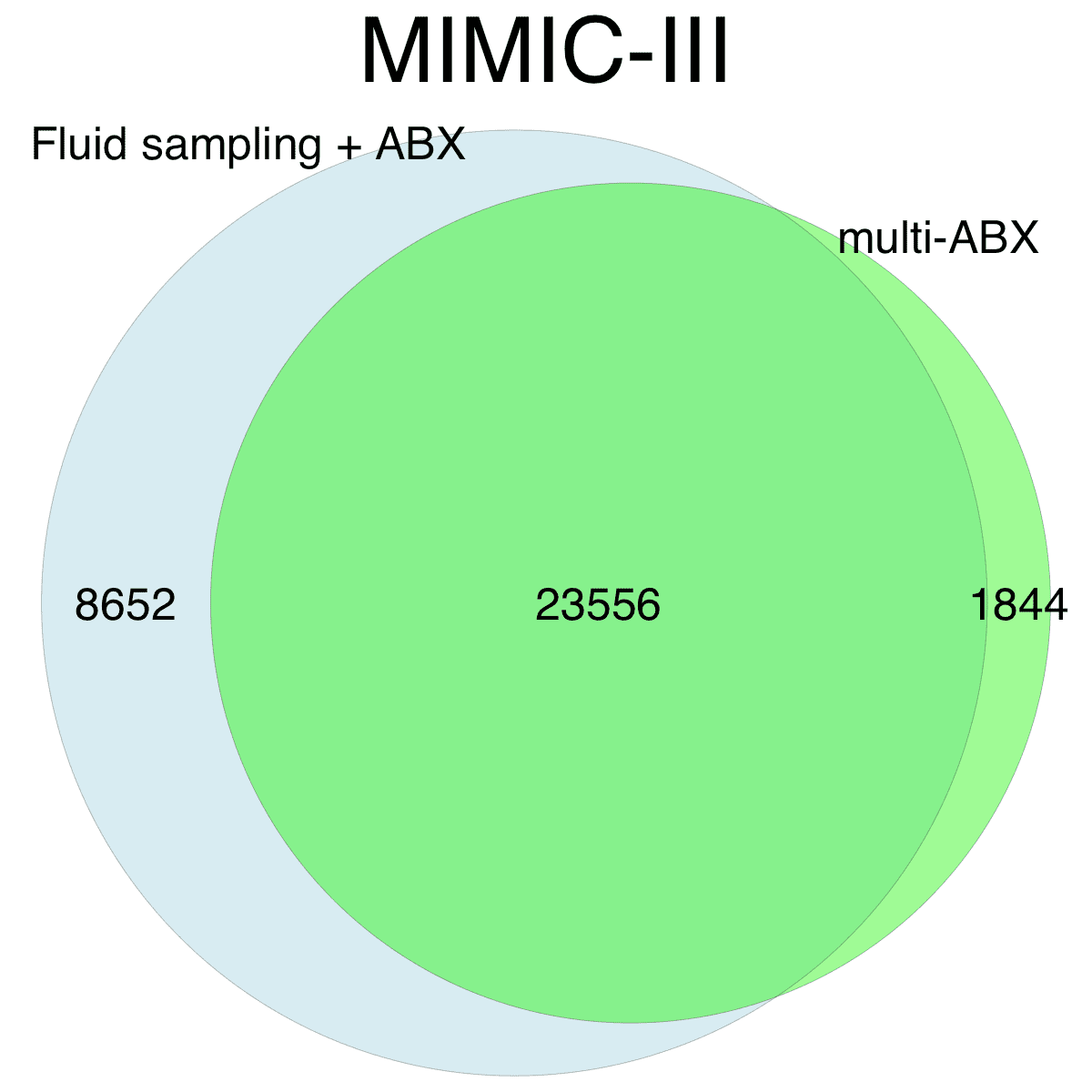}
    }
    \subcaptionbox{}{
        \includegraphics[width=0.30\linewidth]{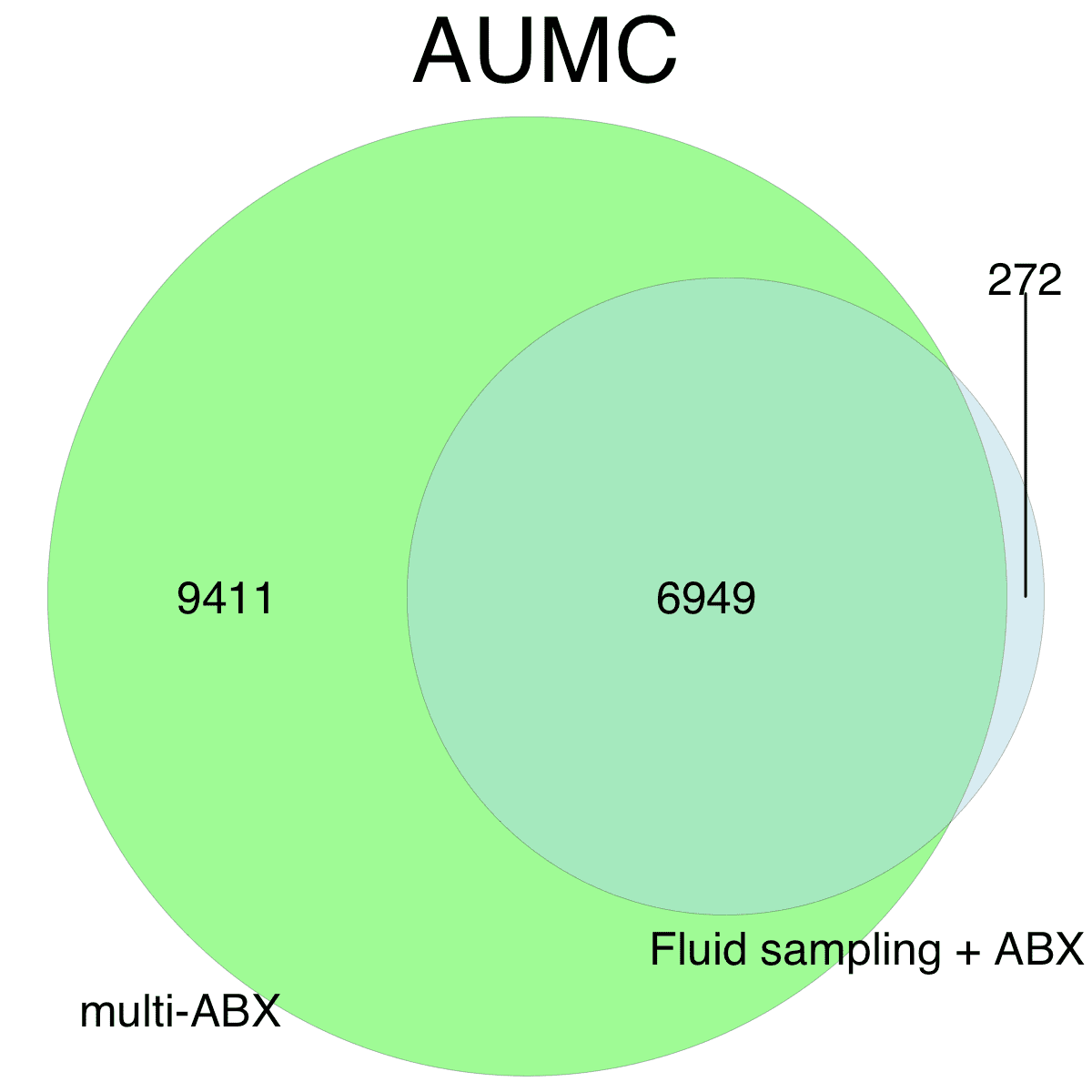}
    }%
    \subcaptionbox{}{
        \includegraphics[width=0.3\linewidth]{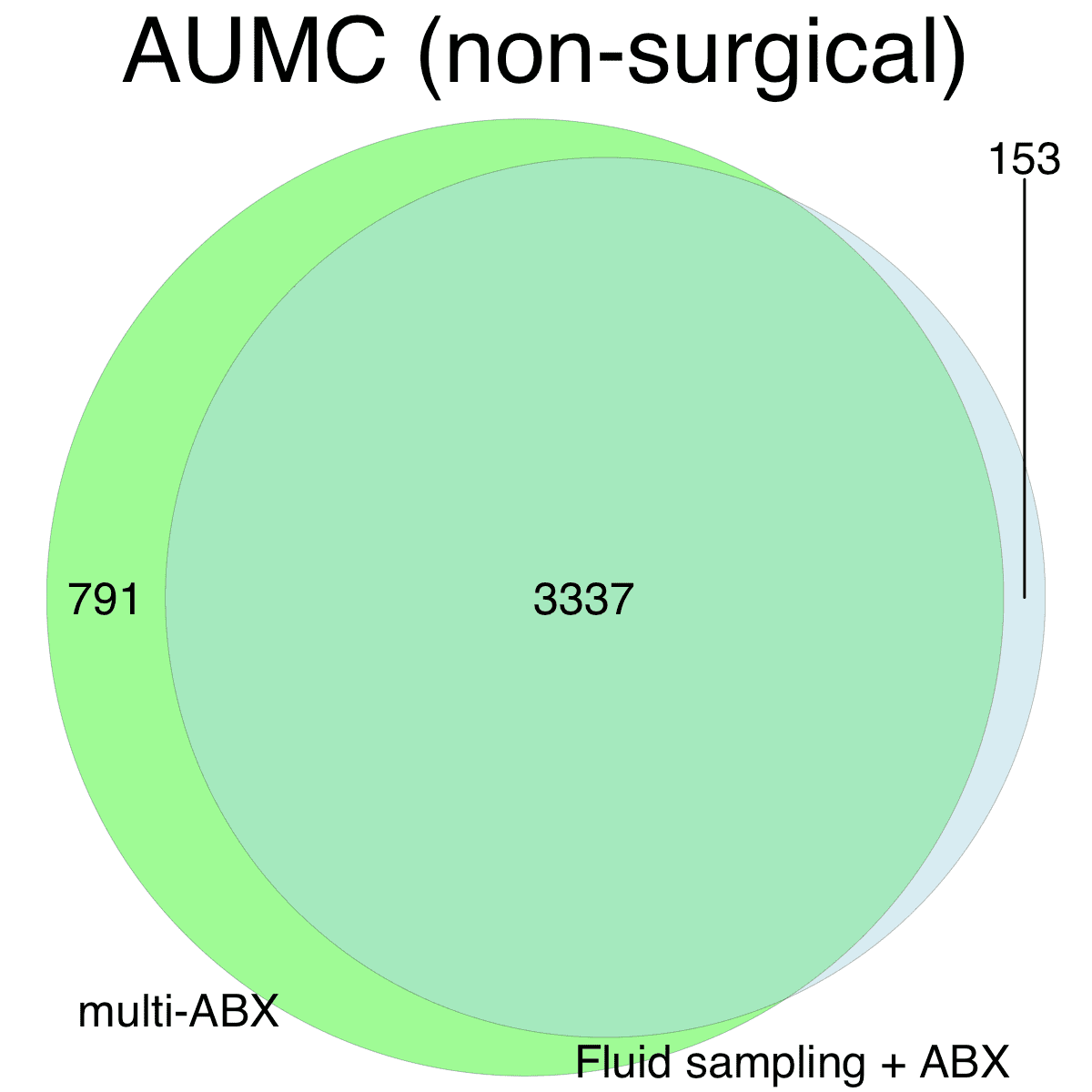}
    }
    \caption{The comparison of multiple antibiotics definition of suspected infection with the original definition showed a Jaccard similarity of (a) 0.69 in the MIMIC-III database; (b) 0.42 in the AUMC database; (c) 0.78 in the non-surgical cohort of the AUMC database. }
    \label{fig:siaumcnonsurg}
\end{figure}
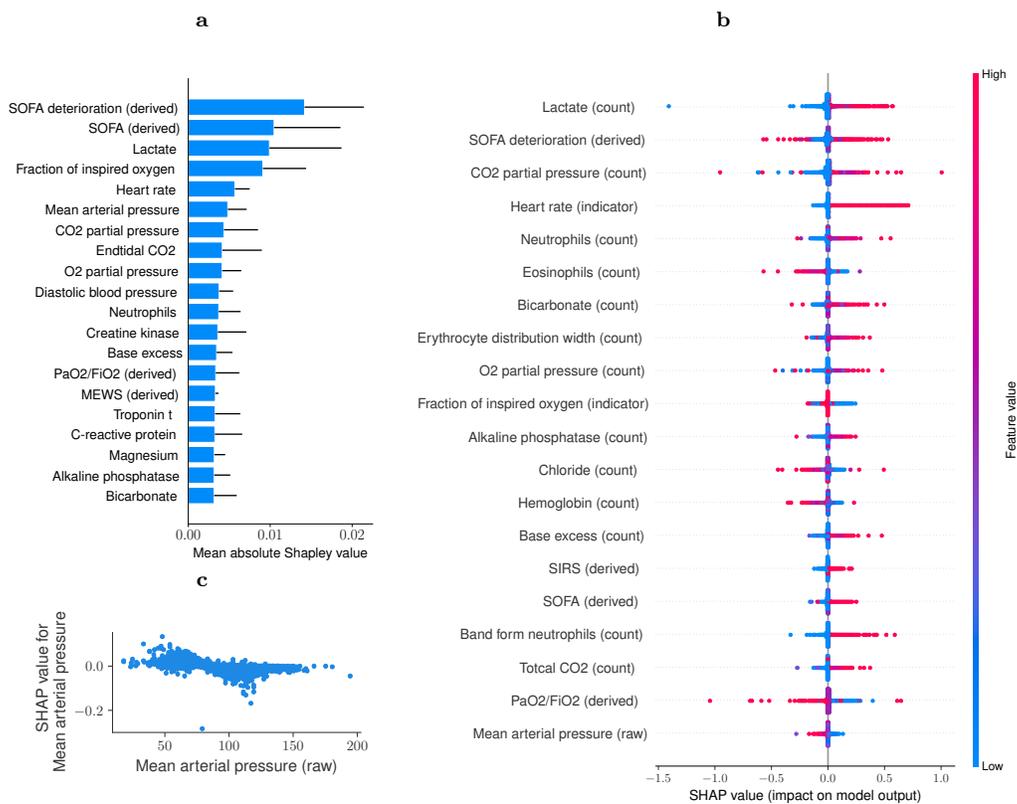
\begin{figure}[tbp]
  \centering
  \newsavebox{\imgd}
  \newsavebox{\imge}
  \newsavebox{\imgf}
  \sbox{\imgd}{%
    \subcaptionbox{\label{fig:shap_all_all_vars}}{%
      \resizebox{0.35\linewidth}{!}{%
        \input{./figures/shapley/shapley_16h_bar.pgf}
      }
      \vspace{-0.75cm}
    }%
  }%
  \sbox{\imge}{%
    \subcaptionbox{\label{fig:shap_beeplot_all_vars}}{%
      \resizebox{0.60\linewidth}{!}{%
        \graphicspath{{figures/shapley/}}
        \input{figures/shapley/shapley_vx8vbt08_16h_EICU_dot.pgf}
      }
    }
  }%
  \sbox{\imgf}{%
    \subcaptionbox{\label{fig:shap_scatter_all_vars}}{%
      \resizebox{0.35\linewidth}{!}{%
        \graphicspath{{figures/shapley/}}
        \input{figures/shapley/shapley_vx8vbt08_16h_EICU_scatter_map.pgf}
      }
    }%
  }%
  %
  \begin{minipage}[t][\ht\imge]{0.35\linewidth}
    \usebox{\imgd}
    \usebox{\imgf}
  \end{minipage}%
  \begin{minipage}[b][\ht\imge]{0.60\linewidth}
    \usebox{\imge}
  \end{minipage}%
  \caption{%
    A repetition of the Shapley value analysis presented in
    \autoref{fig:Shapley}  that considers all features types used in the deep model, i.e., raw,
    count, indicator, and derived features.
  }
  \label{fig:Shapley full}
\end{figure}

\FloatBarrier

\begin{longtable}[t]{lllllll}
\caption{\label{tab:variables}Variables used for sepsis prediction.}\\
\toprule
Name & Description & MI-III & eICU & HiRID & AUMC & Emory\\
\midrule
\endfirsthead
\caption[]{Variables used for sepsis prediction. \textit{(continued)}}\\
\toprule
Name & Description & MI-III & eICU & HiRID & AUMC & Emory\\
\midrule
\endhead

\endfoot
\bottomrule
\endlastfoot
age & patient age & \ding{51} & \ding{51} & \ding{51} & \ding{51} & \ding{51}\\
alb & albumin & \ding{51} & \ding{51} & \ding{51} & \ding{51} & \ding{55}\\
alp & alkaline phosphatase & \ding{51} & \ding{51} & \ding{51} & \ding{51} & \ding{51}\\
alt & alanine aminotransferase & \ding{51} & \ding{51} & \ding{51} & \ding{51} & \ding{55}\\
ast & aspartate aminotransferase & \ding{51} & \ding{51} & \ding{51} & \ding{51} & \ding{51}\\
\addlinespace
basos & basophils & \ding{51} & \ding{51} & \ding{55} & \ding{51} & \ding{55}\\
be & base excess & \ding{51} & \ding{51} & \ding{51} & \ding{51} & \ding{51}\\
bicar & bicarbonate & \ding{51} & \ding{51} & \ding{51} & \ding{51} & \ding{51}\\
bili & total bilirubin & \ding{51} & \ding{51} & \ding{51} & \ding{51} & \ding{51}\\
bili\_dir & bilirubin direct & \ding{51} & \ding{51} & \ding{51} & \ding{51} & \ding{51}\\
\addlinespace
bnd & band form neutrophils & \ding{51} & \ding{51} & \ding{51} & \ding{51} & \ding{55}\\
bun & blood urea nitrogen & \ding{51} & \ding{51} & \ding{51} & \ding{51} & \ding{51}\\
ca & calcium & \ding{51} & \ding{51} & \ding{51} & \ding{51} & \ding{51}\\
cai & calcium ionized & \ding{51} & \ding{51} & \ding{51} & \ding{51} & \ding{55}\\
ck & creatine kinase & \ding{51} & \ding{51} & \ding{51} & \ding{51} & \ding{55}\\
\addlinespace
ckmb & creatine kinase MB & \ding{51} & \ding{51} & \ding{51} & \ding{51} & \ding{55}\\
cl & chloride & \ding{51} & \ding{51} & \ding{51} & \ding{51} & \ding{51}\\
crea & creatinine & \ding{51} & \ding{51} & \ding{51} & \ding{51} & \ding{51}\\
crp & C-reactive protein & \ding{51} & \ding{51} & \ding{51} & \ding{51} & \ding{55}\\
dbp & diastolic blood pressure & \ding{51} & \ding{51} & \ding{51} & \ding{51} & \ding{51}\\
\addlinespace
eos & eosinophils & \ding{51} & \ding{51} & \ding{55} & \ding{51} & \ding{55}\\
esr & erythrocyte sedimentation rate & \ding{51} & \ding{55} & \ding{51} & \ding{51} & \ding{55}\\
etco2 & endtidal CO2 & \ding{51} & \ding{55} & \ding{51} & \ding{51} & \ding{51}\\
fgn & fibrinogen & \ding{51} & \ding{51} & \ding{51} & \ding{51} & \ding{51}\\
fio2 & fraction of inspired oxygen & \ding{51} & \ding{51} & \ding{51} & \ding{51} & \ding{51}\\
\addlinespace
glu & glucose & \ding{51} & \ding{51} & \ding{51} & \ding{51} & \ding{51}\\
hbco & carboxyhemoglobin & \ding{55} & \ding{51} & \ding{51} & \ding{51} & \ding{55}\\
hct & hematocrit & \ding{51} & \ding{51} & \ding{55} & \ding{51} & \ding{51}\\
height & patient height & \ding{51} & \ding{51} & \ding{51} & \ding{51} & \ding{55}\\
hgb & hemoglobin & \ding{51} & \ding{51} & \ding{51} & \ding{51} & \ding{51}\\
\addlinespace
hr & heart rate & \ding{51} & \ding{51} & \ding{51} & \ding{51} & \ding{51}\\
inr\_pt & \makecell[l]{prothrombin time/international\\normalized ratio} & \ding{51} & \ding{51} & \ding{51} & \ding{51} & \ding{55}\\
k & potassium & \ding{51} & \ding{51} & \ding{51} & \ding{51} & \ding{51}\\
lact & lactate & \ding{51} & \ding{51} & \ding{51} & \ding{51} & \ding{51}\\
lymph & lymphocytes & \ding{51} & \ding{51} & \ding{51} & \ding{51} & \ding{55}\\
\addlinespace
map & mean arterial pressure & \ding{51} & \ding{51} & \ding{51} & \ding{51} & \ding{51}\\
mch & mean cell hemoglobin & \ding{51} & \ding{51} & \ding{51} & \ding{51} & \ding{55}\\
mchc & \makecell[l]{mean corpuscular hemoglobin\\concentration} & \ding{51} & \ding{51} & \ding{51} & \ding{51} & \ding{55}\\
mcv & mean corpuscular volume & \ding{51} & \ding{51} & \ding{51} & \ding{51} & \ding{55}\\
methb & methemoglobin & \ding{51} & \ding{51} & \ding{51} & \ding{51} & \ding{55}\\
\addlinespace
mg & magnesium & \ding{51} & \ding{51} & \ding{51} & \ding{51} & \ding{51}\\
na & sodium & \ding{51} & \ding{51} & \ding{51} & \ding{51} & \ding{55}\\
neut & neutrophils & \ding{51} & \ding{51} & \ding{51} & \ding{51} & \ding{55}\\
o2sat & oxygen saturation & \ding{51} & \ding{51} & \ding{51} & \ding{51} & \ding{51}\\
pco2 & CO2 partial pressure & \ding{51} & \ding{51} & \ding{51} & \ding{51} & \ding{51}\\
\addlinespace
ph & pH of blood & \ding{51} & \ding{51} & \ding{51} & \ding{51} & \ding{51}\\
phos & phosphate & \ding{51} & \ding{51} & \ding{51} & \ding{51} & \ding{51}\\
plt & platelet count & \ding{51} & \ding{51} & \ding{51} & \ding{51} & \ding{51}\\
po2 & O2 partial pressure & \ding{51} & \ding{51} & \ding{51} & \ding{51} & \ding{55}\\
pt & prothrombine time & \ding{51} & \ding{51} & \ding{55} & \ding{51} & \ding{55}\\
\addlinespace
ptt & partial thromboplastin time & \ding{51} & \ding{51} & \ding{51} & \ding{51} & \ding{51}\\
rbc & red blood cell count & \ding{51} & \ding{51} & \ding{55} & \ding{51} & \ding{55}\\
rdw & erythrocyte distribution width & \ding{51} & \ding{51} & \ding{55} & \ding{51} & \ding{55}\\
resp & respiratory rate & \ding{51} & \ding{51} & \ding{51} & \ding{51} & \ding{51}\\
sbp & systolic blood pressure & \ding{51} & \ding{51} & \ding{51} & \ding{51} & \ding{51}\\
\addlinespace
sex & patient sex & \ding{51} & \ding{51} & \ding{51} & \ding{51} & \ding{51}\\
tco2 & totcal CO2 & \ding{51} & \ding{51} & \ding{55} & \ding{55} & \ding{55}\\
temp & temperature & \ding{51} & \ding{51} & \ding{51} & \ding{51} & \ding{51}\\
tnt & troponin t & \ding{51} & \ding{51} & \ding{51} & \ding{51} & \ding{55}\\
tri & troponin I & \ding{51} & \ding{51} & \ding{55} & \ding{55} & \ding{51}\\
\addlinespace
urine & urine output & \ding{51} & \ding{51} & \ding{51} & \ding{51} & \ding{55}\\
wbc & white blood cell count & \ding{51} & \ding{51} & \ding{51} & \ding{51} & \ding{51}\\
weight & patient weight & \ding{51} & \ding{51} & \ding{51} & \ding{51} & \ding{55}\\*
\end{longtable}

\begin{longtable}[t]{llllllll}
\caption{\label{tab:baselines}Variables used for baseline scores}\\
\toprule
 &  & Name & Description & MI-III & eICU & HiRID & AUMC\\
\midrule
\endfirsthead
\caption[]{Variables used for baseline scores \textit{(continued)}}\\
\toprule
 &  & Name & Description & MI-III & eICU & HiRID & AUMC\\
\midrule
\endhead

\endfoot
\bottomrule
\endlastfoot
\addlinespace[0.3em]
\multicolumn{8}{l}{\textbf{Modified early warning score}}\\
\hspace{1em} & mews & sbp & systolic blood pressure & \ding{51} & \ding{51} & \ding{51} & \ding{51}\\
\cmidrule{3-8}\nopagebreak
\hspace{1em}\hspace{1em}\hspace{1em} &  & hr & heart rate & \ding{51} & \ding{51} & \ding{51} & \ding{51}\\
\cmidrule{3-8}\nopagebreak
\hspace{1em}\hspace{1em}\hspace{1em} &  & resp & respiratory rate & \ding{51} & \ding{51} & \ding{51} & \ding{51}\\
\cmidrule{3-8}\nopagebreak
\hspace{1em}\hspace{1em} &  & temp & temperature & \ding{51} & \ding{51} & \ding{51} & \ding{51}\\
\cmidrule{3-8}\nopagebreak
\hspace{1em}\hspace{1em} &  & egcs & GCS eye & \ding{51} & \ding{51} & \ding{51} & \ding{51}\\
\cmidrule{3-8}\nopagebreak
\hspace{1em}\hspace{1em}\hspace{1em}\hspace{1em} &  & mgcs & GCS motor & \ding{51} & \ding{51} & \ding{51} & \ding{51}\\
\cmidrule{3-8}\nopagebreak
\hspace{1em}\hspace{1em}\hspace{1em}\hspace{1em} &  & vgcs & GCS verbal & \ding{51} & \ding{51} & \ding{51} & \ding{51}\\
\cmidrule{3-8}\nopagebreak
\hspace{1em}\hspace{1em}\hspace{1em}\hspace{1em} &  & tgcs & GCS total & \ding{51} & \ding{51} & \ding{55} & \ding{55}\\
\cmidrule{3-8}\nopagebreak
\hspace{1em}\hspace{1em}\hspace{1em}\hspace{1em} &  & trach & tracheostomy & \ding{51} & \ding{55} & \ding{51} & \ding{51}\\
\cmidrule{3-8}\nopagebreak
\hspace{1em}\hspace{1em}\hspace{1em}\hspace{1em} &  & rass & \makecell[l]{Richmond agitation\\sedation scale} & \ding{51} & \ding{51} & \ding{51} & \ding{51}\\
\cmidrule{1-8}\pagebreak[0]
\addlinespace[0.3em]
\multicolumn{8}{l}{\textbf{National early warning score}}\\
\hspace{1em} & news & resp & respiratory rate & \ding{51} & \ding{51} & \ding{51} & \ding{51}\\
\cmidrule{3-8}\nopagebreak
\hspace{1em} &  & o2sat & oxygen saturation & \ding{51} & \ding{51} & \ding{51} & \ding{51}\\
\cmidrule{3-8}\nopagebreak
\hspace{1em}\hspace{1em} &  & vent\_start & ventilation start & \ding{51} & \ding{51} & \ding{51} & \ding{51}\\
\cmidrule{3-8}\nopagebreak
\hspace{1em}\hspace{1em} &  & vent\_end & ventilation end & \ding{51} & \ding{51} & \ding{51} & \ding{51}\\
\cmidrule{3-8}\nopagebreak
\hspace{1em}\hspace{1em} &  & fio2 & \makecell[l]{fraction of inspired\\oxygen} & \ding{51} & \ding{51} & \ding{51} & \ding{51}\\
\cmidrule{3-8}\nopagebreak
 &  & temp & temperature & \ding{51} & \ding{51} & \ding{51} & \ding{51}\\
\cmidrule{3-8}\nopagebreak
\hspace{1em}\hspace{1em} &  & sbp & systolic blood pressure & \ding{51} & \ding{51} & \ding{51} & \ding{51}\\
\cmidrule{3-8}\nopagebreak
 &  & hr & heart rate & \ding{51} & \ding{51} & \ding{51} & \ding{51}\\
\cmidrule{3-8}\nopagebreak
 &  & egcs & GCS eye & \ding{51} & \ding{51} & \ding{51} & \ding{51}\\
\cmidrule{3-8}\nopagebreak
 &  & mgcs & GCS motor & \ding{51} & \ding{51} & \ding{51} & \ding{51}\\
\cmidrule{3-8}\nopagebreak
 &  & vgcs & GCS verbal & \ding{51} & \ding{51} & \ding{51} & \ding{51}\\
\cmidrule{3-8}\nopagebreak
 &  & tgcs & GCS total & \ding{51} & \ding{51} & \ding{55} & \ding{55}\\
\cmidrule{3-8}\nopagebreak
 &  & trach & tracheostomy & \ding{51} & \ding{55} & \ding{51} & \ding{51}\\
\cmidrule{3-8}\nopagebreak
 &  & rass & \makecell[l]{Richmond agitation\\sedation scale} & \ding{51} & \ding{51} & \ding{51} & \ding{51}\\
\cmidrule{1-8}\pagebreak[0]
\addlinespace[0.3em]
\multicolumn{8}{l}{\textbf{Quick SOFA score}}\\
\hspace{1em} & qsofa & egcs & GCS eye & \ding{51} & \ding{51} & \ding{51} & \ding{51}\\
\cmidrule{3-8}\nopagebreak
 &  & mgcs & GCS motor & \ding{51} & \ding{51} & \ding{51} & \ding{51}\\
\cmidrule{3-8}\nopagebreak
 &  & vgcs & GCS verbal & \ding{51} & \ding{51} & \ding{51} & \ding{51}\\
\cmidrule{3-8}\nopagebreak
 &  & tgcs & GCS total & \ding{51} & \ding{51} & \ding{55} & \ding{55}\\
\cmidrule{3-8}\nopagebreak
 &  & trach & tracheostomy & \ding{51} & \ding{55} & \ding{51} & \ding{51}\\
\cmidrule{3-8}\nopagebreak
 &  & rass & \makecell[l]{Richmond agitation\\sedation scale} & \ding{51} & \ding{51} & \ding{51} & \ding{51}\\
\cmidrule{3-8}\nopagebreak
 &  & sbp & systolic blood pressure & \ding{51} & \ding{51} & \ding{51} & \ding{51}\\
\cmidrule{3-8}\nopagebreak
 &  & resp & respiratory rate & \ding{51} & \ding{51} & \ding{51} & \ding{51}\\
\cmidrule{1-8}\pagebreak[0]
\addlinespace[0.3em]
\multicolumn{8}{l}{\textbf{Sequential organ failure assessment score}}\\
\hspace{1em} & sofa & sresp & \makecell[l]{SOFA respiratory\\component} & - & - & - & -\\
\cmidrule{3-8}\nopagebreak
\hspace{1em} &  & scoag & \makecell[l]{SOFA coagulation\\component} & - & - & - & -\\
\cmidrule{3-8}\nopagebreak
\hspace{1em} &  & sliver & SOFA liver component & - & - & - & -\\
\cmidrule{3-8}\nopagebreak
\hspace{1em} &  & scardio & \makecell[l]{SOFA cardiovascular\\component} & - & - & - & -\\
\cmidrule{3-8}\nopagebreak
\hspace{1em} &  & scns & \makecell[l]{SOFA central nervous\\system component} & - & - & - & -\\
\cmidrule{3-8}\nopagebreak
\hspace{1em} &  & srenal & SOFA renal component & - & - & - & -\\
\cmidrule{1-8}\pagebreak[0]
\addlinespace[0.3em]
\multicolumn{8}{l}{\textbf{SOFA components}}\\
\hspace{1em} & sresp & po2 & O2 partial pressure & \ding{51} & \ding{51} & \ding{51} & \ding{51}\\
\cmidrule{3-8}\nopagebreak
 &  & fio2 & \makecell[l]{fraction of inspired\\oxygen} & \ding{51} & \ding{51} & \ding{51} & \ding{51}\\
\cmidrule{3-8}\nopagebreak
 &  & vent\_start & ventilation start & \ding{51} & \ding{51} & \ding{51} & \ding{51}\\
\cmidrule{3-8}\nopagebreak
 &  & vent\_end & ventilation end & \ding{51} & \ding{51} & \ding{51} & \ding{51}\\
\cmidrule{2-8}\nopagebreak
\hspace{1em} & scoag & plt & platelet count & \ding{51} & \ding{51} & \ding{51} & \ding{51}\\
\cmidrule{2-8}\nopagebreak
\hspace{1em} & sliver & bili & total bilirubin & \ding{51} & \ding{51} & \ding{51} & \ding{51}\\
\cmidrule{2-8}\nopagebreak
\hspace{1em} & scardio & map & mean arterial pressure & \ding{51} & \ding{51} & \ding{51} & \ding{51}\\
\cmidrule{3-8}\nopagebreak
\hspace{1em} &  & dopa\_rate & dopamine rate & \ding{51} & \ding{51} & \ding{55} & \ding{51}\\
\cmidrule{3-8}\nopagebreak
\hspace{1em} &  & dopa\_dur & dopamine duration & \ding{51} & \ding{51} & \ding{55} & \ding{51}\\
\cmidrule{3-8}\nopagebreak
\hspace{1em} &  & norepi\_rate & norepinephrine rate & \ding{51} & \ding{51} & \ding{51} & \ding{51}\\
\cmidrule{3-8}\nopagebreak
\hspace{1em} &  & norepi\_dur & norepinephrine duration & \ding{51} & \ding{51} & \ding{51} & \ding{51}\\
\cmidrule{3-8}\nopagebreak
\hspace{1em} &  & dobu\_rate & dobutamine rate & \ding{51} & \ding{51} & \ding{51} & \ding{51}\\
\cmidrule{3-8}\nopagebreak
\hspace{1em} &  & dobu\_dur & dobutamine duration & \ding{51} & \ding{51} & \ding{51} & \ding{51}\\
\cmidrule{3-8}\nopagebreak
\hspace{1em} &  & epi\_rate & epinephrine rate & \ding{51} & \ding{51} & \ding{51} & \ding{51}\\
\cmidrule{3-8}\nopagebreak
\hspace{1em} &  & epi\_dur & epinephrine duration & \ding{51} & \ding{51} & \ding{51} & \ding{51}\\
\cmidrule{2-8}\nopagebreak
\hspace{1em} & scns & egcs & GCS eye & \ding{51} & \ding{51} & \ding{51} & \ding{51}\\
\cmidrule{3-8}\nopagebreak
 &  & mgcs & GCS motor & \ding{51} & \ding{51} & \ding{51} & \ding{51}\\
\cmidrule{3-8}\nopagebreak
 &  & vgcs & GCS verbal & \ding{51} & \ding{51} & \ding{51} & \ding{51}\\
\cmidrule{3-8}\nopagebreak
 &  & tgcs & GCS total & \ding{51} & \ding{51} & \ding{55} & \ding{55}\\
\cmidrule{3-8}\nopagebreak
 &  & trach & tracheostomy & \ding{51} & \ding{55} & \ding{51} & \ding{51}\\
\cmidrule{3-8}\nopagebreak
 &  & rass & \makecell[l]{Richmond agitation\\sedation scale} & \ding{51} & \ding{51} & \ding{51} & \ding{51}\\
\cmidrule{2-8}\nopagebreak
\hspace{1em} & srenal & crea & creatinine & \ding{51} & \ding{51} & \ding{51} & \ding{51}\\
\cmidrule{3-8}\nopagebreak
\hspace{1em} &  & urine & urine output & \ding{51} & \ding{51} & \ding{51} & \ding{51}\\
\cmidrule{1-8}\pagebreak[0]
\addlinespace[0.3em]
\multicolumn{8}{l}{\textbf{Systemic inflammatory response syndrome score}}\\
\hspace{1em} & sirs & temp & temperature & \ding{51} & \ding{51} & \ding{51} & \ding{51}\\
\cmidrule{3-8}\nopagebreak
 &  & hr & heart rate & \ding{51} & \ding{51} & \ding{51} & \ding{51}\\
\cmidrule{3-8}\nopagebreak
 &  & resp & respiratory rate & \ding{51} & \ding{51} & \ding{51} & \ding{51}\\
\cmidrule{3-8}\nopagebreak
\hspace{1em} &  & pco2 & CO2 partial pressure & \ding{51} & \ding{51} & \ding{51} & \ding{51}\\
\cmidrule{3-8}\nopagebreak
\hspace{1em} &  & wbc & white blood cell count & \ding{51} & \ding{51} & \ding{51} & \ding{51}\\
\cmidrule{3-8}\nopagebreak
\hspace{1em} &  & bnd & band form neutrophils & \ding{51} & \ding{51} & \ding{51} & \ding{51}\\*
\end{longtable}

\begin{table}
\begin{tabular}{|m{4cm}|m{10cm}|}
    \hline
    Component & Concepts used \\\hline
    SOFA Respiratory & FiO$_2$, PaO$_2$, mechanical ventilation \\\hline
    SOFA Renal & creatinine, urine output \\\hline
    SOFA Cardio & MAP, norepinephrine, epinephrine, dopamine, dobutamine \\\hline
    SOFA Liver & bilirubin \\\hline
    SOFA Coagulation & platelets \\\hline
    SOFA CNS & Glasgow Coma Scale, sedation, RASS scale\\\hline
\end{tabular}
    \caption{Concepts utilized for computing the SOFA score.}
    \label{tab:sofa}
\end{table}

\begin{table}
\begin{tabular}{ll}
    \toprule
    Hyperparameter & Coarse search values \\\midrule
    Depth &  gru: $1$, $2$, $3$, attn: $2$\\
    Width & $32$, $64$, $128$, $256$\\
    Learning rate & log uniformly in range $e^{-9}$ -- $e^{-7}$ \\
    Dropout & $0.3$, $0.4$, $0.5$, $0.6$, $0.7$\\
    Weight decay &  $0.0001$, $0.001$,  $0.01$, $0.1$\\\bottomrule
\end{tabular}
    \caption{Coarse hyperparameter ranges used for the first step of the hyperparameter search of the recurrent neural network employing Gated Recurrent Units~(gru) and the self-attention model (attn).}
    \label{tab:deep_hparams}
\end{table}

\FloatBarrier

\subsection{Additional Validation Results}\label{sec:additional_results}

This section contains additional figures depicting validation results. Please refer to Supplementary Figures~\ref{fig:ex_aumc}--\ref{fig:ex_emory} for
ROC curves of validation scenarios in which we train on a specific dataset~(e.g., eICU) and test on another~(e.g., HiRID). Supplementary Figures~\ref{fig:ex_scatter_aumc}--\ref{fig:ex_scatter_physionet2019} contain the corresponding scatter plots, depicting median earliness~(in hours before onset) against precision~(at 80\% recall) in order to provide a more detailed description of model performance.  

Supplementary Figure~\ref{fig:emory_roc} depicts an internal validation on the Emory dataset, which comprises a smaller variable set. Notice that this dataset did not permit the extraction of clinical baseline scores, which is why they are \emph{not} depicted in the respective subfigures.

Last, Supplementary Figure~\ref{fig:heat_extended} depicts a heatmap of AUROC values for comparing the transferability of models from one dataset to another. Since this figure includes the Emory dataset~(with a smaller variable set), all models were trained on the smaller variable set from this database to ensure consistent comparison.

\subsection{Used variables}
\autoref{tab:variables} enumerates the set of variables used for the prediction task, while indicating availability per data source. The clinical baseline scores used throughout are constructed from the respective variable sets shown in \autoref{tab:baselines}, again displaying availability over data sources.

\subsection{Feature selection}
In \autoref{fig:explore}, we display how to the currently included feature set was determined.
Besides the included features, we also investigated further features, including wavelet scatterings and signature transforms and observed no relevant gain in performance.

\subsection{Suspected infection cohorts}
In \autoref{fig:siaumcnonsurg} the two alternative definitions of our suspicion of infection cohort are displayed for the two datasets (MIMIC-III and AUMC) where the data availability allowed the comparison.

\subsection{Dataset differences}
The empirical results presented in the manuscript show that external validation of models is an important consideration for automated prediction models. It is tempting to try to explain exactly why a trained model suffers in performance, that is to try to quantify distributional differences across datasets. This section is intended to highlight some of the possible reasons that might constitute the observed distributional shift.
\begin{enumerate}[A)]
    \item \textit{Admission composition:} Table \ref{tab:datasets} shows that the admission type varies substantially between datasets, with the eICU dataset having 19\% of surgical admissions, compared to 80\% of surgical admission in the AUMC dataset,
    \item \textit{Definition of suspected infection: } We discussed in Section \ref{sec:sep3} that suspected infection is an integral component of the Sepsis-3 definition; however, due to missingness of body-fluid sampling information, eICU and HiRID datasets have an alternative suspected infection definition that uses multiple antibiotics administrations,
    \item \textit{Data availability before ICU admission:} AUMC and HiRID datasets have very few data points before ICU admission, whereas MIMIC-III and eICU datasets have both vitals and laboratory parameters measured during hospital stay; such a difference in data availability might have an effect on distribution of Sepsis-3 onsets and the ability to predict them.
\end{enumerate}
This short discussion shows that dataset differences are multifaceted and therefore explaining domain shift can be challenging in such a setting.

\end{document}

%% file: figures/shapley/shapley_16h_raw_bar.pgf
\begingroup%
\makeatletter%
\begin{pgfpicture}%
\pgfpathrectangle{\pgfpointorigin}{\pgfqpoint{6.000000in}{9.000000in}}%
\pgfusepath{use as bounding box, clip}%
\begin{pgfscope}%
\pgfsetbuttcap%
\pgfsetmiterjoin%
\definecolor{currentfill}{rgb}{1.000000,1.000000,1.000000}%
\pgfsetfillcolor{currentfill}%
\pgfsetlinewidth{0.000000pt}%
\definecolor{currentstroke}{rgb}{1.000000,1.000000,1.000000}%
\pgfsetstrokecolor{currentstroke}%
\pgfsetdash{}{0pt}%
\pgfpathmoveto{\pgfqpoint{0.000000in}{0.000000in}}%
\pgfpathlineto{\pgfqpoint{6.000000in}{0.000000in}}%
\pgfpathlineto{\pgfqpoint{6.000000in}{9.000000in}}%
\pgfpathlineto{\pgfqpoint{0.000000in}{9.000000in}}%
\pgfpathclose%
\pgfusepath{fill}%
\end{pgfscope}%
\begin{pgfscope}%
\pgfsetbuttcap%
\pgfsetmiterjoin%
\definecolor{currentfill}{rgb}{1.000000,1.000000,1.000000}%
\pgfsetfillcolor{currentfill}%
\pgfsetlinewidth{0.000000pt}%
\definecolor{currentstroke}{rgb}{0.000000,0.000000,0.000000}%
\pgfsetstrokecolor{currentstroke}%
\pgfsetstrokeopacity{0.000000}%
\pgfsetdash{}{0pt}%
\pgfpathmoveto{\pgfqpoint{2.768744in}{1.536516in}}%
\pgfpathlineto{\pgfqpoint{5.775000in}{1.536516in}}%
\pgfpathlineto{\pgfqpoint{5.775000in}{8.550000in}}%
\pgfpathlineto{\pgfqpoint{2.768744in}{8.550000in}}%
\pgfpathclose%
\pgfusepath{fill}%
\end{pgfscope}%
\begin{pgfscope}%
\pgfpathrectangle{\pgfqpoint{2.768744in}{1.536516in}}{\pgfqpoint{3.006256in}{7.013484in}}%
\pgfusepath{clip}%
\pgfsetbuttcap%
\pgfsetmiterjoin%
\definecolor{currentfill}{rgb}{0.000000,0.543378,0.983379}%
\pgfsetfillcolor{currentfill}%
\pgfsetlinewidth{1.003750pt}%
\definecolor{currentstroke}{rgb}{1.000000,1.000000,1.000000}%
\pgfsetstrokecolor{currentstroke}%
\pgfsetstrokeopacity{0.800000}%
\pgfsetdash{}{0pt}%
\pgfpathmoveto{\pgfqpoint{2.768744in}{8.231205in}}%
\pgfpathlineto{\pgfqpoint{4.682266in}{8.231205in}}%
\pgfpathlineto{\pgfqpoint{4.682266in}{7.973593in}}%
\pgfpathlineto{\pgfqpoint{2.768744in}{7.973593in}}%
\pgfpathclose%
\pgfusepath{stroke,fill}%
\end{pgfscope}%
\begin{pgfscope}%
\pgfpathrectangle{\pgfqpoint{2.768744in}{1.536516in}}{\pgfqpoint{3.006256in}{7.013484in}}%
\pgfusepath{clip}%
\pgfsetbuttcap%
\pgfsetmiterjoin%
\definecolor{currentfill}{rgb}{0.000000,0.543378,0.983379}%
\pgfsetfillcolor{currentfill}%
\pgfsetlinewidth{1.003750pt}%
\definecolor{currentstroke}{rgb}{1.000000,1.000000,1.000000}%
\pgfsetstrokecolor{currentstroke}%
\pgfsetstrokeopacity{0.800000}%
\pgfsetdash{}{0pt}%
\pgfpathmoveto{\pgfqpoint{2.768744in}{7.909190in}}%
\pgfpathlineto{\pgfqpoint{4.329178in}{7.909190in}}%
\pgfpathlineto{\pgfqpoint{4.329178in}{7.651578in}}%
\pgfpathlineto{\pgfqpoint{2.768744in}{7.651578in}}%
\pgfpathclose%
\pgfusepath{stroke,fill}%
\end{pgfscope}%
\begin{pgfscope}%
\pgfpathrectangle{\pgfqpoint{2.768744in}{1.536516in}}{\pgfqpoint{3.006256in}{7.013484in}}%
\pgfusepath{clip}%
\pgfsetbuttcap%
\pgfsetmiterjoin%
\definecolor{currentfill}{rgb}{0.000000,0.543378,0.983379}%
\pgfsetfillcolor{currentfill}%
\pgfsetlinewidth{1.003750pt}%
\definecolor{currentstroke}{rgb}{1.000000,1.000000,1.000000}%
\pgfsetstrokecolor{currentstroke}%
\pgfsetstrokeopacity{0.800000}%
\pgfsetdash{}{0pt}%
\pgfpathmoveto{\pgfqpoint{2.768744in}{7.587175in}}%
\pgfpathlineto{\pgfqpoint{3.998960in}{7.587175in}}%
\pgfpathlineto{\pgfqpoint{3.998960in}{7.329564in}}%
\pgfpathlineto{\pgfqpoint{2.768744in}{7.329564in}}%
\pgfpathclose%
\pgfusepath{stroke,fill}%
\end{pgfscope}%
\begin{pgfscope}%
\pgfpathrectangle{\pgfqpoint{2.768744in}{1.536516in}}{\pgfqpoint{3.006256in}{7.013484in}}%
\pgfusepath{clip}%
\pgfsetbuttcap%
\pgfsetmiterjoin%
\definecolor{currentfill}{rgb}{0.000000,0.543378,0.983379}%
\pgfsetfillcolor{currentfill}%
\pgfsetlinewidth{1.003750pt}%
\definecolor{currentstroke}{rgb}{1.000000,1.000000,1.000000}%
\pgfsetstrokecolor{currentstroke}%
\pgfsetstrokeopacity{0.800000}%
\pgfsetdash{}{0pt}%
\pgfpathmoveto{\pgfqpoint{2.768744in}{7.265161in}}%
\pgfpathlineto{\pgfqpoint{3.616817in}{7.265161in}}%
\pgfpathlineto{\pgfqpoint{3.616817in}{7.007549in}}%
\pgfpathlineto{\pgfqpoint{2.768744in}{7.007549in}}%
\pgfpathclose%
\pgfusepath{stroke,fill}%
\end{pgfscope}%
\begin{pgfscope}%
\pgfpathrectangle{\pgfqpoint{2.768744in}{1.536516in}}{\pgfqpoint{3.006256in}{7.013484in}}%
\pgfusepath{clip}%
\pgfsetbuttcap%
\pgfsetmiterjoin%
\definecolor{currentfill}{rgb}{0.000000,0.543378,0.983379}%
\pgfsetfillcolor{currentfill}%
\pgfsetlinewidth{1.003750pt}%
\definecolor{currentstroke}{rgb}{1.000000,1.000000,1.000000}%
\pgfsetstrokecolor{currentstroke}%
\pgfsetstrokeopacity{0.800000}%
\pgfsetdash{}{0pt}%
\pgfpathmoveto{\pgfqpoint{2.768744in}{6.943146in}}%
\pgfpathlineto{\pgfqpoint{3.616744in}{6.943146in}}%
\pgfpathlineto{\pgfqpoint{3.616744in}{6.685534in}}%
\pgfpathlineto{\pgfqpoint{2.768744in}{6.685534in}}%
\pgfpathclose%
\pgfusepath{stroke,fill}%
\end{pgfscope}%
\begin{pgfscope}%
\pgfpathrectangle{\pgfqpoint{2.768744in}{1.536516in}}{\pgfqpoint{3.006256in}{7.013484in}}%
\pgfusepath{clip}%
\pgfsetbuttcap%
\pgfsetmiterjoin%
\definecolor{currentfill}{rgb}{0.000000,0.543378,0.983379}%
\pgfsetfillcolor{currentfill}%
\pgfsetlinewidth{1.003750pt}%
\definecolor{currentstroke}{rgb}{1.000000,1.000000,1.000000}%
\pgfsetstrokecolor{currentstroke}%
\pgfsetstrokeopacity{0.800000}%
\pgfsetdash{}{0pt}%
\pgfpathmoveto{\pgfqpoint{2.768744in}{6.621131in}}%
\pgfpathlineto{\pgfqpoint{3.593693in}{6.621131in}}%
\pgfpathlineto{\pgfqpoint{3.593693in}{6.363519in}}%
\pgfpathlineto{\pgfqpoint{2.768744in}{6.363519in}}%
\pgfpathclose%
\pgfusepath{stroke,fill}%
\end{pgfscope}%
\begin{pgfscope}%
\pgfpathrectangle{\pgfqpoint{2.768744in}{1.536516in}}{\pgfqpoint{3.006256in}{7.013484in}}%
\pgfusepath{clip}%
\pgfsetbuttcap%
\pgfsetmiterjoin%
\definecolor{currentfill}{rgb}{0.000000,0.543378,0.983379}%
\pgfsetfillcolor{currentfill}%
\pgfsetlinewidth{1.003750pt}%
\definecolor{currentstroke}{rgb}{1.000000,1.000000,1.000000}%
\pgfsetstrokecolor{currentstroke}%
\pgfsetstrokeopacity{0.800000}%
\pgfsetdash{}{0pt}%
\pgfpathmoveto{\pgfqpoint{2.768744in}{6.299116in}}%
\pgfpathlineto{\pgfqpoint{3.502805in}{6.299116in}}%
\pgfpathlineto{\pgfqpoint{3.502805in}{6.041504in}}%
\pgfpathlineto{\pgfqpoint{2.768744in}{6.041504in}}%
\pgfpathclose%
\pgfusepath{stroke,fill}%
\end{pgfscope}%
\begin{pgfscope}%
\pgfpathrectangle{\pgfqpoint{2.768744in}{1.536516in}}{\pgfqpoint{3.006256in}{7.013484in}}%
\pgfusepath{clip}%
\pgfsetbuttcap%
\pgfsetmiterjoin%
\definecolor{currentfill}{rgb}{0.000000,0.543378,0.983379}%
\pgfsetfillcolor{currentfill}%
\pgfsetlinewidth{1.003750pt}%
\definecolor{currentstroke}{rgb}{1.000000,1.000000,1.000000}%
\pgfsetstrokecolor{currentstroke}%
\pgfsetstrokeopacity{0.800000}%
\pgfsetdash{}{0pt}%
\pgfpathmoveto{\pgfqpoint{2.768744in}{5.977101in}}%
\pgfpathlineto{\pgfqpoint{3.502354in}{5.977101in}}%
\pgfpathlineto{\pgfqpoint{3.502354in}{5.719489in}}%
\pgfpathlineto{\pgfqpoint{2.768744in}{5.719489in}}%
\pgfpathclose%
\pgfusepath{stroke,fill}%
\end{pgfscope}%
\begin{pgfscope}%
\pgfpathrectangle{\pgfqpoint{2.768744in}{1.536516in}}{\pgfqpoint{3.006256in}{7.013484in}}%
\pgfusepath{clip}%
\pgfsetbuttcap%
\pgfsetmiterjoin%
\definecolor{currentfill}{rgb}{0.000000,0.543378,0.983379}%
\pgfsetfillcolor{currentfill}%
\pgfsetlinewidth{1.003750pt}%
\definecolor{currentstroke}{rgb}{1.000000,1.000000,1.000000}%
\pgfsetstrokecolor{currentstroke}%
\pgfsetstrokeopacity{0.800000}%
\pgfsetdash{}{0pt}%
\pgfpathmoveto{\pgfqpoint{2.768744in}{5.655086in}}%
\pgfpathlineto{\pgfqpoint{3.476718in}{5.655086in}}%
\pgfpathlineto{\pgfqpoint{3.476718in}{5.397474in}}%
\pgfpathlineto{\pgfqpoint{2.768744in}{5.397474in}}%
\pgfpathclose%
\pgfusepath{stroke,fill}%
\end{pgfscope}%
\begin{pgfscope}%
\pgfpathrectangle{\pgfqpoint{2.768744in}{1.536516in}}{\pgfqpoint{3.006256in}{7.013484in}}%
\pgfusepath{clip}%
\pgfsetbuttcap%
\pgfsetmiterjoin%
\definecolor{currentfill}{rgb}{0.000000,0.543378,0.983379}%
\pgfsetfillcolor{currentfill}%
\pgfsetlinewidth{1.003750pt}%
\definecolor{currentstroke}{rgb}{1.000000,1.000000,1.000000}%
\pgfsetstrokecolor{currentstroke}%
\pgfsetstrokeopacity{0.800000}%
\pgfsetdash{}{0pt}%
\pgfpathmoveto{\pgfqpoint{2.768744in}{5.333071in}}%
\pgfpathlineto{\pgfqpoint{3.431092in}{5.333071in}}%
\pgfpathlineto{\pgfqpoint{3.431092in}{5.075459in}}%
\pgfpathlineto{\pgfqpoint{2.768744in}{5.075459in}}%
\pgfpathclose%
\pgfusepath{stroke,fill}%
\end{pgfscope}%
\begin{pgfscope}%
\pgfpathrectangle{\pgfqpoint{2.768744in}{1.536516in}}{\pgfqpoint{3.006256in}{7.013484in}}%
\pgfusepath{clip}%
\pgfsetbuttcap%
\pgfsetmiterjoin%
\definecolor{currentfill}{rgb}{0.000000,0.543378,0.983379}%
\pgfsetfillcolor{currentfill}%
\pgfsetlinewidth{1.003750pt}%
\definecolor{currentstroke}{rgb}{1.000000,1.000000,1.000000}%
\pgfsetstrokecolor{currentstroke}%
\pgfsetstrokeopacity{0.800000}%
\pgfsetdash{}{0pt}%
\pgfpathmoveto{\pgfqpoint{2.768744in}{5.011056in}}%
\pgfpathlineto{\pgfqpoint{3.349868in}{5.011056in}}%
\pgfpathlineto{\pgfqpoint{3.349868in}{4.753444in}}%
\pgfpathlineto{\pgfqpoint{2.768744in}{4.753444in}}%
\pgfpathclose%
\pgfusepath{stroke,fill}%
\end{pgfscope}%
\begin{pgfscope}%
\pgfpathrectangle{\pgfqpoint{2.768744in}{1.536516in}}{\pgfqpoint{3.006256in}{7.013484in}}%
\pgfusepath{clip}%
\pgfsetbuttcap%
\pgfsetmiterjoin%
\definecolor{currentfill}{rgb}{0.000000,0.543378,0.983379}%
\pgfsetfillcolor{currentfill}%
\pgfsetlinewidth{1.003750pt}%
\definecolor{currentstroke}{rgb}{1.000000,1.000000,1.000000}%
\pgfsetstrokecolor{currentstroke}%
\pgfsetstrokeopacity{0.800000}%
\pgfsetdash{}{0pt}%
\pgfpathmoveto{\pgfqpoint{2.768744in}{4.689041in}}%
\pgfpathlineto{\pgfqpoint{3.334749in}{4.689041in}}%
\pgfpathlineto{\pgfqpoint{3.334749in}{4.431430in}}%
\pgfpathlineto{\pgfqpoint{2.768744in}{4.431430in}}%
\pgfpathclose%
\pgfusepath{stroke,fill}%
\end{pgfscope}%
\begin{pgfscope}%
\pgfpathrectangle{\pgfqpoint{2.768744in}{1.536516in}}{\pgfqpoint{3.006256in}{7.013484in}}%
\pgfusepath{clip}%
\pgfsetbuttcap%
\pgfsetmiterjoin%
\definecolor{currentfill}{rgb}{0.000000,0.543378,0.983379}%
\pgfsetfillcolor{currentfill}%
\pgfsetlinewidth{1.003750pt}%
\definecolor{currentstroke}{rgb}{1.000000,1.000000,1.000000}%
\pgfsetstrokecolor{currentstroke}%
\pgfsetstrokeopacity{0.800000}%
\pgfsetdash{}{0pt}%
\pgfpathmoveto{\pgfqpoint{2.768744in}{4.367027in}}%
\pgfpathlineto{\pgfqpoint{3.208861in}{4.367027in}}%
\pgfpathlineto{\pgfqpoint{3.208861in}{4.109415in}}%
\pgfpathlineto{\pgfqpoint{2.768744in}{4.109415in}}%
\pgfpathclose%
\pgfusepath{stroke,fill}%
\end{pgfscope}%
\begin{pgfscope}%
\pgfpathrectangle{\pgfqpoint{2.768744in}{1.536516in}}{\pgfqpoint{3.006256in}{7.013484in}}%
\pgfusepath{clip}%
\pgfsetbuttcap%
\pgfsetmiterjoin%
\definecolor{currentfill}{rgb}{0.000000,0.543378,0.983379}%
\pgfsetfillcolor{currentfill}%
\pgfsetlinewidth{1.003750pt}%
\definecolor{currentstroke}{rgb}{1.000000,1.000000,1.000000}%
\pgfsetstrokecolor{currentstroke}%
\pgfsetstrokeopacity{0.800000}%
\pgfsetdash{}{0pt}%
\pgfpathmoveto{\pgfqpoint{2.768744in}{4.045012in}}%
\pgfpathlineto{\pgfqpoint{3.203957in}{4.045012in}}%
\pgfpathlineto{\pgfqpoint{3.203957in}{3.787400in}}%
\pgfpathlineto{\pgfqpoint{2.768744in}{3.787400in}}%
\pgfpathclose%
\pgfusepath{stroke,fill}%
\end{pgfscope}%
\begin{pgfscope}%
\pgfpathrectangle{\pgfqpoint{2.768744in}{1.536516in}}{\pgfqpoint{3.006256in}{7.013484in}}%
\pgfusepath{clip}%
\pgfsetbuttcap%
\pgfsetmiterjoin%
\definecolor{currentfill}{rgb}{0.000000,0.543378,0.983379}%
\pgfsetfillcolor{currentfill}%
\pgfsetlinewidth{1.003750pt}%
\definecolor{currentstroke}{rgb}{1.000000,1.000000,1.000000}%
\pgfsetstrokecolor{currentstroke}%
\pgfsetstrokeopacity{0.800000}%
\pgfsetdash{}{0pt}%
\pgfpathmoveto{\pgfqpoint{2.768744in}{3.722997in}}%
\pgfpathlineto{\pgfqpoint{3.136858in}{3.722997in}}%
\pgfpathlineto{\pgfqpoint{3.136858in}{3.465385in}}%
\pgfpathlineto{\pgfqpoint{2.768744in}{3.465385in}}%
\pgfpathclose%
\pgfusepath{stroke,fill}%
\end{pgfscope}%
\begin{pgfscope}%
\pgfpathrectangle{\pgfqpoint{2.768744in}{1.536516in}}{\pgfqpoint{3.006256in}{7.013484in}}%
\pgfusepath{clip}%
\pgfsetbuttcap%
\pgfsetmiterjoin%
\definecolor{currentfill}{rgb}{0.000000,0.543378,0.983379}%
\pgfsetfillcolor{currentfill}%
\pgfsetlinewidth{1.003750pt}%
\definecolor{currentstroke}{rgb}{1.000000,1.000000,1.000000}%
\pgfsetstrokecolor{currentstroke}%
\pgfsetstrokeopacity{0.800000}%
\pgfsetdash{}{0pt}%
\pgfpathmoveto{\pgfqpoint{2.768744in}{3.400982in}}%
\pgfpathlineto{\pgfqpoint{3.128250in}{3.400982in}}%
\pgfpathlineto{\pgfqpoint{3.128250in}{3.143370in}}%
\pgfpathlineto{\pgfqpoint{2.768744in}{3.143370in}}%
\pgfpathclose%
\pgfusepath{stroke,fill}%
\end{pgfscope}%
\begin{pgfscope}%
\pgfpathrectangle{\pgfqpoint{2.768744in}{1.536516in}}{\pgfqpoint{3.006256in}{7.013484in}}%
\pgfusepath{clip}%
\pgfsetbuttcap%
\pgfsetmiterjoin%
\definecolor{currentfill}{rgb}{0.000000,0.543378,0.983379}%
\pgfsetfillcolor{currentfill}%
\pgfsetlinewidth{1.003750pt}%
\definecolor{currentstroke}{rgb}{1.000000,1.000000,1.000000}%
\pgfsetstrokecolor{currentstroke}%
\pgfsetstrokeopacity{0.800000}%
\pgfsetdash{}{0pt}%
\pgfpathmoveto{\pgfqpoint{2.768744in}{3.078967in}}%
\pgfpathlineto{\pgfqpoint{3.104436in}{3.078967in}}%
\pgfpathlineto{\pgfqpoint{3.104436in}{2.821355in}}%
\pgfpathlineto{\pgfqpoint{2.768744in}{2.821355in}}%
\pgfpathclose%
\pgfusepath{stroke,fill}%
\end{pgfscope}%
\begin{pgfscope}%
\pgfpathrectangle{\pgfqpoint{2.768744in}{1.536516in}}{\pgfqpoint{3.006256in}{7.013484in}}%
\pgfusepath{clip}%
\pgfsetbuttcap%
\pgfsetmiterjoin%
\definecolor{currentfill}{rgb}{0.000000,0.543378,0.983379}%
\pgfsetfillcolor{currentfill}%
\pgfsetlinewidth{1.003750pt}%
\definecolor{currentstroke}{rgb}{1.000000,1.000000,1.000000}%
\pgfsetstrokecolor{currentstroke}%
\pgfsetstrokeopacity{0.800000}%
\pgfsetdash{}{0pt}%
\pgfpathmoveto{\pgfqpoint{2.768744in}{2.756952in}}%
\pgfpathlineto{\pgfqpoint{3.094592in}{2.756952in}}%
\pgfpathlineto{\pgfqpoint{3.094592in}{2.499340in}}%
\pgfpathlineto{\pgfqpoint{2.768744in}{2.499340in}}%
\pgfpathclose%
\pgfusepath{stroke,fill}%
\end{pgfscope}%
\begin{pgfscope}%
\pgfpathrectangle{\pgfqpoint{2.768744in}{1.536516in}}{\pgfqpoint{3.006256in}{7.013484in}}%
\pgfusepath{clip}%
\pgfsetbuttcap%
\pgfsetmiterjoin%
\definecolor{currentfill}{rgb}{0.000000,0.543378,0.983379}%
\pgfsetfillcolor{currentfill}%
\pgfsetlinewidth{1.003750pt}%
\definecolor{currentstroke}{rgb}{1.000000,1.000000,1.000000}%
\pgfsetstrokecolor{currentstroke}%
\pgfsetstrokeopacity{0.800000}%
\pgfsetdash{}{0pt}%
\pgfpathmoveto{\pgfqpoint{2.768744in}{2.434937in}}%
\pgfpathlineto{\pgfqpoint{3.088875in}{2.434937in}}%
\pgfpathlineto{\pgfqpoint{3.088875in}{2.177325in}}%
\pgfpathlineto{\pgfqpoint{2.768744in}{2.177325in}}%
\pgfpathclose%
\pgfusepath{stroke,fill}%
\end{pgfscope}%
\begin{pgfscope}%
\pgfpathrectangle{\pgfqpoint{2.768744in}{1.536516in}}{\pgfqpoint{3.006256in}{7.013484in}}%
\pgfusepath{clip}%
\pgfsetbuttcap%
\pgfsetmiterjoin%
\definecolor{currentfill}{rgb}{0.000000,0.543378,0.983379}%
\pgfsetfillcolor{currentfill}%
\pgfsetlinewidth{1.003750pt}%
\definecolor{currentstroke}{rgb}{1.000000,1.000000,1.000000}%
\pgfsetstrokecolor{currentstroke}%
\pgfsetstrokeopacity{0.800000}%
\pgfsetdash{}{0pt}%
\pgfpathmoveto{\pgfqpoint{2.768744in}{2.112922in}}%
\pgfpathlineto{\pgfqpoint{3.082932in}{2.112922in}}%
\pgfpathlineto{\pgfqpoint{3.082932in}{1.855310in}}%
\pgfpathlineto{\pgfqpoint{2.768744in}{1.855310in}}%
\pgfpathclose%
\pgfusepath{stroke,fill}%
\end{pgfscope}%
\begin{pgfscope}%
\pgfsetbuttcap%
\pgfsetroundjoin%
\definecolor{currentfill}{rgb}{0.000000,0.000000,0.000000}%
\pgfsetfillcolor{currentfill}%
\pgfsetlinewidth{0.803000pt}%
\definecolor{currentstroke}{rgb}{0.000000,0.000000,0.000000}%
\pgfsetstrokecolor{currentstroke}%
\pgfsetdash{}{0pt}%
\pgfsys@defobject{currentmarker}{\pgfqpoint{0.000000in}{-0.048611in}}{\pgfqpoint{0.000000in}{0.000000in}}{%
\pgfpathmoveto{\pgfqpoint{0.000000in}{0.000000in}}%
\pgfpathlineto{\pgfqpoint{0.000000in}{-0.048611in}}%
\pgfusepath{stroke,fill}%
}%
\begin{pgfscope}%
\pgfsys@transformshift{2.768744in}{1.536516in}%
\pgfsys@useobject{currentmarker}{}%
\end{pgfscope}%
\end{pgfscope}%
\begin{pgfscope}%
\definecolor{textcolor}{rgb}{0.000000,0.000000,0.000000}%
\pgfsetstrokecolor{textcolor}%
\pgfsetfillcolor{textcolor}%
\pgftext[x=2.768744in,y=1.439293in,,top]{\color{textcolor}\sffamily\fontsize{16.000000}{19.200000}\selectfont \(\displaystyle {0.000}\)}%
\end{pgfscope}%
\begin{pgfscope}%
\pgfsetbuttcap%
\pgfsetroundjoin%
\definecolor{currentfill}{rgb}{0.000000,0.000000,0.000000}%
\pgfsetfillcolor{currentfill}%
\pgfsetlinewidth{0.803000pt}%
\definecolor{currentstroke}{rgb}{0.000000,0.000000,0.000000}%
\pgfsetstrokecolor{currentstroke}%
\pgfsetdash{}{0pt}%
\pgfsys@defobject{currentmarker}{\pgfqpoint{0.000000in}{-0.048611in}}{\pgfqpoint{0.000000in}{0.000000in}}{%
\pgfpathmoveto{\pgfqpoint{0.000000in}{0.000000in}}%
\pgfpathlineto{\pgfqpoint{0.000000in}{-0.048611in}}%
\pgfusepath{stroke,fill}%
}%
\begin{pgfscope}%
\pgfsys@transformshift{3.568869in}{1.536516in}%
\pgfsys@useobject{currentmarker}{}%
\end{pgfscope}%
\end{pgfscope}%
\begin{pgfscope}%
\definecolor{textcolor}{rgb}{0.000000,0.000000,0.000000}%
\pgfsetstrokecolor{textcolor}%
\pgfsetfillcolor{textcolor}%
\pgftext[x=3.568869in,y=1.439293in,,top]{\color{textcolor}\sffamily\fontsize{16.000000}{19.200000}\selectfont \(\displaystyle {0.002}\)}%
\end{pgfscope}%
\begin{pgfscope}%
\pgfsetbuttcap%
\pgfsetroundjoin%
\definecolor{currentfill}{rgb}{0.000000,0.000000,0.000000}%
\pgfsetfillcolor{currentfill}%
\pgfsetlinewidth{0.803000pt}%
\definecolor{currentstroke}{rgb}{0.000000,0.000000,0.000000}%
\pgfsetstrokecolor{currentstroke}%
\pgfsetdash{}{0pt}%
\pgfsys@defobject{currentmarker}{\pgfqpoint{0.000000in}{-0.048611in}}{\pgfqpoint{0.000000in}{0.000000in}}{%
\pgfpathmoveto{\pgfqpoint{0.000000in}{0.000000in}}%
\pgfpathlineto{\pgfqpoint{0.000000in}{-0.048611in}}%
\pgfusepath{stroke,fill}%
}%
\begin{pgfscope}%
\pgfsys@transformshift{4.368993in}{1.536516in}%
\pgfsys@useobject{currentmarker}{}%
\end{pgfscope}%
\end{pgfscope}%
\begin{pgfscope}%
\definecolor{textcolor}{rgb}{0.000000,0.000000,0.000000}%
\pgfsetstrokecolor{textcolor}%
\pgfsetfillcolor{textcolor}%
\pgftext[x=4.368993in,y=1.439293in,,top]{\color{textcolor}\sffamily\fontsize{16.000000}{19.200000}\selectfont \(\displaystyle {0.004}\)}%
\end{pgfscope}%
\begin{pgfscope}%
\pgfsetbuttcap%
\pgfsetroundjoin%
\definecolor{currentfill}{rgb}{0.000000,0.000000,0.000000}%
\pgfsetfillcolor{currentfill}%
\pgfsetlinewidth{0.803000pt}%
\definecolor{currentstroke}{rgb}{0.000000,0.000000,0.000000}%
\pgfsetstrokecolor{currentstroke}%
\pgfsetdash{}{0pt}%
\pgfsys@defobject{currentmarker}{\pgfqpoint{0.000000in}{-0.048611in}}{\pgfqpoint{0.000000in}{0.000000in}}{%
\pgfpathmoveto{\pgfqpoint{0.000000in}{0.000000in}}%
\pgfpathlineto{\pgfqpoint{0.000000in}{-0.048611in}}%
\pgfusepath{stroke,fill}%
}%
\begin{pgfscope}%
\pgfsys@transformshift{5.169118in}{1.536516in}%
\pgfsys@useobject{currentmarker}{}%
\end{pgfscope}%
\end{pgfscope}%
\begin{pgfscope}%
\definecolor{textcolor}{rgb}{0.000000,0.000000,0.000000}%
\pgfsetstrokecolor{textcolor}%
\pgfsetfillcolor{textcolor}%
\pgftext[x=5.169118in,y=1.439293in,,top]{\color{textcolor}\sffamily\fontsize{16.000000}{19.200000}\selectfont \(\displaystyle {0.006}\)}%
\end{pgfscope}%
\begin{pgfscope}%
\definecolor{textcolor}{rgb}{0.000000,0.000000,0.000000}%
\pgfsetstrokecolor{textcolor}%
\pgfsetfillcolor{textcolor}%
\pgftext[x=4.271872in,y=1.170389in,,top]{\color{textcolor}\sffamily\fontsize{16.000000}{19.200000}\selectfont Mean absolute Shapley value}%
\end{pgfscope}%
\begin{pgfscope}%
\definecolor{textcolor}{rgb}{0.000000,0.000000,0.000000}%
\pgfsetstrokecolor{textcolor}%
\pgfsetfillcolor{textcolor}%
\pgftext[x=0.518472in, y=8.019066in, left, base]{\color{textcolor}\sffamily\fontsize{16.000000}{19.200000}\selectfont Mean arterial pressure }%
\end{pgfscope}%
\begin{pgfscope}%
\definecolor{textcolor}{rgb}{0.000000,0.000000,0.000000}%
\pgfsetstrokecolor{textcolor}%
\pgfsetfillcolor{textcolor}%
\pgftext[x=1.634397in, y=7.697051in, left, base]{\color{textcolor}\sffamily\fontsize{16.000000}{19.200000}\selectfont Heart rate }%
\end{pgfscope}%
\begin{pgfscope}%
\definecolor{textcolor}{rgb}{0.000000,0.000000,0.000000}%
\pgfsetstrokecolor{textcolor}%
\pgfsetfillcolor{textcolor}%
\pgftext[x=0.356147in, y=7.375036in, left, base]{\color{textcolor}\sffamily\fontsize{16.000000}{19.200000}\selectfont Diastolic blood pressure }%
\end{pgfscope}%
\begin{pgfscope}%
\definecolor{textcolor}{rgb}{0.000000,0.000000,0.000000}%
\pgfsetstrokecolor{textcolor}%
\pgfsetfillcolor{textcolor}%
\pgftext[x=1.395587in, y=7.053021in, left, base]{\color{textcolor}\sffamily\fontsize{16.000000}{19.200000}\selectfont Urine output }%
\end{pgfscope}%
\begin{pgfscope}%
\definecolor{textcolor}{rgb}{0.000000,0.000000,0.000000}%
\pgfsetstrokecolor{textcolor}%
\pgfsetfillcolor{textcolor}%
\pgftext[x=0.876976in, y=6.731006in, left, base]{\color{textcolor}\sffamily\fontsize{16.000000}{19.200000}\selectfont Oxygen saturation }%
\end{pgfscope}%
\begin{pgfscope}%
\definecolor{textcolor}{rgb}{0.000000,0.000000,0.000000}%
\pgfsetstrokecolor{textcolor}%
\pgfsetfillcolor{textcolor}%
\pgftext[x=0.055417in, y=6.408991in, left, base]{\color{textcolor}\sffamily\fontsize{16.000000}{19.200000}\selectfont Fraction of inspired oxygen }%
\end{pgfscope}%
\begin{pgfscope}%
\definecolor{textcolor}{rgb}{0.000000,0.000000,0.000000}%
\pgfsetstrokecolor{textcolor}%
\pgfsetfillcolor{textcolor}%
\pgftext[x=0.451536in, y=6.086977in, left, base]{\color{textcolor}\sffamily\fontsize{16.000000}{19.200000}\selectfont Systolic blood pressure }%
\end{pgfscope}%
\begin{pgfscope}%
\definecolor{textcolor}{rgb}{0.000000,0.000000,0.000000}%
\pgfsetstrokecolor{textcolor}%
\pgfsetfillcolor{textcolor}%
\pgftext[x=0.926165in, y=5.764962in, left, base]{\color{textcolor}\sffamily\fontsize{16.000000}{19.200000}\selectfont C-reactive protein }%
\end{pgfscope}%
\begin{pgfscope}%
\definecolor{textcolor}{rgb}{0.000000,0.000000,0.000000}%
\pgfsetstrokecolor{textcolor}%
\pgfsetfillcolor{textcolor}%
\pgftext[x=1.098618in, y=5.442947in, left, base]{\color{textcolor}\sffamily\fontsize{16.000000}{19.200000}\selectfont Respiratory rate }%
\end{pgfscope}%
\begin{pgfscope}%
\definecolor{textcolor}{rgb}{0.000000,0.000000,0.000000}%
\pgfsetstrokecolor{textcolor}%
\pgfsetfillcolor{textcolor}%
\pgftext[x=0.821807in, y=5.120932in, left, base]{\color{textcolor}\sffamily\fontsize{16.000000}{19.200000}\selectfont O2 partial pressure }%
\end{pgfscope}%
\begin{pgfscope}%
\definecolor{textcolor}{rgb}{0.000000,0.000000,0.000000}%
\pgfsetstrokecolor{textcolor}%
\pgfsetfillcolor{textcolor}%
\pgftext[x=1.399638in, y=4.798917in, left, base]{\color{textcolor}\sffamily\fontsize{16.000000}{19.200000}\selectfont Temperature }%
\end{pgfscope}%
\begin{pgfscope}%
\definecolor{textcolor}{rgb}{0.000000,0.000000,0.000000}%
\pgfsetstrokecolor{textcolor}%
\pgfsetfillcolor{textcolor}%
\pgftext[x=1.480367in, y=4.476902in, left, base]{\color{textcolor}\sffamily\fontsize{16.000000}{19.200000}\selectfont Hemoglobin }%
\end{pgfscope}%
\begin{pgfscope}%
\definecolor{textcolor}{rgb}{0.000000,0.000000,0.000000}%
\pgfsetstrokecolor{textcolor}%
\pgfsetfillcolor{textcolor}%
\pgftext[x=1.887288in, y=4.154887in, left, base]{\color{textcolor}\sffamily\fontsize{16.000000}{19.200000}\selectfont Sodium }%
\end{pgfscope}%
\begin{pgfscope}%
\definecolor{textcolor}{rgb}{0.000000,0.000000,0.000000}%
\pgfsetstrokecolor{textcolor}%
\pgfsetfillcolor{textcolor}%
\pgftext[x=1.876293in, y=3.832872in, left, base]{\color{textcolor}\sffamily\fontsize{16.000000}{19.200000}\selectfont Glucose }%
\end{pgfscope}%
\begin{pgfscope}%
\definecolor{textcolor}{rgb}{0.000000,0.000000,0.000000}%
\pgfsetstrokecolor{textcolor}%
\pgfsetfillcolor{textcolor}%
\pgftext[x=1.472169in, y=3.510857in, left, base]{\color{textcolor}\sffamily\fontsize{16.000000}{19.200000}\selectfont Bicarbonate }%
\end{pgfscope}%
\begin{pgfscope}%
\definecolor{textcolor}{rgb}{0.000000,0.000000,0.000000}%
\pgfsetstrokecolor{textcolor}%
\pgfsetfillcolor{textcolor}%
\pgftext[x=1.894136in, y=3.188843in, left, base]{\color{textcolor}\sffamily\fontsize{16.000000}{19.200000}\selectfont Lactate }%
\end{pgfscope}%
\begin{pgfscope}%
\definecolor{textcolor}{rgb}{0.000000,0.000000,0.000000}%
\pgfsetstrokecolor{textcolor}%
\pgfsetfillcolor{textcolor}%
\pgftext[x=1.829226in, y=2.866828in, left, base]{\color{textcolor}\sffamily\fontsize{16.000000}{19.200000}\selectfont Chloride }%
\end{pgfscope}%
\begin{pgfscope}%
\definecolor{textcolor}{rgb}{0.000000,0.000000,0.000000}%
\pgfsetstrokecolor{textcolor}%
\pgfsetfillcolor{textcolor}%
\pgftext[x=1.116268in, y=2.544813in, left, base]{\color{textcolor}\sffamily\fontsize{16.000000}{19.200000}\selectfont Calcium ionized }%
\end{pgfscope}%
\begin{pgfscope}%
\definecolor{textcolor}{rgb}{0.000000,0.000000,0.000000}%
\pgfsetstrokecolor{textcolor}%
\pgfsetfillcolor{textcolor}%
\pgftext[x=1.634686in, y=2.222798in, left, base]{\color{textcolor}\sffamily\fontsize{16.000000}{19.200000}\selectfont Potassium }%
\end{pgfscope}%
\begin{pgfscope}%
\definecolor{textcolor}{rgb}{0.000000,0.000000,0.000000}%
\pgfsetstrokecolor{textcolor}%
\pgfsetfillcolor{textcolor}%
\pgftext[x=0.544031in, y=1.900783in, left, base]{\color{textcolor}\sffamily\fontsize{16.000000}{19.200000}\selectfont Mean cell hemoglobin }%
\end{pgfscope}%
\begin{pgfscope}%
\pgfpathrectangle{\pgfqpoint{2.768744in}{1.536516in}}{\pgfqpoint{3.006256in}{7.013484in}}%
\pgfusepath{clip}%
\pgfsetbuttcap%
\pgfsetroundjoin%
\pgfsetlinewidth{1.505625pt}%
\definecolor{currentstroke}{rgb}{0.000000,0.000000,0.000000}%
\pgfsetstrokecolor{currentstroke}%
\pgfsetdash{}{0pt}%
\pgfpathmoveto{\pgfqpoint{4.682266in}{8.102399in}}%
\pgfpathlineto{\pgfqpoint{5.631845in}{8.102399in}}%
\pgfusepath{stroke}%
\end{pgfscope}%
\begin{pgfscope}%
\pgfpathrectangle{\pgfqpoint{2.768744in}{1.536516in}}{\pgfqpoint{3.006256in}{7.013484in}}%
\pgfusepath{clip}%
\pgfsetbuttcap%
\pgfsetroundjoin%
\pgfsetlinewidth{1.505625pt}%
\definecolor{currentstroke}{rgb}{0.000000,0.000000,0.000000}%
\pgfsetstrokecolor{currentstroke}%
\pgfsetdash{}{0pt}%
\pgfpathmoveto{\pgfqpoint{4.329178in}{7.780384in}}%
\pgfpathlineto{\pgfqpoint{5.258041in}{7.780384in}}%
\pgfusepath{stroke}%
\end{pgfscope}%
\begin{pgfscope}%
\pgfpathrectangle{\pgfqpoint{2.768744in}{1.536516in}}{\pgfqpoint{3.006256in}{7.013484in}}%
\pgfusepath{clip}%
\pgfsetbuttcap%
\pgfsetroundjoin%
\pgfsetlinewidth{1.505625pt}%
\definecolor{currentstroke}{rgb}{0.000000,0.000000,0.000000}%
\pgfsetstrokecolor{currentstroke}%
\pgfsetdash{}{0pt}%
\pgfpathmoveto{\pgfqpoint{3.998960in}{7.458370in}}%
\pgfpathlineto{\pgfqpoint{4.918747in}{7.458370in}}%
\pgfusepath{stroke}%
\end{pgfscope}%
\begin{pgfscope}%
\pgfpathrectangle{\pgfqpoint{2.768744in}{1.536516in}}{\pgfqpoint{3.006256in}{7.013484in}}%
\pgfusepath{clip}%
\pgfsetbuttcap%
\pgfsetroundjoin%
\pgfsetlinewidth{1.505625pt}%
\definecolor{currentstroke}{rgb}{0.000000,0.000000,0.000000}%
\pgfsetstrokecolor{currentstroke}%
\pgfsetdash{}{0pt}%
\pgfpathmoveto{\pgfqpoint{3.616817in}{7.136355in}}%
\pgfpathlineto{\pgfqpoint{4.079936in}{7.136355in}}%
\pgfusepath{stroke}%
\end{pgfscope}%
\begin{pgfscope}%
\pgfpathrectangle{\pgfqpoint{2.768744in}{1.536516in}}{\pgfqpoint{3.006256in}{7.013484in}}%
\pgfusepath{clip}%
\pgfsetbuttcap%
\pgfsetroundjoin%
\pgfsetlinewidth{1.505625pt}%
\definecolor{currentstroke}{rgb}{0.000000,0.000000,0.000000}%
\pgfsetstrokecolor{currentstroke}%
\pgfsetdash{}{0pt}%
\pgfpathmoveto{\pgfqpoint{3.616744in}{6.814340in}}%
\pgfpathlineto{\pgfqpoint{4.188269in}{6.814340in}}%
\pgfusepath{stroke}%
\end{pgfscope}%
\begin{pgfscope}%
\pgfpathrectangle{\pgfqpoint{2.768744in}{1.536516in}}{\pgfqpoint{3.006256in}{7.013484in}}%
\pgfusepath{clip}%
\pgfsetbuttcap%
\pgfsetroundjoin%
\pgfsetlinewidth{1.505625pt}%
\definecolor{currentstroke}{rgb}{0.000000,0.000000,0.000000}%
\pgfsetstrokecolor{currentstroke}%
\pgfsetdash{}{0pt}%
\pgfpathmoveto{\pgfqpoint{3.593693in}{6.492325in}}%
\pgfpathlineto{\pgfqpoint{4.169491in}{6.492325in}}%
\pgfusepath{stroke}%
\end{pgfscope}%
\begin{pgfscope}%
\pgfpathrectangle{\pgfqpoint{2.768744in}{1.536516in}}{\pgfqpoint{3.006256in}{7.013484in}}%
\pgfusepath{clip}%
\pgfsetbuttcap%
\pgfsetroundjoin%
\pgfsetlinewidth{1.505625pt}%
\definecolor{currentstroke}{rgb}{0.000000,0.000000,0.000000}%
\pgfsetstrokecolor{currentstroke}%
\pgfsetdash{}{0pt}%
\pgfpathmoveto{\pgfqpoint{3.502805in}{6.170310in}}%
\pgfpathlineto{\pgfqpoint{4.042614in}{6.170310in}}%
\pgfusepath{stroke}%
\end{pgfscope}%
\begin{pgfscope}%
\pgfpathrectangle{\pgfqpoint{2.768744in}{1.536516in}}{\pgfqpoint{3.006256in}{7.013484in}}%
\pgfusepath{clip}%
\pgfsetbuttcap%
\pgfsetroundjoin%
\pgfsetlinewidth{1.505625pt}%
\definecolor{currentstroke}{rgb}{0.000000,0.000000,0.000000}%
\pgfsetstrokecolor{currentstroke}%
\pgfsetdash{}{0pt}%
\pgfpathmoveto{\pgfqpoint{3.502354in}{5.848295in}}%
\pgfpathlineto{\pgfqpoint{4.767788in}{5.848295in}}%
\pgfusepath{stroke}%
\end{pgfscope}%
\begin{pgfscope}%
\pgfpathrectangle{\pgfqpoint{2.768744in}{1.536516in}}{\pgfqpoint{3.006256in}{7.013484in}}%
\pgfusepath{clip}%
\pgfsetbuttcap%
\pgfsetroundjoin%
\pgfsetlinewidth{1.505625pt}%
\definecolor{currentstroke}{rgb}{0.000000,0.000000,0.000000}%
\pgfsetstrokecolor{currentstroke}%
\pgfsetdash{}{0pt}%
\pgfpathmoveto{\pgfqpoint{3.476718in}{5.526280in}}%
\pgfpathlineto{\pgfqpoint{3.807933in}{5.526280in}}%
\pgfusepath{stroke}%
\end{pgfscope}%
\begin{pgfscope}%
\pgfpathrectangle{\pgfqpoint{2.768744in}{1.536516in}}{\pgfqpoint{3.006256in}{7.013484in}}%
\pgfusepath{clip}%
\pgfsetbuttcap%
\pgfsetroundjoin%
\pgfsetlinewidth{1.505625pt}%
\definecolor{currentstroke}{rgb}{0.000000,0.000000,0.000000}%
\pgfsetstrokecolor{currentstroke}%
\pgfsetdash{}{0pt}%
\pgfpathmoveto{\pgfqpoint{3.431092in}{5.204265in}}%
\pgfpathlineto{\pgfqpoint{3.999153in}{5.204265in}}%
\pgfusepath{stroke}%
\end{pgfscope}%
\begin{pgfscope}%
\pgfpathrectangle{\pgfqpoint{2.768744in}{1.536516in}}{\pgfqpoint{3.006256in}{7.013484in}}%
\pgfusepath{clip}%
\pgfsetbuttcap%
\pgfsetroundjoin%
\pgfsetlinewidth{1.505625pt}%
\definecolor{currentstroke}{rgb}{0.000000,0.000000,0.000000}%
\pgfsetstrokecolor{currentstroke}%
\pgfsetdash{}{0pt}%
\pgfpathmoveto{\pgfqpoint{3.349868in}{4.882250in}}%
\pgfpathlineto{\pgfqpoint{3.891869in}{4.882250in}}%
\pgfusepath{stroke}%
\end{pgfscope}%
\begin{pgfscope}%
\pgfpathrectangle{\pgfqpoint{2.768744in}{1.536516in}}{\pgfqpoint{3.006256in}{7.013484in}}%
\pgfusepath{clip}%
\pgfsetbuttcap%
\pgfsetroundjoin%
\pgfsetlinewidth{1.505625pt}%
\definecolor{currentstroke}{rgb}{0.000000,0.000000,0.000000}%
\pgfsetstrokecolor{currentstroke}%
\pgfsetdash{}{0pt}%
\pgfpathmoveto{\pgfqpoint{3.334749in}{4.560235in}}%
\pgfpathlineto{\pgfqpoint{3.849445in}{4.560235in}}%
\pgfusepath{stroke}%
\end{pgfscope}%
\begin{pgfscope}%
\pgfpathrectangle{\pgfqpoint{2.768744in}{1.536516in}}{\pgfqpoint{3.006256in}{7.013484in}}%
\pgfusepath{clip}%
\pgfsetbuttcap%
\pgfsetroundjoin%
\pgfsetlinewidth{1.505625pt}%
\definecolor{currentstroke}{rgb}{0.000000,0.000000,0.000000}%
\pgfsetstrokecolor{currentstroke}%
\pgfsetdash{}{0pt}%
\pgfpathmoveto{\pgfqpoint{3.208861in}{4.238221in}}%
\pgfpathlineto{\pgfqpoint{3.289124in}{4.238221in}}%
\pgfusepath{stroke}%
\end{pgfscope}%
\begin{pgfscope}%
\pgfpathrectangle{\pgfqpoint{2.768744in}{1.536516in}}{\pgfqpoint{3.006256in}{7.013484in}}%
\pgfusepath{clip}%
\pgfsetbuttcap%
\pgfsetroundjoin%
\pgfsetlinewidth{1.505625pt}%
\definecolor{currentstroke}{rgb}{0.000000,0.000000,0.000000}%
\pgfsetstrokecolor{currentstroke}%
\pgfsetdash{}{0pt}%
\pgfpathmoveto{\pgfqpoint{3.203957in}{3.916206in}}%
\pgfpathlineto{\pgfqpoint{3.534939in}{3.916206in}}%
\pgfusepath{stroke}%
\end{pgfscope}%
\begin{pgfscope}%
\pgfpathrectangle{\pgfqpoint{2.768744in}{1.536516in}}{\pgfqpoint{3.006256in}{7.013484in}}%
\pgfusepath{clip}%
\pgfsetbuttcap%
\pgfsetroundjoin%
\pgfsetlinewidth{1.505625pt}%
\definecolor{currentstroke}{rgb}{0.000000,0.000000,0.000000}%
\pgfsetstrokecolor{currentstroke}%
\pgfsetdash{}{0pt}%
\pgfpathmoveto{\pgfqpoint{3.136858in}{3.594191in}}%
\pgfpathlineto{\pgfqpoint{3.393963in}{3.594191in}}%
\pgfusepath{stroke}%
\end{pgfscope}%
\begin{pgfscope}%
\pgfpathrectangle{\pgfqpoint{2.768744in}{1.536516in}}{\pgfqpoint{3.006256in}{7.013484in}}%
\pgfusepath{clip}%
\pgfsetbuttcap%
\pgfsetroundjoin%
\pgfsetlinewidth{1.505625pt}%
\definecolor{currentstroke}{rgb}{0.000000,0.000000,0.000000}%
\pgfsetstrokecolor{currentstroke}%
\pgfsetdash{}{0pt}%
\pgfpathmoveto{\pgfqpoint{3.128250in}{3.272176in}}%
\pgfpathlineto{\pgfqpoint{3.486920in}{3.272176in}}%
\pgfusepath{stroke}%
\end{pgfscope}%
\begin{pgfscope}%
\pgfpathrectangle{\pgfqpoint{2.768744in}{1.536516in}}{\pgfqpoint{3.006256in}{7.013484in}}%
\pgfusepath{clip}%
\pgfsetbuttcap%
\pgfsetroundjoin%
\pgfsetlinewidth{1.505625pt}%
\definecolor{currentstroke}{rgb}{0.000000,0.000000,0.000000}%
\pgfsetstrokecolor{currentstroke}%
\pgfsetdash{}{0pt}%
\pgfpathmoveto{\pgfqpoint{3.104436in}{2.950161in}}%
\pgfpathlineto{\pgfqpoint{3.297157in}{2.950161in}}%
\pgfusepath{stroke}%
\end{pgfscope}%
\begin{pgfscope}%
\pgfpathrectangle{\pgfqpoint{2.768744in}{1.536516in}}{\pgfqpoint{3.006256in}{7.013484in}}%
\pgfusepath{clip}%
\pgfsetbuttcap%
\pgfsetroundjoin%
\pgfsetlinewidth{1.505625pt}%
\definecolor{currentstroke}{rgb}{0.000000,0.000000,0.000000}%
\pgfsetstrokecolor{currentstroke}%
\pgfsetdash{}{0pt}%
\pgfpathmoveto{\pgfqpoint{3.094592in}{2.628146in}}%
\pgfpathlineto{\pgfqpoint{3.400254in}{2.628146in}}%
\pgfusepath{stroke}%
\end{pgfscope}%
\begin{pgfscope}%
\pgfpathrectangle{\pgfqpoint{2.768744in}{1.536516in}}{\pgfqpoint{3.006256in}{7.013484in}}%
\pgfusepath{clip}%
\pgfsetbuttcap%
\pgfsetroundjoin%
\pgfsetlinewidth{1.505625pt}%
\definecolor{currentstroke}{rgb}{0.000000,0.000000,0.000000}%
\pgfsetstrokecolor{currentstroke}%
\pgfsetdash{}{0pt}%
\pgfpathmoveto{\pgfqpoint{3.088875in}{2.306131in}}%
\pgfpathlineto{\pgfqpoint{3.333436in}{2.306131in}}%
\pgfusepath{stroke}%
\end{pgfscope}%
\begin{pgfscope}%
\pgfpathrectangle{\pgfqpoint{2.768744in}{1.536516in}}{\pgfqpoint{3.006256in}{7.013484in}}%
\pgfusepath{clip}%
\pgfsetbuttcap%
\pgfsetroundjoin%
\pgfsetlinewidth{1.505625pt}%
\definecolor{currentstroke}{rgb}{0.000000,0.000000,0.000000}%
\pgfsetstrokecolor{currentstroke}%
\pgfsetdash{}{0pt}%
\pgfpathmoveto{\pgfqpoint{3.082932in}{1.984116in}}%
\pgfpathlineto{\pgfqpoint{3.366996in}{1.984116in}}%
\pgfusepath{stroke}%
\end{pgfscope}%
\begin{pgfscope}%
\pgfsetrectcap%
\pgfsetmiterjoin%
\pgfsetlinewidth{0.803000pt}%
\definecolor{currentstroke}{rgb}{0.000000,0.000000,0.000000}%
\pgfsetstrokecolor{currentstroke}%
\pgfsetdash{}{0pt}%
\pgfpathmoveto{\pgfqpoint{2.768744in}{1.536516in}}%
\pgfpathlineto{\pgfqpoint{2.768744in}{8.550000in}}%
\pgfusepath{stroke}%
\end{pgfscope}%
\begin{pgfscope}%
\pgfsetrectcap%
\pgfsetmiterjoin%
\pgfsetlinewidth{0.803000pt}%
\definecolor{currentstroke}{rgb}{0.000000,0.000000,0.000000}%
\pgfsetstrokecolor{currentstroke}%
\pgfsetdash{}{0pt}%
\pgfpathmoveto{\pgfqpoint{2.768744in}{1.536516in}}%
\pgfpathlineto{\pgfqpoint{5.775000in}{1.536516in}}%
\pgfusepath{stroke}%
\end{pgfscope}%
\end{pgfpicture}%
\makeatother%
\endgroup%

%% file: figures/shapley/shapley_vx8vbt08_16h_raw_EICU_dot.pgf
\begingroup%
\makeatletter%
\begin{pgfpicture}%
\pgfpathrectangle{\pgfpointorigin}{\pgfqpoint{8.000000in}{9.500000in}}%
\pgfusepath{use as bounding box, clip}%
\begin{pgfscope}%
\pgfsetbuttcap%
\pgfsetmiterjoin%
\definecolor{currentfill}{rgb}{1.000000,1.000000,1.000000}%
\pgfsetfillcolor{currentfill}%
\pgfsetlinewidth{0.000000pt}%
\definecolor{currentstroke}{rgb}{1.000000,1.000000,1.000000}%
\pgfsetstrokecolor{currentstroke}%
\pgfsetdash{}{0pt}%
\pgfpathmoveto{\pgfqpoint{0.000000in}{0.000000in}}%
\pgfpathlineto{\pgfqpoint{8.000000in}{0.000000in}}%
\pgfpathlineto{\pgfqpoint{8.000000in}{9.500000in}}%
\pgfpathlineto{\pgfqpoint{0.000000in}{9.500000in}}%
\pgfpathclose%
\pgfusepath{fill}%
\end{pgfscope}%
\begin{pgfscope}%
\pgfsetbuttcap%
\pgfsetmiterjoin%
\definecolor{currentfill}{rgb}{1.000000,1.000000,1.000000}%
\pgfsetfillcolor{currentfill}%
\pgfsetlinewidth{0.000000pt}%
\definecolor{currentstroke}{rgb}{0.000000,0.000000,0.000000}%
\pgfsetstrokecolor{currentstroke}%
\pgfsetstrokeopacity{0.000000}%
\pgfsetdash{}{0pt}%
\pgfpathmoveto{\pgfqpoint{3.560335in}{0.694630in}}%
\pgfpathlineto{\pgfqpoint{6.920067in}{0.694630in}}%
\pgfpathlineto{\pgfqpoint{6.920067in}{9.207193in}}%
\pgfpathlineto{\pgfqpoint{3.560335in}{9.207193in}}%
\pgfpathclose%
\pgfusepath{fill}%
\end{pgfscope}%
\begin{pgfscope}%
\pgfpathrectangle{\pgfqpoint{3.560335in}{0.694630in}}{\pgfqpoint{3.359732in}{8.512564in}}%
\pgfusepath{clip}%
\pgfsetrectcap%
\pgfsetroundjoin%
\pgfsetlinewidth{1.505625pt}%
\definecolor{currentstroke}{rgb}{0.600000,0.600000,0.600000}%
\pgfsetstrokecolor{currentstroke}%
\pgfsetdash{}{0pt}%
\pgfpathmoveto{\pgfqpoint{4.909568in}{0.694630in}}%
\pgfpathlineto{\pgfqpoint{4.909568in}{9.207193in}}%
\pgfusepath{stroke}%
\end{pgfscope}%
\begin{pgfscope}%
\pgfpathrectangle{\pgfqpoint{3.560335in}{0.694630in}}{\pgfqpoint{3.359732in}{8.512564in}}%
\pgfusepath{clip}%
\pgfsetbuttcap%
\pgfsetroundjoin%
\pgfsetlinewidth{0.501875pt}%
\definecolor{currentstroke}{rgb}{0.800000,0.800000,0.800000}%
\pgfsetstrokecolor{currentstroke}%
\pgfsetdash{{0.500000pt}{2.500000pt}}{0.000000pt}%
\pgfpathmoveto{\pgfqpoint{3.560335in}{1.099990in}}%
\pgfpathlineto{\pgfqpoint{6.920067in}{1.099990in}}%
\pgfusepath{stroke}%
\end{pgfscope}%
\begin{pgfscope}%
\pgfpathrectangle{\pgfqpoint{3.560335in}{0.694630in}}{\pgfqpoint{3.359732in}{8.512564in}}%
\pgfusepath{clip}%
\pgfsetbuttcap%
\pgfsetroundjoin%
\pgfsetlinewidth{0.501875pt}%
\definecolor{currentstroke}{rgb}{0.800000,0.800000,0.800000}%
\pgfsetstrokecolor{currentstroke}%
\pgfsetdash{{0.500000pt}{2.500000pt}}{0.000000pt}%
\pgfpathmoveto{\pgfqpoint{3.560335in}{1.505350in}}%
\pgfpathlineto{\pgfqpoint{6.920067in}{1.505350in}}%
\pgfusepath{stroke}%
\end{pgfscope}%
\begin{pgfscope}%
\pgfpathrectangle{\pgfqpoint{3.560335in}{0.694630in}}{\pgfqpoint{3.359732in}{8.512564in}}%
\pgfusepath{clip}%
\pgfsetbuttcap%
\pgfsetroundjoin%
\pgfsetlinewidth{0.501875pt}%
\definecolor{currentstroke}{rgb}{0.800000,0.800000,0.800000}%
\pgfsetstrokecolor{currentstroke}%
\pgfsetdash{{0.500000pt}{2.500000pt}}{0.000000pt}%
\pgfpathmoveto{\pgfqpoint{3.560335in}{1.910710in}}%
\pgfpathlineto{\pgfqpoint{6.920067in}{1.910710in}}%
\pgfusepath{stroke}%
\end{pgfscope}%
\begin{pgfscope}%
\pgfpathrectangle{\pgfqpoint{3.560335in}{0.694630in}}{\pgfqpoint{3.359732in}{8.512564in}}%
\pgfusepath{clip}%
\pgfsetbuttcap%
\pgfsetroundjoin%
\pgfsetlinewidth{0.501875pt}%
\definecolor{currentstroke}{rgb}{0.800000,0.800000,0.800000}%
\pgfsetstrokecolor{currentstroke}%
\pgfsetdash{{0.500000pt}{2.500000pt}}{0.000000pt}%
\pgfpathmoveto{\pgfqpoint{3.560335in}{2.316070in}}%
\pgfpathlineto{\pgfqpoint{6.920067in}{2.316070in}}%
\pgfusepath{stroke}%
\end{pgfscope}%
\begin{pgfscope}%
\pgfpathrectangle{\pgfqpoint{3.560335in}{0.694630in}}{\pgfqpoint{3.359732in}{8.512564in}}%
\pgfusepath{clip}%
\pgfsetbuttcap%
\pgfsetroundjoin%
\pgfsetlinewidth{0.501875pt}%
\definecolor{currentstroke}{rgb}{0.800000,0.800000,0.800000}%
\pgfsetstrokecolor{currentstroke}%
\pgfsetdash{{0.500000pt}{2.500000pt}}{0.000000pt}%
\pgfpathmoveto{\pgfqpoint{3.560335in}{2.721430in}}%
\pgfpathlineto{\pgfqpoint{6.920067in}{2.721430in}}%
\pgfusepath{stroke}%
\end{pgfscope}%
\begin{pgfscope}%
\pgfpathrectangle{\pgfqpoint{3.560335in}{0.694630in}}{\pgfqpoint{3.359732in}{8.512564in}}%
\pgfusepath{clip}%
\pgfsetbuttcap%
\pgfsetroundjoin%
\pgfsetlinewidth{0.501875pt}%
\definecolor{currentstroke}{rgb}{0.800000,0.800000,0.800000}%
\pgfsetstrokecolor{currentstroke}%
\pgfsetdash{{0.500000pt}{2.500000pt}}{0.000000pt}%
\pgfpathmoveto{\pgfqpoint{3.560335in}{3.126791in}}%
\pgfpathlineto{\pgfqpoint{6.920067in}{3.126791in}}%
\pgfusepath{stroke}%
\end{pgfscope}%
\begin{pgfscope}%
\pgfpathrectangle{\pgfqpoint{3.560335in}{0.694630in}}{\pgfqpoint{3.359732in}{8.512564in}}%
\pgfusepath{clip}%
\pgfsetbuttcap%
\pgfsetroundjoin%
\pgfsetlinewidth{0.501875pt}%
\definecolor{currentstroke}{rgb}{0.800000,0.800000,0.800000}%
\pgfsetstrokecolor{currentstroke}%
\pgfsetdash{{0.500000pt}{2.500000pt}}{0.000000pt}%
\pgfpathmoveto{\pgfqpoint{3.560335in}{3.532151in}}%
\pgfpathlineto{\pgfqpoint{6.920067in}{3.532151in}}%
\pgfusepath{stroke}%
\end{pgfscope}%
\begin{pgfscope}%
\pgfpathrectangle{\pgfqpoint{3.560335in}{0.694630in}}{\pgfqpoint{3.359732in}{8.512564in}}%
\pgfusepath{clip}%
\pgfsetbuttcap%
\pgfsetroundjoin%
\pgfsetlinewidth{0.501875pt}%
\definecolor{currentstroke}{rgb}{0.800000,0.800000,0.800000}%
\pgfsetstrokecolor{currentstroke}%
\pgfsetdash{{0.500000pt}{2.500000pt}}{0.000000pt}%
\pgfpathmoveto{\pgfqpoint{3.560335in}{3.937511in}}%
\pgfpathlineto{\pgfqpoint{6.920067in}{3.937511in}}%
\pgfusepath{stroke}%
\end{pgfscope}%
\begin{pgfscope}%
\pgfpathrectangle{\pgfqpoint{3.560335in}{0.694630in}}{\pgfqpoint{3.359732in}{8.512564in}}%
\pgfusepath{clip}%
\pgfsetbuttcap%
\pgfsetroundjoin%
\pgfsetlinewidth{0.501875pt}%
\definecolor{currentstroke}{rgb}{0.800000,0.800000,0.800000}%
\pgfsetstrokecolor{currentstroke}%
\pgfsetdash{{0.500000pt}{2.500000pt}}{0.000000pt}%
\pgfpathmoveto{\pgfqpoint{3.560335in}{4.342871in}}%
\pgfpathlineto{\pgfqpoint{6.920067in}{4.342871in}}%
\pgfusepath{stroke}%
\end{pgfscope}%
\begin{pgfscope}%
\pgfpathrectangle{\pgfqpoint{3.560335in}{0.694630in}}{\pgfqpoint{3.359732in}{8.512564in}}%
\pgfusepath{clip}%
\pgfsetbuttcap%
\pgfsetroundjoin%
\pgfsetlinewidth{0.501875pt}%
\definecolor{currentstroke}{rgb}{0.800000,0.800000,0.800000}%
\pgfsetstrokecolor{currentstroke}%
\pgfsetdash{{0.500000pt}{2.500000pt}}{0.000000pt}%
\pgfpathmoveto{\pgfqpoint{3.560335in}{4.748231in}}%
\pgfpathlineto{\pgfqpoint{6.920067in}{4.748231in}}%
\pgfusepath{stroke}%
\end{pgfscope}%
\begin{pgfscope}%
\pgfpathrectangle{\pgfqpoint{3.560335in}{0.694630in}}{\pgfqpoint{3.359732in}{8.512564in}}%
\pgfusepath{clip}%
\pgfsetbuttcap%
\pgfsetroundjoin%
\pgfsetlinewidth{0.501875pt}%
\definecolor{currentstroke}{rgb}{0.800000,0.800000,0.800000}%
\pgfsetstrokecolor{currentstroke}%
\pgfsetdash{{0.500000pt}{2.500000pt}}{0.000000pt}%
\pgfpathmoveto{\pgfqpoint{3.560335in}{5.153592in}}%
\pgfpathlineto{\pgfqpoint{6.920067in}{5.153592in}}%
\pgfusepath{stroke}%
\end{pgfscope}%
\begin{pgfscope}%
\pgfpathrectangle{\pgfqpoint{3.560335in}{0.694630in}}{\pgfqpoint{3.359732in}{8.512564in}}%
\pgfusepath{clip}%
\pgfsetbuttcap%
\pgfsetroundjoin%
\pgfsetlinewidth{0.501875pt}%
\definecolor{currentstroke}{rgb}{0.800000,0.800000,0.800000}%
\pgfsetstrokecolor{currentstroke}%
\pgfsetdash{{0.500000pt}{2.500000pt}}{0.000000pt}%
\pgfpathmoveto{\pgfqpoint{3.560335in}{5.558952in}}%
\pgfpathlineto{\pgfqpoint{6.920067in}{5.558952in}}%
\pgfusepath{stroke}%
\end{pgfscope}%
\begin{pgfscope}%
\pgfpathrectangle{\pgfqpoint{3.560335in}{0.694630in}}{\pgfqpoint{3.359732in}{8.512564in}}%
\pgfusepath{clip}%
\pgfsetbuttcap%
\pgfsetroundjoin%
\pgfsetlinewidth{0.501875pt}%
\definecolor{currentstroke}{rgb}{0.800000,0.800000,0.800000}%
\pgfsetstrokecolor{currentstroke}%
\pgfsetdash{{0.500000pt}{2.500000pt}}{0.000000pt}%
\pgfpathmoveto{\pgfqpoint{3.560335in}{5.964312in}}%
\pgfpathlineto{\pgfqpoint{6.920067in}{5.964312in}}%
\pgfusepath{stroke}%
\end{pgfscope}%
\begin{pgfscope}%
\pgfpathrectangle{\pgfqpoint{3.560335in}{0.694630in}}{\pgfqpoint{3.359732in}{8.512564in}}%
\pgfusepath{clip}%
\pgfsetbuttcap%
\pgfsetroundjoin%
\pgfsetlinewidth{0.501875pt}%
\definecolor{currentstroke}{rgb}{0.800000,0.800000,0.800000}%
\pgfsetstrokecolor{currentstroke}%
\pgfsetdash{{0.500000pt}{2.500000pt}}{0.000000pt}%
\pgfpathmoveto{\pgfqpoint{3.560335in}{6.369672in}}%
\pgfpathlineto{\pgfqpoint{6.920067in}{6.369672in}}%
\pgfusepath{stroke}%
\end{pgfscope}%
\begin{pgfscope}%
\pgfpathrectangle{\pgfqpoint{3.560335in}{0.694630in}}{\pgfqpoint{3.359732in}{8.512564in}}%
\pgfusepath{clip}%
\pgfsetbuttcap%
\pgfsetroundjoin%
\pgfsetlinewidth{0.501875pt}%
\definecolor{currentstroke}{rgb}{0.800000,0.800000,0.800000}%
\pgfsetstrokecolor{currentstroke}%
\pgfsetdash{{0.500000pt}{2.500000pt}}{0.000000pt}%
\pgfpathmoveto{\pgfqpoint{3.560335in}{6.775032in}}%
\pgfpathlineto{\pgfqpoint{6.920067in}{6.775032in}}%
\pgfusepath{stroke}%
\end{pgfscope}%
\begin{pgfscope}%
\pgfpathrectangle{\pgfqpoint{3.560335in}{0.694630in}}{\pgfqpoint{3.359732in}{8.512564in}}%
\pgfusepath{clip}%
\pgfsetbuttcap%
\pgfsetroundjoin%
\pgfsetlinewidth{0.501875pt}%
\definecolor{currentstroke}{rgb}{0.800000,0.800000,0.800000}%
\pgfsetstrokecolor{currentstroke}%
\pgfsetdash{{0.500000pt}{2.500000pt}}{0.000000pt}%
\pgfpathmoveto{\pgfqpoint{3.560335in}{7.180392in}}%
\pgfpathlineto{\pgfqpoint{6.920067in}{7.180392in}}%
\pgfusepath{stroke}%
\end{pgfscope}%
\begin{pgfscope}%
\pgfpathrectangle{\pgfqpoint{3.560335in}{0.694630in}}{\pgfqpoint{3.359732in}{8.512564in}}%
\pgfusepath{clip}%
\pgfsetbuttcap%
\pgfsetroundjoin%
\pgfsetlinewidth{0.501875pt}%
\definecolor{currentstroke}{rgb}{0.800000,0.800000,0.800000}%
\pgfsetstrokecolor{currentstroke}%
\pgfsetdash{{0.500000pt}{2.500000pt}}{0.000000pt}%
\pgfpathmoveto{\pgfqpoint{3.560335in}{7.585753in}}%
\pgfpathlineto{\pgfqpoint{6.920067in}{7.585753in}}%
\pgfusepath{stroke}%
\end{pgfscope}%
\begin{pgfscope}%
\pgfpathrectangle{\pgfqpoint{3.560335in}{0.694630in}}{\pgfqpoint{3.359732in}{8.512564in}}%
\pgfusepath{clip}%
\pgfsetbuttcap%
\pgfsetroundjoin%
\pgfsetlinewidth{0.501875pt}%
\definecolor{currentstroke}{rgb}{0.800000,0.800000,0.800000}%
\pgfsetstrokecolor{currentstroke}%
\pgfsetdash{{0.500000pt}{2.500000pt}}{0.000000pt}%
\pgfpathmoveto{\pgfqpoint{3.560335in}{7.991113in}}%
\pgfpathlineto{\pgfqpoint{6.920067in}{7.991113in}}%
\pgfusepath{stroke}%
\end{pgfscope}%
\begin{pgfscope}%
\pgfpathrectangle{\pgfqpoint{3.560335in}{0.694630in}}{\pgfqpoint{3.359732in}{8.512564in}}%
\pgfusepath{clip}%
\pgfsetbuttcap%
\pgfsetroundjoin%
\pgfsetlinewidth{0.501875pt}%
\definecolor{currentstroke}{rgb}{0.800000,0.800000,0.800000}%
\pgfsetstrokecolor{currentstroke}%
\pgfsetdash{{0.500000pt}{2.500000pt}}{0.000000pt}%
\pgfpathmoveto{\pgfqpoint{3.560335in}{8.396473in}}%
\pgfpathlineto{\pgfqpoint{6.920067in}{8.396473in}}%
\pgfusepath{stroke}%
\end{pgfscope}%
\begin{pgfscope}%
\pgfpathrectangle{\pgfqpoint{3.560335in}{0.694630in}}{\pgfqpoint{3.359732in}{8.512564in}}%
\pgfusepath{clip}%
\pgfsetbuttcap%
\pgfsetroundjoin%
\pgfsetlinewidth{0.501875pt}%
\definecolor{currentstroke}{rgb}{0.800000,0.800000,0.800000}%
\pgfsetstrokecolor{currentstroke}%
\pgfsetdash{{0.500000pt}{2.500000pt}}{0.000000pt}%
\pgfpathmoveto{\pgfqpoint{3.560335in}{8.801833in}}%
\pgfpathlineto{\pgfqpoint{6.920067in}{8.801833in}}%
\pgfusepath{stroke}%
\end{pgfscope}%
\begin{pgfscope}%
\pgfsetbuttcap%
\pgfsetroundjoin%
\definecolor{currentfill}{rgb}{0.200000,0.200000,0.200000}%
\pgfsetfillcolor{currentfill}%
\pgfsetlinewidth{0.803000pt}%
\definecolor{currentstroke}{rgb}{0.200000,0.200000,0.200000}%
\pgfsetstrokecolor{currentstroke}%
\pgfsetdash{}{0pt}%
\pgfsys@defobject{currentmarker}{\pgfqpoint{0.000000in}{-0.048611in}}{\pgfqpoint{0.000000in}{0.000000in}}{%
\pgfpathmoveto{\pgfqpoint{0.000000in}{0.000000in}}%
\pgfpathlineto{\pgfqpoint{0.000000in}{-0.048611in}}%
\pgfusepath{stroke,fill}%
}%
\begin{pgfscope}%
\pgfsys@transformshift{3.888698in}{0.694630in}%
\pgfsys@useobject{currentmarker}{}%
\end{pgfscope}%
\end{pgfscope}%
\begin{pgfscope}%
\definecolor{textcolor}{rgb}{0.200000,0.200000,0.200000}%
\pgfsetstrokecolor{textcolor}%
\pgfsetfillcolor{textcolor}%
\pgftext[x=3.888698in,y=0.597407in,,top]{\color{textcolor}\sffamily\fontsize{11.000000}{13.200000}\selectfont \(\displaystyle {\ensuremath{-}0.4}\)}%
\end{pgfscope}%
\begin{pgfscope}%
\pgfsetbuttcap%
\pgfsetroundjoin%
\definecolor{currentfill}{rgb}{0.200000,0.200000,0.200000}%
\pgfsetfillcolor{currentfill}%
\pgfsetlinewidth{0.803000pt}%
\definecolor{currentstroke}{rgb}{0.200000,0.200000,0.200000}%
\pgfsetstrokecolor{currentstroke}%
\pgfsetdash{}{0pt}%
\pgfsys@defobject{currentmarker}{\pgfqpoint{0.000000in}{-0.048611in}}{\pgfqpoint{0.000000in}{0.000000in}}{%
\pgfpathmoveto{\pgfqpoint{0.000000in}{0.000000in}}%
\pgfpathlineto{\pgfqpoint{0.000000in}{-0.048611in}}%
\pgfusepath{stroke,fill}%
}%
\begin{pgfscope}%
\pgfsys@transformshift{4.399133in}{0.694630in}%
\pgfsys@useobject{currentmarker}{}%
\end{pgfscope}%
\end{pgfscope}%
\begin{pgfscope}%
\definecolor{textcolor}{rgb}{0.200000,0.200000,0.200000}%
\pgfsetstrokecolor{textcolor}%
\pgfsetfillcolor{textcolor}%
\pgftext[x=4.399133in,y=0.597407in,,top]{\color{textcolor}\sffamily\fontsize{11.000000}{13.200000}\selectfont \(\displaystyle {\ensuremath{-}0.2}\)}%
\end{pgfscope}%
\begin{pgfscope}%
\pgfsetbuttcap%
\pgfsetroundjoin%
\definecolor{currentfill}{rgb}{0.200000,0.200000,0.200000}%
\pgfsetfillcolor{currentfill}%
\pgfsetlinewidth{0.803000pt}%
\definecolor{currentstroke}{rgb}{0.200000,0.200000,0.200000}%
\pgfsetstrokecolor{currentstroke}%
\pgfsetdash{}{0pt}%
\pgfsys@defobject{currentmarker}{\pgfqpoint{0.000000in}{-0.048611in}}{\pgfqpoint{0.000000in}{0.000000in}}{%
\pgfpathmoveto{\pgfqpoint{0.000000in}{0.000000in}}%
\pgfpathlineto{\pgfqpoint{0.000000in}{-0.048611in}}%
\pgfusepath{stroke,fill}%
}%
\begin{pgfscope}%
\pgfsys@transformshift{4.909568in}{0.694630in}%
\pgfsys@useobject{currentmarker}{}%
\end{pgfscope}%
\end{pgfscope}%
\begin{pgfscope}%
\definecolor{textcolor}{rgb}{0.200000,0.200000,0.200000}%
\pgfsetstrokecolor{textcolor}%
\pgfsetfillcolor{textcolor}%
\pgftext[x=4.909568in,y=0.597407in,,top]{\color{textcolor}\sffamily\fontsize{11.000000}{13.200000}\selectfont \(\displaystyle {0.0}\)}%
\end{pgfscope}%
\begin{pgfscope}%
\pgfsetbuttcap%
\pgfsetroundjoin%
\definecolor{currentfill}{rgb}{0.200000,0.200000,0.200000}%
\pgfsetfillcolor{currentfill}%
\pgfsetlinewidth{0.803000pt}%
\definecolor{currentstroke}{rgb}{0.200000,0.200000,0.200000}%
\pgfsetstrokecolor{currentstroke}%
\pgfsetdash{}{0pt}%
\pgfsys@defobject{currentmarker}{\pgfqpoint{0.000000in}{-0.048611in}}{\pgfqpoint{0.000000in}{0.000000in}}{%
\pgfpathmoveto{\pgfqpoint{0.000000in}{0.000000in}}%
\pgfpathlineto{\pgfqpoint{0.000000in}{-0.048611in}}%
\pgfusepath{stroke,fill}%
}%
\begin{pgfscope}%
\pgfsys@transformshift{5.420003in}{0.694630in}%
\pgfsys@useobject{currentmarker}{}%
\end{pgfscope}%
\end{pgfscope}%
\begin{pgfscope}%
\definecolor{textcolor}{rgb}{0.200000,0.200000,0.200000}%
\pgfsetstrokecolor{textcolor}%
\pgfsetfillcolor{textcolor}%
\pgftext[x=5.420003in,y=0.597407in,,top]{\color{textcolor}\sffamily\fontsize{11.000000}{13.200000}\selectfont \(\displaystyle {0.2}\)}%
\end{pgfscope}%
\begin{pgfscope}%
\pgfsetbuttcap%
\pgfsetroundjoin%
\definecolor{currentfill}{rgb}{0.200000,0.200000,0.200000}%
\pgfsetfillcolor{currentfill}%
\pgfsetlinewidth{0.803000pt}%
\definecolor{currentstroke}{rgb}{0.200000,0.200000,0.200000}%
\pgfsetstrokecolor{currentstroke}%
\pgfsetdash{}{0pt}%
\pgfsys@defobject{currentmarker}{\pgfqpoint{0.000000in}{-0.048611in}}{\pgfqpoint{0.000000in}{0.000000in}}{%
\pgfpathmoveto{\pgfqpoint{0.000000in}{0.000000in}}%
\pgfpathlineto{\pgfqpoint{0.000000in}{-0.048611in}}%
\pgfusepath{stroke,fill}%
}%
\begin{pgfscope}%
\pgfsys@transformshift{5.930438in}{0.694630in}%
\pgfsys@useobject{currentmarker}{}%
\end{pgfscope}%
\end{pgfscope}%
\begin{pgfscope}%
\definecolor{textcolor}{rgb}{0.200000,0.200000,0.200000}%
\pgfsetstrokecolor{textcolor}%
\pgfsetfillcolor{textcolor}%
\pgftext[x=5.930438in,y=0.597407in,,top]{\color{textcolor}\sffamily\fontsize{11.000000}{13.200000}\selectfont \(\displaystyle {0.4}\)}%
\end{pgfscope}%
\begin{pgfscope}%
\pgfsetbuttcap%
\pgfsetroundjoin%
\definecolor{currentfill}{rgb}{0.200000,0.200000,0.200000}%
\pgfsetfillcolor{currentfill}%
\pgfsetlinewidth{0.803000pt}%
\definecolor{currentstroke}{rgb}{0.200000,0.200000,0.200000}%
\pgfsetstrokecolor{currentstroke}%
\pgfsetdash{}{0pt}%
\pgfsys@defobject{currentmarker}{\pgfqpoint{0.000000in}{-0.048611in}}{\pgfqpoint{0.000000in}{0.000000in}}{%
\pgfpathmoveto{\pgfqpoint{0.000000in}{0.000000in}}%
\pgfpathlineto{\pgfqpoint{0.000000in}{-0.048611in}}%
\pgfusepath{stroke,fill}%
}%
\begin{pgfscope}%
\pgfsys@transformshift{6.440873in}{0.694630in}%
\pgfsys@useobject{currentmarker}{}%
\end{pgfscope}%
\end{pgfscope}%
\begin{pgfscope}%
\definecolor{textcolor}{rgb}{0.200000,0.200000,0.200000}%
\pgfsetstrokecolor{textcolor}%
\pgfsetfillcolor{textcolor}%
\pgftext[x=6.440873in,y=0.597407in,,top]{\color{textcolor}\sffamily\fontsize{11.000000}{13.200000}\selectfont \(\displaystyle {0.6}\)}%
\end{pgfscope}%
\begin{pgfscope}%
\definecolor{textcolor}{rgb}{0.000000,0.000000,0.000000}%
\pgfsetstrokecolor{textcolor}%
\pgfsetfillcolor{textcolor}%
\pgftext[x=5.240201in,y=0.406667in,,top]{\color{textcolor}\sffamily\fontsize{13.000000}{15.600000}\selectfont SHAP value (impact on model output)}%
\end{pgfscope}%
\begin{pgfscope}%
\definecolor{textcolor}{rgb}{0.200000,0.200000,0.200000}%
\pgfsetstrokecolor{textcolor}%
\pgfsetfillcolor{textcolor}%
\pgftext[x=1.510529in, y=1.042120in, left, base]{\color{textcolor}\sffamily\fontsize{13.000000}{15.600000}\selectfont Mean corpuscular volume}%
\end{pgfscope}%
\begin{pgfscope}%
\definecolor{textcolor}{rgb}{0.200000,0.200000,0.200000}%
\pgfsetstrokecolor{textcolor}%
\pgfsetfillcolor{textcolor}%
\pgftext[x=1.876709in, y=1.447480in, left, base]{\color{textcolor}\sffamily\fontsize{13.000000}{15.600000}\selectfont Blood urea nitrogen}%
\end{pgfscope}%
\begin{pgfscope}%
\definecolor{textcolor}{rgb}{0.200000,0.200000,0.200000}%
\pgfsetstrokecolor{textcolor}%
\pgfsetfillcolor{textcolor}%
\pgftext[x=1.685373in, y=1.852840in, left, base]{\color{textcolor}\sffamily\fontsize{13.000000}{15.600000}\selectfont Systolic blood pressure}%
\end{pgfscope}%
\begin{pgfscope}%
\definecolor{textcolor}{rgb}{0.200000,0.200000,0.200000}%
\pgfsetstrokecolor{textcolor}%
\pgfsetfillcolor{textcolor}%
\pgftext[x=2.277457in, y=2.258200in, left, base]{\color{textcolor}\sffamily\fontsize{13.000000}{15.600000}\selectfont Platelet count}%
\end{pgfscope}%
\begin{pgfscope}%
\definecolor{textcolor}{rgb}{0.200000,0.200000,0.200000}%
\pgfsetstrokecolor{textcolor}%
\pgfsetfillcolor{textcolor}%
\pgftext[x=2.669139in, y=2.663560in, left, base]{\color{textcolor}\sffamily\fontsize{13.000000}{15.600000}\selectfont Albumin}%
\end{pgfscope}%
\begin{pgfscope}%
\definecolor{textcolor}{rgb}{0.200000,0.200000,0.200000}%
\pgfsetstrokecolor{textcolor}%
\pgfsetfillcolor{textcolor}%
\pgftext[x=2.428979in, y=3.068920in, left, base]{\color{textcolor}\sffamily\fontsize{13.000000}{15.600000}\selectfont Hemoglobin}%
\end{pgfscope}%
\begin{pgfscope}%
\definecolor{textcolor}{rgb}{0.200000,0.200000,0.200000}%
\pgfsetstrokecolor{textcolor}%
\pgfsetfillcolor{textcolor}%
\pgftext[x=1.683618in, y=3.474281in, left, base]{\color{textcolor}\sffamily\fontsize{13.000000}{15.600000}\selectfont White blood cell count}%
\end{pgfscope}%
\begin{pgfscope}%
\definecolor{textcolor}{rgb}{0.200000,0.200000,0.200000}%
\pgfsetstrokecolor{textcolor}%
\pgfsetfillcolor{textcolor}%
\pgftext[x=2.347325in, y=3.879641in, left, base]{\color{textcolor}\sffamily\fontsize{13.000000}{15.600000}\selectfont Lymphocytes}%
\end{pgfscope}%
\begin{pgfscope}%
\definecolor{textcolor}{rgb}{0.200000,0.200000,0.200000}%
\pgfsetstrokecolor{textcolor}%
\pgfsetfillcolor{textcolor}%
\pgftext[x=2.469487in, y=4.285001in, left, base]{\color{textcolor}\sffamily\fontsize{13.000000}{15.600000}\selectfont Hematocrit}%
\end{pgfscope}%
\begin{pgfscope}%
\definecolor{textcolor}{rgb}{0.200000,0.200000,0.200000}%
\pgfsetstrokecolor{textcolor}%
\pgfsetfillcolor{textcolor}%
\pgftext[x=2.722282in, y=4.690361in, left, base]{\color{textcolor}\sffamily\fontsize{13.000000}{15.600000}\selectfont Sodium}%
\end{pgfscope}%
\begin{pgfscope}%
\definecolor{textcolor}{rgb}{0.200000,0.200000,0.200000}%
\pgfsetstrokecolor{textcolor}%
\pgfsetfillcolor{textcolor}%
\pgftext[x=2.547901in, y=5.095721in, left, base]{\color{textcolor}\sffamily\fontsize{13.000000}{15.600000}\selectfont Creatinine}%
\end{pgfscope}%
\begin{pgfscope}%
\definecolor{textcolor}{rgb}{0.200000,0.200000,0.200000}%
\pgfsetstrokecolor{textcolor}%
\pgfsetfillcolor{textcolor}%
\pgftext[x=1.753061in, y=5.501082in, left, base]{\color{textcolor}\sffamily\fontsize{13.000000}{15.600000}\selectfont Mean cell hemoglobin}%
\end{pgfscope}%
\begin{pgfscope}%
\definecolor{textcolor}{rgb}{0.200000,0.200000,0.200000}%
\pgfsetstrokecolor{textcolor}%
\pgfsetfillcolor{textcolor}%
\pgftext[x=2.680423in, y=5.906442in, left, base]{\color{textcolor}\sffamily\fontsize{13.000000}{15.600000}\selectfont Chloride}%
\end{pgfscope}%
\begin{pgfscope}%
\definecolor{textcolor}{rgb}{0.200000,0.200000,0.200000}%
\pgfsetstrokecolor{textcolor}%
\pgfsetfillcolor{textcolor}%
\pgftext[x=2.689489in, y=6.311802in, left, base]{\color{textcolor}\sffamily\fontsize{13.000000}{15.600000}\selectfont Calcium}%
\end{pgfscope}%
\begin{pgfscope}%
\definecolor{textcolor}{rgb}{0.200000,0.200000,0.200000}%
\pgfsetstrokecolor{textcolor}%
\pgfsetfillcolor{textcolor}%
\pgftext[x=1.992487in, y=6.717162in, left, base]{\color{textcolor}\sffamily\fontsize{13.000000}{15.600000}\selectfont Oxygen saturation}%
\end{pgfscope}%
\begin{pgfscope}%
\definecolor{textcolor}{rgb}{0.200000,0.200000,0.200000}%
\pgfsetstrokecolor{textcolor}%
\pgfsetfillcolor{textcolor}%
\pgftext[x=0.240000in, y=7.122522in, left, base]{\color{textcolor}\sffamily\fontsize{13.000000}{15.600000}\selectfont Mean corpuscular hemoglobin concentration}%
\end{pgfscope}%
\begin{pgfscope}%
\definecolor{textcolor}{rgb}{0.200000,0.200000,0.200000}%
\pgfsetstrokecolor{textcolor}%
\pgfsetfillcolor{textcolor}%
\pgftext[x=2.367733in, y=7.527882in, left, base]{\color{textcolor}\sffamily\fontsize{13.000000}{15.600000}\selectfont Urine output}%
\end{pgfscope}%
\begin{pgfscope}%
\definecolor{textcolor}{rgb}{0.200000,0.200000,0.200000}%
\pgfsetstrokecolor{textcolor}%
\pgfsetfillcolor{textcolor}%
\pgftext[x=2.539605in, y=7.933243in, left, base]{\color{textcolor}\sffamily\fontsize{13.000000}{15.600000}\selectfont Heart rate}%
\end{pgfscope}%
\begin{pgfscope}%
\definecolor{textcolor}{rgb}{0.200000,0.200000,0.200000}%
\pgfsetstrokecolor{textcolor}%
\pgfsetfillcolor{textcolor}%
\pgftext[x=1.399709in, y=8.338603in, left, base]{\color{textcolor}\sffamily\fontsize{13.000000}{15.600000}\selectfont Fraction of inspired oxygen}%
\end{pgfscope}%
\begin{pgfscope}%
\definecolor{textcolor}{rgb}{0.200000,0.200000,0.200000}%
\pgfsetstrokecolor{textcolor}%
\pgfsetfillcolor{textcolor}%
\pgftext[x=1.734330in, y=8.743963in, left, base]{\color{textcolor}\sffamily\fontsize{13.000000}{15.600000}\selectfont Mean arterial pressure}%
\end{pgfscope}%
\begin{pgfscope}%
\pgfsetrectcap%
\pgfsetmiterjoin%
\pgfsetlinewidth{0.803000pt}%
\definecolor{currentstroke}{rgb}{0.000000,0.000000,0.000000}%
\pgfsetstrokecolor{currentstroke}%
\pgfsetdash{}{0pt}%
\pgfpathmoveto{\pgfqpoint{3.560335in}{0.694630in}}%
\pgfpathlineto{\pgfqpoint{6.920067in}{0.694630in}}%
\pgfusepath{stroke}%
\end{pgfscope}%
\begin{pgfscope}%
\pgfsys@transformshift{3.683333in}{0.923333in}%
\pgftext[left,bottom]{\includegraphics[interpolate=true,width=3.113333in,height=8.053333in]{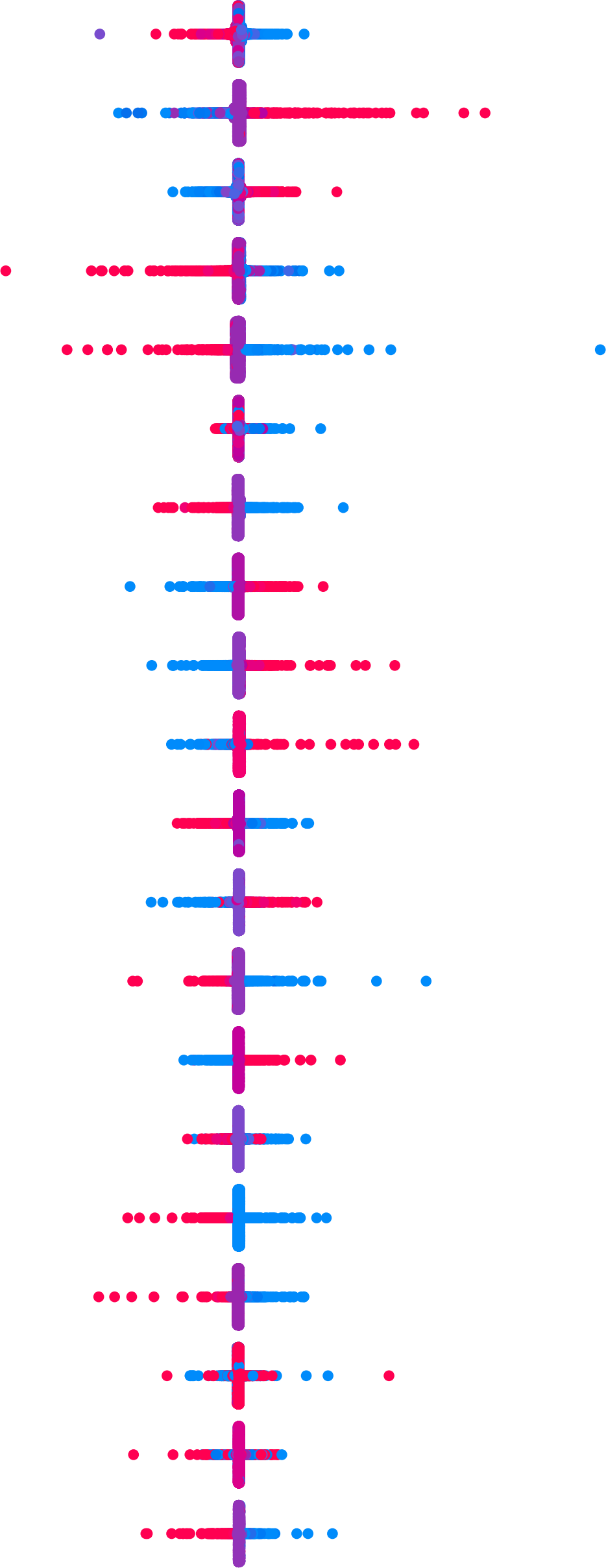}}%
\end{pgfscope}%
\begin{pgfscope}%
\pgfsetbuttcap%
\pgfsetmiterjoin%
\definecolor{currentfill}{rgb}{1.000000,1.000000,1.000000}%
\pgfsetfillcolor{currentfill}%
\pgfsetlinewidth{0.000000pt}%
\definecolor{currentstroke}{rgb}{0.000000,0.000000,0.000000}%
\pgfsetstrokecolor{currentstroke}%
\pgfsetstrokeopacity{0.000000}%
\pgfsetdash{}{0pt}%
\pgfpathmoveto{\pgfqpoint{7.130050in}{0.694630in}}%
\pgfpathlineto{\pgfqpoint{7.187708in}{0.694630in}}%
\pgfpathlineto{\pgfqpoint{7.187708in}{9.207193in}}%
\pgfpathlineto{\pgfqpoint{7.130050in}{9.207193in}}%
\pgfpathclose%
\pgfusepath{fill}%
\end{pgfscope}%
\begin{pgfscope}%
\pgfpathrectangle{\pgfqpoint{7.130050in}{0.694630in}}{\pgfqpoint{0.057658in}{8.512564in}}%
\pgfusepath{clip}%
\pgfsetbuttcap%
\pgfsetmiterjoin%
\definecolor{currentfill}{rgb}{1.000000,1.000000,1.000000}%
\pgfsetfillcolor{currentfill}%
\pgfsetlinewidth{0.010037pt}%
\definecolor{currentstroke}{rgb}{1.000000,1.000000,1.000000}%
\pgfsetstrokecolor{currentstroke}%
\pgfsetdash{}{0pt}%
\pgfpathmoveto{\pgfqpoint{7.130050in}{0.694630in}}%
\pgfpathlineto{\pgfqpoint{7.130050in}{0.727882in}}%
\pgfpathlineto{\pgfqpoint{7.130050in}{9.173941in}}%
\pgfpathlineto{\pgfqpoint{7.130050in}{9.207193in}}%
\pgfpathlineto{\pgfqpoint{7.187708in}{9.207193in}}%
\pgfpathlineto{\pgfqpoint{7.187708in}{9.173941in}}%
\pgfpathlineto{\pgfqpoint{7.187708in}{0.727882in}}%
\pgfpathlineto{\pgfqpoint{7.187708in}{0.694630in}}%
\pgfpathlineto{\pgfqpoint{7.187708in}{0.694630in}}%
\pgfpathclose%
\pgfusepath{stroke,fill}%
\end{pgfscope}%
\begin{pgfscope}%
\pgfsys@transformshift{7.130000in}{0.693333in}%
\pgftext[left,bottom]{\includegraphics[interpolate=true,width=0.056667in,height=8.513333in]{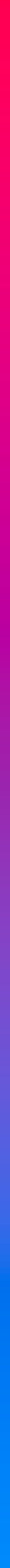}}%
\end{pgfscope}%
\begin{pgfscope}%
\definecolor{textcolor}{rgb}{0.000000,0.000000,0.000000}%
\pgfsetstrokecolor{textcolor}%
\pgfsetfillcolor{textcolor}%
\pgftext[x=7.236319in, y=0.641823in, left, base]{\color{textcolor}\sffamily\fontsize{11.000000}{13.200000}\selectfont Low}%
\end{pgfscope}%
\begin{pgfscope}%
\definecolor{textcolor}{rgb}{0.000000,0.000000,0.000000}%
\pgfsetstrokecolor{textcolor}%
\pgfsetfillcolor{textcolor}%
\pgftext[x=7.236319in, y=9.154387in, left, base]{\color{textcolor}\sffamily\fontsize{11.000000}{13.200000}\selectfont High}%
\end{pgfscope}%
\begin{pgfscope}%
\definecolor{textcolor}{rgb}{0.000000,0.000000,0.000000}%
\pgfsetstrokecolor{textcolor}%
\pgfsetfillcolor{textcolor}%
\pgftext[x=7.534994in,y=4.950911in,,top,rotate=90.000000]{\color{textcolor}\sffamily\fontsize{12.000000}{14.400000}\selectfont Feature value}%
\end{pgfscope}%
\end{pgfpicture}%
\makeatother%
\endgroup%

%% file: figures/shapley/shapley_vx8vbt08_16h_raw_EICU_scatter_map.pgf
\begingroup%
\makeatletter%
\begin{pgfpicture}%
\pgfpathrectangle{\pgfpointorigin}{\pgfqpoint{4.000000in}{2.000000in}}%
\pgfusepath{use as bounding box, clip}%
\begin{pgfscope}%
\pgfsetbuttcap%
\pgfsetmiterjoin%
\definecolor{currentfill}{rgb}{1.000000,1.000000,1.000000}%
\pgfsetfillcolor{currentfill}%
\pgfsetlinewidth{0.000000pt}%
\definecolor{currentstroke}{rgb}{1.000000,1.000000,1.000000}%
\pgfsetstrokecolor{currentstroke}%
\pgfsetdash{}{0pt}%
\pgfpathmoveto{\pgfqpoint{0.000000in}{0.000000in}}%
\pgfpathlineto{\pgfqpoint{4.000000in}{0.000000in}}%
\pgfpathlineto{\pgfqpoint{4.000000in}{2.000000in}}%
\pgfpathlineto{\pgfqpoint{0.000000in}{2.000000in}}%
\pgfpathclose%
\pgfusepath{fill}%
\end{pgfscope}%
\begin{pgfscope}%
\pgfsetbuttcap%
\pgfsetmiterjoin%
\definecolor{currentfill}{rgb}{1.000000,1.000000,1.000000}%
\pgfsetfillcolor{currentfill}%
\pgfsetlinewidth{0.000000pt}%
\definecolor{currentstroke}{rgb}{0.000000,0.000000,0.000000}%
\pgfsetstrokecolor{currentstroke}%
\pgfsetstrokeopacity{0.000000}%
\pgfsetdash{}{0pt}%
\pgfpathmoveto{\pgfqpoint{1.024837in}{0.676110in}}%
\pgfpathlineto{\pgfqpoint{3.690782in}{0.676110in}}%
\pgfpathlineto{\pgfqpoint{3.690782in}{1.760000in}}%
\pgfpathlineto{\pgfqpoint{1.024837in}{1.760000in}}%
\pgfpathclose%
\pgfusepath{fill}%
\end{pgfscope}%
\begin{pgfscope}%
\pgfsys@transformshift{1.120000in}{0.696667in}%
\pgftext[left,bottom]{\includegraphics[interpolate=true,width=2.480000in,height=1.040000in]{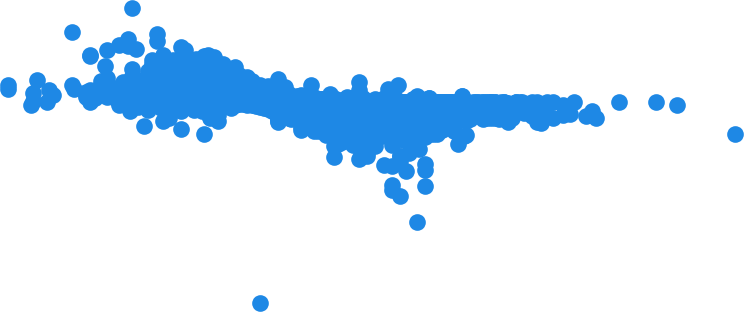}}%
\end{pgfscope}%
\begin{pgfscope}%
\pgfpathrectangle{\pgfqpoint{1.024837in}{0.676110in}}{\pgfqpoint{2.665945in}{1.083890in}}%
\pgfusepath{clip}%
\pgfsetbuttcap%
\pgfsetroundjoin%
\definecolor{currentfill}{rgb}{0.117647,0.533333,0.898039}%
\pgfsetfillcolor{currentfill}%
\pgfsetlinewidth{2.007500pt}%
\definecolor{currentstroke}{rgb}{0.117647,0.533333,0.898039}%
\pgfsetstrokecolor{currentstroke}%
\pgfsetdash{}{0pt}%
\pgfsys@defobject{currentmarker}{\pgfqpoint{0.000000in}{0.000000in}}{\pgfqpoint{0.083333in}{0.000000in}}{%
\pgfpathmoveto{\pgfqpoint{0.000000in}{0.000000in}}%
\pgfpathlineto{\pgfqpoint{0.083333in}{0.000000in}}%
\pgfusepath{stroke,fill}%
}%
\end{pgfscope}%
\begin{pgfscope}%
\pgfsetbuttcap%
\pgfsetroundjoin%
\definecolor{currentfill}{rgb}{0.200000,0.200000,0.200000}%
\pgfsetfillcolor{currentfill}%
\pgfsetlinewidth{0.803000pt}%
\definecolor{currentstroke}{rgb}{0.200000,0.200000,0.200000}%
\pgfsetstrokecolor{currentstroke}%
\pgfsetdash{}{0pt}%
\pgfsys@defobject{currentmarker}{\pgfqpoint{0.000000in}{-0.048611in}}{\pgfqpoint{0.000000in}{0.000000in}}{%
\pgfpathmoveto{\pgfqpoint{0.000000in}{0.000000in}}%
\pgfpathlineto{\pgfqpoint{0.000000in}{-0.048611in}}%
\pgfusepath{stroke,fill}%
}%
\begin{pgfscope}%
\pgfsys@transformshift{1.586662in}{0.676110in}%
\pgfsys@useobject{currentmarker}{}%
\end{pgfscope}%
\end{pgfscope}%
\begin{pgfscope}%
\definecolor{textcolor}{rgb}{0.200000,0.200000,0.200000}%
\pgfsetstrokecolor{textcolor}%
\pgfsetfillcolor{textcolor}%
\pgftext[x=1.586662in,y=0.578888in,,top]{\color{textcolor}\sffamily\fontsize{11.000000}{13.200000}\selectfont \(\displaystyle {50}\)}%
\end{pgfscope}%
\begin{pgfscope}%
\pgfsetbuttcap%
\pgfsetroundjoin%
\definecolor{currentfill}{rgb}{0.200000,0.200000,0.200000}%
\pgfsetfillcolor{currentfill}%
\pgfsetlinewidth{0.803000pt}%
\definecolor{currentstroke}{rgb}{0.200000,0.200000,0.200000}%
\pgfsetstrokecolor{currentstroke}%
\pgfsetdash{}{0pt}%
\pgfsys@defobject{currentmarker}{\pgfqpoint{0.000000in}{-0.048611in}}{\pgfqpoint{0.000000in}{0.000000in}}{%
\pgfpathmoveto{\pgfqpoint{0.000000in}{0.000000in}}%
\pgfpathlineto{\pgfqpoint{0.000000in}{-0.048611in}}%
\pgfusepath{stroke,fill}%
}%
\begin{pgfscope}%
\pgfsys@transformshift{2.275180in}{0.676110in}%
\pgfsys@useobject{currentmarker}{}%
\end{pgfscope}%
\end{pgfscope}%
\begin{pgfscope}%
\definecolor{textcolor}{rgb}{0.200000,0.200000,0.200000}%
\pgfsetstrokecolor{textcolor}%
\pgfsetfillcolor{textcolor}%
\pgftext[x=2.275180in,y=0.578888in,,top]{\color{textcolor}\sffamily\fontsize{11.000000}{13.200000}\selectfont \(\displaystyle {100}\)}%
\end{pgfscope}%
\begin{pgfscope}%
\pgfsetbuttcap%
\pgfsetroundjoin%
\definecolor{currentfill}{rgb}{0.200000,0.200000,0.200000}%
\pgfsetfillcolor{currentfill}%
\pgfsetlinewidth{0.803000pt}%
\definecolor{currentstroke}{rgb}{0.200000,0.200000,0.200000}%
\pgfsetstrokecolor{currentstroke}%
\pgfsetdash{}{0pt}%
\pgfsys@defobject{currentmarker}{\pgfqpoint{0.000000in}{-0.048611in}}{\pgfqpoint{0.000000in}{0.000000in}}{%
\pgfpathmoveto{\pgfqpoint{0.000000in}{0.000000in}}%
\pgfpathlineto{\pgfqpoint{0.000000in}{-0.048611in}}%
\pgfusepath{stroke,fill}%
}%
\begin{pgfscope}%
\pgfsys@transformshift{2.963698in}{0.676110in}%
\pgfsys@useobject{currentmarker}{}%
\end{pgfscope}%
\end{pgfscope}%
\begin{pgfscope}%
\definecolor{textcolor}{rgb}{0.200000,0.200000,0.200000}%
\pgfsetstrokecolor{textcolor}%
\pgfsetfillcolor{textcolor}%
\pgftext[x=2.963698in,y=0.578888in,,top]{\color{textcolor}\sffamily\fontsize{11.000000}{13.200000}\selectfont \(\displaystyle {150}\)}%
\end{pgfscope}%
\begin{pgfscope}%
\pgfsetbuttcap%
\pgfsetroundjoin%
\definecolor{currentfill}{rgb}{0.200000,0.200000,0.200000}%
\pgfsetfillcolor{currentfill}%
\pgfsetlinewidth{0.803000pt}%
\definecolor{currentstroke}{rgb}{0.200000,0.200000,0.200000}%
\pgfsetstrokecolor{currentstroke}%
\pgfsetdash{}{0pt}%
\pgfsys@defobject{currentmarker}{\pgfqpoint{0.000000in}{-0.048611in}}{\pgfqpoint{0.000000in}{0.000000in}}{%
\pgfpathmoveto{\pgfqpoint{0.000000in}{0.000000in}}%
\pgfpathlineto{\pgfqpoint{0.000000in}{-0.048611in}}%
\pgfusepath{stroke,fill}%
}%
\begin{pgfscope}%
\pgfsys@transformshift{3.652216in}{0.676110in}%
\pgfsys@useobject{currentmarker}{}%
\end{pgfscope}%
\end{pgfscope}%
\begin{pgfscope}%
\definecolor{textcolor}{rgb}{0.200000,0.200000,0.200000}%
\pgfsetstrokecolor{textcolor}%
\pgfsetfillcolor{textcolor}%
\pgftext[x=3.652216in,y=0.578888in,,top]{\color{textcolor}\sffamily\fontsize{11.000000}{13.200000}\selectfont \(\displaystyle {200}\)}%
\end{pgfscope}%
\begin{pgfscope}%
\definecolor{textcolor}{rgb}{0.200000,0.200000,0.200000}%
\pgfsetstrokecolor{textcolor}%
\pgfsetfillcolor{textcolor}%
\pgftext[x=2.357810in,y=0.388148in,,top]{\color{textcolor}\sffamily\fontsize{13.000000}{15.600000}\selectfont Mean arterial pressure}%
\end{pgfscope}%
\begin{pgfscope}%
\pgfsetbuttcap%
\pgfsetroundjoin%
\definecolor{currentfill}{rgb}{0.200000,0.200000,0.200000}%
\pgfsetfillcolor{currentfill}%
\pgfsetlinewidth{0.803000pt}%
\definecolor{currentstroke}{rgb}{0.200000,0.200000,0.200000}%
\pgfsetstrokecolor{currentstroke}%
\pgfsetdash{}{0pt}%
\pgfsys@defobject{currentmarker}{\pgfqpoint{-0.048611in}{0.000000in}}{\pgfqpoint{-0.000000in}{0.000000in}}{%
\pgfpathmoveto{\pgfqpoint{-0.000000in}{0.000000in}}%
\pgfpathlineto{\pgfqpoint{-0.048611in}{0.000000in}}%
\pgfusepath{stroke,fill}%
}%
\begin{pgfscope}%
\pgfsys@transformshift{1.024837in}{0.915984in}%
\pgfsys@useobject{currentmarker}{}%
\end{pgfscope}%
\end{pgfscope}%
\begin{pgfscope}%
\definecolor{textcolor}{rgb}{0.200000,0.200000,0.200000}%
\pgfsetstrokecolor{textcolor}%
\pgfsetfillcolor{textcolor}%
\pgftext[x=0.614999in, y=0.863177in, left, base]{\color{textcolor}\sffamily\fontsize{11.000000}{13.200000}\selectfont \(\displaystyle {\ensuremath{-}0.2}\)}%
\end{pgfscope}%
\begin{pgfscope}%
\pgfsetbuttcap%
\pgfsetroundjoin%
\definecolor{currentfill}{rgb}{0.200000,0.200000,0.200000}%
\pgfsetfillcolor{currentfill}%
\pgfsetlinewidth{0.803000pt}%
\definecolor{currentstroke}{rgb}{0.200000,0.200000,0.200000}%
\pgfsetstrokecolor{currentstroke}%
\pgfsetdash{}{0pt}%
\pgfsys@defobject{currentmarker}{\pgfqpoint{-0.048611in}{0.000000in}}{\pgfqpoint{-0.000000in}{0.000000in}}{%
\pgfpathmoveto{\pgfqpoint{-0.000000in}{0.000000in}}%
\pgfpathlineto{\pgfqpoint{-0.048611in}{0.000000in}}%
\pgfusepath{stroke,fill}%
}%
\begin{pgfscope}%
\pgfsys@transformshift{1.024837in}{1.395136in}%
\pgfsys@useobject{currentmarker}{}%
\end{pgfscope}%
\end{pgfscope}%
\begin{pgfscope}%
\definecolor{textcolor}{rgb}{0.200000,0.200000,0.200000}%
\pgfsetstrokecolor{textcolor}%
\pgfsetfillcolor{textcolor}%
\pgftext[x=0.733286in, y=1.342329in, left, base]{\color{textcolor}\sffamily\fontsize{11.000000}{13.200000}\selectfont \(\displaystyle {0.0}\)}%
\end{pgfscope}%
\begin{pgfscope}%
\definecolor{textcolor}{rgb}{0.200000,0.200000,0.200000}%
\pgfsetstrokecolor{textcolor}%
\pgfsetfillcolor{textcolor}%
\pgftext[x=0.355740in, y=0.690428in, left, base,rotate=90.000000]{\color{textcolor}\sffamily\fontsize{13.000000}{15.600000}\selectfont SHAP value for}%
\end{pgfscope}%
\begin{pgfscope}%
\definecolor{textcolor}{rgb}{0.200000,0.200000,0.200000}%
\pgfsetstrokecolor{textcolor}%
\pgfsetfillcolor{textcolor}%
\pgftext[x=0.527036in, y=0.468247in, left,
base,rotate=90.000000]{\color{textcolor}\sffamily\fontsize{13.000000}{15.600000}\selectfont MAP}%
\end{pgfscope}%
\begin{pgfscope}%
\pgfsetrectcap%
\pgfsetmiterjoin%
\pgfsetlinewidth{0.803000pt}%
\definecolor{currentstroke}{rgb}{0.200000,0.200000,0.200000}%
\pgfsetstrokecolor{currentstroke}%
\pgfsetdash{}{0pt}%
\pgfpathmoveto{\pgfqpoint{1.024837in}{0.676110in}}%
\pgfpathlineto{\pgfqpoint{1.024837in}{1.760000in}}%
\pgfusepath{stroke}%
\end{pgfscope}%
\begin{pgfscope}%
\pgfsetrectcap%
\pgfsetmiterjoin%
\pgfsetlinewidth{0.803000pt}%
\definecolor{currentstroke}{rgb}{0.200000,0.200000,0.200000}%
\pgfsetstrokecolor{currentstroke}%
\pgfsetdash{}{0pt}%
\pgfpathmoveto{\pgfqpoint{1.024837in}{0.676110in}}%
\pgfpathlineto{\pgfqpoint{3.690782in}{0.676110in}}%
\pgfusepath{stroke}%
\end{pgfscope}%
\end{pgfpicture}%
\makeatother%
\endgroup%

%% file: figures/shapley/shapley_j76ft4wm_16h_raw_AUMC_dot.pgf
\begingroup%
\makeatletter%
\begin{pgfpicture}%
\pgfpathrectangle{\pgfpointorigin}{\pgfqpoint{8.000000in}{9.500000in}}%
\pgfusepath{use as bounding box, clip}%
\begin{pgfscope}%
\pgfsetbuttcap%
\pgfsetmiterjoin%
\definecolor{currentfill}{rgb}{1.000000,1.000000,1.000000}%
\pgfsetfillcolor{currentfill}%
\pgfsetlinewidth{0.000000pt}%
\definecolor{currentstroke}{rgb}{1.000000,1.000000,1.000000}%
\pgfsetstrokecolor{currentstroke}%
\pgfsetdash{}{0pt}%
\pgfpathmoveto{\pgfqpoint{0.000000in}{0.000000in}}%
\pgfpathlineto{\pgfqpoint{8.000000in}{0.000000in}}%
\pgfpathlineto{\pgfqpoint{8.000000in}{9.500000in}}%
\pgfpathlineto{\pgfqpoint{0.000000in}{9.500000in}}%
\pgfpathclose%
\pgfusepath{fill}%
\end{pgfscope}%
\begin{pgfscope}%
\pgfsetbuttcap%
\pgfsetmiterjoin%
\definecolor{currentfill}{rgb}{1.000000,1.000000,1.000000}%
\pgfsetfillcolor{currentfill}%
\pgfsetlinewidth{0.000000pt}%
\definecolor{currentstroke}{rgb}{0.000000,0.000000,0.000000}%
\pgfsetstrokecolor{currentstroke}%
\pgfsetstrokeopacity{0.000000}%
\pgfsetdash{}{0pt}%
\pgfpathmoveto{\pgfqpoint{2.400626in}{0.694630in}}%
\pgfpathlineto{\pgfqpoint{6.688125in}{0.694630in}}%
\pgfpathlineto{\pgfqpoint{6.688125in}{9.207193in}}%
\pgfpathlineto{\pgfqpoint{2.400626in}{9.207193in}}%
\pgfpathclose%
\pgfusepath{fill}%
\end{pgfscope}%
\begin{pgfscope}%
\pgfpathrectangle{\pgfqpoint{2.400626in}{0.694630in}}{\pgfqpoint{4.287499in}{8.512564in}}%
\pgfusepath{clip}%
\pgfsetrectcap%
\pgfsetroundjoin%
\pgfsetlinewidth{1.505625pt}%
\definecolor{currentstroke}{rgb}{0.600000,0.600000,0.600000}%
\pgfsetstrokecolor{currentstroke}%
\pgfsetdash{}{0pt}%
\pgfpathmoveto{\pgfqpoint{3.923913in}{0.694630in}}%
\pgfpathlineto{\pgfqpoint{3.923913in}{9.207193in}}%
\pgfusepath{stroke}%
\end{pgfscope}%
\begin{pgfscope}%
\pgfpathrectangle{\pgfqpoint{2.400626in}{0.694630in}}{\pgfqpoint{4.287499in}{8.512564in}}%
\pgfusepath{clip}%
\pgfsetbuttcap%
\pgfsetroundjoin%
\pgfsetlinewidth{0.501875pt}%
\definecolor{currentstroke}{rgb}{0.800000,0.800000,0.800000}%
\pgfsetstrokecolor{currentstroke}%
\pgfsetdash{{0.500000pt}{2.500000pt}}{0.000000pt}%
\pgfpathmoveto{\pgfqpoint{2.400626in}{1.099990in}}%
\pgfpathlineto{\pgfqpoint{6.688125in}{1.099990in}}%
\pgfusepath{stroke}%
\end{pgfscope}%
\begin{pgfscope}%
\pgfpathrectangle{\pgfqpoint{2.400626in}{0.694630in}}{\pgfqpoint{4.287499in}{8.512564in}}%
\pgfusepath{clip}%
\pgfsetbuttcap%
\pgfsetroundjoin%
\pgfsetlinewidth{0.501875pt}%
\definecolor{currentstroke}{rgb}{0.800000,0.800000,0.800000}%
\pgfsetstrokecolor{currentstroke}%
\pgfsetdash{{0.500000pt}{2.500000pt}}{0.000000pt}%
\pgfpathmoveto{\pgfqpoint{2.400626in}{1.505350in}}%
\pgfpathlineto{\pgfqpoint{6.688125in}{1.505350in}}%
\pgfusepath{stroke}%
\end{pgfscope}%
\begin{pgfscope}%
\pgfpathrectangle{\pgfqpoint{2.400626in}{0.694630in}}{\pgfqpoint{4.287499in}{8.512564in}}%
\pgfusepath{clip}%
\pgfsetbuttcap%
\pgfsetroundjoin%
\pgfsetlinewidth{0.501875pt}%
\definecolor{currentstroke}{rgb}{0.800000,0.800000,0.800000}%
\pgfsetstrokecolor{currentstroke}%
\pgfsetdash{{0.500000pt}{2.500000pt}}{0.000000pt}%
\pgfpathmoveto{\pgfqpoint{2.400626in}{1.910710in}}%
\pgfpathlineto{\pgfqpoint{6.688125in}{1.910710in}}%
\pgfusepath{stroke}%
\end{pgfscope}%
\begin{pgfscope}%
\pgfpathrectangle{\pgfqpoint{2.400626in}{0.694630in}}{\pgfqpoint{4.287499in}{8.512564in}}%
\pgfusepath{clip}%
\pgfsetbuttcap%
\pgfsetroundjoin%
\pgfsetlinewidth{0.501875pt}%
\definecolor{currentstroke}{rgb}{0.800000,0.800000,0.800000}%
\pgfsetstrokecolor{currentstroke}%
\pgfsetdash{{0.500000pt}{2.500000pt}}{0.000000pt}%
\pgfpathmoveto{\pgfqpoint{2.400626in}{2.316070in}}%
\pgfpathlineto{\pgfqpoint{6.688125in}{2.316070in}}%
\pgfusepath{stroke}%
\end{pgfscope}%
\begin{pgfscope}%
\pgfpathrectangle{\pgfqpoint{2.400626in}{0.694630in}}{\pgfqpoint{4.287499in}{8.512564in}}%
\pgfusepath{clip}%
\pgfsetbuttcap%
\pgfsetroundjoin%
\pgfsetlinewidth{0.501875pt}%
\definecolor{currentstroke}{rgb}{0.800000,0.800000,0.800000}%
\pgfsetstrokecolor{currentstroke}%
\pgfsetdash{{0.500000pt}{2.500000pt}}{0.000000pt}%
\pgfpathmoveto{\pgfqpoint{2.400626in}{2.721430in}}%
\pgfpathlineto{\pgfqpoint{6.688125in}{2.721430in}}%
\pgfusepath{stroke}%
\end{pgfscope}%
\begin{pgfscope}%
\pgfpathrectangle{\pgfqpoint{2.400626in}{0.694630in}}{\pgfqpoint{4.287499in}{8.512564in}}%
\pgfusepath{clip}%
\pgfsetbuttcap%
\pgfsetroundjoin%
\pgfsetlinewidth{0.501875pt}%
\definecolor{currentstroke}{rgb}{0.800000,0.800000,0.800000}%
\pgfsetstrokecolor{currentstroke}%
\pgfsetdash{{0.500000pt}{2.500000pt}}{0.000000pt}%
\pgfpathmoveto{\pgfqpoint{2.400626in}{3.126791in}}%
\pgfpathlineto{\pgfqpoint{6.688125in}{3.126791in}}%
\pgfusepath{stroke}%
\end{pgfscope}%
\begin{pgfscope}%
\pgfpathrectangle{\pgfqpoint{2.400626in}{0.694630in}}{\pgfqpoint{4.287499in}{8.512564in}}%
\pgfusepath{clip}%
\pgfsetbuttcap%
\pgfsetroundjoin%
\pgfsetlinewidth{0.501875pt}%
\definecolor{currentstroke}{rgb}{0.800000,0.800000,0.800000}%
\pgfsetstrokecolor{currentstroke}%
\pgfsetdash{{0.500000pt}{2.500000pt}}{0.000000pt}%
\pgfpathmoveto{\pgfqpoint{2.400626in}{3.532151in}}%
\pgfpathlineto{\pgfqpoint{6.688125in}{3.532151in}}%
\pgfusepath{stroke}%
\end{pgfscope}%
\begin{pgfscope}%
\pgfpathrectangle{\pgfqpoint{2.400626in}{0.694630in}}{\pgfqpoint{4.287499in}{8.512564in}}%
\pgfusepath{clip}%
\pgfsetbuttcap%
\pgfsetroundjoin%
\pgfsetlinewidth{0.501875pt}%
\definecolor{currentstroke}{rgb}{0.800000,0.800000,0.800000}%
\pgfsetstrokecolor{currentstroke}%
\pgfsetdash{{0.500000pt}{2.500000pt}}{0.000000pt}%
\pgfpathmoveto{\pgfqpoint{2.400626in}{3.937511in}}%
\pgfpathlineto{\pgfqpoint{6.688125in}{3.937511in}}%
\pgfusepath{stroke}%
\end{pgfscope}%
\begin{pgfscope}%
\pgfpathrectangle{\pgfqpoint{2.400626in}{0.694630in}}{\pgfqpoint{4.287499in}{8.512564in}}%
\pgfusepath{clip}%
\pgfsetbuttcap%
\pgfsetroundjoin%
\pgfsetlinewidth{0.501875pt}%
\definecolor{currentstroke}{rgb}{0.800000,0.800000,0.800000}%
\pgfsetstrokecolor{currentstroke}%
\pgfsetdash{{0.500000pt}{2.500000pt}}{0.000000pt}%
\pgfpathmoveto{\pgfqpoint{2.400626in}{4.342871in}}%
\pgfpathlineto{\pgfqpoint{6.688125in}{4.342871in}}%
\pgfusepath{stroke}%
\end{pgfscope}%
\begin{pgfscope}%
\pgfpathrectangle{\pgfqpoint{2.400626in}{0.694630in}}{\pgfqpoint{4.287499in}{8.512564in}}%
\pgfusepath{clip}%
\pgfsetbuttcap%
\pgfsetroundjoin%
\pgfsetlinewidth{0.501875pt}%
\definecolor{currentstroke}{rgb}{0.800000,0.800000,0.800000}%
\pgfsetstrokecolor{currentstroke}%
\pgfsetdash{{0.500000pt}{2.500000pt}}{0.000000pt}%
\pgfpathmoveto{\pgfqpoint{2.400626in}{4.748231in}}%
\pgfpathlineto{\pgfqpoint{6.688125in}{4.748231in}}%
\pgfusepath{stroke}%
\end{pgfscope}%
\begin{pgfscope}%
\pgfpathrectangle{\pgfqpoint{2.400626in}{0.694630in}}{\pgfqpoint{4.287499in}{8.512564in}}%
\pgfusepath{clip}%
\pgfsetbuttcap%
\pgfsetroundjoin%
\pgfsetlinewidth{0.501875pt}%
\definecolor{currentstroke}{rgb}{0.800000,0.800000,0.800000}%
\pgfsetstrokecolor{currentstroke}%
\pgfsetdash{{0.500000pt}{2.500000pt}}{0.000000pt}%
\pgfpathmoveto{\pgfqpoint{2.400626in}{5.153592in}}%
\pgfpathlineto{\pgfqpoint{6.688125in}{5.153592in}}%
\pgfusepath{stroke}%
\end{pgfscope}%
\begin{pgfscope}%
\pgfpathrectangle{\pgfqpoint{2.400626in}{0.694630in}}{\pgfqpoint{4.287499in}{8.512564in}}%
\pgfusepath{clip}%
\pgfsetbuttcap%
\pgfsetroundjoin%
\pgfsetlinewidth{0.501875pt}%
\definecolor{currentstroke}{rgb}{0.800000,0.800000,0.800000}%
\pgfsetstrokecolor{currentstroke}%
\pgfsetdash{{0.500000pt}{2.500000pt}}{0.000000pt}%
\pgfpathmoveto{\pgfqpoint{2.400626in}{5.558952in}}%
\pgfpathlineto{\pgfqpoint{6.688125in}{5.558952in}}%
\pgfusepath{stroke}%
\end{pgfscope}%
\begin{pgfscope}%
\pgfpathrectangle{\pgfqpoint{2.400626in}{0.694630in}}{\pgfqpoint{4.287499in}{8.512564in}}%
\pgfusepath{clip}%
\pgfsetbuttcap%
\pgfsetroundjoin%
\pgfsetlinewidth{0.501875pt}%
\definecolor{currentstroke}{rgb}{0.800000,0.800000,0.800000}%
\pgfsetstrokecolor{currentstroke}%
\pgfsetdash{{0.500000pt}{2.500000pt}}{0.000000pt}%
\pgfpathmoveto{\pgfqpoint{2.400626in}{5.964312in}}%
\pgfpathlineto{\pgfqpoint{6.688125in}{5.964312in}}%
\pgfusepath{stroke}%
\end{pgfscope}%
\begin{pgfscope}%
\pgfpathrectangle{\pgfqpoint{2.400626in}{0.694630in}}{\pgfqpoint{4.287499in}{8.512564in}}%
\pgfusepath{clip}%
\pgfsetbuttcap%
\pgfsetroundjoin%
\pgfsetlinewidth{0.501875pt}%
\definecolor{currentstroke}{rgb}{0.800000,0.800000,0.800000}%
\pgfsetstrokecolor{currentstroke}%
\pgfsetdash{{0.500000pt}{2.500000pt}}{0.000000pt}%
\pgfpathmoveto{\pgfqpoint{2.400626in}{6.369672in}}%
\pgfpathlineto{\pgfqpoint{6.688125in}{6.369672in}}%
\pgfusepath{stroke}%
\end{pgfscope}%
\begin{pgfscope}%
\pgfpathrectangle{\pgfqpoint{2.400626in}{0.694630in}}{\pgfqpoint{4.287499in}{8.512564in}}%
\pgfusepath{clip}%
\pgfsetbuttcap%
\pgfsetroundjoin%
\pgfsetlinewidth{0.501875pt}%
\definecolor{currentstroke}{rgb}{0.800000,0.800000,0.800000}%
\pgfsetstrokecolor{currentstroke}%
\pgfsetdash{{0.500000pt}{2.500000pt}}{0.000000pt}%
\pgfpathmoveto{\pgfqpoint{2.400626in}{6.775032in}}%
\pgfpathlineto{\pgfqpoint{6.688125in}{6.775032in}}%
\pgfusepath{stroke}%
\end{pgfscope}%
\begin{pgfscope}%
\pgfpathrectangle{\pgfqpoint{2.400626in}{0.694630in}}{\pgfqpoint{4.287499in}{8.512564in}}%
\pgfusepath{clip}%
\pgfsetbuttcap%
\pgfsetroundjoin%
\pgfsetlinewidth{0.501875pt}%
\definecolor{currentstroke}{rgb}{0.800000,0.800000,0.800000}%
\pgfsetstrokecolor{currentstroke}%
\pgfsetdash{{0.500000pt}{2.500000pt}}{0.000000pt}%
\pgfpathmoveto{\pgfqpoint{2.400626in}{7.180392in}}%
\pgfpathlineto{\pgfqpoint{6.688125in}{7.180392in}}%
\pgfusepath{stroke}%
\end{pgfscope}%
\begin{pgfscope}%
\pgfpathrectangle{\pgfqpoint{2.400626in}{0.694630in}}{\pgfqpoint{4.287499in}{8.512564in}}%
\pgfusepath{clip}%
\pgfsetbuttcap%
\pgfsetroundjoin%
\pgfsetlinewidth{0.501875pt}%
\definecolor{currentstroke}{rgb}{0.800000,0.800000,0.800000}%
\pgfsetstrokecolor{currentstroke}%
\pgfsetdash{{0.500000pt}{2.500000pt}}{0.000000pt}%
\pgfpathmoveto{\pgfqpoint{2.400626in}{7.585753in}}%
\pgfpathlineto{\pgfqpoint{6.688125in}{7.585753in}}%
\pgfusepath{stroke}%
\end{pgfscope}%
\begin{pgfscope}%
\pgfpathrectangle{\pgfqpoint{2.400626in}{0.694630in}}{\pgfqpoint{4.287499in}{8.512564in}}%
\pgfusepath{clip}%
\pgfsetbuttcap%
\pgfsetroundjoin%
\pgfsetlinewidth{0.501875pt}%
\definecolor{currentstroke}{rgb}{0.800000,0.800000,0.800000}%
\pgfsetstrokecolor{currentstroke}%
\pgfsetdash{{0.500000pt}{2.500000pt}}{0.000000pt}%
\pgfpathmoveto{\pgfqpoint{2.400626in}{7.991113in}}%
\pgfpathlineto{\pgfqpoint{6.688125in}{7.991113in}}%
\pgfusepath{stroke}%
\end{pgfscope}%
\begin{pgfscope}%
\pgfpathrectangle{\pgfqpoint{2.400626in}{0.694630in}}{\pgfqpoint{4.287499in}{8.512564in}}%
\pgfusepath{clip}%
\pgfsetbuttcap%
\pgfsetroundjoin%
\pgfsetlinewidth{0.501875pt}%
\definecolor{currentstroke}{rgb}{0.800000,0.800000,0.800000}%
\pgfsetstrokecolor{currentstroke}%
\pgfsetdash{{0.500000pt}{2.500000pt}}{0.000000pt}%
\pgfpathmoveto{\pgfqpoint{2.400626in}{8.396473in}}%
\pgfpathlineto{\pgfqpoint{6.688125in}{8.396473in}}%
\pgfusepath{stroke}%
\end{pgfscope}%
\begin{pgfscope}%
\pgfpathrectangle{\pgfqpoint{2.400626in}{0.694630in}}{\pgfqpoint{4.287499in}{8.512564in}}%
\pgfusepath{clip}%
\pgfsetbuttcap%
\pgfsetroundjoin%
\pgfsetlinewidth{0.501875pt}%
\definecolor{currentstroke}{rgb}{0.800000,0.800000,0.800000}%
\pgfsetstrokecolor{currentstroke}%
\pgfsetdash{{0.500000pt}{2.500000pt}}{0.000000pt}%
\pgfpathmoveto{\pgfqpoint{2.400626in}{8.801833in}}%
\pgfpathlineto{\pgfqpoint{6.688125in}{8.801833in}}%
\pgfusepath{stroke}%
\end{pgfscope}%
\begin{pgfscope}%
\pgfsetbuttcap%
\pgfsetroundjoin%
\definecolor{currentfill}{rgb}{0.200000,0.200000,0.200000}%
\pgfsetfillcolor{currentfill}%
\pgfsetlinewidth{0.803000pt}%
\definecolor{currentstroke}{rgb}{0.200000,0.200000,0.200000}%
\pgfsetstrokecolor{currentstroke}%
\pgfsetdash{}{0pt}%
\pgfsys@defobject{currentmarker}{\pgfqpoint{0.000000in}{-0.048611in}}{\pgfqpoint{0.000000in}{0.000000in}}{%
\pgfpathmoveto{\pgfqpoint{0.000000in}{0.000000in}}%
\pgfpathlineto{\pgfqpoint{0.000000in}{-0.048611in}}%
\pgfusepath{stroke,fill}%
}%
\begin{pgfscope}%
\pgfsys@transformshift{2.762179in}{0.694630in}%
\pgfsys@useobject{currentmarker}{}%
\end{pgfscope}%
\end{pgfscope}%
\begin{pgfscope}%
\definecolor{textcolor}{rgb}{0.200000,0.200000,0.200000}%
\pgfsetstrokecolor{textcolor}%
\pgfsetfillcolor{textcolor}%
\pgftext[x=2.762179in,y=0.597407in,,top]{\color{textcolor}\sffamily\fontsize{11.000000}{13.200000}\selectfont \(\displaystyle {\ensuremath{-}0.50}\)}%
\end{pgfscope}%
\begin{pgfscope}%
\pgfsetbuttcap%
\pgfsetroundjoin%
\definecolor{currentfill}{rgb}{0.200000,0.200000,0.200000}%
\pgfsetfillcolor{currentfill}%
\pgfsetlinewidth{0.803000pt}%
\definecolor{currentstroke}{rgb}{0.200000,0.200000,0.200000}%
\pgfsetstrokecolor{currentstroke}%
\pgfsetdash{}{0pt}%
\pgfsys@defobject{currentmarker}{\pgfqpoint{0.000000in}{-0.048611in}}{\pgfqpoint{0.000000in}{0.000000in}}{%
\pgfpathmoveto{\pgfqpoint{0.000000in}{0.000000in}}%
\pgfpathlineto{\pgfqpoint{0.000000in}{-0.048611in}}%
\pgfusepath{stroke,fill}%
}%
\begin{pgfscope}%
\pgfsys@transformshift{3.343046in}{0.694630in}%
\pgfsys@useobject{currentmarker}{}%
\end{pgfscope}%
\end{pgfscope}%
\begin{pgfscope}%
\definecolor{textcolor}{rgb}{0.200000,0.200000,0.200000}%
\pgfsetstrokecolor{textcolor}%
\pgfsetfillcolor{textcolor}%
\pgftext[x=3.343046in,y=0.597407in,,top]{\color{textcolor}\sffamily\fontsize{11.000000}{13.200000}\selectfont \(\displaystyle {\ensuremath{-}0.25}\)}%
\end{pgfscope}%
\begin{pgfscope}%
\pgfsetbuttcap%
\pgfsetroundjoin%
\definecolor{currentfill}{rgb}{0.200000,0.200000,0.200000}%
\pgfsetfillcolor{currentfill}%
\pgfsetlinewidth{0.803000pt}%
\definecolor{currentstroke}{rgb}{0.200000,0.200000,0.200000}%
\pgfsetstrokecolor{currentstroke}%
\pgfsetdash{}{0pt}%
\pgfsys@defobject{currentmarker}{\pgfqpoint{0.000000in}{-0.048611in}}{\pgfqpoint{0.000000in}{0.000000in}}{%
\pgfpathmoveto{\pgfqpoint{0.000000in}{0.000000in}}%
\pgfpathlineto{\pgfqpoint{0.000000in}{-0.048611in}}%
\pgfusepath{stroke,fill}%
}%
\begin{pgfscope}%
\pgfsys@transformshift{3.923913in}{0.694630in}%
\pgfsys@useobject{currentmarker}{}%
\end{pgfscope}%
\end{pgfscope}%
\begin{pgfscope}%
\definecolor{textcolor}{rgb}{0.200000,0.200000,0.200000}%
\pgfsetstrokecolor{textcolor}%
\pgfsetfillcolor{textcolor}%
\pgftext[x=3.923913in,y=0.597407in,,top]{\color{textcolor}\sffamily\fontsize{11.000000}{13.200000}\selectfont \(\displaystyle {0.00}\)}%
\end{pgfscope}%
\begin{pgfscope}%
\pgfsetbuttcap%
\pgfsetroundjoin%
\definecolor{currentfill}{rgb}{0.200000,0.200000,0.200000}%
\pgfsetfillcolor{currentfill}%
\pgfsetlinewidth{0.803000pt}%
\definecolor{currentstroke}{rgb}{0.200000,0.200000,0.200000}%
\pgfsetstrokecolor{currentstroke}%
\pgfsetdash{}{0pt}%
\pgfsys@defobject{currentmarker}{\pgfqpoint{0.000000in}{-0.048611in}}{\pgfqpoint{0.000000in}{0.000000in}}{%
\pgfpathmoveto{\pgfqpoint{0.000000in}{0.000000in}}%
\pgfpathlineto{\pgfqpoint{0.000000in}{-0.048611in}}%
\pgfusepath{stroke,fill}%
}%
\begin{pgfscope}%
\pgfsys@transformshift{4.504781in}{0.694630in}%
\pgfsys@useobject{currentmarker}{}%
\end{pgfscope}%
\end{pgfscope}%
\begin{pgfscope}%
\definecolor{textcolor}{rgb}{0.200000,0.200000,0.200000}%
\pgfsetstrokecolor{textcolor}%
\pgfsetfillcolor{textcolor}%
\pgftext[x=4.504781in,y=0.597407in,,top]{\color{textcolor}\sffamily\fontsize{11.000000}{13.200000}\selectfont \(\displaystyle {0.25}\)}%
\end{pgfscope}%
\begin{pgfscope}%
\pgfsetbuttcap%
\pgfsetroundjoin%
\definecolor{currentfill}{rgb}{0.200000,0.200000,0.200000}%
\pgfsetfillcolor{currentfill}%
\pgfsetlinewidth{0.803000pt}%
\definecolor{currentstroke}{rgb}{0.200000,0.200000,0.200000}%
\pgfsetstrokecolor{currentstroke}%
\pgfsetdash{}{0pt}%
\pgfsys@defobject{currentmarker}{\pgfqpoint{0.000000in}{-0.048611in}}{\pgfqpoint{0.000000in}{0.000000in}}{%
\pgfpathmoveto{\pgfqpoint{0.000000in}{0.000000in}}%
\pgfpathlineto{\pgfqpoint{0.000000in}{-0.048611in}}%
\pgfusepath{stroke,fill}%
}%
\begin{pgfscope}%
\pgfsys@transformshift{5.085648in}{0.694630in}%
\pgfsys@useobject{currentmarker}{}%
\end{pgfscope}%
\end{pgfscope}%
\begin{pgfscope}%
\definecolor{textcolor}{rgb}{0.200000,0.200000,0.200000}%
\pgfsetstrokecolor{textcolor}%
\pgfsetfillcolor{textcolor}%
\pgftext[x=5.085648in,y=0.597407in,,top]{\color{textcolor}\sffamily\fontsize{11.000000}{13.200000}\selectfont \(\displaystyle {0.50}\)}%
\end{pgfscope}%
\begin{pgfscope}%
\pgfsetbuttcap%
\pgfsetroundjoin%
\definecolor{currentfill}{rgb}{0.200000,0.200000,0.200000}%
\pgfsetfillcolor{currentfill}%
\pgfsetlinewidth{0.803000pt}%
\definecolor{currentstroke}{rgb}{0.200000,0.200000,0.200000}%
\pgfsetstrokecolor{currentstroke}%
\pgfsetdash{}{0pt}%
\pgfsys@defobject{currentmarker}{\pgfqpoint{0.000000in}{-0.048611in}}{\pgfqpoint{0.000000in}{0.000000in}}{%
\pgfpathmoveto{\pgfqpoint{0.000000in}{0.000000in}}%
\pgfpathlineto{\pgfqpoint{0.000000in}{-0.048611in}}%
\pgfusepath{stroke,fill}%
}%
\begin{pgfscope}%
\pgfsys@transformshift{5.666515in}{0.694630in}%
\pgfsys@useobject{currentmarker}{}%
\end{pgfscope}%
\end{pgfscope}%
\begin{pgfscope}%
\definecolor{textcolor}{rgb}{0.200000,0.200000,0.200000}%
\pgfsetstrokecolor{textcolor}%
\pgfsetfillcolor{textcolor}%
\pgftext[x=5.666515in,y=0.597407in,,top]{\color{textcolor}\sffamily\fontsize{11.000000}{13.200000}\selectfont \(\displaystyle {0.75}\)}%
\end{pgfscope}%
\begin{pgfscope}%
\pgfsetbuttcap%
\pgfsetroundjoin%
\definecolor{currentfill}{rgb}{0.200000,0.200000,0.200000}%
\pgfsetfillcolor{currentfill}%
\pgfsetlinewidth{0.803000pt}%
\definecolor{currentstroke}{rgb}{0.200000,0.200000,0.200000}%
\pgfsetstrokecolor{currentstroke}%
\pgfsetdash{}{0pt}%
\pgfsys@defobject{currentmarker}{\pgfqpoint{0.000000in}{-0.048611in}}{\pgfqpoint{0.000000in}{0.000000in}}{%
\pgfpathmoveto{\pgfqpoint{0.000000in}{0.000000in}}%
\pgfpathlineto{\pgfqpoint{0.000000in}{-0.048611in}}%
\pgfusepath{stroke,fill}%
}%
\begin{pgfscope}%
\pgfsys@transformshift{6.247382in}{0.694630in}%
\pgfsys@useobject{currentmarker}{}%
\end{pgfscope}%
\end{pgfscope}%
\begin{pgfscope}%
\definecolor{textcolor}{rgb}{0.200000,0.200000,0.200000}%
\pgfsetstrokecolor{textcolor}%
\pgfsetfillcolor{textcolor}%
\pgftext[x=6.247382in,y=0.597407in,,top]{\color{textcolor}\sffamily\fontsize{11.000000}{13.200000}\selectfont \(\displaystyle {1.00}\)}%
\end{pgfscope}%
\begin{pgfscope}%
\definecolor{textcolor}{rgb}{0.000000,0.000000,0.000000}%
\pgfsetstrokecolor{textcolor}%
\pgfsetfillcolor{textcolor}%
\pgftext[x=4.544375in,y=0.406667in,,top]{\color{textcolor}\sffamily\fontsize{13.000000}{15.600000}\selectfont SHAP value (impact on model output)}%
\end{pgfscope}%
\begin{pgfscope}%
\definecolor{textcolor}{rgb}{0.200000,0.200000,0.200000}%
\pgfsetstrokecolor{textcolor}%
\pgfsetfillcolor{textcolor}%
\pgftext[x=1.566912in, y=1.042120in, left, base]{\color{textcolor}\sffamily\fontsize{13.000000}{15.600000}\selectfont Lactate}%
\end{pgfscope}%
\begin{pgfscope}%
\definecolor{textcolor}{rgb}{0.200000,0.200000,0.200000}%
\pgfsetstrokecolor{textcolor}%
\pgfsetfillcolor{textcolor}%
\pgftext[x=1.280959in, y=1.447480in, left, base]{\color{textcolor}\sffamily\fontsize{13.000000}{15.600000}\selectfont Base excess}%
\end{pgfscope}%
\begin{pgfscope}%
\definecolor{textcolor}{rgb}{0.200000,0.200000,0.200000}%
\pgfsetstrokecolor{textcolor}%
\pgfsetfillcolor{textcolor}%
\pgftext[x=1.309778in, y=1.852840in, left, base]{\color{textcolor}\sffamily\fontsize{13.000000}{15.600000}\selectfont Hematocrit}%
\end{pgfscope}%
\begin{pgfscope}%
\definecolor{textcolor}{rgb}{0.200000,0.200000,0.200000}%
\pgfsetstrokecolor{textcolor}%
\pgfsetfillcolor{textcolor}%
\pgftext[x=0.688624in, y=2.258200in, left, base]{\color{textcolor}\sffamily\fontsize{13.000000}{15.600000}\selectfont CO2 partial pressure}%
\end{pgfscope}%
\begin{pgfscope}%
\definecolor{textcolor}{rgb}{0.200000,0.200000,0.200000}%
\pgfsetstrokecolor{textcolor}%
\pgfsetfillcolor{textcolor}%
\pgftext[x=0.525663in, y=2.663560in, left, base]{\color{textcolor}\sffamily\fontsize{13.000000}{15.600000}\selectfont Systolic blood pressure}%
\end{pgfscope}%
\begin{pgfscope}%
\definecolor{textcolor}{rgb}{0.200000,0.200000,0.200000}%
\pgfsetstrokecolor{textcolor}%
\pgfsetfillcolor{textcolor}%
\pgftext[x=1.237731in, y=3.068920in, left, base]{\color{textcolor}\sffamily\fontsize{13.000000}{15.600000}\selectfont PH of blood}%
\end{pgfscope}%
\begin{pgfscope}%
\definecolor{textcolor}{rgb}{0.200000,0.200000,0.200000}%
\pgfsetstrokecolor{textcolor}%
\pgfsetfillcolor{textcolor}%
\pgftext[x=1.117748in, y=3.474281in, left, base]{\color{textcolor}\sffamily\fontsize{13.000000}{15.600000}\selectfont Platelet count}%
\end{pgfscope}%
\begin{pgfscope}%
\definecolor{textcolor}{rgb}{0.200000,0.200000,0.200000}%
\pgfsetstrokecolor{textcolor}%
\pgfsetfillcolor{textcolor}%
\pgftext[x=0.792886in, y=3.879641in, left, base]{\color{textcolor}\sffamily\fontsize{13.000000}{15.600000}\selectfont O2 partial pressure}%
\end{pgfscope}%
\begin{pgfscope}%
\definecolor{textcolor}{rgb}{0.200000,0.200000,0.200000}%
\pgfsetstrokecolor{textcolor}%
\pgfsetfillcolor{textcolor}%
\pgftext[x=0.868232in, y=4.285001in, left, base]{\color{textcolor}\sffamily\fontsize{13.000000}{15.600000}\selectfont C-reactive protein}%
\end{pgfscope}%
\begin{pgfscope}%
\definecolor{textcolor}{rgb}{0.200000,0.200000,0.200000}%
\pgfsetstrokecolor{textcolor}%
\pgfsetfillcolor{textcolor}%
\pgftext[x=1.269270in, y=4.690361in, left, base]{\color{textcolor}\sffamily\fontsize{13.000000}{15.600000}\selectfont Hemoglobin}%
\end{pgfscope}%
\begin{pgfscope}%
\definecolor{textcolor}{rgb}{0.200000,0.200000,0.200000}%
\pgfsetstrokecolor{textcolor}%
\pgfsetfillcolor{textcolor}%
\pgftext[x=1.379896in, y=5.095721in, left, base]{\color{textcolor}\sffamily\fontsize{13.000000}{15.600000}\selectfont Heart rate}%
\end{pgfscope}%
\begin{pgfscope}%
\definecolor{textcolor}{rgb}{0.200000,0.200000,0.200000}%
\pgfsetstrokecolor{textcolor}%
\pgfsetfillcolor{textcolor}%
\pgftext[x=1.208024in, y=5.501082in, left, base]{\color{textcolor}\sffamily\fontsize{13.000000}{15.600000}\selectfont Urine output}%
\end{pgfscope}%
\begin{pgfscope}%
\definecolor{textcolor}{rgb}{0.200000,0.200000,0.200000}%
\pgfsetstrokecolor{textcolor}%
\pgfsetfillcolor{textcolor}%
\pgftext[x=1.562573in, y=5.906442in, left, base]{\color{textcolor}\sffamily\fontsize{13.000000}{15.600000}\selectfont Sodium}%
\end{pgfscope}%
\begin{pgfscope}%
\definecolor{textcolor}{rgb}{0.200000,0.200000,0.200000}%
\pgfsetstrokecolor{textcolor}%
\pgfsetfillcolor{textcolor}%
\pgftext[x=1.209953in, y=6.311802in, left, base]{\color{textcolor}\sffamily\fontsize{13.000000}{15.600000}\selectfont Temperature}%
\end{pgfscope}%
\begin{pgfscope}%
\definecolor{textcolor}{rgb}{0.200000,0.200000,0.200000}%
\pgfsetstrokecolor{textcolor}%
\pgfsetfillcolor{textcolor}%
\pgftext[x=1.554123in, y=6.717162in, left, base]{\color{textcolor}\sffamily\fontsize{13.000000}{15.600000}\selectfont Glucose}%
\end{pgfscope}%
\begin{pgfscope}%
\definecolor{textcolor}{rgb}{0.200000,0.200000,0.200000}%
\pgfsetstrokecolor{textcolor}%
\pgfsetfillcolor{textcolor}%
\pgftext[x=0.457088in, y=7.122522in, left, base]{\color{textcolor}\sffamily\fontsize{13.000000}{15.600000}\selectfont Diastolic blood pressure}%
\end{pgfscope}%
\begin{pgfscope}%
\definecolor{textcolor}{rgb}{0.200000,0.200000,0.200000}%
\pgfsetstrokecolor{textcolor}%
\pgfsetfillcolor{textcolor}%
\pgftext[x=1.006253in, y=7.527882in, left, base]{\color{textcolor}\sffamily\fontsize{13.000000}{15.600000}\selectfont Calcium ionized}%
\end{pgfscope}%
\begin{pgfscope}%
\definecolor{textcolor}{rgb}{0.200000,0.200000,0.200000}%
\pgfsetstrokecolor{textcolor}%
\pgfsetfillcolor{textcolor}%
\pgftext[x=0.240000in, y=7.933243in, left, base]{\color{textcolor}\sffamily\fontsize{13.000000}{15.600000}\selectfont Fraction of inspired oxygen}%
\end{pgfscope}%
\begin{pgfscope}%
\definecolor{textcolor}{rgb}{0.200000,0.200000,0.200000}%
\pgfsetstrokecolor{textcolor}%
\pgfsetfillcolor{textcolor}%
\pgftext[x=0.574621in, y=8.338603in, left, base]{\color{textcolor}\sffamily\fontsize{13.000000}{15.600000}\selectfont Mean arterial pressure}%
\end{pgfscope}%
\begin{pgfscope}%
\definecolor{textcolor}{rgb}{0.200000,0.200000,0.200000}%
\pgfsetstrokecolor{textcolor}%
\pgfsetfillcolor{textcolor}%
\pgftext[x=0.992884in, y=8.743963in, left, base]{\color{textcolor}\sffamily\fontsize{13.000000}{15.600000}\selectfont Respiratory rate}%
\end{pgfscope}%
\begin{pgfscope}%
\pgfsetrectcap%
\pgfsetmiterjoin%
\pgfsetlinewidth{0.803000pt}%
\definecolor{currentstroke}{rgb}{0.000000,0.000000,0.000000}%
\pgfsetstrokecolor{currentstroke}%
\pgfsetdash{}{0pt}%
\pgfpathmoveto{\pgfqpoint{2.400626in}{0.694630in}}%
\pgfpathlineto{\pgfqpoint{6.688125in}{0.694630in}}%
\pgfusepath{stroke}%
\end{pgfscope}%
\begin{pgfscope}%
\pgfsys@transformshift{2.566667in}{0.923333in}%
\pgftext[left,bottom]{\includegraphics[interpolate=true,width=3.956667in,height=8.053333in]{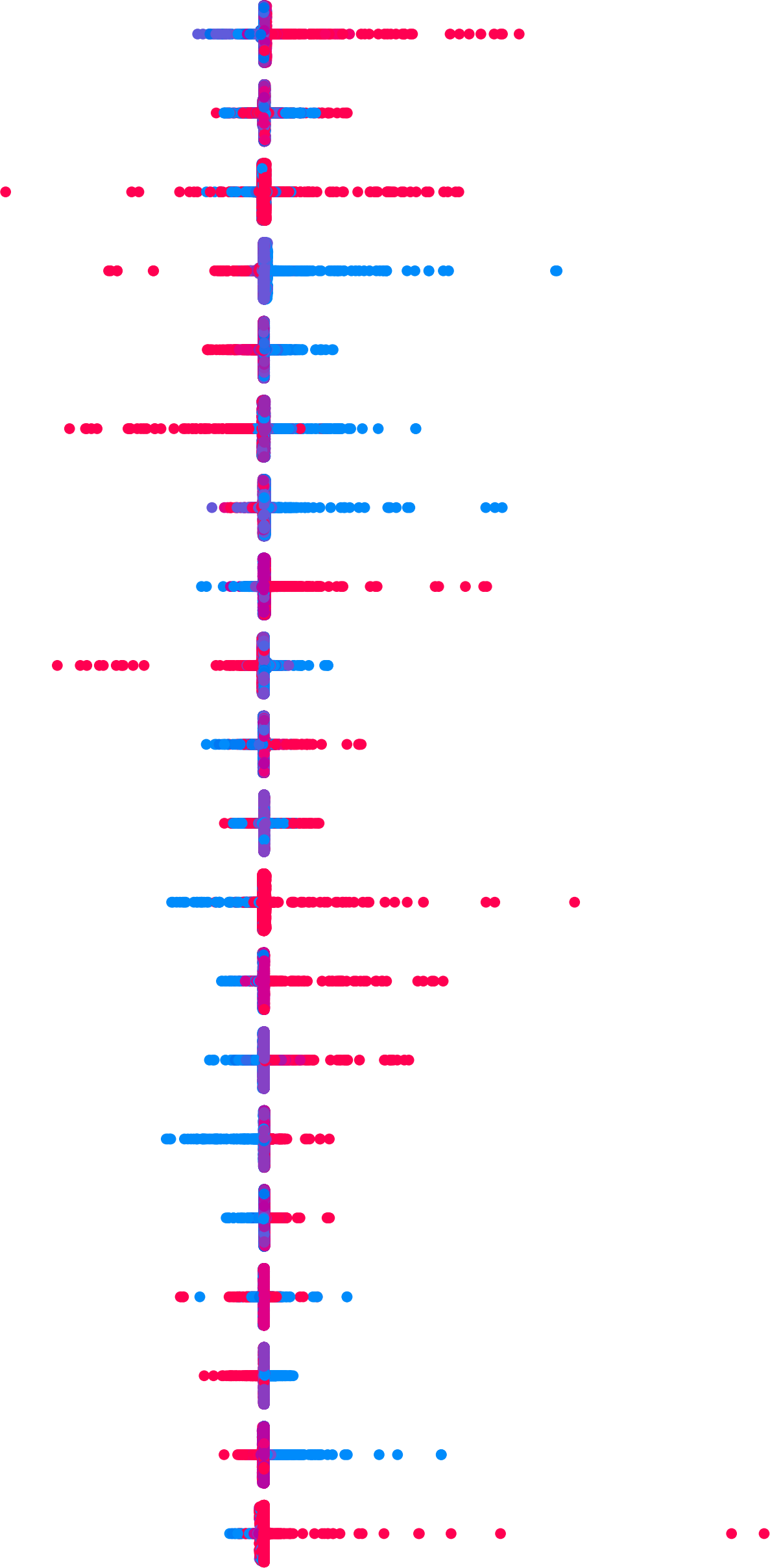}}%
\end{pgfscope}%
\begin{pgfscope}%
\pgfsetbuttcap%
\pgfsetmiterjoin%
\definecolor{currentfill}{rgb}{1.000000,1.000000,1.000000}%
\pgfsetfillcolor{currentfill}%
\pgfsetlinewidth{0.000000pt}%
\definecolor{currentstroke}{rgb}{0.000000,0.000000,0.000000}%
\pgfsetstrokecolor{currentstroke}%
\pgfsetstrokeopacity{0.000000}%
\pgfsetdash{}{0pt}%
\pgfpathmoveto{\pgfqpoint{6.956094in}{0.694630in}}%
\pgfpathlineto{\pgfqpoint{7.013751in}{0.694630in}}%
\pgfpathlineto{\pgfqpoint{7.013751in}{9.207193in}}%
\pgfpathlineto{\pgfqpoint{6.956094in}{9.207193in}}%
\pgfpathclose%
\pgfusepath{fill}%
\end{pgfscope}%
\begin{pgfscope}%
\pgfpathrectangle{\pgfqpoint{6.956094in}{0.694630in}}{\pgfqpoint{0.057658in}{8.512564in}}%
\pgfusepath{clip}%
\pgfsetbuttcap%
\pgfsetmiterjoin%
\definecolor{currentfill}{rgb}{1.000000,1.000000,1.000000}%
\pgfsetfillcolor{currentfill}%
\pgfsetlinewidth{0.010037pt}%
\definecolor{currentstroke}{rgb}{1.000000,1.000000,1.000000}%
\pgfsetstrokecolor{currentstroke}%
\pgfsetdash{}{0pt}%
\pgfpathmoveto{\pgfqpoint{6.956094in}{0.694630in}}%
\pgfpathlineto{\pgfqpoint{6.956094in}{0.727882in}}%
\pgfpathlineto{\pgfqpoint{6.956094in}{9.173941in}}%
\pgfpathlineto{\pgfqpoint{6.956094in}{9.207193in}}%
\pgfpathlineto{\pgfqpoint{7.013751in}{9.207193in}}%
\pgfpathlineto{\pgfqpoint{7.013751in}{9.173941in}}%
\pgfpathlineto{\pgfqpoint{7.013751in}{0.727882in}}%
\pgfpathlineto{\pgfqpoint{7.013751in}{0.694630in}}%
\pgfpathlineto{\pgfqpoint{7.013751in}{0.694630in}}%
\pgfpathclose%
\pgfusepath{stroke,fill}%
\end{pgfscope}%
\begin{pgfscope}%
\pgfsys@transformshift{6.956667in}{0.693333in}%
\pgftext[left,bottom]{\includegraphics[interpolate=true,width=0.056667in,height=8.513333in]{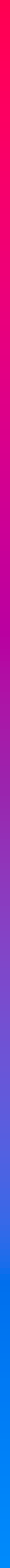}}%
\end{pgfscope}%
\begin{pgfscope}%
\definecolor{textcolor}{rgb}{0.000000,0.000000,0.000000}%
\pgfsetstrokecolor{textcolor}%
\pgfsetfillcolor{textcolor}%
\pgftext[x=7.062363in, y=0.641823in, left, base]{\color{textcolor}\sffamily\fontsize{11.000000}{13.200000}\selectfont Low}%
\end{pgfscope}%
\begin{pgfscope}%
\definecolor{textcolor}{rgb}{0.000000,0.000000,0.000000}%
\pgfsetstrokecolor{textcolor}%
\pgfsetfillcolor{textcolor}%
\pgftext[x=7.062363in, y=9.154387in, left, base]{\color{textcolor}\sffamily\fontsize{11.000000}{13.200000}\selectfont High}%
\end{pgfscope}%
\begin{pgfscope}%
\definecolor{textcolor}{rgb}{0.000000,0.000000,0.000000}%
\pgfsetstrokecolor{textcolor}%
\pgfsetfillcolor{textcolor}%
\pgftext[x=7.361038in,y=4.950911in,,top,rotate=90.000000]{\color{textcolor}\sffamily\fontsize{12.000000}{14.400000}\selectfont Feature value}%
\end{pgfscope}%
\end{pgfpicture}%
\makeatother%
\endgroup%

%% file: figures/shapley/shapley_gjtf48im_16h_raw_Hirid_dot.pgf
\begingroup%
\makeatletter%
\begin{pgfpicture}%
\pgfpathrectangle{\pgfpointorigin}{\pgfqpoint{8.000000in}{9.500000in}}%
\pgfusepath{use as bounding box, clip}%
\begin{pgfscope}%
\pgfsetbuttcap%
\pgfsetmiterjoin%
\definecolor{currentfill}{rgb}{1.000000,1.000000,1.000000}%
\pgfsetfillcolor{currentfill}%
\pgfsetlinewidth{0.000000pt}%
\definecolor{currentstroke}{rgb}{1.000000,1.000000,1.000000}%
\pgfsetstrokecolor{currentstroke}%
\pgfsetdash{}{0pt}%
\pgfpathmoveto{\pgfqpoint{0.000000in}{0.000000in}}%
\pgfpathlineto{\pgfqpoint{8.000000in}{0.000000in}}%
\pgfpathlineto{\pgfqpoint{8.000000in}{9.500000in}}%
\pgfpathlineto{\pgfqpoint{0.000000in}{9.500000in}}%
\pgfpathclose%
\pgfusepath{fill}%
\end{pgfscope}%
\begin{pgfscope}%
\pgfsetbuttcap%
\pgfsetmiterjoin%
\definecolor{currentfill}{rgb}{1.000000,1.000000,1.000000}%
\pgfsetfillcolor{currentfill}%
\pgfsetlinewidth{0.000000pt}%
\definecolor{currentstroke}{rgb}{0.000000,0.000000,0.000000}%
\pgfsetstrokecolor{currentstroke}%
\pgfsetstrokeopacity{0.000000}%
\pgfsetdash{}{0pt}%
\pgfpathmoveto{\pgfqpoint{2.400626in}{0.694630in}}%
\pgfpathlineto{\pgfqpoint{6.688125in}{0.694630in}}%
\pgfpathlineto{\pgfqpoint{6.688125in}{9.207193in}}%
\pgfpathlineto{\pgfqpoint{2.400626in}{9.207193in}}%
\pgfpathclose%
\pgfusepath{fill}%
\end{pgfscope}%
\begin{pgfscope}%
\pgfpathrectangle{\pgfqpoint{2.400626in}{0.694630in}}{\pgfqpoint{4.287499in}{8.512564in}}%
\pgfusepath{clip}%
\pgfsetrectcap%
\pgfsetroundjoin%
\pgfsetlinewidth{1.505625pt}%
\definecolor{currentstroke}{rgb}{0.600000,0.600000,0.600000}%
\pgfsetstrokecolor{currentstroke}%
\pgfsetdash{}{0pt}%
\pgfpathmoveto{\pgfqpoint{3.917690in}{0.694630in}}%
\pgfpathlineto{\pgfqpoint{3.917690in}{9.207193in}}%
\pgfusepath{stroke}%
\end{pgfscope}%
\begin{pgfscope}%
\pgfpathrectangle{\pgfqpoint{2.400626in}{0.694630in}}{\pgfqpoint{4.287499in}{8.512564in}}%
\pgfusepath{clip}%
\pgfsetbuttcap%
\pgfsetroundjoin%
\pgfsetlinewidth{0.501875pt}%
\definecolor{currentstroke}{rgb}{0.800000,0.800000,0.800000}%
\pgfsetstrokecolor{currentstroke}%
\pgfsetdash{{0.500000pt}{2.500000pt}}{0.000000pt}%
\pgfpathmoveto{\pgfqpoint{2.400626in}{1.099990in}}%
\pgfpathlineto{\pgfqpoint{6.688125in}{1.099990in}}%
\pgfusepath{stroke}%
\end{pgfscope}%
\begin{pgfscope}%
\pgfpathrectangle{\pgfqpoint{2.400626in}{0.694630in}}{\pgfqpoint{4.287499in}{8.512564in}}%
\pgfusepath{clip}%
\pgfsetbuttcap%
\pgfsetroundjoin%
\pgfsetlinewidth{0.501875pt}%
\definecolor{currentstroke}{rgb}{0.800000,0.800000,0.800000}%
\pgfsetstrokecolor{currentstroke}%
\pgfsetdash{{0.500000pt}{2.500000pt}}{0.000000pt}%
\pgfpathmoveto{\pgfqpoint{2.400626in}{1.505350in}}%
\pgfpathlineto{\pgfqpoint{6.688125in}{1.505350in}}%
\pgfusepath{stroke}%
\end{pgfscope}%
\begin{pgfscope}%
\pgfpathrectangle{\pgfqpoint{2.400626in}{0.694630in}}{\pgfqpoint{4.287499in}{8.512564in}}%
\pgfusepath{clip}%
\pgfsetbuttcap%
\pgfsetroundjoin%
\pgfsetlinewidth{0.501875pt}%
\definecolor{currentstroke}{rgb}{0.800000,0.800000,0.800000}%
\pgfsetstrokecolor{currentstroke}%
\pgfsetdash{{0.500000pt}{2.500000pt}}{0.000000pt}%
\pgfpathmoveto{\pgfqpoint{2.400626in}{1.910710in}}%
\pgfpathlineto{\pgfqpoint{6.688125in}{1.910710in}}%
\pgfusepath{stroke}%
\end{pgfscope}%
\begin{pgfscope}%
\pgfpathrectangle{\pgfqpoint{2.400626in}{0.694630in}}{\pgfqpoint{4.287499in}{8.512564in}}%
\pgfusepath{clip}%
\pgfsetbuttcap%
\pgfsetroundjoin%
\pgfsetlinewidth{0.501875pt}%
\definecolor{currentstroke}{rgb}{0.800000,0.800000,0.800000}%
\pgfsetstrokecolor{currentstroke}%
\pgfsetdash{{0.500000pt}{2.500000pt}}{0.000000pt}%
\pgfpathmoveto{\pgfqpoint{2.400626in}{2.316070in}}%
\pgfpathlineto{\pgfqpoint{6.688125in}{2.316070in}}%
\pgfusepath{stroke}%
\end{pgfscope}%
\begin{pgfscope}%
\pgfpathrectangle{\pgfqpoint{2.400626in}{0.694630in}}{\pgfqpoint{4.287499in}{8.512564in}}%
\pgfusepath{clip}%
\pgfsetbuttcap%
\pgfsetroundjoin%
\pgfsetlinewidth{0.501875pt}%
\definecolor{currentstroke}{rgb}{0.800000,0.800000,0.800000}%
\pgfsetstrokecolor{currentstroke}%
\pgfsetdash{{0.500000pt}{2.500000pt}}{0.000000pt}%
\pgfpathmoveto{\pgfqpoint{2.400626in}{2.721430in}}%
\pgfpathlineto{\pgfqpoint{6.688125in}{2.721430in}}%
\pgfusepath{stroke}%
\end{pgfscope}%
\begin{pgfscope}%
\pgfpathrectangle{\pgfqpoint{2.400626in}{0.694630in}}{\pgfqpoint{4.287499in}{8.512564in}}%
\pgfusepath{clip}%
\pgfsetbuttcap%
\pgfsetroundjoin%
\pgfsetlinewidth{0.501875pt}%
\definecolor{currentstroke}{rgb}{0.800000,0.800000,0.800000}%
\pgfsetstrokecolor{currentstroke}%
\pgfsetdash{{0.500000pt}{2.500000pt}}{0.000000pt}%
\pgfpathmoveto{\pgfqpoint{2.400626in}{3.126791in}}%
\pgfpathlineto{\pgfqpoint{6.688125in}{3.126791in}}%
\pgfusepath{stroke}%
\end{pgfscope}%
\begin{pgfscope}%
\pgfpathrectangle{\pgfqpoint{2.400626in}{0.694630in}}{\pgfqpoint{4.287499in}{8.512564in}}%
\pgfusepath{clip}%
\pgfsetbuttcap%
\pgfsetroundjoin%
\pgfsetlinewidth{0.501875pt}%
\definecolor{currentstroke}{rgb}{0.800000,0.800000,0.800000}%
\pgfsetstrokecolor{currentstroke}%
\pgfsetdash{{0.500000pt}{2.500000pt}}{0.000000pt}%
\pgfpathmoveto{\pgfqpoint{2.400626in}{3.532151in}}%
\pgfpathlineto{\pgfqpoint{6.688125in}{3.532151in}}%
\pgfusepath{stroke}%
\end{pgfscope}%
\begin{pgfscope}%
\pgfpathrectangle{\pgfqpoint{2.400626in}{0.694630in}}{\pgfqpoint{4.287499in}{8.512564in}}%
\pgfusepath{clip}%
\pgfsetbuttcap%
\pgfsetroundjoin%
\pgfsetlinewidth{0.501875pt}%
\definecolor{currentstroke}{rgb}{0.800000,0.800000,0.800000}%
\pgfsetstrokecolor{currentstroke}%
\pgfsetdash{{0.500000pt}{2.500000pt}}{0.000000pt}%
\pgfpathmoveto{\pgfqpoint{2.400626in}{3.937511in}}%
\pgfpathlineto{\pgfqpoint{6.688125in}{3.937511in}}%
\pgfusepath{stroke}%
\end{pgfscope}%
\begin{pgfscope}%
\pgfpathrectangle{\pgfqpoint{2.400626in}{0.694630in}}{\pgfqpoint{4.287499in}{8.512564in}}%
\pgfusepath{clip}%
\pgfsetbuttcap%
\pgfsetroundjoin%
\pgfsetlinewidth{0.501875pt}%
\definecolor{currentstroke}{rgb}{0.800000,0.800000,0.800000}%
\pgfsetstrokecolor{currentstroke}%
\pgfsetdash{{0.500000pt}{2.500000pt}}{0.000000pt}%
\pgfpathmoveto{\pgfqpoint{2.400626in}{4.342871in}}%
\pgfpathlineto{\pgfqpoint{6.688125in}{4.342871in}}%
\pgfusepath{stroke}%
\end{pgfscope}%
\begin{pgfscope}%
\pgfpathrectangle{\pgfqpoint{2.400626in}{0.694630in}}{\pgfqpoint{4.287499in}{8.512564in}}%
\pgfusepath{clip}%
\pgfsetbuttcap%
\pgfsetroundjoin%
\pgfsetlinewidth{0.501875pt}%
\definecolor{currentstroke}{rgb}{0.800000,0.800000,0.800000}%
\pgfsetstrokecolor{currentstroke}%
\pgfsetdash{{0.500000pt}{2.500000pt}}{0.000000pt}%
\pgfpathmoveto{\pgfqpoint{2.400626in}{4.748231in}}%
\pgfpathlineto{\pgfqpoint{6.688125in}{4.748231in}}%
\pgfusepath{stroke}%
\end{pgfscope}%
\begin{pgfscope}%
\pgfpathrectangle{\pgfqpoint{2.400626in}{0.694630in}}{\pgfqpoint{4.287499in}{8.512564in}}%
\pgfusepath{clip}%
\pgfsetbuttcap%
\pgfsetroundjoin%
\pgfsetlinewidth{0.501875pt}%
\definecolor{currentstroke}{rgb}{0.800000,0.800000,0.800000}%
\pgfsetstrokecolor{currentstroke}%
\pgfsetdash{{0.500000pt}{2.500000pt}}{0.000000pt}%
\pgfpathmoveto{\pgfqpoint{2.400626in}{5.153592in}}%
\pgfpathlineto{\pgfqpoint{6.688125in}{5.153592in}}%
\pgfusepath{stroke}%
\end{pgfscope}%
\begin{pgfscope}%
\pgfpathrectangle{\pgfqpoint{2.400626in}{0.694630in}}{\pgfqpoint{4.287499in}{8.512564in}}%
\pgfusepath{clip}%
\pgfsetbuttcap%
\pgfsetroundjoin%
\pgfsetlinewidth{0.501875pt}%
\definecolor{currentstroke}{rgb}{0.800000,0.800000,0.800000}%
\pgfsetstrokecolor{currentstroke}%
\pgfsetdash{{0.500000pt}{2.500000pt}}{0.000000pt}%
\pgfpathmoveto{\pgfqpoint{2.400626in}{5.558952in}}%
\pgfpathlineto{\pgfqpoint{6.688125in}{5.558952in}}%
\pgfusepath{stroke}%
\end{pgfscope}%
\begin{pgfscope}%
\pgfpathrectangle{\pgfqpoint{2.400626in}{0.694630in}}{\pgfqpoint{4.287499in}{8.512564in}}%
\pgfusepath{clip}%
\pgfsetbuttcap%
\pgfsetroundjoin%
\pgfsetlinewidth{0.501875pt}%
\definecolor{currentstroke}{rgb}{0.800000,0.800000,0.800000}%
\pgfsetstrokecolor{currentstroke}%
\pgfsetdash{{0.500000pt}{2.500000pt}}{0.000000pt}%
\pgfpathmoveto{\pgfqpoint{2.400626in}{5.964312in}}%
\pgfpathlineto{\pgfqpoint{6.688125in}{5.964312in}}%
\pgfusepath{stroke}%
\end{pgfscope}%
\begin{pgfscope}%
\pgfpathrectangle{\pgfqpoint{2.400626in}{0.694630in}}{\pgfqpoint{4.287499in}{8.512564in}}%
\pgfusepath{clip}%
\pgfsetbuttcap%
\pgfsetroundjoin%
\pgfsetlinewidth{0.501875pt}%
\definecolor{currentstroke}{rgb}{0.800000,0.800000,0.800000}%
\pgfsetstrokecolor{currentstroke}%
\pgfsetdash{{0.500000pt}{2.500000pt}}{0.000000pt}%
\pgfpathmoveto{\pgfqpoint{2.400626in}{6.369672in}}%
\pgfpathlineto{\pgfqpoint{6.688125in}{6.369672in}}%
\pgfusepath{stroke}%
\end{pgfscope}%
\begin{pgfscope}%
\pgfpathrectangle{\pgfqpoint{2.400626in}{0.694630in}}{\pgfqpoint{4.287499in}{8.512564in}}%
\pgfusepath{clip}%
\pgfsetbuttcap%
\pgfsetroundjoin%
\pgfsetlinewidth{0.501875pt}%
\definecolor{currentstroke}{rgb}{0.800000,0.800000,0.800000}%
\pgfsetstrokecolor{currentstroke}%
\pgfsetdash{{0.500000pt}{2.500000pt}}{0.000000pt}%
\pgfpathmoveto{\pgfqpoint{2.400626in}{6.775032in}}%
\pgfpathlineto{\pgfqpoint{6.688125in}{6.775032in}}%
\pgfusepath{stroke}%
\end{pgfscope}%
\begin{pgfscope}%
\pgfpathrectangle{\pgfqpoint{2.400626in}{0.694630in}}{\pgfqpoint{4.287499in}{8.512564in}}%
\pgfusepath{clip}%
\pgfsetbuttcap%
\pgfsetroundjoin%
\pgfsetlinewidth{0.501875pt}%
\definecolor{currentstroke}{rgb}{0.800000,0.800000,0.800000}%
\pgfsetstrokecolor{currentstroke}%
\pgfsetdash{{0.500000pt}{2.500000pt}}{0.000000pt}%
\pgfpathmoveto{\pgfqpoint{2.400626in}{7.180392in}}%
\pgfpathlineto{\pgfqpoint{6.688125in}{7.180392in}}%
\pgfusepath{stroke}%
\end{pgfscope}%
\begin{pgfscope}%
\pgfpathrectangle{\pgfqpoint{2.400626in}{0.694630in}}{\pgfqpoint{4.287499in}{8.512564in}}%
\pgfusepath{clip}%
\pgfsetbuttcap%
\pgfsetroundjoin%
\pgfsetlinewidth{0.501875pt}%
\definecolor{currentstroke}{rgb}{0.800000,0.800000,0.800000}%
\pgfsetstrokecolor{currentstroke}%
\pgfsetdash{{0.500000pt}{2.500000pt}}{0.000000pt}%
\pgfpathmoveto{\pgfqpoint{2.400626in}{7.585753in}}%
\pgfpathlineto{\pgfqpoint{6.688125in}{7.585753in}}%
\pgfusepath{stroke}%
\end{pgfscope}%
\begin{pgfscope}%
\pgfpathrectangle{\pgfqpoint{2.400626in}{0.694630in}}{\pgfqpoint{4.287499in}{8.512564in}}%
\pgfusepath{clip}%
\pgfsetbuttcap%
\pgfsetroundjoin%
\pgfsetlinewidth{0.501875pt}%
\definecolor{currentstroke}{rgb}{0.800000,0.800000,0.800000}%
\pgfsetstrokecolor{currentstroke}%
\pgfsetdash{{0.500000pt}{2.500000pt}}{0.000000pt}%
\pgfpathmoveto{\pgfqpoint{2.400626in}{7.991113in}}%
\pgfpathlineto{\pgfqpoint{6.688125in}{7.991113in}}%
\pgfusepath{stroke}%
\end{pgfscope}%
\begin{pgfscope}%
\pgfpathrectangle{\pgfqpoint{2.400626in}{0.694630in}}{\pgfqpoint{4.287499in}{8.512564in}}%
\pgfusepath{clip}%
\pgfsetbuttcap%
\pgfsetroundjoin%
\pgfsetlinewidth{0.501875pt}%
\definecolor{currentstroke}{rgb}{0.800000,0.800000,0.800000}%
\pgfsetstrokecolor{currentstroke}%
\pgfsetdash{{0.500000pt}{2.500000pt}}{0.000000pt}%
\pgfpathmoveto{\pgfqpoint{2.400626in}{8.396473in}}%
\pgfpathlineto{\pgfqpoint{6.688125in}{8.396473in}}%
\pgfusepath{stroke}%
\end{pgfscope}%
\begin{pgfscope}%
\pgfpathrectangle{\pgfqpoint{2.400626in}{0.694630in}}{\pgfqpoint{4.287499in}{8.512564in}}%
\pgfusepath{clip}%
\pgfsetbuttcap%
\pgfsetroundjoin%
\pgfsetlinewidth{0.501875pt}%
\definecolor{currentstroke}{rgb}{0.800000,0.800000,0.800000}%
\pgfsetstrokecolor{currentstroke}%
\pgfsetdash{{0.500000pt}{2.500000pt}}{0.000000pt}%
\pgfpathmoveto{\pgfqpoint{2.400626in}{8.801833in}}%
\pgfpathlineto{\pgfqpoint{6.688125in}{8.801833in}}%
\pgfusepath{stroke}%
\end{pgfscope}%
\begin{pgfscope}%
\pgfsetbuttcap%
\pgfsetroundjoin%
\definecolor{currentfill}{rgb}{0.200000,0.200000,0.200000}%
\pgfsetfillcolor{currentfill}%
\pgfsetlinewidth{0.803000pt}%
\definecolor{currentstroke}{rgb}{0.200000,0.200000,0.200000}%
\pgfsetstrokecolor{currentstroke}%
\pgfsetdash{}{0pt}%
\pgfsys@defobject{currentmarker}{\pgfqpoint{0.000000in}{-0.048611in}}{\pgfqpoint{0.000000in}{0.000000in}}{%
\pgfpathmoveto{\pgfqpoint{0.000000in}{0.000000in}}%
\pgfpathlineto{\pgfqpoint{0.000000in}{-0.048611in}}%
\pgfusepath{stroke,fill}%
}%
\begin{pgfscope}%
\pgfsys@transformshift{2.890124in}{0.694630in}%
\pgfsys@useobject{currentmarker}{}%
\end{pgfscope}%
\end{pgfscope}%
\begin{pgfscope}%
\definecolor{textcolor}{rgb}{0.200000,0.200000,0.200000}%
\pgfsetstrokecolor{textcolor}%
\pgfsetfillcolor{textcolor}%
\pgftext[x=2.890124in,y=0.597407in,,top]{\color{textcolor}\sffamily\fontsize{11.000000}{13.200000}\selectfont \(\displaystyle {\ensuremath{-}0.50}\)}%
\end{pgfscope}%
\begin{pgfscope}%
\pgfsetbuttcap%
\pgfsetroundjoin%
\definecolor{currentfill}{rgb}{0.200000,0.200000,0.200000}%
\pgfsetfillcolor{currentfill}%
\pgfsetlinewidth{0.803000pt}%
\definecolor{currentstroke}{rgb}{0.200000,0.200000,0.200000}%
\pgfsetstrokecolor{currentstroke}%
\pgfsetdash{}{0pt}%
\pgfsys@defobject{currentmarker}{\pgfqpoint{0.000000in}{-0.048611in}}{\pgfqpoint{0.000000in}{0.000000in}}{%
\pgfpathmoveto{\pgfqpoint{0.000000in}{0.000000in}}%
\pgfpathlineto{\pgfqpoint{0.000000in}{-0.048611in}}%
\pgfusepath{stroke,fill}%
}%
\begin{pgfscope}%
\pgfsys@transformshift{3.403907in}{0.694630in}%
\pgfsys@useobject{currentmarker}{}%
\end{pgfscope}%
\end{pgfscope}%
\begin{pgfscope}%
\definecolor{textcolor}{rgb}{0.200000,0.200000,0.200000}%
\pgfsetstrokecolor{textcolor}%
\pgfsetfillcolor{textcolor}%
\pgftext[x=3.403907in,y=0.597407in,,top]{\color{textcolor}\sffamily\fontsize{11.000000}{13.200000}\selectfont \(\displaystyle {\ensuremath{-}0.25}\)}%
\end{pgfscope}%
\begin{pgfscope}%
\pgfsetbuttcap%
\pgfsetroundjoin%
\definecolor{currentfill}{rgb}{0.200000,0.200000,0.200000}%
\pgfsetfillcolor{currentfill}%
\pgfsetlinewidth{0.803000pt}%
\definecolor{currentstroke}{rgb}{0.200000,0.200000,0.200000}%
\pgfsetstrokecolor{currentstroke}%
\pgfsetdash{}{0pt}%
\pgfsys@defobject{currentmarker}{\pgfqpoint{0.000000in}{-0.048611in}}{\pgfqpoint{0.000000in}{0.000000in}}{%
\pgfpathmoveto{\pgfqpoint{0.000000in}{0.000000in}}%
\pgfpathlineto{\pgfqpoint{0.000000in}{-0.048611in}}%
\pgfusepath{stroke,fill}%
}%
\begin{pgfscope}%
\pgfsys@transformshift{3.917690in}{0.694630in}%
\pgfsys@useobject{currentmarker}{}%
\end{pgfscope}%
\end{pgfscope}%
\begin{pgfscope}%
\definecolor{textcolor}{rgb}{0.200000,0.200000,0.200000}%
\pgfsetstrokecolor{textcolor}%
\pgfsetfillcolor{textcolor}%
\pgftext[x=3.917690in,y=0.597407in,,top]{\color{textcolor}\sffamily\fontsize{11.000000}{13.200000}\selectfont \(\displaystyle {0.00}\)}%
\end{pgfscope}%
\begin{pgfscope}%
\pgfsetbuttcap%
\pgfsetroundjoin%
\definecolor{currentfill}{rgb}{0.200000,0.200000,0.200000}%
\pgfsetfillcolor{currentfill}%
\pgfsetlinewidth{0.803000pt}%
\definecolor{currentstroke}{rgb}{0.200000,0.200000,0.200000}%
\pgfsetstrokecolor{currentstroke}%
\pgfsetdash{}{0pt}%
\pgfsys@defobject{currentmarker}{\pgfqpoint{0.000000in}{-0.048611in}}{\pgfqpoint{0.000000in}{0.000000in}}{%
\pgfpathmoveto{\pgfqpoint{0.000000in}{0.000000in}}%
\pgfpathlineto{\pgfqpoint{0.000000in}{-0.048611in}}%
\pgfusepath{stroke,fill}%
}%
\begin{pgfscope}%
\pgfsys@transformshift{4.431473in}{0.694630in}%
\pgfsys@useobject{currentmarker}{}%
\end{pgfscope}%
\end{pgfscope}%
\begin{pgfscope}%
\definecolor{textcolor}{rgb}{0.200000,0.200000,0.200000}%
\pgfsetstrokecolor{textcolor}%
\pgfsetfillcolor{textcolor}%
\pgftext[x=4.431473in,y=0.597407in,,top]{\color{textcolor}\sffamily\fontsize{11.000000}{13.200000}\selectfont \(\displaystyle {0.25}\)}%
\end{pgfscope}%
\begin{pgfscope}%
\pgfsetbuttcap%
\pgfsetroundjoin%
\definecolor{currentfill}{rgb}{0.200000,0.200000,0.200000}%
\pgfsetfillcolor{currentfill}%
\pgfsetlinewidth{0.803000pt}%
\definecolor{currentstroke}{rgb}{0.200000,0.200000,0.200000}%
\pgfsetstrokecolor{currentstroke}%
\pgfsetdash{}{0pt}%
\pgfsys@defobject{currentmarker}{\pgfqpoint{0.000000in}{-0.048611in}}{\pgfqpoint{0.000000in}{0.000000in}}{%
\pgfpathmoveto{\pgfqpoint{0.000000in}{0.000000in}}%
\pgfpathlineto{\pgfqpoint{0.000000in}{-0.048611in}}%
\pgfusepath{stroke,fill}%
}%
\begin{pgfscope}%
\pgfsys@transformshift{4.945256in}{0.694630in}%
\pgfsys@useobject{currentmarker}{}%
\end{pgfscope}%
\end{pgfscope}%
\begin{pgfscope}%
\definecolor{textcolor}{rgb}{0.200000,0.200000,0.200000}%
\pgfsetstrokecolor{textcolor}%
\pgfsetfillcolor{textcolor}%
\pgftext[x=4.945256in,y=0.597407in,,top]{\color{textcolor}\sffamily\fontsize{11.000000}{13.200000}\selectfont \(\displaystyle {0.50}\)}%
\end{pgfscope}%
\begin{pgfscope}%
\pgfsetbuttcap%
\pgfsetroundjoin%
\definecolor{currentfill}{rgb}{0.200000,0.200000,0.200000}%
\pgfsetfillcolor{currentfill}%
\pgfsetlinewidth{0.803000pt}%
\definecolor{currentstroke}{rgb}{0.200000,0.200000,0.200000}%
\pgfsetstrokecolor{currentstroke}%
\pgfsetdash{}{0pt}%
\pgfsys@defobject{currentmarker}{\pgfqpoint{0.000000in}{-0.048611in}}{\pgfqpoint{0.000000in}{0.000000in}}{%
\pgfpathmoveto{\pgfqpoint{0.000000in}{0.000000in}}%
\pgfpathlineto{\pgfqpoint{0.000000in}{-0.048611in}}%
\pgfusepath{stroke,fill}%
}%
\begin{pgfscope}%
\pgfsys@transformshift{5.459038in}{0.694630in}%
\pgfsys@useobject{currentmarker}{}%
\end{pgfscope}%
\end{pgfscope}%
\begin{pgfscope}%
\definecolor{textcolor}{rgb}{0.200000,0.200000,0.200000}%
\pgfsetstrokecolor{textcolor}%
\pgfsetfillcolor{textcolor}%
\pgftext[x=5.459038in,y=0.597407in,,top]{\color{textcolor}\sffamily\fontsize{11.000000}{13.200000}\selectfont \(\displaystyle {0.75}\)}%
\end{pgfscope}%
\begin{pgfscope}%
\pgfsetbuttcap%
\pgfsetroundjoin%
\definecolor{currentfill}{rgb}{0.200000,0.200000,0.200000}%
\pgfsetfillcolor{currentfill}%
\pgfsetlinewidth{0.803000pt}%
\definecolor{currentstroke}{rgb}{0.200000,0.200000,0.200000}%
\pgfsetstrokecolor{currentstroke}%
\pgfsetdash{}{0pt}%
\pgfsys@defobject{currentmarker}{\pgfqpoint{0.000000in}{-0.048611in}}{\pgfqpoint{0.000000in}{0.000000in}}{%
\pgfpathmoveto{\pgfqpoint{0.000000in}{0.000000in}}%
\pgfpathlineto{\pgfqpoint{0.000000in}{-0.048611in}}%
\pgfusepath{stroke,fill}%
}%
\begin{pgfscope}%
\pgfsys@transformshift{5.972821in}{0.694630in}%
\pgfsys@useobject{currentmarker}{}%
\end{pgfscope}%
\end{pgfscope}%
\begin{pgfscope}%
\definecolor{textcolor}{rgb}{0.200000,0.200000,0.200000}%
\pgfsetstrokecolor{textcolor}%
\pgfsetfillcolor{textcolor}%
\pgftext[x=5.972821in,y=0.597407in,,top]{\color{textcolor}\sffamily\fontsize{11.000000}{13.200000}\selectfont \(\displaystyle {1.00}\)}%
\end{pgfscope}%
\begin{pgfscope}%
\pgfsetbuttcap%
\pgfsetroundjoin%
\definecolor{currentfill}{rgb}{0.200000,0.200000,0.200000}%
\pgfsetfillcolor{currentfill}%
\pgfsetlinewidth{0.803000pt}%
\definecolor{currentstroke}{rgb}{0.200000,0.200000,0.200000}%
\pgfsetstrokecolor{currentstroke}%
\pgfsetdash{}{0pt}%
\pgfsys@defobject{currentmarker}{\pgfqpoint{0.000000in}{-0.048611in}}{\pgfqpoint{0.000000in}{0.000000in}}{%
\pgfpathmoveto{\pgfqpoint{0.000000in}{0.000000in}}%
\pgfpathlineto{\pgfqpoint{0.000000in}{-0.048611in}}%
\pgfusepath{stroke,fill}%
}%
\begin{pgfscope}%
\pgfsys@transformshift{6.486604in}{0.694630in}%
\pgfsys@useobject{currentmarker}{}%
\end{pgfscope}%
\end{pgfscope}%
\begin{pgfscope}%
\definecolor{textcolor}{rgb}{0.200000,0.200000,0.200000}%
\pgfsetstrokecolor{textcolor}%
\pgfsetfillcolor{textcolor}%
\pgftext[x=6.486604in,y=0.597407in,,top]{\color{textcolor}\sffamily\fontsize{11.000000}{13.200000}\selectfont \(\displaystyle {1.25}\)}%
\end{pgfscope}%
\begin{pgfscope}%
\definecolor{textcolor}{rgb}{0.000000,0.000000,0.000000}%
\pgfsetstrokecolor{textcolor}%
\pgfsetfillcolor{textcolor}%
\pgftext[x=4.544375in,y=0.406667in,,top]{\color{textcolor}\sffamily\fontsize{13.000000}{15.600000}\selectfont SHAP value (impact on model output)}%
\end{pgfscope}%
\begin{pgfscope}%
\definecolor{textcolor}{rgb}{0.200000,0.200000,0.200000}%
\pgfsetstrokecolor{textcolor}%
\pgfsetfillcolor{textcolor}%
\pgftext[x=1.026892in, y=1.042120in, left, base]{\color{textcolor}\sffamily\fontsize{13.000000}{15.600000}\selectfont Methemoglobin}%
\end{pgfscope}%
\begin{pgfscope}%
\definecolor{textcolor}{rgb}{0.200000,0.200000,0.200000}%
\pgfsetstrokecolor{textcolor}%
\pgfsetfillcolor{textcolor}%
\pgftext[x=1.562573in, y=1.447480in, left, base]{\color{textcolor}\sffamily\fontsize{13.000000}{15.600000}\selectfont Sodium}%
\end{pgfscope}%
\begin{pgfscope}%
\definecolor{textcolor}{rgb}{0.200000,0.200000,0.200000}%
\pgfsetstrokecolor{textcolor}%
\pgfsetfillcolor{textcolor}%
\pgftext[x=0.717000in, y=1.852840in, left, base]{\color{textcolor}\sffamily\fontsize{13.000000}{15.600000}\selectfont Blood urea nitrogen}%
\end{pgfscope}%
\begin{pgfscope}%
\definecolor{textcolor}{rgb}{0.200000,0.200000,0.200000}%
\pgfsetstrokecolor{textcolor}%
\pgfsetfillcolor{textcolor}%
\pgftext[x=1.006253in, y=2.258200in, left, base]{\color{textcolor}\sffamily\fontsize{13.000000}{15.600000}\selectfont Calcium ionized}%
\end{pgfscope}%
\begin{pgfscope}%
\definecolor{textcolor}{rgb}{0.200000,0.200000,0.200000}%
\pgfsetstrokecolor{textcolor}%
\pgfsetfillcolor{textcolor}%
\pgftext[x=0.593352in, y=2.663560in, left, base]{\color{textcolor}\sffamily\fontsize{13.000000}{15.600000}\selectfont Mean cell hemoglobin}%
\end{pgfscope}%
\begin{pgfscope}%
\definecolor{textcolor}{rgb}{0.200000,0.200000,0.200000}%
\pgfsetstrokecolor{textcolor}%
\pgfsetfillcolor{textcolor}%
\pgftext[x=0.350820in, y=3.068920in, left, base]{\color{textcolor}\sffamily\fontsize{13.000000}{15.600000}\selectfont Mean corpuscular volume}%
\end{pgfscope}%
\begin{pgfscope}%
\definecolor{textcolor}{rgb}{0.200000,0.200000,0.200000}%
\pgfsetstrokecolor{textcolor}%
\pgfsetfillcolor{textcolor}%
\pgftext[x=1.380071in, y=3.474281in, left, base]{\color{textcolor}\sffamily\fontsize{13.000000}{15.600000}\selectfont Potassium}%
\end{pgfscope}%
\begin{pgfscope}%
\definecolor{textcolor}{rgb}{0.200000,0.200000,0.200000}%
\pgfsetstrokecolor{textcolor}%
\pgfsetfillcolor{textcolor}%
\pgftext[x=1.262808in, y=3.879641in, left, base]{\color{textcolor}\sffamily\fontsize{13.000000}{15.600000}\selectfont Bicarbonate}%
\end{pgfscope}%
\begin{pgfscope}%
\definecolor{textcolor}{rgb}{0.200000,0.200000,0.200000}%
\pgfsetstrokecolor{textcolor}%
\pgfsetfillcolor{textcolor}%
\pgftext[x=1.554123in, y=4.285001in, left, base]{\color{textcolor}\sffamily\fontsize{13.000000}{15.600000}\selectfont Glucose}%
\end{pgfscope}%
\begin{pgfscope}%
\definecolor{textcolor}{rgb}{0.200000,0.200000,0.200000}%
\pgfsetstrokecolor{textcolor}%
\pgfsetfillcolor{textcolor}%
\pgftext[x=1.566912in, y=4.690361in, left, base]{\color{textcolor}\sffamily\fontsize{13.000000}{15.600000}\selectfont Lactate}%
\end{pgfscope}%
\begin{pgfscope}%
\definecolor{textcolor}{rgb}{0.200000,0.200000,0.200000}%
\pgfsetstrokecolor{textcolor}%
\pgfsetfillcolor{textcolor}%
\pgftext[x=0.992884in, y=5.095721in, left, base]{\color{textcolor}\sffamily\fontsize{13.000000}{15.600000}\selectfont Respiratory rate}%
\end{pgfscope}%
\begin{pgfscope}%
\definecolor{textcolor}{rgb}{0.200000,0.200000,0.200000}%
\pgfsetstrokecolor{textcolor}%
\pgfsetfillcolor{textcolor}%
\pgftext[x=0.240000in, y=5.501082in, left, base]{\color{textcolor}\sffamily\fontsize{13.000000}{15.600000}\selectfont Fraction of inspired oxygen}%
\end{pgfscope}%
\begin{pgfscope}%
\definecolor{textcolor}{rgb}{0.200000,0.200000,0.200000}%
\pgfsetstrokecolor{textcolor}%
\pgfsetfillcolor{textcolor}%
\pgftext[x=0.832778in, y=5.906442in, left, base]{\color{textcolor}\sffamily\fontsize{13.000000}{15.600000}\selectfont Oxygen saturation}%
\end{pgfscope}%
\begin{pgfscope}%
\definecolor{textcolor}{rgb}{0.200000,0.200000,0.200000}%
\pgfsetstrokecolor{textcolor}%
\pgfsetfillcolor{textcolor}%
\pgftext[x=1.269270in, y=6.311802in, left, base]{\color{textcolor}\sffamily\fontsize{13.000000}{15.600000}\selectfont Hemoglobin}%
\end{pgfscope}%
\begin{pgfscope}%
\definecolor{textcolor}{rgb}{0.200000,0.200000,0.200000}%
\pgfsetstrokecolor{textcolor}%
\pgfsetfillcolor{textcolor}%
\pgftext[x=0.792886in, y=6.717162in, left, base]{\color{textcolor}\sffamily\fontsize{13.000000}{15.600000}\selectfont O2 partial pressure}%
\end{pgfscope}%
\begin{pgfscope}%
\definecolor{textcolor}{rgb}{0.200000,0.200000,0.200000}%
\pgfsetstrokecolor{textcolor}%
\pgfsetfillcolor{textcolor}%
\pgftext[x=0.525663in, y=7.122522in, left, base]{\color{textcolor}\sffamily\fontsize{13.000000}{15.600000}\selectfont Systolic blood pressure}%
\end{pgfscope}%
\begin{pgfscope}%
\definecolor{textcolor}{rgb}{0.200000,0.200000,0.200000}%
\pgfsetstrokecolor{textcolor}%
\pgfsetfillcolor{textcolor}%
\pgftext[x=0.457088in, y=7.527882in, left, base]{\color{textcolor}\sffamily\fontsize{13.000000}{15.600000}\selectfont Diastolic blood pressure}%
\end{pgfscope}%
\begin{pgfscope}%
\definecolor{textcolor}{rgb}{0.200000,0.200000,0.200000}%
\pgfsetstrokecolor{textcolor}%
\pgfsetfillcolor{textcolor}%
\pgftext[x=0.574621in, y=7.933243in, left, base]{\color{textcolor}\sffamily\fontsize{13.000000}{15.600000}\selectfont Mean arterial pressure}%
\end{pgfscope}%
\begin{pgfscope}%
\definecolor{textcolor}{rgb}{0.200000,0.200000,0.200000}%
\pgfsetstrokecolor{textcolor}%
\pgfsetfillcolor{textcolor}%
\pgftext[x=0.868232in, y=8.338603in, left, base]{\color{textcolor}\sffamily\fontsize{13.000000}{15.600000}\selectfont C-reactive protein}%
\end{pgfscope}%
\begin{pgfscope}%
\definecolor{textcolor}{rgb}{0.200000,0.200000,0.200000}%
\pgfsetstrokecolor{textcolor}%
\pgfsetfillcolor{textcolor}%
\pgftext[x=1.379896in, y=8.743963in, left, base]{\color{textcolor}\sffamily\fontsize{13.000000}{15.600000}\selectfont Heart rate}%
\end{pgfscope}%
\begin{pgfscope}%
\pgfsetrectcap%
\pgfsetmiterjoin%
\pgfsetlinewidth{0.803000pt}%
\definecolor{currentstroke}{rgb}{0.000000,0.000000,0.000000}%
\pgfsetstrokecolor{currentstroke}%
\pgfsetdash{}{0pt}%
\pgfpathmoveto{\pgfqpoint{2.400626in}{0.694630in}}%
\pgfpathlineto{\pgfqpoint{6.688125in}{0.694630in}}%
\pgfusepath{stroke}%
\end{pgfscope}%
\begin{pgfscope}%
\pgfsys@transformshift{2.566667in}{0.923333in}%
\pgftext[left,bottom]{\includegraphics[interpolate=true,width=3.956667in,height=8.053333in]{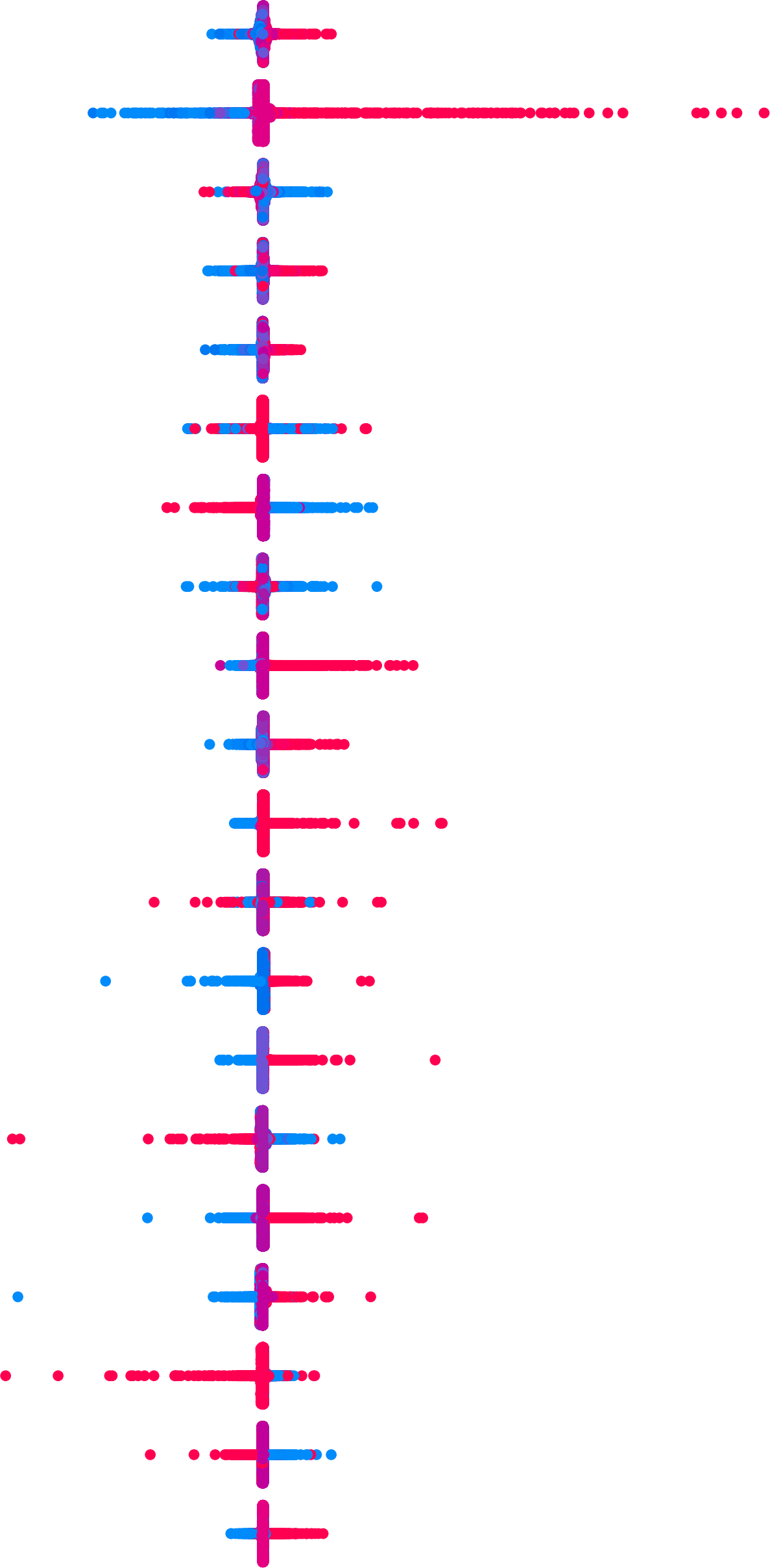}}%
\end{pgfscope}%
\begin{pgfscope}%
\pgfsetbuttcap%
\pgfsetmiterjoin%
\definecolor{currentfill}{rgb}{1.000000,1.000000,1.000000}%
\pgfsetfillcolor{currentfill}%
\pgfsetlinewidth{0.000000pt}%
\definecolor{currentstroke}{rgb}{0.000000,0.000000,0.000000}%
\pgfsetstrokecolor{currentstroke}%
\pgfsetstrokeopacity{0.000000}%
\pgfsetdash{}{0pt}%
\pgfpathmoveto{\pgfqpoint{6.956094in}{0.694630in}}%
\pgfpathlineto{\pgfqpoint{7.013751in}{0.694630in}}%
\pgfpathlineto{\pgfqpoint{7.013751in}{9.207193in}}%
\pgfpathlineto{\pgfqpoint{6.956094in}{9.207193in}}%
\pgfpathclose%
\pgfusepath{fill}%
\end{pgfscope}%
\begin{pgfscope}%
\pgfpathrectangle{\pgfqpoint{6.956094in}{0.694630in}}{\pgfqpoint{0.057658in}{8.512564in}}%
\pgfusepath{clip}%
\pgfsetbuttcap%
\pgfsetmiterjoin%
\definecolor{currentfill}{rgb}{1.000000,1.000000,1.000000}%
\pgfsetfillcolor{currentfill}%
\pgfsetlinewidth{0.010037pt}%
\definecolor{currentstroke}{rgb}{1.000000,1.000000,1.000000}%
\pgfsetstrokecolor{currentstroke}%
\pgfsetdash{}{0pt}%
\pgfpathmoveto{\pgfqpoint{6.956094in}{0.694630in}}%
\pgfpathlineto{\pgfqpoint{6.956094in}{0.727882in}}%
\pgfpathlineto{\pgfqpoint{6.956094in}{9.173941in}}%
\pgfpathlineto{\pgfqpoint{6.956094in}{9.207193in}}%
\pgfpathlineto{\pgfqpoint{7.013751in}{9.207193in}}%
\pgfpathlineto{\pgfqpoint{7.013751in}{9.173941in}}%
\pgfpathlineto{\pgfqpoint{7.013751in}{0.727882in}}%
\pgfpathlineto{\pgfqpoint{7.013751in}{0.694630in}}%
\pgfpathlineto{\pgfqpoint{7.013751in}{0.694630in}}%
\pgfpathclose%
\pgfusepath{stroke,fill}%
\end{pgfscope}%
\begin{pgfscope}%
\pgfsys@transformshift{6.956667in}{0.693333in}%
\pgftext[left,bottom]{\includegraphics[interpolate=true,width=0.056667in,height=8.513333in]{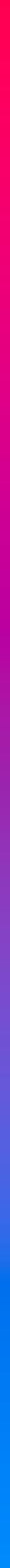}}%
\end{pgfscope}%
\begin{pgfscope}%
\definecolor{textcolor}{rgb}{0.000000,0.000000,0.000000}%
\pgfsetstrokecolor{textcolor}%
\pgfsetfillcolor{textcolor}%
\pgftext[x=7.062363in, y=0.641823in, left, base]{\color{textcolor}\sffamily\fontsize{11.000000}{13.200000}\selectfont Low}%
\end{pgfscope}%
\begin{pgfscope}%
\definecolor{textcolor}{rgb}{0.000000,0.000000,0.000000}%
\pgfsetstrokecolor{textcolor}%
\pgfsetfillcolor{textcolor}%
\pgftext[x=7.062363in, y=9.154387in, left, base]{\color{textcolor}\sffamily\fontsize{11.000000}{13.200000}\selectfont High}%
\end{pgfscope}%
\begin{pgfscope}%
\definecolor{textcolor}{rgb}{0.000000,0.000000,0.000000}%
\pgfsetstrokecolor{textcolor}%
\pgfsetfillcolor{textcolor}%
\pgftext[x=7.361038in,y=4.950911in,,top,rotate=90.000000]{\color{textcolor}\sffamily\fontsize{12.000000}{14.400000}\selectfont Feature value}%
\end{pgfscope}%
\end{pgfpicture}%
\makeatother%
\endgroup%

%% file: figures/shapley/shapley_pr9pa8oa_16h_raw_MIMIC_dot.pgf
\begingroup%
\makeatletter%
\begin{pgfpicture}%
\pgfpathrectangle{\pgfpointorigin}{\pgfqpoint{8.000000in}{9.500000in}}%
\pgfusepath{use as bounding box, clip}%
\begin{pgfscope}%
\pgfsetbuttcap%
\pgfsetmiterjoin%
\definecolor{currentfill}{rgb}{1.000000,1.000000,1.000000}%
\pgfsetfillcolor{currentfill}%
\pgfsetlinewidth{0.000000pt}%
\definecolor{currentstroke}{rgb}{1.000000,1.000000,1.000000}%
\pgfsetstrokecolor{currentstroke}%
\pgfsetdash{}{0pt}%
\pgfpathmoveto{\pgfqpoint{0.000000in}{0.000000in}}%
\pgfpathlineto{\pgfqpoint{8.000000in}{0.000000in}}%
\pgfpathlineto{\pgfqpoint{8.000000in}{9.500000in}}%
\pgfpathlineto{\pgfqpoint{0.000000in}{9.500000in}}%
\pgfpathclose%
\pgfusepath{fill}%
\end{pgfscope}%
\begin{pgfscope}%
\pgfsetbuttcap%
\pgfsetmiterjoin%
\definecolor{currentfill}{rgb}{1.000000,1.000000,1.000000}%
\pgfsetfillcolor{currentfill}%
\pgfsetlinewidth{0.000000pt}%
\definecolor{currentstroke}{rgb}{0.000000,0.000000,0.000000}%
\pgfsetstrokecolor{currentstroke}%
\pgfsetstrokeopacity{0.000000}%
\pgfsetdash{}{0pt}%
\pgfpathmoveto{\pgfqpoint{3.560335in}{0.694630in}}%
\pgfpathlineto{\pgfqpoint{6.920067in}{0.694630in}}%
\pgfpathlineto{\pgfqpoint{6.920067in}{9.207193in}}%
\pgfpathlineto{\pgfqpoint{3.560335in}{9.207193in}}%
\pgfpathclose%
\pgfusepath{fill}%
\end{pgfscope}%
\begin{pgfscope}%
\pgfpathrectangle{\pgfqpoint{3.560335in}{0.694630in}}{\pgfqpoint{3.359732in}{8.512564in}}%
\pgfusepath{clip}%
\pgfsetrectcap%
\pgfsetroundjoin%
\pgfsetlinewidth{1.505625pt}%
\definecolor{currentstroke}{rgb}{0.600000,0.600000,0.600000}%
\pgfsetstrokecolor{currentstroke}%
\pgfsetdash{}{0pt}%
\pgfpathmoveto{\pgfqpoint{4.669269in}{0.694630in}}%
\pgfpathlineto{\pgfqpoint{4.669269in}{9.207193in}}%
\pgfusepath{stroke}%
\end{pgfscope}%
\begin{pgfscope}%
\pgfpathrectangle{\pgfqpoint{3.560335in}{0.694630in}}{\pgfqpoint{3.359732in}{8.512564in}}%
\pgfusepath{clip}%
\pgfsetbuttcap%
\pgfsetroundjoin%
\pgfsetlinewidth{0.501875pt}%
\definecolor{currentstroke}{rgb}{0.800000,0.800000,0.800000}%
\pgfsetstrokecolor{currentstroke}%
\pgfsetdash{{0.500000pt}{2.500000pt}}{0.000000pt}%
\pgfpathmoveto{\pgfqpoint{3.560335in}{1.099990in}}%
\pgfpathlineto{\pgfqpoint{6.920067in}{1.099990in}}%
\pgfusepath{stroke}%
\end{pgfscope}%
\begin{pgfscope}%
\pgfpathrectangle{\pgfqpoint{3.560335in}{0.694630in}}{\pgfqpoint{3.359732in}{8.512564in}}%
\pgfusepath{clip}%
\pgfsetbuttcap%
\pgfsetroundjoin%
\pgfsetlinewidth{0.501875pt}%
\definecolor{currentstroke}{rgb}{0.800000,0.800000,0.800000}%
\pgfsetstrokecolor{currentstroke}%
\pgfsetdash{{0.500000pt}{2.500000pt}}{0.000000pt}%
\pgfpathmoveto{\pgfqpoint{3.560335in}{1.505350in}}%
\pgfpathlineto{\pgfqpoint{6.920067in}{1.505350in}}%
\pgfusepath{stroke}%
\end{pgfscope}%
\begin{pgfscope}%
\pgfpathrectangle{\pgfqpoint{3.560335in}{0.694630in}}{\pgfqpoint{3.359732in}{8.512564in}}%
\pgfusepath{clip}%
\pgfsetbuttcap%
\pgfsetroundjoin%
\pgfsetlinewidth{0.501875pt}%
\definecolor{currentstroke}{rgb}{0.800000,0.800000,0.800000}%
\pgfsetstrokecolor{currentstroke}%
\pgfsetdash{{0.500000pt}{2.500000pt}}{0.000000pt}%
\pgfpathmoveto{\pgfqpoint{3.560335in}{1.910710in}}%
\pgfpathlineto{\pgfqpoint{6.920067in}{1.910710in}}%
\pgfusepath{stroke}%
\end{pgfscope}%
\begin{pgfscope}%
\pgfpathrectangle{\pgfqpoint{3.560335in}{0.694630in}}{\pgfqpoint{3.359732in}{8.512564in}}%
\pgfusepath{clip}%
\pgfsetbuttcap%
\pgfsetroundjoin%
\pgfsetlinewidth{0.501875pt}%
\definecolor{currentstroke}{rgb}{0.800000,0.800000,0.800000}%
\pgfsetstrokecolor{currentstroke}%
\pgfsetdash{{0.500000pt}{2.500000pt}}{0.000000pt}%
\pgfpathmoveto{\pgfqpoint{3.560335in}{2.316070in}}%
\pgfpathlineto{\pgfqpoint{6.920067in}{2.316070in}}%
\pgfusepath{stroke}%
\end{pgfscope}%
\begin{pgfscope}%
\pgfpathrectangle{\pgfqpoint{3.560335in}{0.694630in}}{\pgfqpoint{3.359732in}{8.512564in}}%
\pgfusepath{clip}%
\pgfsetbuttcap%
\pgfsetroundjoin%
\pgfsetlinewidth{0.501875pt}%
\definecolor{currentstroke}{rgb}{0.800000,0.800000,0.800000}%
\pgfsetstrokecolor{currentstroke}%
\pgfsetdash{{0.500000pt}{2.500000pt}}{0.000000pt}%
\pgfpathmoveto{\pgfqpoint{3.560335in}{2.721430in}}%
\pgfpathlineto{\pgfqpoint{6.920067in}{2.721430in}}%
\pgfusepath{stroke}%
\end{pgfscope}%
\begin{pgfscope}%
\pgfpathrectangle{\pgfqpoint{3.560335in}{0.694630in}}{\pgfqpoint{3.359732in}{8.512564in}}%
\pgfusepath{clip}%
\pgfsetbuttcap%
\pgfsetroundjoin%
\pgfsetlinewidth{0.501875pt}%
\definecolor{currentstroke}{rgb}{0.800000,0.800000,0.800000}%
\pgfsetstrokecolor{currentstroke}%
\pgfsetdash{{0.500000pt}{2.500000pt}}{0.000000pt}%
\pgfpathmoveto{\pgfqpoint{3.560335in}{3.126791in}}%
\pgfpathlineto{\pgfqpoint{6.920067in}{3.126791in}}%
\pgfusepath{stroke}%
\end{pgfscope}%
\begin{pgfscope}%
\pgfpathrectangle{\pgfqpoint{3.560335in}{0.694630in}}{\pgfqpoint{3.359732in}{8.512564in}}%
\pgfusepath{clip}%
\pgfsetbuttcap%
\pgfsetroundjoin%
\pgfsetlinewidth{0.501875pt}%
\definecolor{currentstroke}{rgb}{0.800000,0.800000,0.800000}%
\pgfsetstrokecolor{currentstroke}%
\pgfsetdash{{0.500000pt}{2.500000pt}}{0.000000pt}%
\pgfpathmoveto{\pgfqpoint{3.560335in}{3.532151in}}%
\pgfpathlineto{\pgfqpoint{6.920067in}{3.532151in}}%
\pgfusepath{stroke}%
\end{pgfscope}%
\begin{pgfscope}%
\pgfpathrectangle{\pgfqpoint{3.560335in}{0.694630in}}{\pgfqpoint{3.359732in}{8.512564in}}%
\pgfusepath{clip}%
\pgfsetbuttcap%
\pgfsetroundjoin%
\pgfsetlinewidth{0.501875pt}%
\definecolor{currentstroke}{rgb}{0.800000,0.800000,0.800000}%
\pgfsetstrokecolor{currentstroke}%
\pgfsetdash{{0.500000pt}{2.500000pt}}{0.000000pt}%
\pgfpathmoveto{\pgfqpoint{3.560335in}{3.937511in}}%
\pgfpathlineto{\pgfqpoint{6.920067in}{3.937511in}}%
\pgfusepath{stroke}%
\end{pgfscope}%
\begin{pgfscope}%
\pgfpathrectangle{\pgfqpoint{3.560335in}{0.694630in}}{\pgfqpoint{3.359732in}{8.512564in}}%
\pgfusepath{clip}%
\pgfsetbuttcap%
\pgfsetroundjoin%
\pgfsetlinewidth{0.501875pt}%
\definecolor{currentstroke}{rgb}{0.800000,0.800000,0.800000}%
\pgfsetstrokecolor{currentstroke}%
\pgfsetdash{{0.500000pt}{2.500000pt}}{0.000000pt}%
\pgfpathmoveto{\pgfqpoint{3.560335in}{4.342871in}}%
\pgfpathlineto{\pgfqpoint{6.920067in}{4.342871in}}%
\pgfusepath{stroke}%
\end{pgfscope}%
\begin{pgfscope}%
\pgfpathrectangle{\pgfqpoint{3.560335in}{0.694630in}}{\pgfqpoint{3.359732in}{8.512564in}}%
\pgfusepath{clip}%
\pgfsetbuttcap%
\pgfsetroundjoin%
\pgfsetlinewidth{0.501875pt}%
\definecolor{currentstroke}{rgb}{0.800000,0.800000,0.800000}%
\pgfsetstrokecolor{currentstroke}%
\pgfsetdash{{0.500000pt}{2.500000pt}}{0.000000pt}%
\pgfpathmoveto{\pgfqpoint{3.560335in}{4.748231in}}%
\pgfpathlineto{\pgfqpoint{6.920067in}{4.748231in}}%
\pgfusepath{stroke}%
\end{pgfscope}%
\begin{pgfscope}%
\pgfpathrectangle{\pgfqpoint{3.560335in}{0.694630in}}{\pgfqpoint{3.359732in}{8.512564in}}%
\pgfusepath{clip}%
\pgfsetbuttcap%
\pgfsetroundjoin%
\pgfsetlinewidth{0.501875pt}%
\definecolor{currentstroke}{rgb}{0.800000,0.800000,0.800000}%
\pgfsetstrokecolor{currentstroke}%
\pgfsetdash{{0.500000pt}{2.500000pt}}{0.000000pt}%
\pgfpathmoveto{\pgfqpoint{3.560335in}{5.153592in}}%
\pgfpathlineto{\pgfqpoint{6.920067in}{5.153592in}}%
\pgfusepath{stroke}%
\end{pgfscope}%
\begin{pgfscope}%
\pgfpathrectangle{\pgfqpoint{3.560335in}{0.694630in}}{\pgfqpoint{3.359732in}{8.512564in}}%
\pgfusepath{clip}%
\pgfsetbuttcap%
\pgfsetroundjoin%
\pgfsetlinewidth{0.501875pt}%
\definecolor{currentstroke}{rgb}{0.800000,0.800000,0.800000}%
\pgfsetstrokecolor{currentstroke}%
\pgfsetdash{{0.500000pt}{2.500000pt}}{0.000000pt}%
\pgfpathmoveto{\pgfqpoint{3.560335in}{5.558952in}}%
\pgfpathlineto{\pgfqpoint{6.920067in}{5.558952in}}%
\pgfusepath{stroke}%
\end{pgfscope}%
\begin{pgfscope}%
\pgfpathrectangle{\pgfqpoint{3.560335in}{0.694630in}}{\pgfqpoint{3.359732in}{8.512564in}}%
\pgfusepath{clip}%
\pgfsetbuttcap%
\pgfsetroundjoin%
\pgfsetlinewidth{0.501875pt}%
\definecolor{currentstroke}{rgb}{0.800000,0.800000,0.800000}%
\pgfsetstrokecolor{currentstroke}%
\pgfsetdash{{0.500000pt}{2.500000pt}}{0.000000pt}%
\pgfpathmoveto{\pgfqpoint{3.560335in}{5.964312in}}%
\pgfpathlineto{\pgfqpoint{6.920067in}{5.964312in}}%
\pgfusepath{stroke}%
\end{pgfscope}%
\begin{pgfscope}%
\pgfpathrectangle{\pgfqpoint{3.560335in}{0.694630in}}{\pgfqpoint{3.359732in}{8.512564in}}%
\pgfusepath{clip}%
\pgfsetbuttcap%
\pgfsetroundjoin%
\pgfsetlinewidth{0.501875pt}%
\definecolor{currentstroke}{rgb}{0.800000,0.800000,0.800000}%
\pgfsetstrokecolor{currentstroke}%
\pgfsetdash{{0.500000pt}{2.500000pt}}{0.000000pt}%
\pgfpathmoveto{\pgfqpoint{3.560335in}{6.369672in}}%
\pgfpathlineto{\pgfqpoint{6.920067in}{6.369672in}}%
\pgfusepath{stroke}%
\end{pgfscope}%
\begin{pgfscope}%
\pgfpathrectangle{\pgfqpoint{3.560335in}{0.694630in}}{\pgfqpoint{3.359732in}{8.512564in}}%
\pgfusepath{clip}%
\pgfsetbuttcap%
\pgfsetroundjoin%
\pgfsetlinewidth{0.501875pt}%
\definecolor{currentstroke}{rgb}{0.800000,0.800000,0.800000}%
\pgfsetstrokecolor{currentstroke}%
\pgfsetdash{{0.500000pt}{2.500000pt}}{0.000000pt}%
\pgfpathmoveto{\pgfqpoint{3.560335in}{6.775032in}}%
\pgfpathlineto{\pgfqpoint{6.920067in}{6.775032in}}%
\pgfusepath{stroke}%
\end{pgfscope}%
\begin{pgfscope}%
\pgfpathrectangle{\pgfqpoint{3.560335in}{0.694630in}}{\pgfqpoint{3.359732in}{8.512564in}}%
\pgfusepath{clip}%
\pgfsetbuttcap%
\pgfsetroundjoin%
\pgfsetlinewidth{0.501875pt}%
\definecolor{currentstroke}{rgb}{0.800000,0.800000,0.800000}%
\pgfsetstrokecolor{currentstroke}%
\pgfsetdash{{0.500000pt}{2.500000pt}}{0.000000pt}%
\pgfpathmoveto{\pgfqpoint{3.560335in}{7.180392in}}%
\pgfpathlineto{\pgfqpoint{6.920067in}{7.180392in}}%
\pgfusepath{stroke}%
\end{pgfscope}%
\begin{pgfscope}%
\pgfpathrectangle{\pgfqpoint{3.560335in}{0.694630in}}{\pgfqpoint{3.359732in}{8.512564in}}%
\pgfusepath{clip}%
\pgfsetbuttcap%
\pgfsetroundjoin%
\pgfsetlinewidth{0.501875pt}%
\definecolor{currentstroke}{rgb}{0.800000,0.800000,0.800000}%
\pgfsetstrokecolor{currentstroke}%
\pgfsetdash{{0.500000pt}{2.500000pt}}{0.000000pt}%
\pgfpathmoveto{\pgfqpoint{3.560335in}{7.585753in}}%
\pgfpathlineto{\pgfqpoint{6.920067in}{7.585753in}}%
\pgfusepath{stroke}%
\end{pgfscope}%
\begin{pgfscope}%
\pgfpathrectangle{\pgfqpoint{3.560335in}{0.694630in}}{\pgfqpoint{3.359732in}{8.512564in}}%
\pgfusepath{clip}%
\pgfsetbuttcap%
\pgfsetroundjoin%
\pgfsetlinewidth{0.501875pt}%
\definecolor{currentstroke}{rgb}{0.800000,0.800000,0.800000}%
\pgfsetstrokecolor{currentstroke}%
\pgfsetdash{{0.500000pt}{2.500000pt}}{0.000000pt}%
\pgfpathmoveto{\pgfqpoint{3.560335in}{7.991113in}}%
\pgfpathlineto{\pgfqpoint{6.920067in}{7.991113in}}%
\pgfusepath{stroke}%
\end{pgfscope}%
\begin{pgfscope}%
\pgfpathrectangle{\pgfqpoint{3.560335in}{0.694630in}}{\pgfqpoint{3.359732in}{8.512564in}}%
\pgfusepath{clip}%
\pgfsetbuttcap%
\pgfsetroundjoin%
\pgfsetlinewidth{0.501875pt}%
\definecolor{currentstroke}{rgb}{0.800000,0.800000,0.800000}%
\pgfsetstrokecolor{currentstroke}%
\pgfsetdash{{0.500000pt}{2.500000pt}}{0.000000pt}%
\pgfpathmoveto{\pgfqpoint{3.560335in}{8.396473in}}%
\pgfpathlineto{\pgfqpoint{6.920067in}{8.396473in}}%
\pgfusepath{stroke}%
\end{pgfscope}%
\begin{pgfscope}%
\pgfpathrectangle{\pgfqpoint{3.560335in}{0.694630in}}{\pgfqpoint{3.359732in}{8.512564in}}%
\pgfusepath{clip}%
\pgfsetbuttcap%
\pgfsetroundjoin%
\pgfsetlinewidth{0.501875pt}%
\definecolor{currentstroke}{rgb}{0.800000,0.800000,0.800000}%
\pgfsetstrokecolor{currentstroke}%
\pgfsetdash{{0.500000pt}{2.500000pt}}{0.000000pt}%
\pgfpathmoveto{\pgfqpoint{3.560335in}{8.801833in}}%
\pgfpathlineto{\pgfqpoint{6.920067in}{8.801833in}}%
\pgfusepath{stroke}%
\end{pgfscope}%
\begin{pgfscope}%
\pgfsetbuttcap%
\pgfsetroundjoin%
\definecolor{currentfill}{rgb}{0.200000,0.200000,0.200000}%
\pgfsetfillcolor{currentfill}%
\pgfsetlinewidth{0.803000pt}%
\definecolor{currentstroke}{rgb}{0.200000,0.200000,0.200000}%
\pgfsetstrokecolor{currentstroke}%
\pgfsetdash{}{0pt}%
\pgfsys@defobject{currentmarker}{\pgfqpoint{0.000000in}{-0.048611in}}{\pgfqpoint{0.000000in}{0.000000in}}{%
\pgfpathmoveto{\pgfqpoint{0.000000in}{0.000000in}}%
\pgfpathlineto{\pgfqpoint{0.000000in}{-0.048611in}}%
\pgfusepath{stroke,fill}%
}%
\begin{pgfscope}%
\pgfsys@transformshift{3.601492in}{0.694630in}%
\pgfsys@useobject{currentmarker}{}%
\end{pgfscope}%
\end{pgfscope}%
\begin{pgfscope}%
\definecolor{textcolor}{rgb}{0.200000,0.200000,0.200000}%
\pgfsetstrokecolor{textcolor}%
\pgfsetfillcolor{textcolor}%
\pgftext[x=3.601492in,y=0.597407in,,top]{\color{textcolor}\sffamily\fontsize{11.000000}{13.200000}\selectfont \(\displaystyle {\ensuremath{-}0.4}\)}%
\end{pgfscope}%
\begin{pgfscope}%
\pgfsetbuttcap%
\pgfsetroundjoin%
\definecolor{currentfill}{rgb}{0.200000,0.200000,0.200000}%
\pgfsetfillcolor{currentfill}%
\pgfsetlinewidth{0.803000pt}%
\definecolor{currentstroke}{rgb}{0.200000,0.200000,0.200000}%
\pgfsetstrokecolor{currentstroke}%
\pgfsetdash{}{0pt}%
\pgfsys@defobject{currentmarker}{\pgfqpoint{0.000000in}{-0.048611in}}{\pgfqpoint{0.000000in}{0.000000in}}{%
\pgfpathmoveto{\pgfqpoint{0.000000in}{0.000000in}}%
\pgfpathlineto{\pgfqpoint{0.000000in}{-0.048611in}}%
\pgfusepath{stroke,fill}%
}%
\begin{pgfscope}%
\pgfsys@transformshift{4.135381in}{0.694630in}%
\pgfsys@useobject{currentmarker}{}%
\end{pgfscope}%
\end{pgfscope}%
\begin{pgfscope}%
\definecolor{textcolor}{rgb}{0.200000,0.200000,0.200000}%
\pgfsetstrokecolor{textcolor}%
\pgfsetfillcolor{textcolor}%
\pgftext[x=4.135381in,y=0.597407in,,top]{\color{textcolor}\sffamily\fontsize{11.000000}{13.200000}\selectfont \(\displaystyle {\ensuremath{-}0.2}\)}%
\end{pgfscope}%
\begin{pgfscope}%
\pgfsetbuttcap%
\pgfsetroundjoin%
\definecolor{currentfill}{rgb}{0.200000,0.200000,0.200000}%
\pgfsetfillcolor{currentfill}%
\pgfsetlinewidth{0.803000pt}%
\definecolor{currentstroke}{rgb}{0.200000,0.200000,0.200000}%
\pgfsetstrokecolor{currentstroke}%
\pgfsetdash{}{0pt}%
\pgfsys@defobject{currentmarker}{\pgfqpoint{0.000000in}{-0.048611in}}{\pgfqpoint{0.000000in}{0.000000in}}{%
\pgfpathmoveto{\pgfqpoint{0.000000in}{0.000000in}}%
\pgfpathlineto{\pgfqpoint{0.000000in}{-0.048611in}}%
\pgfusepath{stroke,fill}%
}%
\begin{pgfscope}%
\pgfsys@transformshift{4.669269in}{0.694630in}%
\pgfsys@useobject{currentmarker}{}%
\end{pgfscope}%
\end{pgfscope}%
\begin{pgfscope}%
\definecolor{textcolor}{rgb}{0.200000,0.200000,0.200000}%
\pgfsetstrokecolor{textcolor}%
\pgfsetfillcolor{textcolor}%
\pgftext[x=4.669269in,y=0.597407in,,top]{\color{textcolor}\sffamily\fontsize{11.000000}{13.200000}\selectfont \(\displaystyle {0.0}\)}%
\end{pgfscope}%
\begin{pgfscope}%
\pgfsetbuttcap%
\pgfsetroundjoin%
\definecolor{currentfill}{rgb}{0.200000,0.200000,0.200000}%
\pgfsetfillcolor{currentfill}%
\pgfsetlinewidth{0.803000pt}%
\definecolor{currentstroke}{rgb}{0.200000,0.200000,0.200000}%
\pgfsetstrokecolor{currentstroke}%
\pgfsetdash{}{0pt}%
\pgfsys@defobject{currentmarker}{\pgfqpoint{0.000000in}{-0.048611in}}{\pgfqpoint{0.000000in}{0.000000in}}{%
\pgfpathmoveto{\pgfqpoint{0.000000in}{0.000000in}}%
\pgfpathlineto{\pgfqpoint{0.000000in}{-0.048611in}}%
\pgfusepath{stroke,fill}%
}%
\begin{pgfscope}%
\pgfsys@transformshift{5.203158in}{0.694630in}%
\pgfsys@useobject{currentmarker}{}%
\end{pgfscope}%
\end{pgfscope}%
\begin{pgfscope}%
\definecolor{textcolor}{rgb}{0.200000,0.200000,0.200000}%
\pgfsetstrokecolor{textcolor}%
\pgfsetfillcolor{textcolor}%
\pgftext[x=5.203158in,y=0.597407in,,top]{\color{textcolor}\sffamily\fontsize{11.000000}{13.200000}\selectfont \(\displaystyle {0.2}\)}%
\end{pgfscope}%
\begin{pgfscope}%
\pgfsetbuttcap%
\pgfsetroundjoin%
\definecolor{currentfill}{rgb}{0.200000,0.200000,0.200000}%
\pgfsetfillcolor{currentfill}%
\pgfsetlinewidth{0.803000pt}%
\definecolor{currentstroke}{rgb}{0.200000,0.200000,0.200000}%
\pgfsetstrokecolor{currentstroke}%
\pgfsetdash{}{0pt}%
\pgfsys@defobject{currentmarker}{\pgfqpoint{0.000000in}{-0.048611in}}{\pgfqpoint{0.000000in}{0.000000in}}{%
\pgfpathmoveto{\pgfqpoint{0.000000in}{0.000000in}}%
\pgfpathlineto{\pgfqpoint{0.000000in}{-0.048611in}}%
\pgfusepath{stroke,fill}%
}%
\begin{pgfscope}%
\pgfsys@transformshift{5.737046in}{0.694630in}%
\pgfsys@useobject{currentmarker}{}%
\end{pgfscope}%
\end{pgfscope}%
\begin{pgfscope}%
\definecolor{textcolor}{rgb}{0.200000,0.200000,0.200000}%
\pgfsetstrokecolor{textcolor}%
\pgfsetfillcolor{textcolor}%
\pgftext[x=5.737046in,y=0.597407in,,top]{\color{textcolor}\sffamily\fontsize{11.000000}{13.200000}\selectfont \(\displaystyle {0.4}\)}%
\end{pgfscope}%
\begin{pgfscope}%
\pgfsetbuttcap%
\pgfsetroundjoin%
\definecolor{currentfill}{rgb}{0.200000,0.200000,0.200000}%
\pgfsetfillcolor{currentfill}%
\pgfsetlinewidth{0.803000pt}%
\definecolor{currentstroke}{rgb}{0.200000,0.200000,0.200000}%
\pgfsetstrokecolor{currentstroke}%
\pgfsetdash{}{0pt}%
\pgfsys@defobject{currentmarker}{\pgfqpoint{0.000000in}{-0.048611in}}{\pgfqpoint{0.000000in}{0.000000in}}{%
\pgfpathmoveto{\pgfqpoint{0.000000in}{0.000000in}}%
\pgfpathlineto{\pgfqpoint{0.000000in}{-0.048611in}}%
\pgfusepath{stroke,fill}%
}%
\begin{pgfscope}%
\pgfsys@transformshift{6.270935in}{0.694630in}%
\pgfsys@useobject{currentmarker}{}%
\end{pgfscope}%
\end{pgfscope}%
\begin{pgfscope}%
\definecolor{textcolor}{rgb}{0.200000,0.200000,0.200000}%
\pgfsetstrokecolor{textcolor}%
\pgfsetfillcolor{textcolor}%
\pgftext[x=6.270935in,y=0.597407in,,top]{\color{textcolor}\sffamily\fontsize{11.000000}{13.200000}\selectfont \(\displaystyle {0.6}\)}%
\end{pgfscope}%
\begin{pgfscope}%
\pgfsetbuttcap%
\pgfsetroundjoin%
\definecolor{currentfill}{rgb}{0.200000,0.200000,0.200000}%
\pgfsetfillcolor{currentfill}%
\pgfsetlinewidth{0.803000pt}%
\definecolor{currentstroke}{rgb}{0.200000,0.200000,0.200000}%
\pgfsetstrokecolor{currentstroke}%
\pgfsetdash{}{0pt}%
\pgfsys@defobject{currentmarker}{\pgfqpoint{0.000000in}{-0.048611in}}{\pgfqpoint{0.000000in}{0.000000in}}{%
\pgfpathmoveto{\pgfqpoint{0.000000in}{0.000000in}}%
\pgfpathlineto{\pgfqpoint{0.000000in}{-0.048611in}}%
\pgfusepath{stroke,fill}%
}%
\begin{pgfscope}%
\pgfsys@transformshift{6.804823in}{0.694630in}%
\pgfsys@useobject{currentmarker}{}%
\end{pgfscope}%
\end{pgfscope}%
\begin{pgfscope}%
\definecolor{textcolor}{rgb}{0.200000,0.200000,0.200000}%
\pgfsetstrokecolor{textcolor}%
\pgfsetfillcolor{textcolor}%
\pgftext[x=6.804823in,y=0.597407in,,top]{\color{textcolor}\sffamily\fontsize{11.000000}{13.200000}\selectfont \(\displaystyle {0.8}\)}%
\end{pgfscope}%
\begin{pgfscope}%
\definecolor{textcolor}{rgb}{0.000000,0.000000,0.000000}%
\pgfsetstrokecolor{textcolor}%
\pgfsetfillcolor{textcolor}%
\pgftext[x=5.240201in,y=0.406667in,,top]{\color{textcolor}\sffamily\fontsize{13.000000}{15.600000}\selectfont SHAP value (impact on model output)}%
\end{pgfscope}%
\begin{pgfscope}%
\definecolor{textcolor}{rgb}{0.200000,0.200000,0.200000}%
\pgfsetstrokecolor{textcolor}%
\pgfsetfillcolor{textcolor}%
\pgftext[x=1.952712in, y=1.042120in, left, base]{\color{textcolor}\sffamily\fontsize{13.000000}{15.600000}\selectfont Prothrombine time}%
\end{pgfscope}%
\begin{pgfscope}%
\definecolor{textcolor}{rgb}{0.200000,0.200000,0.200000}%
\pgfsetstrokecolor{textcolor}%
\pgfsetfillcolor{textcolor}%
\pgftext[x=2.462678in, y=1.447480in, left, base]{\color{textcolor}\sffamily\fontsize{13.000000}{15.600000}\selectfont Magnesium}%
\end{pgfscope}%
\begin{pgfscope}%
\definecolor{textcolor}{rgb}{0.200000,0.200000,0.200000}%
\pgfsetstrokecolor{textcolor}%
\pgfsetfillcolor{textcolor}%
\pgftext[x=1.683618in, y=1.852840in, left, base]{\color{textcolor}\sffamily\fontsize{13.000000}{15.600000}\selectfont White blood cell count}%
\end{pgfscope}%
\begin{pgfscope}%
\definecolor{textcolor}{rgb}{0.200000,0.200000,0.200000}%
\pgfsetstrokecolor{textcolor}%
\pgfsetfillcolor{textcolor}%
\pgftext[x=2.277457in, y=2.258200in, left, base]{\color{textcolor}\sffamily\fontsize{13.000000}{15.600000}\selectfont Platelet count}%
\end{pgfscope}%
\begin{pgfscope}%
\definecolor{textcolor}{rgb}{0.200000,0.200000,0.200000}%
\pgfsetstrokecolor{textcolor}%
\pgfsetfillcolor{textcolor}%
\pgftext[x=2.422517in, y=2.663560in, left, base]{\color{textcolor}\sffamily\fontsize{13.000000}{15.600000}\selectfont Bicarbonate}%
\end{pgfscope}%
\begin{pgfscope}%
\definecolor{textcolor}{rgb}{0.200000,0.200000,0.200000}%
\pgfsetstrokecolor{textcolor}%
\pgfsetfillcolor{textcolor}%
\pgftext[x=1.829063in, y=3.068920in, left, base]{\color{textcolor}\sffamily\fontsize{13.000000}{15.600000}\selectfont Red blood cell count}%
\end{pgfscope}%
\begin{pgfscope}%
\definecolor{textcolor}{rgb}{0.200000,0.200000,0.200000}%
\pgfsetstrokecolor{textcolor}%
\pgfsetfillcolor{textcolor}%
\pgftext[x=2.722282in, y=3.474281in, left, base]{\color{textcolor}\sffamily\fontsize{13.000000}{15.600000}\selectfont Sodium}%
\end{pgfscope}%
\begin{pgfscope}%
\definecolor{textcolor}{rgb}{0.200000,0.200000,0.200000}%
\pgfsetstrokecolor{textcolor}%
\pgfsetfillcolor{textcolor}%
\pgftext[x=2.680423in, y=3.879641in, left, base]{\color{textcolor}\sffamily\fontsize{13.000000}{15.600000}\selectfont Chloride}%
\end{pgfscope}%
\begin{pgfscope}%
\definecolor{textcolor}{rgb}{0.200000,0.200000,0.200000}%
\pgfsetstrokecolor{textcolor}%
\pgfsetfillcolor{textcolor}%
\pgftext[x=2.469487in, y=4.285001in, left, base]{\color{textcolor}\sffamily\fontsize{13.000000}{15.600000}\selectfont Hematocrit}%
\end{pgfscope}%
\begin{pgfscope}%
\definecolor{textcolor}{rgb}{0.200000,0.200000,0.200000}%
\pgfsetstrokecolor{textcolor}%
\pgfsetfillcolor{textcolor}%
\pgftext[x=0.240000in, y=4.690361in, left, base]{\color{textcolor}\sffamily\fontsize{13.000000}{15.600000}\selectfont Mean corpuscular hemoglobin concentration}%
\end{pgfscope}%
\begin{pgfscope}%
\definecolor{textcolor}{rgb}{0.200000,0.200000,0.200000}%
\pgfsetstrokecolor{textcolor}%
\pgfsetfillcolor{textcolor}%
\pgftext[x=2.689489in, y=5.095721in, left, base]{\color{textcolor}\sffamily\fontsize{13.000000}{15.600000}\selectfont Calcium}%
\end{pgfscope}%
\begin{pgfscope}%
\definecolor{textcolor}{rgb}{0.200000,0.200000,0.200000}%
\pgfsetstrokecolor{textcolor}%
\pgfsetfillcolor{textcolor}%
\pgftext[x=2.152593in, y=5.501082in, left, base]{\color{textcolor}\sffamily\fontsize{13.000000}{15.600000}\selectfont Respiratory rate}%
\end{pgfscope}%
\begin{pgfscope}%
\definecolor{textcolor}{rgb}{0.200000,0.200000,0.200000}%
\pgfsetstrokecolor{textcolor}%
\pgfsetfillcolor{textcolor}%
\pgftext[x=1.952595in, y=5.906442in, left, base]{\color{textcolor}\sffamily\fontsize{13.000000}{15.600000}\selectfont O2 partial pressure}%
\end{pgfscope}%
\begin{pgfscope}%
\definecolor{textcolor}{rgb}{0.200000,0.200000,0.200000}%
\pgfsetstrokecolor{textcolor}%
\pgfsetfillcolor{textcolor}%
\pgftext[x=1.685373in, y=6.311802in, left, base]{\color{textcolor}\sffamily\fontsize{13.000000}{15.600000}\selectfont Systolic blood pressure}%
\end{pgfscope}%
\begin{pgfscope}%
\definecolor{textcolor}{rgb}{0.200000,0.200000,0.200000}%
\pgfsetstrokecolor{textcolor}%
\pgfsetfillcolor{textcolor}%
\pgftext[x=2.367733in, y=6.717162in, left, base]{\color{textcolor}\sffamily\fontsize{13.000000}{15.600000}\selectfont Urine output}%
\end{pgfscope}%
\begin{pgfscope}%
\definecolor{textcolor}{rgb}{0.200000,0.200000,0.200000}%
\pgfsetstrokecolor{textcolor}%
\pgfsetfillcolor{textcolor}%
\pgftext[x=2.369662in, y=7.122522in, left, base]{\color{textcolor}\sffamily\fontsize{13.000000}{15.600000}\selectfont Temperature}%
\end{pgfscope}%
\begin{pgfscope}%
\definecolor{textcolor}{rgb}{0.200000,0.200000,0.200000}%
\pgfsetstrokecolor{textcolor}%
\pgfsetfillcolor{textcolor}%
\pgftext[x=1.992487in, y=7.527882in, left, base]{\color{textcolor}\sffamily\fontsize{13.000000}{15.600000}\selectfont Oxygen saturation}%
\end{pgfscope}%
\begin{pgfscope}%
\definecolor{textcolor}{rgb}{0.200000,0.200000,0.200000}%
\pgfsetstrokecolor{textcolor}%
\pgfsetfillcolor{textcolor}%
\pgftext[x=2.539605in, y=7.933243in, left, base]{\color{textcolor}\sffamily\fontsize{13.000000}{15.600000}\selectfont Heart rate}%
\end{pgfscope}%
\begin{pgfscope}%
\definecolor{textcolor}{rgb}{0.200000,0.200000,0.200000}%
\pgfsetstrokecolor{textcolor}%
\pgfsetfillcolor{textcolor}%
\pgftext[x=1.616797in, y=8.338603in, left, base]{\color{textcolor}\sffamily\fontsize{13.000000}{15.600000}\selectfont Diastolic blood pressure}%
\end{pgfscope}%
\begin{pgfscope}%
\definecolor{textcolor}{rgb}{0.200000,0.200000,0.200000}%
\pgfsetstrokecolor{textcolor}%
\pgfsetfillcolor{textcolor}%
\pgftext[x=1.734330in, y=8.743963in, left, base]{\color{textcolor}\sffamily\fontsize{13.000000}{15.600000}\selectfont Mean arterial pressure}%
\end{pgfscope}%
\begin{pgfscope}%
\pgfsetrectcap%
\pgfsetmiterjoin%
\pgfsetlinewidth{0.803000pt}%
\definecolor{currentstroke}{rgb}{0.000000,0.000000,0.000000}%
\pgfsetstrokecolor{currentstroke}%
\pgfsetdash{}{0pt}%
\pgfpathmoveto{\pgfqpoint{3.560335in}{0.694630in}}%
\pgfpathlineto{\pgfqpoint{6.920067in}{0.694630in}}%
\pgfusepath{stroke}%
\end{pgfscope}%
\begin{pgfscope}%
\pgfsys@transformshift{3.683333in}{0.923333in}%
\pgftext[left,bottom]{\includegraphics[interpolate=true,width=3.113333in,height=8.053333in]{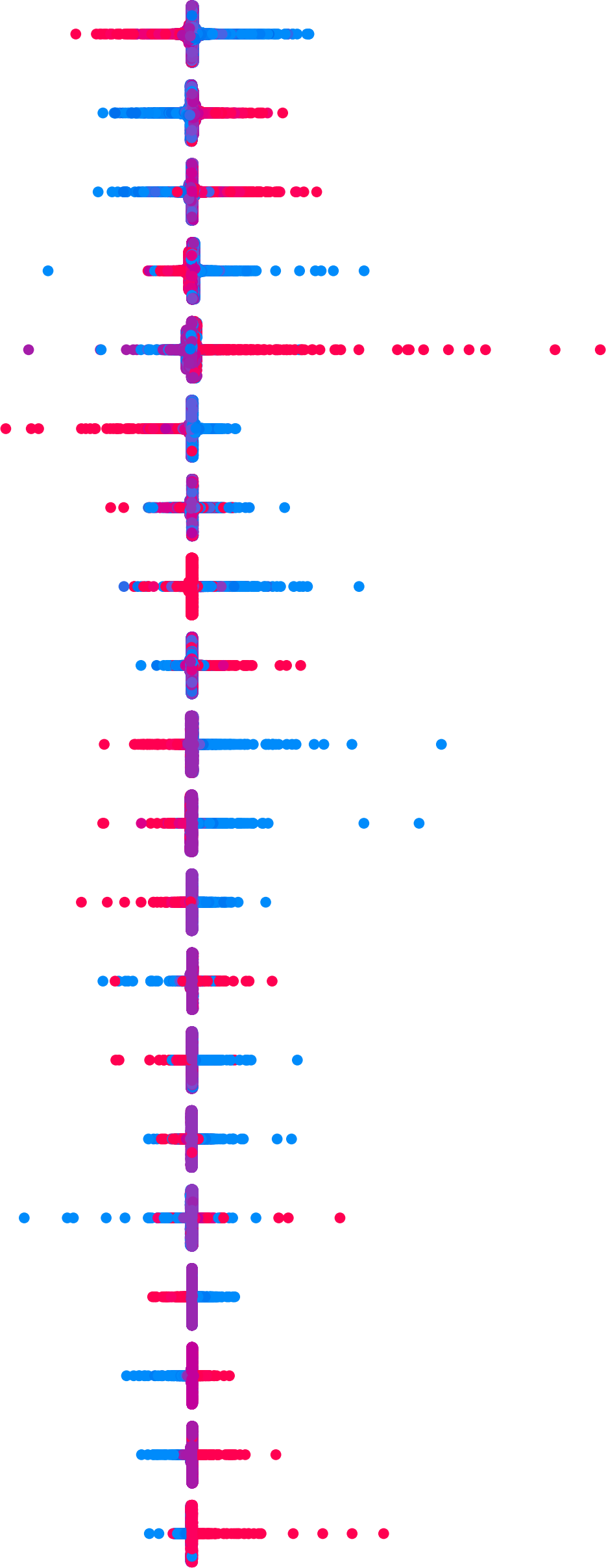}}%
\end{pgfscope}%
\begin{pgfscope}%
\pgfsetbuttcap%
\pgfsetmiterjoin%
\definecolor{currentfill}{rgb}{1.000000,1.000000,1.000000}%
\pgfsetfillcolor{currentfill}%
\pgfsetlinewidth{0.000000pt}%
\definecolor{currentstroke}{rgb}{0.000000,0.000000,0.000000}%
\pgfsetstrokecolor{currentstroke}%
\pgfsetstrokeopacity{0.000000}%
\pgfsetdash{}{0pt}%
\pgfpathmoveto{\pgfqpoint{7.130050in}{0.694630in}}%
\pgfpathlineto{\pgfqpoint{7.187708in}{0.694630in}}%
\pgfpathlineto{\pgfqpoint{7.187708in}{9.207193in}}%
\pgfpathlineto{\pgfqpoint{7.130050in}{9.207193in}}%
\pgfpathclose%
\pgfusepath{fill}%
\end{pgfscope}%
\begin{pgfscope}%
\pgfpathrectangle{\pgfqpoint{7.130050in}{0.694630in}}{\pgfqpoint{0.057658in}{8.512564in}}%
\pgfusepath{clip}%
\pgfsetbuttcap%
\pgfsetmiterjoin%
\definecolor{currentfill}{rgb}{1.000000,1.000000,1.000000}%
\pgfsetfillcolor{currentfill}%
\pgfsetlinewidth{0.010037pt}%
\definecolor{currentstroke}{rgb}{1.000000,1.000000,1.000000}%
\pgfsetstrokecolor{currentstroke}%
\pgfsetdash{}{0pt}%
\pgfpathmoveto{\pgfqpoint{7.130050in}{0.694630in}}%
\pgfpathlineto{\pgfqpoint{7.130050in}{0.727882in}}%
\pgfpathlineto{\pgfqpoint{7.130050in}{9.173941in}}%
\pgfpathlineto{\pgfqpoint{7.130050in}{9.207193in}}%
\pgfpathlineto{\pgfqpoint{7.187708in}{9.207193in}}%
\pgfpathlineto{\pgfqpoint{7.187708in}{9.173941in}}%
\pgfpathlineto{\pgfqpoint{7.187708in}{0.727882in}}%
\pgfpathlineto{\pgfqpoint{7.187708in}{0.694630in}}%
\pgfpathlineto{\pgfqpoint{7.187708in}{0.694630in}}%
\pgfpathclose%
\pgfusepath{stroke,fill}%
\end{pgfscope}%
\begin{pgfscope}%
\pgfsys@transformshift{7.130000in}{0.693333in}%
\pgftext[left,bottom]{\includegraphics[interpolate=true,width=0.056667in,height=8.513333in]{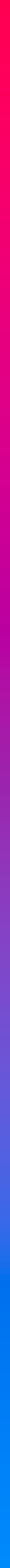}}%
\end{pgfscope}%
\begin{pgfscope}%
\definecolor{textcolor}{rgb}{0.000000,0.000000,0.000000}%
\pgfsetstrokecolor{textcolor}%
\pgfsetfillcolor{textcolor}%
\pgftext[x=7.236319in, y=0.641823in, left, base]{\color{textcolor}\sffamily\fontsize{11.000000}{13.200000}\selectfont Low}%
\end{pgfscope}%
\begin{pgfscope}%
\definecolor{textcolor}{rgb}{0.000000,0.000000,0.000000}%
\pgfsetstrokecolor{textcolor}%
\pgfsetfillcolor{textcolor}%
\pgftext[x=7.236319in, y=9.154387in, left, base]{\color{textcolor}\sffamily\fontsize{11.000000}{13.200000}\selectfont High}%
\end{pgfscope}%
\begin{pgfscope}%
\definecolor{textcolor}{rgb}{0.000000,0.000000,0.000000}%
\pgfsetstrokecolor{textcolor}%
\pgfsetfillcolor{textcolor}%
\pgftext[x=7.534994in,y=4.950911in,,top,rotate=90.000000]{\color{textcolor}\sffamily\fontsize{12.000000}{14.400000}\selectfont Feature value}%
\end{pgfscope}%
\end{pgfpicture}%
\makeatother%
\endgroup%

%% file: figures/shapley/shapley_16h_bar.pgf
\begingroup%
\makeatletter%
\begin{pgfpicture}%
\pgfpathrectangle{\pgfpointorigin}{\pgfqpoint{6.000000in}{9.000000in}}%
\pgfusepath{use as bounding box, clip}%
\begin{pgfscope}%
\pgfsetbuttcap%
\pgfsetmiterjoin%
\definecolor{currentfill}{rgb}{1.000000,1.000000,1.000000}%
\pgfsetfillcolor{currentfill}%
\pgfsetlinewidth{0.000000pt}%
\definecolor{currentstroke}{rgb}{1.000000,1.000000,1.000000}%
\pgfsetstrokecolor{currentstroke}%
\pgfsetdash{}{0pt}%
\pgfpathmoveto{\pgfqpoint{0.000000in}{0.000000in}}%
\pgfpathlineto{\pgfqpoint{6.000000in}{0.000000in}}%
\pgfpathlineto{\pgfqpoint{6.000000in}{9.000000in}}%
\pgfpathlineto{\pgfqpoint{0.000000in}{9.000000in}}%
\pgfpathclose%
\pgfusepath{fill}%
\end{pgfscope}%
\begin{pgfscope}%
\pgfsetbuttcap%
\pgfsetmiterjoin%
\definecolor{currentfill}{rgb}{1.000000,1.000000,1.000000}%
\pgfsetfillcolor{currentfill}%
\pgfsetlinewidth{0.000000pt}%
\definecolor{currentstroke}{rgb}{0.000000,0.000000,0.000000}%
\pgfsetstrokecolor{currentstroke}%
\pgfsetstrokeopacity{0.000000}%
\pgfsetdash{}{0pt}%
\pgfpathmoveto{\pgfqpoint{2.873633in}{1.536516in}}%
\pgfpathlineto{\pgfqpoint{5.775000in}{1.536516in}}%
\pgfpathlineto{\pgfqpoint{5.775000in}{8.550000in}}%
\pgfpathlineto{\pgfqpoint{2.873633in}{8.550000in}}%
\pgfpathclose%
\pgfusepath{fill}%
\end{pgfscope}%
\begin{pgfscope}%
\pgfpathrectangle{\pgfqpoint{2.873633in}{1.536516in}}{\pgfqpoint{2.901367in}{7.013484in}}%
\pgfusepath{clip}%
\pgfsetbuttcap%
\pgfsetmiterjoin%
\definecolor{currentfill}{rgb}{0.000000,0.543378,0.983379}%
\pgfsetfillcolor{currentfill}%
\pgfsetlinewidth{1.003750pt}%
\definecolor{currentstroke}{rgb}{1.000000,1.000000,1.000000}%
\pgfsetstrokecolor{currentstroke}%
\pgfsetstrokeopacity{0.800000}%
\pgfsetdash{}{0pt}%
\pgfpathmoveto{\pgfqpoint{2.873633in}{8.231205in}}%
\pgfpathlineto{\pgfqpoint{4.704290in}{8.231205in}}%
\pgfpathlineto{\pgfqpoint{4.704290in}{7.973593in}}%
\pgfpathlineto{\pgfqpoint{2.873633in}{7.973593in}}%
\pgfpathclose%
\pgfusepath{stroke,fill}%
\end{pgfscope}%
\begin{pgfscope}%
\pgfpathrectangle{\pgfqpoint{2.873633in}{1.536516in}}{\pgfqpoint{2.901367in}{7.013484in}}%
\pgfusepath{clip}%
\pgfsetbuttcap%
\pgfsetmiterjoin%
\definecolor{currentfill}{rgb}{0.000000,0.543378,0.983379}%
\pgfsetfillcolor{currentfill}%
\pgfsetlinewidth{1.003750pt}%
\definecolor{currentstroke}{rgb}{1.000000,1.000000,1.000000}%
\pgfsetstrokecolor{currentstroke}%
\pgfsetstrokeopacity{0.800000}%
\pgfsetdash{}{0pt}%
\pgfpathmoveto{\pgfqpoint{2.873633in}{7.909190in}}%
\pgfpathlineto{\pgfqpoint{4.224502in}{7.909190in}}%
\pgfpathlineto{\pgfqpoint{4.224502in}{7.651578in}}%
\pgfpathlineto{\pgfqpoint{2.873633in}{7.651578in}}%
\pgfpathclose%
\pgfusepath{stroke,fill}%
\end{pgfscope}%
\begin{pgfscope}%
\pgfpathrectangle{\pgfqpoint{2.873633in}{1.536516in}}{\pgfqpoint{2.901367in}{7.013484in}}%
\pgfusepath{clip}%
\pgfsetbuttcap%
\pgfsetmiterjoin%
\definecolor{currentfill}{rgb}{0.000000,0.543378,0.983379}%
\pgfsetfillcolor{currentfill}%
\pgfsetlinewidth{1.003750pt}%
\definecolor{currentstroke}{rgb}{1.000000,1.000000,1.000000}%
\pgfsetstrokecolor{currentstroke}%
\pgfsetstrokeopacity{0.800000}%
\pgfsetdash{}{0pt}%
\pgfpathmoveto{\pgfqpoint{2.873633in}{7.587175in}}%
\pgfpathlineto{\pgfqpoint{4.151705in}{7.587175in}}%
\pgfpathlineto{\pgfqpoint{4.151705in}{7.329564in}}%
\pgfpathlineto{\pgfqpoint{2.873633in}{7.329564in}}%
\pgfpathclose%
\pgfusepath{stroke,fill}%
\end{pgfscope}%
\begin{pgfscope}%
\pgfpathrectangle{\pgfqpoint{2.873633in}{1.536516in}}{\pgfqpoint{2.901367in}{7.013484in}}%
\pgfusepath{clip}%
\pgfsetbuttcap%
\pgfsetmiterjoin%
\definecolor{currentfill}{rgb}{0.000000,0.543378,0.983379}%
\pgfsetfillcolor{currentfill}%
\pgfsetlinewidth{1.003750pt}%
\definecolor{currentstroke}{rgb}{1.000000,1.000000,1.000000}%
\pgfsetstrokecolor{currentstroke}%
\pgfsetstrokeopacity{0.800000}%
\pgfsetdash{}{0pt}%
\pgfpathmoveto{\pgfqpoint{2.873633in}{7.265161in}}%
\pgfpathlineto{\pgfqpoint{4.048340in}{7.265161in}}%
\pgfpathlineto{\pgfqpoint{4.048340in}{7.007549in}}%
\pgfpathlineto{\pgfqpoint{2.873633in}{7.007549in}}%
\pgfpathclose%
\pgfusepath{stroke,fill}%
\end{pgfscope}%
\begin{pgfscope}%
\pgfpathrectangle{\pgfqpoint{2.873633in}{1.536516in}}{\pgfqpoint{2.901367in}{7.013484in}}%
\pgfusepath{clip}%
\pgfsetbuttcap%
\pgfsetmiterjoin%
\definecolor{currentfill}{rgb}{0.000000,0.543378,0.983379}%
\pgfsetfillcolor{currentfill}%
\pgfsetlinewidth{1.003750pt}%
\definecolor{currentstroke}{rgb}{1.000000,1.000000,1.000000}%
\pgfsetstrokecolor{currentstroke}%
\pgfsetstrokeopacity{0.800000}%
\pgfsetdash{}{0pt}%
\pgfpathmoveto{\pgfqpoint{2.873633in}{6.943146in}}%
\pgfpathlineto{\pgfqpoint{3.607061in}{6.943146in}}%
\pgfpathlineto{\pgfqpoint{3.607061in}{6.685534in}}%
\pgfpathlineto{\pgfqpoint{2.873633in}{6.685534in}}%
\pgfpathclose%
\pgfusepath{stroke,fill}%
\end{pgfscope}%
\begin{pgfscope}%
\pgfpathrectangle{\pgfqpoint{2.873633in}{1.536516in}}{\pgfqpoint{2.901367in}{7.013484in}}%
\pgfusepath{clip}%
\pgfsetbuttcap%
\pgfsetmiterjoin%
\definecolor{currentfill}{rgb}{0.000000,0.543378,0.983379}%
\pgfsetfillcolor{currentfill}%
\pgfsetlinewidth{1.003750pt}%
\definecolor{currentstroke}{rgb}{1.000000,1.000000,1.000000}%
\pgfsetstrokecolor{currentstroke}%
\pgfsetstrokeopacity{0.800000}%
\pgfsetdash{}{0pt}%
\pgfpathmoveto{\pgfqpoint{2.873633in}{6.621131in}}%
\pgfpathlineto{\pgfqpoint{3.496013in}{6.621131in}}%
\pgfpathlineto{\pgfqpoint{3.496013in}{6.363519in}}%
\pgfpathlineto{\pgfqpoint{2.873633in}{6.363519in}}%
\pgfpathclose%
\pgfusepath{stroke,fill}%
\end{pgfscope}%
\begin{pgfscope}%
\pgfpathrectangle{\pgfqpoint{2.873633in}{1.536516in}}{\pgfqpoint{2.901367in}{7.013484in}}%
\pgfusepath{clip}%
\pgfsetbuttcap%
\pgfsetmiterjoin%
\definecolor{currentfill}{rgb}{0.000000,0.543378,0.983379}%
\pgfsetfillcolor{currentfill}%
\pgfsetlinewidth{1.003750pt}%
\definecolor{currentstroke}{rgb}{1.000000,1.000000,1.000000}%
\pgfsetstrokecolor{currentstroke}%
\pgfsetstrokeopacity{0.800000}%
\pgfsetdash{}{0pt}%
\pgfpathmoveto{\pgfqpoint{2.873633in}{6.299116in}}%
\pgfpathlineto{\pgfqpoint{3.434925in}{6.299116in}}%
\pgfpathlineto{\pgfqpoint{3.434925in}{6.041504in}}%
\pgfpathlineto{\pgfqpoint{2.873633in}{6.041504in}}%
\pgfpathclose%
\pgfusepath{stroke,fill}%
\end{pgfscope}%
\begin{pgfscope}%
\pgfpathrectangle{\pgfqpoint{2.873633in}{1.536516in}}{\pgfqpoint{2.901367in}{7.013484in}}%
\pgfusepath{clip}%
\pgfsetbuttcap%
\pgfsetmiterjoin%
\definecolor{currentfill}{rgb}{0.000000,0.543378,0.983379}%
\pgfsetfillcolor{currentfill}%
\pgfsetlinewidth{1.003750pt}%
\definecolor{currentstroke}{rgb}{1.000000,1.000000,1.000000}%
\pgfsetstrokecolor{currentstroke}%
\pgfsetstrokeopacity{0.800000}%
\pgfsetdash{}{0pt}%
\pgfpathmoveto{\pgfqpoint{2.873633in}{5.977101in}}%
\pgfpathlineto{\pgfqpoint{3.409696in}{5.977101in}}%
\pgfpathlineto{\pgfqpoint{3.409696in}{5.719489in}}%
\pgfpathlineto{\pgfqpoint{2.873633in}{5.719489in}}%
\pgfpathclose%
\pgfusepath{stroke,fill}%
\end{pgfscope}%
\begin{pgfscope}%
\pgfpathrectangle{\pgfqpoint{2.873633in}{1.536516in}}{\pgfqpoint{2.901367in}{7.013484in}}%
\pgfusepath{clip}%
\pgfsetbuttcap%
\pgfsetmiterjoin%
\definecolor{currentfill}{rgb}{0.000000,0.543378,0.983379}%
\pgfsetfillcolor{currentfill}%
\pgfsetlinewidth{1.003750pt}%
\definecolor{currentstroke}{rgb}{1.000000,1.000000,1.000000}%
\pgfsetstrokecolor{currentstroke}%
\pgfsetstrokeopacity{0.800000}%
\pgfsetdash{}{0pt}%
\pgfpathmoveto{\pgfqpoint{2.873633in}{5.655086in}}%
\pgfpathlineto{\pgfqpoint{3.406444in}{5.655086in}}%
\pgfpathlineto{\pgfqpoint{3.406444in}{5.397474in}}%
\pgfpathlineto{\pgfqpoint{2.873633in}{5.397474in}}%
\pgfpathclose%
\pgfusepath{stroke,fill}%
\end{pgfscope}%
\begin{pgfscope}%
\pgfpathrectangle{\pgfqpoint{2.873633in}{1.536516in}}{\pgfqpoint{2.901367in}{7.013484in}}%
\pgfusepath{clip}%
\pgfsetbuttcap%
\pgfsetmiterjoin%
\definecolor{currentfill}{rgb}{0.000000,0.543378,0.983379}%
\pgfsetfillcolor{currentfill}%
\pgfsetlinewidth{1.003750pt}%
\definecolor{currentstroke}{rgb}{1.000000,1.000000,1.000000}%
\pgfsetstrokecolor{currentstroke}%
\pgfsetstrokeopacity{0.800000}%
\pgfsetdash{}{0pt}%
\pgfpathmoveto{\pgfqpoint{2.873633in}{5.333071in}}%
\pgfpathlineto{\pgfqpoint{3.358895in}{5.333071in}}%
\pgfpathlineto{\pgfqpoint{3.358895in}{5.075459in}}%
\pgfpathlineto{\pgfqpoint{2.873633in}{5.075459in}}%
\pgfpathclose%
\pgfusepath{stroke,fill}%
\end{pgfscope}%
\begin{pgfscope}%
\pgfpathrectangle{\pgfqpoint{2.873633in}{1.536516in}}{\pgfqpoint{2.901367in}{7.013484in}}%
\pgfusepath{clip}%
\pgfsetbuttcap%
\pgfsetmiterjoin%
\definecolor{currentfill}{rgb}{0.000000,0.543378,0.983379}%
\pgfsetfillcolor{currentfill}%
\pgfsetlinewidth{1.003750pt}%
\definecolor{currentstroke}{rgb}{1.000000,1.000000,1.000000}%
\pgfsetstrokecolor{currentstroke}%
\pgfsetstrokeopacity{0.800000}%
\pgfsetdash{}{0pt}%
\pgfpathmoveto{\pgfqpoint{2.873633in}{5.011056in}}%
\pgfpathlineto{\pgfqpoint{3.354100in}{5.011056in}}%
\pgfpathlineto{\pgfqpoint{3.354100in}{4.753444in}}%
\pgfpathlineto{\pgfqpoint{2.873633in}{4.753444in}}%
\pgfpathclose%
\pgfusepath{stroke,fill}%
\end{pgfscope}%
\begin{pgfscope}%
\pgfpathrectangle{\pgfqpoint{2.873633in}{1.536516in}}{\pgfqpoint{2.901367in}{7.013484in}}%
\pgfusepath{clip}%
\pgfsetbuttcap%
\pgfsetmiterjoin%
\definecolor{currentfill}{rgb}{0.000000,0.543378,0.983379}%
\pgfsetfillcolor{currentfill}%
\pgfsetlinewidth{1.003750pt}%
\definecolor{currentstroke}{rgb}{1.000000,1.000000,1.000000}%
\pgfsetstrokecolor{currentstroke}%
\pgfsetstrokeopacity{0.800000}%
\pgfsetdash{}{0pt}%
\pgfpathmoveto{\pgfqpoint{2.873633in}{4.689041in}}%
\pgfpathlineto{\pgfqpoint{3.339380in}{4.689041in}}%
\pgfpathlineto{\pgfqpoint{3.339380in}{4.431430in}}%
\pgfpathlineto{\pgfqpoint{2.873633in}{4.431430in}}%
\pgfpathclose%
\pgfusepath{stroke,fill}%
\end{pgfscope}%
\begin{pgfscope}%
\pgfpathrectangle{\pgfqpoint{2.873633in}{1.536516in}}{\pgfqpoint{2.901367in}{7.013484in}}%
\pgfusepath{clip}%
\pgfsetbuttcap%
\pgfsetmiterjoin%
\definecolor{currentfill}{rgb}{0.000000,0.543378,0.983379}%
\pgfsetfillcolor{currentfill}%
\pgfsetlinewidth{1.003750pt}%
\definecolor{currentstroke}{rgb}{1.000000,1.000000,1.000000}%
\pgfsetstrokecolor{currentstroke}%
\pgfsetstrokeopacity{0.800000}%
\pgfsetdash{}{0pt}%
\pgfpathmoveto{\pgfqpoint{2.873633in}{4.367027in}}%
\pgfpathlineto{\pgfqpoint{3.320465in}{4.367027in}}%
\pgfpathlineto{\pgfqpoint{3.320465in}{4.109415in}}%
\pgfpathlineto{\pgfqpoint{2.873633in}{4.109415in}}%
\pgfpathclose%
\pgfusepath{stroke,fill}%
\end{pgfscope}%
\begin{pgfscope}%
\pgfpathrectangle{\pgfqpoint{2.873633in}{1.536516in}}{\pgfqpoint{2.901367in}{7.013484in}}%
\pgfusepath{clip}%
\pgfsetbuttcap%
\pgfsetmiterjoin%
\definecolor{currentfill}{rgb}{0.000000,0.543378,0.983379}%
\pgfsetfillcolor{currentfill}%
\pgfsetlinewidth{1.003750pt}%
\definecolor{currentstroke}{rgb}{1.000000,1.000000,1.000000}%
\pgfsetstrokecolor{currentstroke}%
\pgfsetstrokeopacity{0.800000}%
\pgfsetdash{}{0pt}%
\pgfpathmoveto{\pgfqpoint{2.873633in}{4.045012in}}%
\pgfpathlineto{\pgfqpoint{3.306199in}{4.045012in}}%
\pgfpathlineto{\pgfqpoint{3.306199in}{3.787400in}}%
\pgfpathlineto{\pgfqpoint{2.873633in}{3.787400in}}%
\pgfpathclose%
\pgfusepath{stroke,fill}%
\end{pgfscope}%
\begin{pgfscope}%
\pgfpathrectangle{\pgfqpoint{2.873633in}{1.536516in}}{\pgfqpoint{2.901367in}{7.013484in}}%
\pgfusepath{clip}%
\pgfsetbuttcap%
\pgfsetmiterjoin%
\definecolor{currentfill}{rgb}{0.000000,0.543378,0.983379}%
\pgfsetfillcolor{currentfill}%
\pgfsetlinewidth{1.003750pt}%
\definecolor{currentstroke}{rgb}{1.000000,1.000000,1.000000}%
\pgfsetstrokecolor{currentstroke}%
\pgfsetstrokeopacity{0.800000}%
\pgfsetdash{}{0pt}%
\pgfpathmoveto{\pgfqpoint{2.873633in}{3.722997in}}%
\pgfpathlineto{\pgfqpoint{3.301053in}{3.722997in}}%
\pgfpathlineto{\pgfqpoint{3.301053in}{3.465385in}}%
\pgfpathlineto{\pgfqpoint{2.873633in}{3.465385in}}%
\pgfpathclose%
\pgfusepath{stroke,fill}%
\end{pgfscope}%
\begin{pgfscope}%
\pgfpathrectangle{\pgfqpoint{2.873633in}{1.536516in}}{\pgfqpoint{2.901367in}{7.013484in}}%
\pgfusepath{clip}%
\pgfsetbuttcap%
\pgfsetmiterjoin%
\definecolor{currentfill}{rgb}{0.000000,0.543378,0.983379}%
\pgfsetfillcolor{currentfill}%
\pgfsetlinewidth{1.003750pt}%
\definecolor{currentstroke}{rgb}{1.000000,1.000000,1.000000}%
\pgfsetstrokecolor{currentstroke}%
\pgfsetstrokeopacity{0.800000}%
\pgfsetdash{}{0pt}%
\pgfpathmoveto{\pgfqpoint{2.873633in}{3.400982in}}%
\pgfpathlineto{\pgfqpoint{3.295372in}{3.400982in}}%
\pgfpathlineto{\pgfqpoint{3.295372in}{3.143370in}}%
\pgfpathlineto{\pgfqpoint{2.873633in}{3.143370in}}%
\pgfpathclose%
\pgfusepath{stroke,fill}%
\end{pgfscope}%
\begin{pgfscope}%
\pgfpathrectangle{\pgfqpoint{2.873633in}{1.536516in}}{\pgfqpoint{2.901367in}{7.013484in}}%
\pgfusepath{clip}%
\pgfsetbuttcap%
\pgfsetmiterjoin%
\definecolor{currentfill}{rgb}{0.000000,0.543378,0.983379}%
\pgfsetfillcolor{currentfill}%
\pgfsetlinewidth{1.003750pt}%
\definecolor{currentstroke}{rgb}{1.000000,1.000000,1.000000}%
\pgfsetstrokecolor{currentstroke}%
\pgfsetstrokeopacity{0.800000}%
\pgfsetdash{}{0pt}%
\pgfpathmoveto{\pgfqpoint{2.873633in}{3.078967in}}%
\pgfpathlineto{\pgfqpoint{3.292601in}{3.078967in}}%
\pgfpathlineto{\pgfqpoint{3.292601in}{2.821355in}}%
\pgfpathlineto{\pgfqpoint{2.873633in}{2.821355in}}%
\pgfpathclose%
\pgfusepath{stroke,fill}%
\end{pgfscope}%
\begin{pgfscope}%
\pgfpathrectangle{\pgfqpoint{2.873633in}{1.536516in}}{\pgfqpoint{2.901367in}{7.013484in}}%
\pgfusepath{clip}%
\pgfsetbuttcap%
\pgfsetmiterjoin%
\definecolor{currentfill}{rgb}{0.000000,0.543378,0.983379}%
\pgfsetfillcolor{currentfill}%
\pgfsetlinewidth{1.003750pt}%
\definecolor{currentstroke}{rgb}{1.000000,1.000000,1.000000}%
\pgfsetstrokecolor{currentstroke}%
\pgfsetstrokeopacity{0.800000}%
\pgfsetdash{}{0pt}%
\pgfpathmoveto{\pgfqpoint{2.873633in}{2.756952in}}%
\pgfpathlineto{\pgfqpoint{3.283710in}{2.756952in}}%
\pgfpathlineto{\pgfqpoint{3.283710in}{2.499340in}}%
\pgfpathlineto{\pgfqpoint{2.873633in}{2.499340in}}%
\pgfpathclose%
\pgfusepath{stroke,fill}%
\end{pgfscope}%
\begin{pgfscope}%
\pgfpathrectangle{\pgfqpoint{2.873633in}{1.536516in}}{\pgfqpoint{2.901367in}{7.013484in}}%
\pgfusepath{clip}%
\pgfsetbuttcap%
\pgfsetmiterjoin%
\definecolor{currentfill}{rgb}{0.000000,0.543378,0.983379}%
\pgfsetfillcolor{currentfill}%
\pgfsetlinewidth{1.003750pt}%
\definecolor{currentstroke}{rgb}{1.000000,1.000000,1.000000}%
\pgfsetstrokecolor{currentstroke}%
\pgfsetstrokeopacity{0.800000}%
\pgfsetdash{}{0pt}%
\pgfpathmoveto{\pgfqpoint{2.873633in}{2.434937in}}%
\pgfpathlineto{\pgfqpoint{3.280529in}{2.434937in}}%
\pgfpathlineto{\pgfqpoint{3.280529in}{2.177325in}}%
\pgfpathlineto{\pgfqpoint{2.873633in}{2.177325in}}%
\pgfpathclose%
\pgfusepath{stroke,fill}%
\end{pgfscope}%
\begin{pgfscope}%
\pgfpathrectangle{\pgfqpoint{2.873633in}{1.536516in}}{\pgfqpoint{2.901367in}{7.013484in}}%
\pgfusepath{clip}%
\pgfsetbuttcap%
\pgfsetmiterjoin%
\definecolor{currentfill}{rgb}{0.000000,0.543378,0.983379}%
\pgfsetfillcolor{currentfill}%
\pgfsetlinewidth{1.003750pt}%
\definecolor{currentstroke}{rgb}{1.000000,1.000000,1.000000}%
\pgfsetstrokecolor{currentstroke}%
\pgfsetstrokeopacity{0.800000}%
\pgfsetdash{}{0pt}%
\pgfpathmoveto{\pgfqpoint{2.873633in}{2.112922in}}%
\pgfpathlineto{\pgfqpoint{3.280392in}{2.112922in}}%
\pgfpathlineto{\pgfqpoint{3.280392in}{1.855310in}}%
\pgfpathlineto{\pgfqpoint{2.873633in}{1.855310in}}%
\pgfpathclose%
\pgfusepath{stroke,fill}%
\end{pgfscope}%
\begin{pgfscope}%
\pgfsetbuttcap%
\pgfsetroundjoin%
\definecolor{currentfill}{rgb}{0.000000,0.000000,0.000000}%
\pgfsetfillcolor{currentfill}%
\pgfsetlinewidth{0.803000pt}%
\definecolor{currentstroke}{rgb}{0.000000,0.000000,0.000000}%
\pgfsetstrokecolor{currentstroke}%
\pgfsetdash{}{0pt}%
\pgfsys@defobject{currentmarker}{\pgfqpoint{0.000000in}{-0.048611in}}{\pgfqpoint{0.000000in}{0.000000in}}{%
\pgfpathmoveto{\pgfqpoint{0.000000in}{0.000000in}}%
\pgfpathlineto{\pgfqpoint{0.000000in}{-0.048611in}}%
\pgfusepath{stroke,fill}%
}%
\begin{pgfscope}%
\pgfsys@transformshift{2.873633in}{1.536516in}%
\pgfsys@useobject{currentmarker}{}%
\end{pgfscope}%
\end{pgfscope}%
\begin{pgfscope}%
\definecolor{textcolor}{rgb}{0.000000,0.000000,0.000000}%
\pgfsetstrokecolor{textcolor}%
\pgfsetfillcolor{textcolor}%
\pgftext[x=2.873633in,y=1.439293in,,top]{\color{textcolor}\sffamily\fontsize{16.000000}{19.200000}\selectfont \(\displaystyle {0.00}\)}%
\end{pgfscope}%
\begin{pgfscope}%
\pgfsetbuttcap%
\pgfsetroundjoin%
\definecolor{currentfill}{rgb}{0.000000,0.000000,0.000000}%
\pgfsetfillcolor{currentfill}%
\pgfsetlinewidth{0.803000pt}%
\definecolor{currentstroke}{rgb}{0.000000,0.000000,0.000000}%
\pgfsetstrokecolor{currentstroke}%
\pgfsetdash{}{0pt}%
\pgfsys@defobject{currentmarker}{\pgfqpoint{0.000000in}{-0.048611in}}{\pgfqpoint{0.000000in}{0.000000in}}{%
\pgfpathmoveto{\pgfqpoint{0.000000in}{0.000000in}}%
\pgfpathlineto{\pgfqpoint{0.000000in}{-0.048611in}}%
\pgfusepath{stroke,fill}%
}%
\begin{pgfscope}%
\pgfsys@transformshift{4.163522in}{1.536516in}%
\pgfsys@useobject{currentmarker}{}%
\end{pgfscope}%
\end{pgfscope}%
\begin{pgfscope}%
\definecolor{textcolor}{rgb}{0.000000,0.000000,0.000000}%
\pgfsetstrokecolor{textcolor}%
\pgfsetfillcolor{textcolor}%
\pgftext[x=4.163522in,y=1.439293in,,top]{\color{textcolor}\sffamily\fontsize{16.000000}{19.200000}\selectfont \(\displaystyle {0.01}\)}%
\end{pgfscope}%
\begin{pgfscope}%
\pgfsetbuttcap%
\pgfsetroundjoin%
\definecolor{currentfill}{rgb}{0.000000,0.000000,0.000000}%
\pgfsetfillcolor{currentfill}%
\pgfsetlinewidth{0.803000pt}%
\definecolor{currentstroke}{rgb}{0.000000,0.000000,0.000000}%
\pgfsetstrokecolor{currentstroke}%
\pgfsetdash{}{0pt}%
\pgfsys@defobject{currentmarker}{\pgfqpoint{0.000000in}{-0.048611in}}{\pgfqpoint{0.000000in}{0.000000in}}{%
\pgfpathmoveto{\pgfqpoint{0.000000in}{0.000000in}}%
\pgfpathlineto{\pgfqpoint{0.000000in}{-0.048611in}}%
\pgfusepath{stroke,fill}%
}%
\begin{pgfscope}%
\pgfsys@transformshift{5.453412in}{1.536516in}%
\pgfsys@useobject{currentmarker}{}%
\end{pgfscope}%
\end{pgfscope}%
\begin{pgfscope}%
\definecolor{textcolor}{rgb}{0.000000,0.000000,0.000000}%
\pgfsetstrokecolor{textcolor}%
\pgfsetfillcolor{textcolor}%
\pgftext[x=5.453412in,y=1.439293in,,top]{\color{textcolor}\sffamily\fontsize{16.000000}{19.200000}\selectfont \(\displaystyle {0.02}\)}%
\end{pgfscope}%
\begin{pgfscope}%
\definecolor{textcolor}{rgb}{0.000000,0.000000,0.000000}%
\pgfsetstrokecolor{textcolor}%
\pgfsetfillcolor{textcolor}%
\pgftext[x=4.324317in,y=1.170389in,,top]{\color{textcolor}\sffamily\fontsize{16.000000}{19.200000}\selectfont Mean absolute Shapley value}%
\end{pgfscope}%
\begin{pgfscope}%
\definecolor{textcolor}{rgb}{0.000000,0.000000,0.000000}%
\pgfsetstrokecolor{textcolor}%
\pgfsetfillcolor{textcolor}%
\pgftext[x=0.048424in, y=8.012507in, left, base]{\color{textcolor}\sffamily\fontsize{16.000000}{19.200000}\selectfont SOFA deterioration (derived)}%
\end{pgfscope}%
\begin{pgfscope}%
\definecolor{textcolor}{rgb}{0.000000,0.000000,0.000000}%
\pgfsetstrokecolor{textcolor}%
\pgfsetfillcolor{textcolor}%
\pgftext[x=1.307770in, y=7.690493in, left, base]{\color{textcolor}\sffamily\fontsize{16.000000}{19.200000}\selectfont SOFA (derived)}%
\end{pgfscope}%
\begin{pgfscope}%
\definecolor{textcolor}{rgb}{0.000000,0.000000,0.000000}%
\pgfsetstrokecolor{textcolor}%
\pgfsetfillcolor{textcolor}%
\pgftext[x=1.999025in, y=7.375036in, left, base]{\color{textcolor}\sffamily\fontsize{16.000000}{19.200000}\selectfont Lactate }%
\end{pgfscope}%
\begin{pgfscope}%
\definecolor{textcolor}{rgb}{0.000000,0.000000,0.000000}%
\pgfsetstrokecolor{textcolor}%
\pgfsetfillcolor{textcolor}%
\pgftext[x=0.160306in, y=7.053021in, left, base]{\color{textcolor}\sffamily\fontsize{16.000000}{19.200000}\selectfont Fraction of inspired oxygen }%
\end{pgfscope}%
\begin{pgfscope}%
\definecolor{textcolor}{rgb}{0.000000,0.000000,0.000000}%
\pgfsetstrokecolor{textcolor}%
\pgfsetfillcolor{textcolor}%
\pgftext[x=1.739285in, y=6.731006in, left, base]{\color{textcolor}\sffamily\fontsize{16.000000}{19.200000}\selectfont Heart rate }%
\end{pgfscope}%
\begin{pgfscope}%
\definecolor{textcolor}{rgb}{0.000000,0.000000,0.000000}%
\pgfsetstrokecolor{textcolor}%
\pgfsetfillcolor{textcolor}%
\pgftext[x=0.623361in, y=6.408991in, left, base]{\color{textcolor}\sffamily\fontsize{16.000000}{19.200000}\selectfont Mean arterial pressure }%
\end{pgfscope}%
\begin{pgfscope}%
\definecolor{textcolor}{rgb}{0.000000,0.000000,0.000000}%
\pgfsetstrokecolor{textcolor}%
\pgfsetfillcolor{textcolor}%
\pgftext[x=0.782503in, y=6.086977in, left, base]{\color{textcolor}\sffamily\fontsize{16.000000}{19.200000}\selectfont CO2 partial pressure }%
\end{pgfscope}%
\begin{pgfscope}%
\definecolor{textcolor}{rgb}{0.000000,0.000000,0.000000}%
\pgfsetstrokecolor{textcolor}%
\pgfsetfillcolor{textcolor}%
\pgftext[x=1.423027in, y=5.764962in, left, base]{\color{textcolor}\sffamily\fontsize{16.000000}{19.200000}\selectfont Endtidal CO2 }%
\end{pgfscope}%
\begin{pgfscope}%
\definecolor{textcolor}{rgb}{0.000000,0.000000,0.000000}%
\pgfsetstrokecolor{textcolor}%
\pgfsetfillcolor{textcolor}%
\pgftext[x=0.926696in, y=5.442947in, left, base]{\color{textcolor}\sffamily\fontsize{16.000000}{19.200000}\selectfont O2 partial pressure }%
\end{pgfscope}%
\begin{pgfscope}%
\definecolor{textcolor}{rgb}{0.000000,0.000000,0.000000}%
\pgfsetstrokecolor{textcolor}%
\pgfsetfillcolor{textcolor}%
\pgftext[x=0.461036in, y=5.120932in, left, base]{\color{textcolor}\sffamily\fontsize{16.000000}{19.200000}\selectfont Diastolic blood pressure }%
\end{pgfscope}%
\begin{pgfscope}%
\definecolor{textcolor}{rgb}{0.000000,0.000000,0.000000}%
\pgfsetstrokecolor{textcolor}%
\pgfsetfillcolor{textcolor}%
\pgftext[x=1.628754in, y=4.798917in, left, base]{\color{textcolor}\sffamily\fontsize{16.000000}{19.200000}\selectfont Neutrophils }%
\end{pgfscope}%
\begin{pgfscope}%
\definecolor{textcolor}{rgb}{0.000000,0.000000,0.000000}%
\pgfsetstrokecolor{textcolor}%
\pgfsetfillcolor{textcolor}%
\pgftext[x=1.270539in, y=4.476902in, left, base]{\color{textcolor}\sffamily\fontsize{16.000000}{19.200000}\selectfont Creatine kinase }%
\end{pgfscope}%
\begin{pgfscope}%
\definecolor{textcolor}{rgb}{0.000000,0.000000,0.000000}%
\pgfsetstrokecolor{textcolor}%
\pgfsetfillcolor{textcolor}%
\pgftext[x=1.603194in, y=4.154887in, left, base]{\color{textcolor}\sffamily\fontsize{16.000000}{19.200000}\selectfont Base excess }%
\end{pgfscope}%
\begin{pgfscope}%
\definecolor{textcolor}{rgb}{0.000000,0.000000,0.000000}%
\pgfsetstrokecolor{textcolor}%
\pgfsetfillcolor{textcolor}%
\pgftext[x=0.760128in, y=3.826314in, left, base]{\color{textcolor}\sffamily\fontsize{16.000000}{19.200000}\selectfont PaO2/FiO2 (derived)}%
\end{pgfscope}%
\begin{pgfscope}%
\definecolor{textcolor}{rgb}{0.000000,0.000000,0.000000}%
\pgfsetstrokecolor{textcolor}%
\pgfsetfillcolor{textcolor}%
\pgftext[x=1.189522in, y=3.504299in, left, base]{\color{textcolor}\sffamily\fontsize{16.000000}{19.200000}\selectfont MEWS (derived)}%
\end{pgfscope}%
\begin{pgfscope}%
\definecolor{textcolor}{rgb}{0.000000,0.000000,0.000000}%
\pgfsetstrokecolor{textcolor}%
\pgfsetfillcolor{textcolor}%
\pgftext[x=1.698873in, y=3.188843in, left, base]{\color{textcolor}\sffamily\fontsize{16.000000}{19.200000}\selectfont Troponin t }%
\end{pgfscope}%
\begin{pgfscope}%
\definecolor{textcolor}{rgb}{0.000000,0.000000,0.000000}%
\pgfsetstrokecolor{textcolor}%
\pgfsetfillcolor{textcolor}%
\pgftext[x=1.031054in, y=2.866828in, left, base]{\color{textcolor}\sffamily\fontsize{16.000000}{19.200000}\selectfont C-reactive protein }%
\end{pgfscope}%
\begin{pgfscope}%
\definecolor{textcolor}{rgb}{0.000000,0.000000,0.000000}%
\pgfsetstrokecolor{textcolor}%
\pgfsetfillcolor{textcolor}%
\pgftext[x=1.632033in, y=2.544813in, left, base]{\color{textcolor}\sffamily\fontsize{16.000000}{19.200000}\selectfont Magnesium }%
\end{pgfscope}%
\begin{pgfscope}%
\definecolor{textcolor}{rgb}{0.000000,0.000000,0.000000}%
\pgfsetstrokecolor{textcolor}%
\pgfsetfillcolor{textcolor}%
\pgftext[x=0.738908in, y=2.222798in, left, base]{\color{textcolor}\sffamily\fontsize{16.000000}{19.200000}\selectfont Alkaline phosphatase }%
\end{pgfscope}%
\begin{pgfscope}%
\definecolor{textcolor}{rgb}{0.000000,0.000000,0.000000}%
\pgfsetstrokecolor{textcolor}%
\pgfsetfillcolor{textcolor}%
\pgftext[x=1.577057in, y=1.900783in, left, base]{\color{textcolor}\sffamily\fontsize{16.000000}{19.200000}\selectfont Bicarbonate }%
\end{pgfscope}%
\begin{pgfscope}%
\pgfpathrectangle{\pgfqpoint{2.873633in}{1.536516in}}{\pgfqpoint{2.901367in}{7.013484in}}%
\pgfusepath{clip}%
\pgfsetbuttcap%
\pgfsetroundjoin%
\pgfsetlinewidth{1.505625pt}%
\definecolor{currentstroke}{rgb}{0.000000,0.000000,0.000000}%
\pgfsetstrokecolor{currentstroke}%
\pgfsetdash{}{0pt}%
\pgfpathmoveto{\pgfqpoint{4.704290in}{8.102399in}}%
\pgfpathlineto{\pgfqpoint{5.636840in}{8.102399in}}%
\pgfusepath{stroke}%
\end{pgfscope}%
\begin{pgfscope}%
\pgfpathrectangle{\pgfqpoint{2.873633in}{1.536516in}}{\pgfqpoint{2.901367in}{7.013484in}}%
\pgfusepath{clip}%
\pgfsetbuttcap%
\pgfsetroundjoin%
\pgfsetlinewidth{1.505625pt}%
\definecolor{currentstroke}{rgb}{0.000000,0.000000,0.000000}%
\pgfsetstrokecolor{currentstroke}%
\pgfsetdash{}{0pt}%
\pgfpathmoveto{\pgfqpoint{4.224502in}{7.780384in}}%
\pgfpathlineto{\pgfqpoint{5.268833in}{7.780384in}}%
\pgfusepath{stroke}%
\end{pgfscope}%
\begin{pgfscope}%
\pgfpathrectangle{\pgfqpoint{2.873633in}{1.536516in}}{\pgfqpoint{2.901367in}{7.013484in}}%
\pgfusepath{clip}%
\pgfsetbuttcap%
\pgfsetroundjoin%
\pgfsetlinewidth{1.505625pt}%
\definecolor{currentstroke}{rgb}{0.000000,0.000000,0.000000}%
\pgfsetstrokecolor{currentstroke}%
\pgfsetdash{}{0pt}%
\pgfpathmoveto{\pgfqpoint{4.151705in}{7.458370in}}%
\pgfpathlineto{\pgfqpoint{5.285407in}{7.458370in}}%
\pgfusepath{stroke}%
\end{pgfscope}%
\begin{pgfscope}%
\pgfpathrectangle{\pgfqpoint{2.873633in}{1.536516in}}{\pgfqpoint{2.901367in}{7.013484in}}%
\pgfusepath{clip}%
\pgfsetbuttcap%
\pgfsetroundjoin%
\pgfsetlinewidth{1.505625pt}%
\definecolor{currentstroke}{rgb}{0.000000,0.000000,0.000000}%
\pgfsetstrokecolor{currentstroke}%
\pgfsetdash{}{0pt}%
\pgfpathmoveto{\pgfqpoint{4.048340in}{7.136355in}}%
\pgfpathlineto{\pgfqpoint{4.728959in}{7.136355in}}%
\pgfusepath{stroke}%
\end{pgfscope}%
\begin{pgfscope}%
\pgfpathrectangle{\pgfqpoint{2.873633in}{1.536516in}}{\pgfqpoint{2.901367in}{7.013484in}}%
\pgfusepath{clip}%
\pgfsetbuttcap%
\pgfsetroundjoin%
\pgfsetlinewidth{1.505625pt}%
\definecolor{currentstroke}{rgb}{0.000000,0.000000,0.000000}%
\pgfsetstrokecolor{currentstroke}%
\pgfsetdash{}{0pt}%
\pgfpathmoveto{\pgfqpoint{3.607061in}{6.814340in}}%
\pgfpathlineto{\pgfqpoint{3.841025in}{6.814340in}}%
\pgfusepath{stroke}%
\end{pgfscope}%
\begin{pgfscope}%
\pgfpathrectangle{\pgfqpoint{2.873633in}{1.536516in}}{\pgfqpoint{2.901367in}{7.013484in}}%
\pgfusepath{clip}%
\pgfsetbuttcap%
\pgfsetroundjoin%
\pgfsetlinewidth{1.505625pt}%
\definecolor{currentstroke}{rgb}{0.000000,0.000000,0.000000}%
\pgfsetstrokecolor{currentstroke}%
\pgfsetdash{}{0pt}%
\pgfpathmoveto{\pgfqpoint{3.496013in}{6.492325in}}%
\pgfpathlineto{\pgfqpoint{3.792565in}{6.492325in}}%
\pgfusepath{stroke}%
\end{pgfscope}%
\begin{pgfscope}%
\pgfpathrectangle{\pgfqpoint{2.873633in}{1.536516in}}{\pgfqpoint{2.901367in}{7.013484in}}%
\pgfusepath{clip}%
\pgfsetbuttcap%
\pgfsetroundjoin%
\pgfsetlinewidth{1.505625pt}%
\definecolor{currentstroke}{rgb}{0.000000,0.000000,0.000000}%
\pgfsetstrokecolor{currentstroke}%
\pgfsetdash{}{0pt}%
\pgfpathmoveto{\pgfqpoint{3.434925in}{6.170310in}}%
\pgfpathlineto{\pgfqpoint{3.971456in}{6.170310in}}%
\pgfusepath{stroke}%
\end{pgfscope}%
\begin{pgfscope}%
\pgfpathrectangle{\pgfqpoint{2.873633in}{1.536516in}}{\pgfqpoint{2.901367in}{7.013484in}}%
\pgfusepath{clip}%
\pgfsetbuttcap%
\pgfsetroundjoin%
\pgfsetlinewidth{1.505625pt}%
\definecolor{currentstroke}{rgb}{0.000000,0.000000,0.000000}%
\pgfsetstrokecolor{currentstroke}%
\pgfsetdash{}{0pt}%
\pgfpathmoveto{\pgfqpoint{3.409696in}{5.848295in}}%
\pgfpathlineto{\pgfqpoint{4.031568in}{5.848295in}}%
\pgfusepath{stroke}%
\end{pgfscope}%
\begin{pgfscope}%
\pgfpathrectangle{\pgfqpoint{2.873633in}{1.536516in}}{\pgfqpoint{2.901367in}{7.013484in}}%
\pgfusepath{clip}%
\pgfsetbuttcap%
\pgfsetroundjoin%
\pgfsetlinewidth{1.505625pt}%
\definecolor{currentstroke}{rgb}{0.000000,0.000000,0.000000}%
\pgfsetstrokecolor{currentstroke}%
\pgfsetdash{}{0pt}%
\pgfpathmoveto{\pgfqpoint{3.406444in}{5.526280in}}%
\pgfpathlineto{\pgfqpoint{3.707817in}{5.526280in}}%
\pgfusepath{stroke}%
\end{pgfscope}%
\begin{pgfscope}%
\pgfpathrectangle{\pgfqpoint{2.873633in}{1.536516in}}{\pgfqpoint{2.901367in}{7.013484in}}%
\pgfusepath{clip}%
\pgfsetbuttcap%
\pgfsetroundjoin%
\pgfsetlinewidth{1.505625pt}%
\definecolor{currentstroke}{rgb}{0.000000,0.000000,0.000000}%
\pgfsetstrokecolor{currentstroke}%
\pgfsetdash{}{0pt}%
\pgfpathmoveto{\pgfqpoint{3.358895in}{5.204265in}}%
\pgfpathlineto{\pgfqpoint{3.580952in}{5.204265in}}%
\pgfusepath{stroke}%
\end{pgfscope}%
\begin{pgfscope}%
\pgfpathrectangle{\pgfqpoint{2.873633in}{1.536516in}}{\pgfqpoint{2.901367in}{7.013484in}}%
\pgfusepath{clip}%
\pgfsetbuttcap%
\pgfsetroundjoin%
\pgfsetlinewidth{1.505625pt}%
\definecolor{currentstroke}{rgb}{0.000000,0.000000,0.000000}%
\pgfsetstrokecolor{currentstroke}%
\pgfsetdash{}{0pt}%
\pgfpathmoveto{\pgfqpoint{3.354100in}{4.882250in}}%
\pgfpathlineto{\pgfqpoint{3.697192in}{4.882250in}}%
\pgfusepath{stroke}%
\end{pgfscope}%
\begin{pgfscope}%
\pgfpathrectangle{\pgfqpoint{2.873633in}{1.536516in}}{\pgfqpoint{2.901367in}{7.013484in}}%
\pgfusepath{clip}%
\pgfsetbuttcap%
\pgfsetroundjoin%
\pgfsetlinewidth{1.505625pt}%
\definecolor{currentstroke}{rgb}{0.000000,0.000000,0.000000}%
\pgfsetstrokecolor{currentstroke}%
\pgfsetdash{}{0pt}%
\pgfpathmoveto{\pgfqpoint{3.339380in}{4.560235in}}%
\pgfpathlineto{\pgfqpoint{3.788849in}{4.560235in}}%
\pgfusepath{stroke}%
\end{pgfscope}%
\begin{pgfscope}%
\pgfpathrectangle{\pgfqpoint{2.873633in}{1.536516in}}{\pgfqpoint{2.901367in}{7.013484in}}%
\pgfusepath{clip}%
\pgfsetbuttcap%
\pgfsetroundjoin%
\pgfsetlinewidth{1.505625pt}%
\definecolor{currentstroke}{rgb}{0.000000,0.000000,0.000000}%
\pgfsetstrokecolor{currentstroke}%
\pgfsetdash{}{0pt}%
\pgfpathmoveto{\pgfqpoint{3.320465in}{4.238221in}}%
\pgfpathlineto{\pgfqpoint{3.569142in}{4.238221in}}%
\pgfusepath{stroke}%
\end{pgfscope}%
\begin{pgfscope}%
\pgfpathrectangle{\pgfqpoint{2.873633in}{1.536516in}}{\pgfqpoint{2.901367in}{7.013484in}}%
\pgfusepath{clip}%
\pgfsetbuttcap%
\pgfsetroundjoin%
\pgfsetlinewidth{1.505625pt}%
\definecolor{currentstroke}{rgb}{0.000000,0.000000,0.000000}%
\pgfsetstrokecolor{currentstroke}%
\pgfsetdash{}{0pt}%
\pgfpathmoveto{\pgfqpoint{3.306199in}{3.916206in}}%
\pgfpathlineto{\pgfqpoint{3.678112in}{3.916206in}}%
\pgfusepath{stroke}%
\end{pgfscope}%
\begin{pgfscope}%
\pgfpathrectangle{\pgfqpoint{2.873633in}{1.536516in}}{\pgfqpoint{2.901367in}{7.013484in}}%
\pgfusepath{clip}%
\pgfsetbuttcap%
\pgfsetroundjoin%
\pgfsetlinewidth{1.505625pt}%
\definecolor{currentstroke}{rgb}{0.000000,0.000000,0.000000}%
\pgfsetstrokecolor{currentstroke}%
\pgfsetdash{}{0pt}%
\pgfpathmoveto{\pgfqpoint{3.301053in}{3.594191in}}%
\pgfpathlineto{\pgfqpoint{3.349203in}{3.594191in}}%
\pgfusepath{stroke}%
\end{pgfscope}%
\begin{pgfscope}%
\pgfpathrectangle{\pgfqpoint{2.873633in}{1.536516in}}{\pgfqpoint{2.901367in}{7.013484in}}%
\pgfusepath{clip}%
\pgfsetbuttcap%
\pgfsetroundjoin%
\pgfsetlinewidth{1.505625pt}%
\definecolor{currentstroke}{rgb}{0.000000,0.000000,0.000000}%
\pgfsetstrokecolor{currentstroke}%
\pgfsetdash{}{0pt}%
\pgfpathmoveto{\pgfqpoint{3.295372in}{3.272176in}}%
\pgfpathlineto{\pgfqpoint{3.693027in}{3.272176in}}%
\pgfusepath{stroke}%
\end{pgfscope}%
\begin{pgfscope}%
\pgfpathrectangle{\pgfqpoint{2.873633in}{1.536516in}}{\pgfqpoint{2.901367in}{7.013484in}}%
\pgfusepath{clip}%
\pgfsetbuttcap%
\pgfsetroundjoin%
\pgfsetlinewidth{1.505625pt}%
\definecolor{currentstroke}{rgb}{0.000000,0.000000,0.000000}%
\pgfsetstrokecolor{currentstroke}%
\pgfsetdash{}{0pt}%
\pgfpathmoveto{\pgfqpoint{3.292601in}{2.950161in}}%
\pgfpathlineto{\pgfqpoint{3.723344in}{2.950161in}}%
\pgfusepath{stroke}%
\end{pgfscope}%
\begin{pgfscope}%
\pgfpathrectangle{\pgfqpoint{2.873633in}{1.536516in}}{\pgfqpoint{2.901367in}{7.013484in}}%
\pgfusepath{clip}%
\pgfsetbuttcap%
\pgfsetroundjoin%
\pgfsetlinewidth{1.505625pt}%
\definecolor{currentstroke}{rgb}{0.000000,0.000000,0.000000}%
\pgfsetstrokecolor{currentstroke}%
\pgfsetdash{}{0pt}%
\pgfpathmoveto{\pgfqpoint{3.283710in}{2.628146in}}%
\pgfpathlineto{\pgfqpoint{3.454844in}{2.628146in}}%
\pgfusepath{stroke}%
\end{pgfscope}%
\begin{pgfscope}%
\pgfpathrectangle{\pgfqpoint{2.873633in}{1.536516in}}{\pgfqpoint{2.901367in}{7.013484in}}%
\pgfusepath{clip}%
\pgfsetbuttcap%
\pgfsetroundjoin%
\pgfsetlinewidth{1.505625pt}%
\definecolor{currentstroke}{rgb}{0.000000,0.000000,0.000000}%
\pgfsetstrokecolor{currentstroke}%
\pgfsetdash{}{0pt}%
\pgfpathmoveto{\pgfqpoint{3.280529in}{2.306131in}}%
\pgfpathlineto{\pgfqpoint{3.534615in}{2.306131in}}%
\pgfusepath{stroke}%
\end{pgfscope}%
\begin{pgfscope}%
\pgfpathrectangle{\pgfqpoint{2.873633in}{1.536516in}}{\pgfqpoint{2.901367in}{7.013484in}}%
\pgfusepath{clip}%
\pgfsetbuttcap%
\pgfsetroundjoin%
\pgfsetlinewidth{1.505625pt}%
\definecolor{currentstroke}{rgb}{0.000000,0.000000,0.000000}%
\pgfsetstrokecolor{currentstroke}%
\pgfsetdash{}{0pt}%
\pgfpathmoveto{\pgfqpoint{3.280392in}{1.984116in}}%
\pgfpathlineto{\pgfqpoint{3.636261in}{1.984116in}}%
\pgfusepath{stroke}%
\end{pgfscope}%
\begin{pgfscope}%
\pgfsetrectcap%
\pgfsetmiterjoin%
\pgfsetlinewidth{0.803000pt}%
\definecolor{currentstroke}{rgb}{0.000000,0.000000,0.000000}%
\pgfsetstrokecolor{currentstroke}%
\pgfsetdash{}{0pt}%
\pgfpathmoveto{\pgfqpoint{2.873633in}{1.536516in}}%
\pgfpathlineto{\pgfqpoint{2.873633in}{8.550000in}}%
\pgfusepath{stroke}%
\end{pgfscope}%
\begin{pgfscope}%
\pgfsetrectcap%
\pgfsetmiterjoin%
\pgfsetlinewidth{0.803000pt}%
\definecolor{currentstroke}{rgb}{0.000000,0.000000,0.000000}%
\pgfsetstrokecolor{currentstroke}%
\pgfsetdash{}{0pt}%
\pgfpathmoveto{\pgfqpoint{2.873633in}{1.536516in}}%
\pgfpathlineto{\pgfqpoint{5.775000in}{1.536516in}}%
\pgfusepath{stroke}%
\end{pgfscope}%
\end{pgfpicture}%
\makeatother%
\endgroup%

%% file: figures/shapley/shapley_vx8vbt08_16h_EICU_dot.pgf
\begingroup%
\makeatletter%
\begin{pgfpicture}%
\pgfpathrectangle{\pgfpointorigin}{\pgfqpoint{8.000000in}{9.500000in}}%
\pgfusepath{use as bounding box, clip}%
\begin{pgfscope}%
\pgfsetbuttcap%
\pgfsetmiterjoin%
\definecolor{currentfill}{rgb}{1.000000,1.000000,1.000000}%
\pgfsetfillcolor{currentfill}%
\pgfsetlinewidth{0.000000pt}%
\definecolor{currentstroke}{rgb}{1.000000,1.000000,1.000000}%
\pgfsetstrokecolor{currentstroke}%
\pgfsetdash{}{0pt}%
\pgfpathmoveto{\pgfqpoint{0.000000in}{0.000000in}}%
\pgfpathlineto{\pgfqpoint{8.000000in}{0.000000in}}%
\pgfpathlineto{\pgfqpoint{8.000000in}{9.500000in}}%
\pgfpathlineto{\pgfqpoint{0.000000in}{9.500000in}}%
\pgfpathclose%
\pgfusepath{fill}%
\end{pgfscope}%
\begin{pgfscope}%
\pgfsetbuttcap%
\pgfsetmiterjoin%
\definecolor{currentfill}{rgb}{1.000000,1.000000,1.000000}%
\pgfsetfillcolor{currentfill}%
\pgfsetlinewidth{0.000000pt}%
\definecolor{currentstroke}{rgb}{0.000000,0.000000,0.000000}%
\pgfsetstrokecolor{currentstroke}%
\pgfsetstrokeopacity{0.000000}%
\pgfsetdash{}{0pt}%
\pgfpathmoveto{\pgfqpoint{3.170388in}{0.694630in}}%
\pgfpathlineto{\pgfqpoint{6.842078in}{0.694630in}}%
\pgfpathlineto{\pgfqpoint{6.842078in}{9.207193in}}%
\pgfpathlineto{\pgfqpoint{3.170388in}{9.207193in}}%
\pgfpathclose%
\pgfusepath{fill}%
\end{pgfscope}%
\begin{pgfscope}%
\pgfpathrectangle{\pgfqpoint{3.170388in}{0.694630in}}{\pgfqpoint{3.671689in}{8.512564in}}%
\pgfusepath{clip}%
\pgfsetrectcap%
\pgfsetroundjoin%
\pgfsetlinewidth{1.505625pt}%
\definecolor{currentstroke}{rgb}{0.600000,0.600000,0.600000}%
\pgfsetstrokecolor{currentstroke}%
\pgfsetdash{}{0pt}%
\pgfpathmoveto{\pgfqpoint{5.284721in}{0.694630in}}%
\pgfpathlineto{\pgfqpoint{5.284721in}{9.207193in}}%
\pgfusepath{stroke}%
\end{pgfscope}%
\begin{pgfscope}%
\pgfpathrectangle{\pgfqpoint{3.170388in}{0.694630in}}{\pgfqpoint{3.671689in}{8.512564in}}%
\pgfusepath{clip}%
\pgfsetbuttcap%
\pgfsetroundjoin%
\pgfsetlinewidth{0.501875pt}%
\definecolor{currentstroke}{rgb}{0.800000,0.800000,0.800000}%
\pgfsetstrokecolor{currentstroke}%
\pgfsetdash{{0.500000pt}{2.500000pt}}{0.000000pt}%
\pgfpathmoveto{\pgfqpoint{3.170388in}{1.099990in}}%
\pgfpathlineto{\pgfqpoint{6.842078in}{1.099990in}}%
\pgfusepath{stroke}%
\end{pgfscope}%
\begin{pgfscope}%
\pgfpathrectangle{\pgfqpoint{3.170388in}{0.694630in}}{\pgfqpoint{3.671689in}{8.512564in}}%
\pgfusepath{clip}%
\pgfsetbuttcap%
\pgfsetroundjoin%
\pgfsetlinewidth{0.501875pt}%
\definecolor{currentstroke}{rgb}{0.800000,0.800000,0.800000}%
\pgfsetstrokecolor{currentstroke}%
\pgfsetdash{{0.500000pt}{2.500000pt}}{0.000000pt}%
\pgfpathmoveto{\pgfqpoint{3.170388in}{1.505350in}}%
\pgfpathlineto{\pgfqpoint{6.842078in}{1.505350in}}%
\pgfusepath{stroke}%
\end{pgfscope}%
\begin{pgfscope}%
\pgfpathrectangle{\pgfqpoint{3.170388in}{0.694630in}}{\pgfqpoint{3.671689in}{8.512564in}}%
\pgfusepath{clip}%
\pgfsetbuttcap%
\pgfsetroundjoin%
\pgfsetlinewidth{0.501875pt}%
\definecolor{currentstroke}{rgb}{0.800000,0.800000,0.800000}%
\pgfsetstrokecolor{currentstroke}%
\pgfsetdash{{0.500000pt}{2.500000pt}}{0.000000pt}%
\pgfpathmoveto{\pgfqpoint{3.170388in}{1.910710in}}%
\pgfpathlineto{\pgfqpoint{6.842078in}{1.910710in}}%
\pgfusepath{stroke}%
\end{pgfscope}%
\begin{pgfscope}%
\pgfpathrectangle{\pgfqpoint{3.170388in}{0.694630in}}{\pgfqpoint{3.671689in}{8.512564in}}%
\pgfusepath{clip}%
\pgfsetbuttcap%
\pgfsetroundjoin%
\pgfsetlinewidth{0.501875pt}%
\definecolor{currentstroke}{rgb}{0.800000,0.800000,0.800000}%
\pgfsetstrokecolor{currentstroke}%
\pgfsetdash{{0.500000pt}{2.500000pt}}{0.000000pt}%
\pgfpathmoveto{\pgfqpoint{3.170388in}{2.316070in}}%
\pgfpathlineto{\pgfqpoint{6.842078in}{2.316070in}}%
\pgfusepath{stroke}%
\end{pgfscope}%
\begin{pgfscope}%
\pgfpathrectangle{\pgfqpoint{3.170388in}{0.694630in}}{\pgfqpoint{3.671689in}{8.512564in}}%
\pgfusepath{clip}%
\pgfsetbuttcap%
\pgfsetroundjoin%
\pgfsetlinewidth{0.501875pt}%
\definecolor{currentstroke}{rgb}{0.800000,0.800000,0.800000}%
\pgfsetstrokecolor{currentstroke}%
\pgfsetdash{{0.500000pt}{2.500000pt}}{0.000000pt}%
\pgfpathmoveto{\pgfqpoint{3.170388in}{2.721430in}}%
\pgfpathlineto{\pgfqpoint{6.842078in}{2.721430in}}%
\pgfusepath{stroke}%
\end{pgfscope}%
\begin{pgfscope}%
\pgfpathrectangle{\pgfqpoint{3.170388in}{0.694630in}}{\pgfqpoint{3.671689in}{8.512564in}}%
\pgfusepath{clip}%
\pgfsetbuttcap%
\pgfsetroundjoin%
\pgfsetlinewidth{0.501875pt}%
\definecolor{currentstroke}{rgb}{0.800000,0.800000,0.800000}%
\pgfsetstrokecolor{currentstroke}%
\pgfsetdash{{0.500000pt}{2.500000pt}}{0.000000pt}%
\pgfpathmoveto{\pgfqpoint{3.170388in}{3.126791in}}%
\pgfpathlineto{\pgfqpoint{6.842078in}{3.126791in}}%
\pgfusepath{stroke}%
\end{pgfscope}%
\begin{pgfscope}%
\pgfpathrectangle{\pgfqpoint{3.170388in}{0.694630in}}{\pgfqpoint{3.671689in}{8.512564in}}%
\pgfusepath{clip}%
\pgfsetbuttcap%
\pgfsetroundjoin%
\pgfsetlinewidth{0.501875pt}%
\definecolor{currentstroke}{rgb}{0.800000,0.800000,0.800000}%
\pgfsetstrokecolor{currentstroke}%
\pgfsetdash{{0.500000pt}{2.500000pt}}{0.000000pt}%
\pgfpathmoveto{\pgfqpoint{3.170388in}{3.532151in}}%
\pgfpathlineto{\pgfqpoint{6.842078in}{3.532151in}}%
\pgfusepath{stroke}%
\end{pgfscope}%
\begin{pgfscope}%
\pgfpathrectangle{\pgfqpoint{3.170388in}{0.694630in}}{\pgfqpoint{3.671689in}{8.512564in}}%
\pgfusepath{clip}%
\pgfsetbuttcap%
\pgfsetroundjoin%
\pgfsetlinewidth{0.501875pt}%
\definecolor{currentstroke}{rgb}{0.800000,0.800000,0.800000}%
\pgfsetstrokecolor{currentstroke}%
\pgfsetdash{{0.500000pt}{2.500000pt}}{0.000000pt}%
\pgfpathmoveto{\pgfqpoint{3.170388in}{3.937511in}}%
\pgfpathlineto{\pgfqpoint{6.842078in}{3.937511in}}%
\pgfusepath{stroke}%
\end{pgfscope}%
\begin{pgfscope}%
\pgfpathrectangle{\pgfqpoint{3.170388in}{0.694630in}}{\pgfqpoint{3.671689in}{8.512564in}}%
\pgfusepath{clip}%
\pgfsetbuttcap%
\pgfsetroundjoin%
\pgfsetlinewidth{0.501875pt}%
\definecolor{currentstroke}{rgb}{0.800000,0.800000,0.800000}%
\pgfsetstrokecolor{currentstroke}%
\pgfsetdash{{0.500000pt}{2.500000pt}}{0.000000pt}%
\pgfpathmoveto{\pgfqpoint{3.170388in}{4.342871in}}%
\pgfpathlineto{\pgfqpoint{6.842078in}{4.342871in}}%
\pgfusepath{stroke}%
\end{pgfscope}%
\begin{pgfscope}%
\pgfpathrectangle{\pgfqpoint{3.170388in}{0.694630in}}{\pgfqpoint{3.671689in}{8.512564in}}%
\pgfusepath{clip}%
\pgfsetbuttcap%
\pgfsetroundjoin%
\pgfsetlinewidth{0.501875pt}%
\definecolor{currentstroke}{rgb}{0.800000,0.800000,0.800000}%
\pgfsetstrokecolor{currentstroke}%
\pgfsetdash{{0.500000pt}{2.500000pt}}{0.000000pt}%
\pgfpathmoveto{\pgfqpoint{3.170388in}{4.748231in}}%
\pgfpathlineto{\pgfqpoint{6.842078in}{4.748231in}}%
\pgfusepath{stroke}%
\end{pgfscope}%
\begin{pgfscope}%
\pgfpathrectangle{\pgfqpoint{3.170388in}{0.694630in}}{\pgfqpoint{3.671689in}{8.512564in}}%
\pgfusepath{clip}%
\pgfsetbuttcap%
\pgfsetroundjoin%
\pgfsetlinewidth{0.501875pt}%
\definecolor{currentstroke}{rgb}{0.800000,0.800000,0.800000}%
\pgfsetstrokecolor{currentstroke}%
\pgfsetdash{{0.500000pt}{2.500000pt}}{0.000000pt}%
\pgfpathmoveto{\pgfqpoint{3.170388in}{5.153592in}}%
\pgfpathlineto{\pgfqpoint{6.842078in}{5.153592in}}%
\pgfusepath{stroke}%
\end{pgfscope}%
\begin{pgfscope}%
\pgfpathrectangle{\pgfqpoint{3.170388in}{0.694630in}}{\pgfqpoint{3.671689in}{8.512564in}}%
\pgfusepath{clip}%
\pgfsetbuttcap%
\pgfsetroundjoin%
\pgfsetlinewidth{0.501875pt}%
\definecolor{currentstroke}{rgb}{0.800000,0.800000,0.800000}%
\pgfsetstrokecolor{currentstroke}%
\pgfsetdash{{0.500000pt}{2.500000pt}}{0.000000pt}%
\pgfpathmoveto{\pgfqpoint{3.170388in}{5.558952in}}%
\pgfpathlineto{\pgfqpoint{6.842078in}{5.558952in}}%
\pgfusepath{stroke}%
\end{pgfscope}%
\begin{pgfscope}%
\pgfpathrectangle{\pgfqpoint{3.170388in}{0.694630in}}{\pgfqpoint{3.671689in}{8.512564in}}%
\pgfusepath{clip}%
\pgfsetbuttcap%
\pgfsetroundjoin%
\pgfsetlinewidth{0.501875pt}%
\definecolor{currentstroke}{rgb}{0.800000,0.800000,0.800000}%
\pgfsetstrokecolor{currentstroke}%
\pgfsetdash{{0.500000pt}{2.500000pt}}{0.000000pt}%
\pgfpathmoveto{\pgfqpoint{3.170388in}{5.964312in}}%
\pgfpathlineto{\pgfqpoint{6.842078in}{5.964312in}}%
\pgfusepath{stroke}%
\end{pgfscope}%
\begin{pgfscope}%
\pgfpathrectangle{\pgfqpoint{3.170388in}{0.694630in}}{\pgfqpoint{3.671689in}{8.512564in}}%
\pgfusepath{clip}%
\pgfsetbuttcap%
\pgfsetroundjoin%
\pgfsetlinewidth{0.501875pt}%
\definecolor{currentstroke}{rgb}{0.800000,0.800000,0.800000}%
\pgfsetstrokecolor{currentstroke}%
\pgfsetdash{{0.500000pt}{2.500000pt}}{0.000000pt}%
\pgfpathmoveto{\pgfqpoint{3.170388in}{6.369672in}}%
\pgfpathlineto{\pgfqpoint{6.842078in}{6.369672in}}%
\pgfusepath{stroke}%
\end{pgfscope}%
\begin{pgfscope}%
\pgfpathrectangle{\pgfqpoint{3.170388in}{0.694630in}}{\pgfqpoint{3.671689in}{8.512564in}}%
\pgfusepath{clip}%
\pgfsetbuttcap%
\pgfsetroundjoin%
\pgfsetlinewidth{0.501875pt}%
\definecolor{currentstroke}{rgb}{0.800000,0.800000,0.800000}%
\pgfsetstrokecolor{currentstroke}%
\pgfsetdash{{0.500000pt}{2.500000pt}}{0.000000pt}%
\pgfpathmoveto{\pgfqpoint{3.170388in}{6.775032in}}%
\pgfpathlineto{\pgfqpoint{6.842078in}{6.775032in}}%
\pgfusepath{stroke}%
\end{pgfscope}%
\begin{pgfscope}%
\pgfpathrectangle{\pgfqpoint{3.170388in}{0.694630in}}{\pgfqpoint{3.671689in}{8.512564in}}%
\pgfusepath{clip}%
\pgfsetbuttcap%
\pgfsetroundjoin%
\pgfsetlinewidth{0.501875pt}%
\definecolor{currentstroke}{rgb}{0.800000,0.800000,0.800000}%
\pgfsetstrokecolor{currentstroke}%
\pgfsetdash{{0.500000pt}{2.500000pt}}{0.000000pt}%
\pgfpathmoveto{\pgfqpoint{3.170388in}{7.180392in}}%
\pgfpathlineto{\pgfqpoint{6.842078in}{7.180392in}}%
\pgfusepath{stroke}%
\end{pgfscope}%
\begin{pgfscope}%
\pgfpathrectangle{\pgfqpoint{3.170388in}{0.694630in}}{\pgfqpoint{3.671689in}{8.512564in}}%
\pgfusepath{clip}%
\pgfsetbuttcap%
\pgfsetroundjoin%
\pgfsetlinewidth{0.501875pt}%
\definecolor{currentstroke}{rgb}{0.800000,0.800000,0.800000}%
\pgfsetstrokecolor{currentstroke}%
\pgfsetdash{{0.500000pt}{2.500000pt}}{0.000000pt}%
\pgfpathmoveto{\pgfqpoint{3.170388in}{7.585753in}}%
\pgfpathlineto{\pgfqpoint{6.842078in}{7.585753in}}%
\pgfusepath{stroke}%
\end{pgfscope}%
\begin{pgfscope}%
\pgfpathrectangle{\pgfqpoint{3.170388in}{0.694630in}}{\pgfqpoint{3.671689in}{8.512564in}}%
\pgfusepath{clip}%
\pgfsetbuttcap%
\pgfsetroundjoin%
\pgfsetlinewidth{0.501875pt}%
\definecolor{currentstroke}{rgb}{0.800000,0.800000,0.800000}%
\pgfsetstrokecolor{currentstroke}%
\pgfsetdash{{0.500000pt}{2.500000pt}}{0.000000pt}%
\pgfpathmoveto{\pgfqpoint{3.170388in}{7.991113in}}%
\pgfpathlineto{\pgfqpoint{6.842078in}{7.991113in}}%
\pgfusepath{stroke}%
\end{pgfscope}%
\begin{pgfscope}%
\pgfpathrectangle{\pgfqpoint{3.170388in}{0.694630in}}{\pgfqpoint{3.671689in}{8.512564in}}%
\pgfusepath{clip}%
\pgfsetbuttcap%
\pgfsetroundjoin%
\pgfsetlinewidth{0.501875pt}%
\definecolor{currentstroke}{rgb}{0.800000,0.800000,0.800000}%
\pgfsetstrokecolor{currentstroke}%
\pgfsetdash{{0.500000pt}{2.500000pt}}{0.000000pt}%
\pgfpathmoveto{\pgfqpoint{3.170388in}{8.396473in}}%
\pgfpathlineto{\pgfqpoint{6.842078in}{8.396473in}}%
\pgfusepath{stroke}%
\end{pgfscope}%
\begin{pgfscope}%
\pgfpathrectangle{\pgfqpoint{3.170388in}{0.694630in}}{\pgfqpoint{3.671689in}{8.512564in}}%
\pgfusepath{clip}%
\pgfsetbuttcap%
\pgfsetroundjoin%
\pgfsetlinewidth{0.501875pt}%
\definecolor{currentstroke}{rgb}{0.800000,0.800000,0.800000}%
\pgfsetstrokecolor{currentstroke}%
\pgfsetdash{{0.500000pt}{2.500000pt}}{0.000000pt}%
\pgfpathmoveto{\pgfqpoint{3.170388in}{8.801833in}}%
\pgfpathlineto{\pgfqpoint{6.842078in}{8.801833in}}%
\pgfusepath{stroke}%
\end{pgfscope}%
\begin{pgfscope}%
\pgfsetbuttcap%
\pgfsetroundjoin%
\definecolor{currentfill}{rgb}{0.200000,0.200000,0.200000}%
\pgfsetfillcolor{currentfill}%
\pgfsetlinewidth{0.803000pt}%
\definecolor{currentstroke}{rgb}{0.200000,0.200000,0.200000}%
\pgfsetstrokecolor{currentstroke}%
\pgfsetdash{}{0pt}%
\pgfsys@defobject{currentmarker}{\pgfqpoint{0.000000in}{-0.048611in}}{\pgfqpoint{0.000000in}{0.000000in}}{%
\pgfpathmoveto{\pgfqpoint{0.000000in}{0.000000in}}%
\pgfpathlineto{\pgfqpoint{0.000000in}{-0.048611in}}%
\pgfusepath{stroke,fill}%
}%
\begin{pgfscope}%
\pgfsys@transformshift{3.200453in}{0.694630in}%
\pgfsys@useobject{currentmarker}{}%
\end{pgfscope}%
\end{pgfscope}%
\begin{pgfscope}%
\definecolor{textcolor}{rgb}{0.200000,0.200000,0.200000}%
\pgfsetstrokecolor{textcolor}%
\pgfsetfillcolor{textcolor}%
\pgftext[x=3.200453in,y=0.597407in,,top]{\color{textcolor}\sffamily\fontsize{11.000000}{13.200000}\selectfont \(\displaystyle {\ensuremath{-}1.5}\)}%
\end{pgfscope}%
\begin{pgfscope}%
\pgfsetbuttcap%
\pgfsetroundjoin%
\definecolor{currentfill}{rgb}{0.200000,0.200000,0.200000}%
\pgfsetfillcolor{currentfill}%
\pgfsetlinewidth{0.803000pt}%
\definecolor{currentstroke}{rgb}{0.200000,0.200000,0.200000}%
\pgfsetstrokecolor{currentstroke}%
\pgfsetdash{}{0pt}%
\pgfsys@defobject{currentmarker}{\pgfqpoint{0.000000in}{-0.048611in}}{\pgfqpoint{0.000000in}{0.000000in}}{%
\pgfpathmoveto{\pgfqpoint{0.000000in}{0.000000in}}%
\pgfpathlineto{\pgfqpoint{0.000000in}{-0.048611in}}%
\pgfusepath{stroke,fill}%
}%
\begin{pgfscope}%
\pgfsys@transformshift{3.895209in}{0.694630in}%
\pgfsys@useobject{currentmarker}{}%
\end{pgfscope}%
\end{pgfscope}%
\begin{pgfscope}%
\definecolor{textcolor}{rgb}{0.200000,0.200000,0.200000}%
\pgfsetstrokecolor{textcolor}%
\pgfsetfillcolor{textcolor}%
\pgftext[x=3.895209in,y=0.597407in,,top]{\color{textcolor}\sffamily\fontsize{11.000000}{13.200000}\selectfont \(\displaystyle {\ensuremath{-}1.0}\)}%
\end{pgfscope}%
\begin{pgfscope}%
\pgfsetbuttcap%
\pgfsetroundjoin%
\definecolor{currentfill}{rgb}{0.200000,0.200000,0.200000}%
\pgfsetfillcolor{currentfill}%
\pgfsetlinewidth{0.803000pt}%
\definecolor{currentstroke}{rgb}{0.200000,0.200000,0.200000}%
\pgfsetstrokecolor{currentstroke}%
\pgfsetdash{}{0pt}%
\pgfsys@defobject{currentmarker}{\pgfqpoint{0.000000in}{-0.048611in}}{\pgfqpoint{0.000000in}{0.000000in}}{%
\pgfpathmoveto{\pgfqpoint{0.000000in}{0.000000in}}%
\pgfpathlineto{\pgfqpoint{0.000000in}{-0.048611in}}%
\pgfusepath{stroke,fill}%
}%
\begin{pgfscope}%
\pgfsys@transformshift{4.589965in}{0.694630in}%
\pgfsys@useobject{currentmarker}{}%
\end{pgfscope}%
\end{pgfscope}%
\begin{pgfscope}%
\definecolor{textcolor}{rgb}{0.200000,0.200000,0.200000}%
\pgfsetstrokecolor{textcolor}%
\pgfsetfillcolor{textcolor}%
\pgftext[x=4.589965in,y=0.597407in,,top]{\color{textcolor}\sffamily\fontsize{11.000000}{13.200000}\selectfont \(\displaystyle {\ensuremath{-}0.5}\)}%
\end{pgfscope}%
\begin{pgfscope}%
\pgfsetbuttcap%
\pgfsetroundjoin%
\definecolor{currentfill}{rgb}{0.200000,0.200000,0.200000}%
\pgfsetfillcolor{currentfill}%
\pgfsetlinewidth{0.803000pt}%
\definecolor{currentstroke}{rgb}{0.200000,0.200000,0.200000}%
\pgfsetstrokecolor{currentstroke}%
\pgfsetdash{}{0pt}%
\pgfsys@defobject{currentmarker}{\pgfqpoint{0.000000in}{-0.048611in}}{\pgfqpoint{0.000000in}{0.000000in}}{%
\pgfpathmoveto{\pgfqpoint{0.000000in}{0.000000in}}%
\pgfpathlineto{\pgfqpoint{0.000000in}{-0.048611in}}%
\pgfusepath{stroke,fill}%
}%
\begin{pgfscope}%
\pgfsys@transformshift{5.284721in}{0.694630in}%
\pgfsys@useobject{currentmarker}{}%
\end{pgfscope}%
\end{pgfscope}%
\begin{pgfscope}%
\definecolor{textcolor}{rgb}{0.200000,0.200000,0.200000}%
\pgfsetstrokecolor{textcolor}%
\pgfsetfillcolor{textcolor}%
\pgftext[x=5.284721in,y=0.597407in,,top]{\color{textcolor}\sffamily\fontsize{11.000000}{13.200000}\selectfont \(\displaystyle {0.0}\)}%
\end{pgfscope}%
\begin{pgfscope}%
\pgfsetbuttcap%
\pgfsetroundjoin%
\definecolor{currentfill}{rgb}{0.200000,0.200000,0.200000}%
\pgfsetfillcolor{currentfill}%
\pgfsetlinewidth{0.803000pt}%
\definecolor{currentstroke}{rgb}{0.200000,0.200000,0.200000}%
\pgfsetstrokecolor{currentstroke}%
\pgfsetdash{}{0pt}%
\pgfsys@defobject{currentmarker}{\pgfqpoint{0.000000in}{-0.048611in}}{\pgfqpoint{0.000000in}{0.000000in}}{%
\pgfpathmoveto{\pgfqpoint{0.000000in}{0.000000in}}%
\pgfpathlineto{\pgfqpoint{0.000000in}{-0.048611in}}%
\pgfusepath{stroke,fill}%
}%
\begin{pgfscope}%
\pgfsys@transformshift{5.979477in}{0.694630in}%
\pgfsys@useobject{currentmarker}{}%
\end{pgfscope}%
\end{pgfscope}%
\begin{pgfscope}%
\definecolor{textcolor}{rgb}{0.200000,0.200000,0.200000}%
\pgfsetstrokecolor{textcolor}%
\pgfsetfillcolor{textcolor}%
\pgftext[x=5.979477in,y=0.597407in,,top]{\color{textcolor}\sffamily\fontsize{11.000000}{13.200000}\selectfont \(\displaystyle {0.5}\)}%
\end{pgfscope}%
\begin{pgfscope}%
\pgfsetbuttcap%
\pgfsetroundjoin%
\definecolor{currentfill}{rgb}{0.200000,0.200000,0.200000}%
\pgfsetfillcolor{currentfill}%
\pgfsetlinewidth{0.803000pt}%
\definecolor{currentstroke}{rgb}{0.200000,0.200000,0.200000}%
\pgfsetstrokecolor{currentstroke}%
\pgfsetdash{}{0pt}%
\pgfsys@defobject{currentmarker}{\pgfqpoint{0.000000in}{-0.048611in}}{\pgfqpoint{0.000000in}{0.000000in}}{%
\pgfpathmoveto{\pgfqpoint{0.000000in}{0.000000in}}%
\pgfpathlineto{\pgfqpoint{0.000000in}{-0.048611in}}%
\pgfusepath{stroke,fill}%
}%
\begin{pgfscope}%
\pgfsys@transformshift{6.674233in}{0.694630in}%
\pgfsys@useobject{currentmarker}{}%
\end{pgfscope}%
\end{pgfscope}%
\begin{pgfscope}%
\definecolor{textcolor}{rgb}{0.200000,0.200000,0.200000}%
\pgfsetstrokecolor{textcolor}%
\pgfsetfillcolor{textcolor}%
\pgftext[x=6.674233in,y=0.597407in,,top]{\color{textcolor}\sffamily\fontsize{11.000000}{13.200000}\selectfont \(\displaystyle {1.0}\)}%
\end{pgfscope}%
\begin{pgfscope}%
\definecolor{textcolor}{rgb}{0.000000,0.000000,0.000000}%
\pgfsetstrokecolor{textcolor}%
\pgfsetfillcolor{textcolor}%
\pgftext[x=5.006233in,y=0.406667in,,top]{\color{textcolor}\sffamily\fontsize{13.000000}{15.600000}\selectfont SHAP value (impact on model output)}%
\end{pgfscope}%
\begin{pgfscope}%
\definecolor{textcolor}{rgb}{0.200000,0.200000,0.200000}%
\pgfsetstrokecolor{textcolor}%
\pgfsetfillcolor{textcolor}%
\pgftext[x=0.923092in, y=1.037490in, left, base]{\color{textcolor}\sffamily\fontsize{13.000000}{15.600000}\selectfont Mean arterial pressure (raw)}%
\end{pgfscope}%
\begin{pgfscope}%
\definecolor{textcolor}{rgb}{0.200000,0.200000,0.200000}%
\pgfsetstrokecolor{textcolor}%
\pgfsetfillcolor{textcolor}%
\pgftext[x=1.387709in, y=1.442850in, left, base]{\color{textcolor}\sffamily\fontsize{13.000000}{15.600000}\selectfont PaO2/FiO2 (derived)}%
\end{pgfscope}%
\begin{pgfscope}%
\definecolor{textcolor}{rgb}{0.200000,0.200000,0.200000}%
\pgfsetstrokecolor{textcolor}%
\pgfsetfillcolor{textcolor}%
\pgftext[x=1.494285in, y=1.848210in, left, base]{\color{textcolor}\sffamily\fontsize{13.000000}{15.600000}\selectfont Totcal CO2 (count)}%
\end{pgfscope}%
\begin{pgfscope}%
\definecolor{textcolor}{rgb}{0.200000,0.200000,0.200000}%
\pgfsetstrokecolor{textcolor}%
\pgfsetfillcolor{textcolor}%
\pgftext[x=0.765359in, y=2.253570in, left, base]{\color{textcolor}\sffamily\fontsize{13.000000}{15.600000}\selectfont Band form neutrophils (count)}%
\end{pgfscope}%
\begin{pgfscope}%
\definecolor{textcolor}{rgb}{0.200000,0.200000,0.200000}%
\pgfsetstrokecolor{textcolor}%
\pgfsetfillcolor{textcolor}%
\pgftext[x=1.783731in, y=2.658930in, left, base]{\color{textcolor}\sffamily\fontsize{13.000000}{15.600000}\selectfont SOFA (derived)}%
\end{pgfscope}%
\begin{pgfscope}%
\definecolor{textcolor}{rgb}{0.200000,0.200000,0.200000}%
\pgfsetstrokecolor{textcolor}%
\pgfsetfillcolor{textcolor}%
\pgftext[x=1.850956in, y=3.064291in, left, base]{\color{textcolor}\sffamily\fontsize{13.000000}{15.600000}\selectfont SIRS (derived)}%
\end{pgfscope}%
\begin{pgfscope}%
\definecolor{textcolor}{rgb}{0.200000,0.200000,0.200000}%
\pgfsetstrokecolor{textcolor}%
\pgfsetfillcolor{textcolor}%
\pgftext[x=1.488903in, y=3.469651in, left, base]{\color{textcolor}\sffamily\fontsize{13.000000}{15.600000}\selectfont Base excess (count)}%
\end{pgfscope}%
\begin{pgfscope}%
\definecolor{textcolor}{rgb}{0.200000,0.200000,0.200000}%
\pgfsetstrokecolor{textcolor}%
\pgfsetfillcolor{textcolor}%
\pgftext[x=1.477214in, y=3.875011in, left, base]{\color{textcolor}\sffamily\fontsize{13.000000}{15.600000}\selectfont Hemoglobin (count)}%
\end{pgfscope}%
\begin{pgfscope}%
\definecolor{textcolor}{rgb}{0.200000,0.200000,0.200000}%
\pgfsetstrokecolor{textcolor}%
\pgfsetfillcolor{textcolor}%
\pgftext[x=1.728658in, y=4.280371in, left, base]{\color{textcolor}\sffamily\fontsize{13.000000}{15.600000}\selectfont Chloride (count)}%
\end{pgfscope}%
\begin{pgfscope}%
\definecolor{textcolor}{rgb}{0.200000,0.200000,0.200000}%
\pgfsetstrokecolor{textcolor}%
\pgfsetfillcolor{textcolor}%
\pgftext[x=0.866477in, y=4.685731in, left, base]{\color{textcolor}\sffamily\fontsize{13.000000}{15.600000}\selectfont Alkaline phosphatase (count)}%
\end{pgfscope}%
\begin{pgfscope}%
\definecolor{textcolor}{rgb}{0.200000,0.200000,0.200000}%
\pgfsetstrokecolor{textcolor}%
\pgfsetfillcolor{textcolor}%
\pgftext[x=0.242121in, y=5.091092in, left, base]{\color{textcolor}\sffamily\fontsize{13.000000}{15.600000}\selectfont Fraction of inspired oxygen (indicator)}%
\end{pgfscope}%
\begin{pgfscope}%
\definecolor{textcolor}{rgb}{0.200000,0.200000,0.200000}%
\pgfsetstrokecolor{textcolor}%
\pgfsetfillcolor{textcolor}%
\pgftext[x=1.000830in, y=5.496452in, left, base]{\color{textcolor}\sffamily\fontsize{13.000000}{15.600000}\selectfont O2 partial pressure (count)}%
\end{pgfscope}%
\begin{pgfscope}%
\definecolor{textcolor}{rgb}{0.200000,0.200000,0.200000}%
\pgfsetstrokecolor{textcolor}%
\pgfsetfillcolor{textcolor}%
\pgftext[x=0.240000in, y=5.901812in, left, base]{\color{textcolor}\sffamily\fontsize{13.000000}{15.600000}\selectfont Erythrocyte distribution width (count)}%
\end{pgfscope}%
\begin{pgfscope}%
\definecolor{textcolor}{rgb}{0.200000,0.200000,0.200000}%
\pgfsetstrokecolor{textcolor}%
\pgfsetfillcolor{textcolor}%
\pgftext[x=1.470752in, y=6.307172in, left, base]{\color{textcolor}\sffamily\fontsize{13.000000}{15.600000}\selectfont Bicarbonate (count)}%
\end{pgfscope}%
\begin{pgfscope}%
\definecolor{textcolor}{rgb}{0.200000,0.200000,0.200000}%
\pgfsetstrokecolor{textcolor}%
\pgfsetfillcolor{textcolor}%
\pgftext[x=1.530339in, y=6.712532in, left, base]{\color{textcolor}\sffamily\fontsize{13.000000}{15.600000}\selectfont Eosinophils (count)}%
\end{pgfscope}%
\begin{pgfscope}%
\definecolor{textcolor}{rgb}{0.200000,0.200000,0.200000}%
\pgfsetstrokecolor{textcolor}%
\pgfsetfillcolor{textcolor}%
\pgftext[x=1.508792in, y=7.117892in, left, base]{\color{textcolor}\sffamily\fontsize{13.000000}{15.600000}\selectfont Neutrophils (count)}%
\end{pgfscope}%
\begin{pgfscope}%
\definecolor{textcolor}{rgb}{0.200000,0.200000,0.200000}%
\pgfsetstrokecolor{textcolor}%
\pgfsetfillcolor{textcolor}%
\pgftext[x=1.382018in, y=7.523253in, left, base]{\color{textcolor}\sffamily\fontsize{13.000000}{15.600000}\selectfont Heart rate (indicator)}%
\end{pgfscope}%
\begin{pgfscope}%
\definecolor{textcolor}{rgb}{0.200000,0.200000,0.200000}%
\pgfsetstrokecolor{textcolor}%
\pgfsetfillcolor{textcolor}%
\pgftext[x=0.896568in, y=7.928613in, left, base]{\color{textcolor}\sffamily\fontsize{13.000000}{15.600000}\selectfont CO2 partial pressure (count)}%
\end{pgfscope}%
\begin{pgfscope}%
\definecolor{textcolor}{rgb}{0.200000,0.200000,0.200000}%
\pgfsetstrokecolor{textcolor}%
\pgfsetfillcolor{textcolor}%
\pgftext[x=0.874502in, y=8.333973in, left, base]{\color{textcolor}\sffamily\fontsize{13.000000}{15.600000}\selectfont SOFA deterioration (derived)}%
\end{pgfscope}%
\begin{pgfscope}%
\definecolor{textcolor}{rgb}{0.200000,0.200000,0.200000}%
\pgfsetstrokecolor{textcolor}%
\pgfsetfillcolor{textcolor}%
\pgftext[x=1.774857in, y=8.739333in, left, base]{\color{textcolor}\sffamily\fontsize{13.000000}{15.600000}\selectfont Lactate (count)}%
\end{pgfscope}%
\begin{pgfscope}%
\pgfsetrectcap%
\pgfsetmiterjoin%
\pgfsetlinewidth{0.803000pt}%
\definecolor{currentstroke}{rgb}{0.000000,0.000000,0.000000}%
\pgfsetstrokecolor{currentstroke}%
\pgfsetdash{}{0pt}%
\pgfpathmoveto{\pgfqpoint{3.170388in}{0.694630in}}%
\pgfpathlineto{\pgfqpoint{6.842078in}{0.694630in}}%
\pgfusepath{stroke}%
\end{pgfscope}%
\begin{pgfscope}%
\pgfsys@transformshift{3.306667in}{0.923333in}%
\pgftext[left,bottom]{\includegraphics[interpolate=true,width=3.396667in,height=8.053333in]{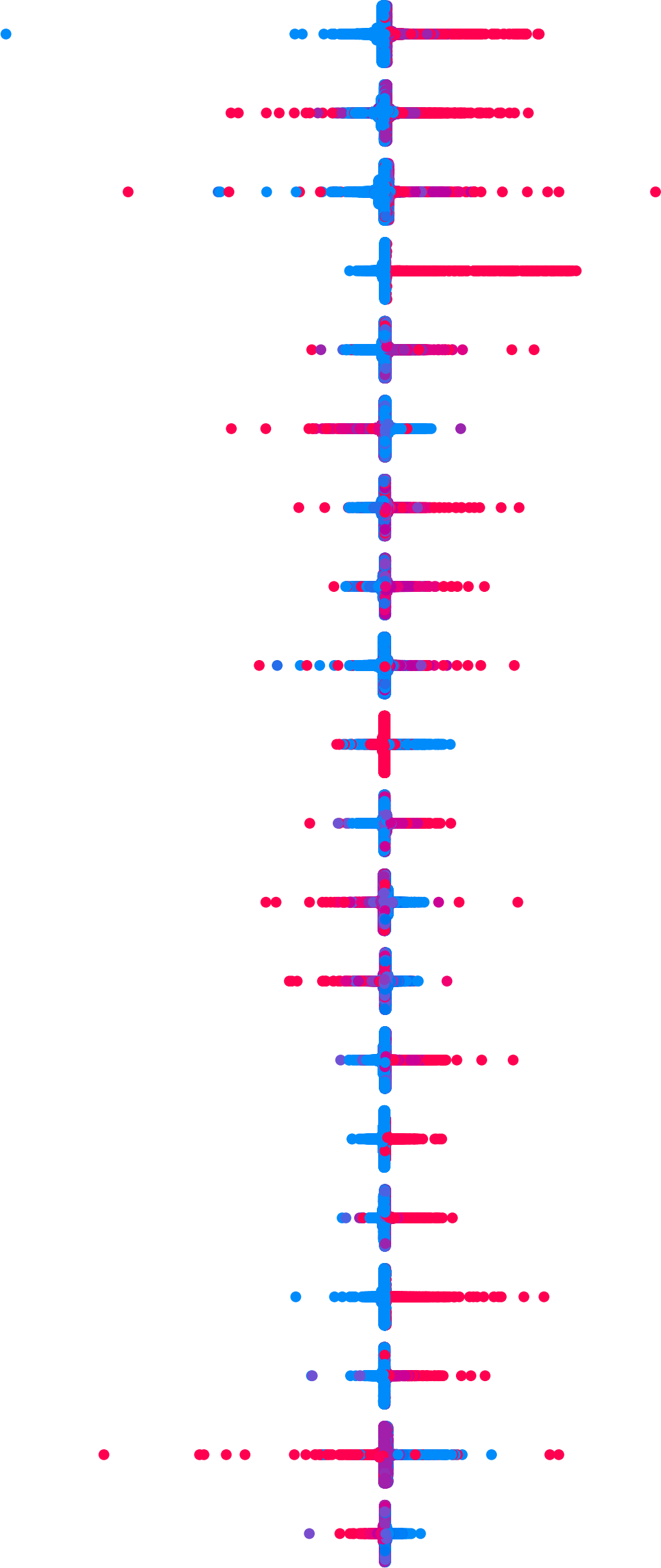}}%
\end{pgfscope}%
\begin{pgfscope}%
\pgfsetbuttcap%
\pgfsetmiterjoin%
\definecolor{currentfill}{rgb}{1.000000,1.000000,1.000000}%
\pgfsetfillcolor{currentfill}%
\pgfsetlinewidth{0.000000pt}%
\definecolor{currentstroke}{rgb}{0.000000,0.000000,0.000000}%
\pgfsetstrokecolor{currentstroke}%
\pgfsetstrokeopacity{0.000000}%
\pgfsetdash{}{0pt}%
\pgfpathmoveto{\pgfqpoint{7.071558in}{0.694630in}}%
\pgfpathlineto{\pgfqpoint{7.129216in}{0.694630in}}%
\pgfpathlineto{\pgfqpoint{7.129216in}{9.207193in}}%
\pgfpathlineto{\pgfqpoint{7.071558in}{9.207193in}}%
\pgfpathclose%
\pgfusepath{fill}%
\end{pgfscope}%
\begin{pgfscope}%
\pgfpathrectangle{\pgfqpoint{7.071558in}{0.694630in}}{\pgfqpoint{0.057658in}{8.512564in}}%
\pgfusepath{clip}%
\pgfsetbuttcap%
\pgfsetmiterjoin%
\definecolor{currentfill}{rgb}{1.000000,1.000000,1.000000}%
\pgfsetfillcolor{currentfill}%
\pgfsetlinewidth{0.010037pt}%
\definecolor{currentstroke}{rgb}{1.000000,1.000000,1.000000}%
\pgfsetstrokecolor{currentstroke}%
\pgfsetdash{}{0pt}%
\pgfpathmoveto{\pgfqpoint{7.071558in}{0.694630in}}%
\pgfpathlineto{\pgfqpoint{7.071558in}{0.727882in}}%
\pgfpathlineto{\pgfqpoint{7.071558in}{9.173941in}}%
\pgfpathlineto{\pgfqpoint{7.071558in}{9.207193in}}%
\pgfpathlineto{\pgfqpoint{7.129216in}{9.207193in}}%
\pgfpathlineto{\pgfqpoint{7.129216in}{9.173941in}}%
\pgfpathlineto{\pgfqpoint{7.129216in}{0.727882in}}%
\pgfpathlineto{\pgfqpoint{7.129216in}{0.694630in}}%
\pgfpathlineto{\pgfqpoint{7.129216in}{0.694630in}}%
\pgfpathclose%
\pgfusepath{stroke,fill}%
\end{pgfscope}%
\begin{pgfscope}%
\pgfsys@transformshift{7.070000in}{0.693333in}%
\pgftext[left,bottom]{\includegraphics[interpolate=true,width=0.060000in,height=8.513333in]{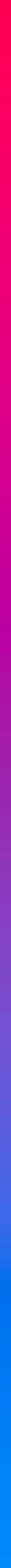}}%
\end{pgfscope}%
\begin{pgfscope}%
\definecolor{textcolor}{rgb}{0.000000,0.000000,0.000000}%
\pgfsetstrokecolor{textcolor}%
\pgfsetfillcolor{textcolor}%
\pgftext[x=7.177827in, y=0.641823in, left, base]{\color{textcolor}\sffamily\fontsize{11.000000}{13.200000}\selectfont Low}%
\end{pgfscope}%
\begin{pgfscope}%
\definecolor{textcolor}{rgb}{0.000000,0.000000,0.000000}%
\pgfsetstrokecolor{textcolor}%
\pgfsetfillcolor{textcolor}%
\pgftext[x=7.177827in, y=9.154387in, left, base]{\color{textcolor}\sffamily\fontsize{11.000000}{13.200000}\selectfont High}%
\end{pgfscope}%
\begin{pgfscope}%
\definecolor{textcolor}{rgb}{0.000000,0.000000,0.000000}%
\pgfsetstrokecolor{textcolor}%
\pgfsetfillcolor{textcolor}%
\pgftext[x=7.476502in,y=4.950911in,,top,rotate=90.000000]{\color{textcolor}\sffamily\fontsize{12.000000}{14.400000}\selectfont Feature value}%
\end{pgfscope}%
\end{pgfpicture}%
\makeatother%
\endgroup%

%% file: figures/shapley/shapley_vx8vbt08_16h_EICU_scatter_map.pgf
\begingroup%
\makeatletter%
\begin{pgfpicture}%
\pgfpathrectangle{\pgfpointorigin}{\pgfqpoint{4.000000in}{2.000000in}}%
\pgfusepath{use as bounding box, clip}%
\begin{pgfscope}%
\pgfsetbuttcap%
\pgfsetmiterjoin%
\definecolor{currentfill}{rgb}{1.000000,1.000000,1.000000}%
\pgfsetfillcolor{currentfill}%
\pgfsetlinewidth{0.000000pt}%
\definecolor{currentstroke}{rgb}{1.000000,1.000000,1.000000}%
\pgfsetstrokecolor{currentstroke}%
\pgfsetdash{}{0pt}%
\pgfpathmoveto{\pgfqpoint{0.000000in}{0.000000in}}%
\pgfpathlineto{\pgfqpoint{4.000000in}{0.000000in}}%
\pgfpathlineto{\pgfqpoint{4.000000in}{2.000000in}}%
\pgfpathlineto{\pgfqpoint{0.000000in}{2.000000in}}%
\pgfpathclose%
\pgfusepath{fill}%
\end{pgfscope}%
\begin{pgfscope}%
\pgfsetbuttcap%
\pgfsetmiterjoin%
\definecolor{currentfill}{rgb}{1.000000,1.000000,1.000000}%
\pgfsetfillcolor{currentfill}%
\pgfsetlinewidth{0.000000pt}%
\definecolor{currentstroke}{rgb}{0.000000,0.000000,0.000000}%
\pgfsetstrokecolor{currentstroke}%
\pgfsetstrokeopacity{0.000000}%
\pgfsetdash{}{0pt}%
\pgfpathmoveto{\pgfqpoint{1.045208in}{0.694630in}}%
\pgfpathlineto{\pgfqpoint{3.690763in}{0.694630in}}%
\pgfpathlineto{\pgfqpoint{3.690763in}{1.760000in}}%
\pgfpathlineto{\pgfqpoint{1.045208in}{1.760000in}}%
\pgfpathclose%
\pgfusepath{fill}%
\end{pgfscope}%
\begin{pgfscope}%
\pgfsys@transformshift{1.140000in}{0.713333in}%
\pgftext[left,bottom]{\includegraphics[interpolate=true,width=2.460000in,height=1.023333in]{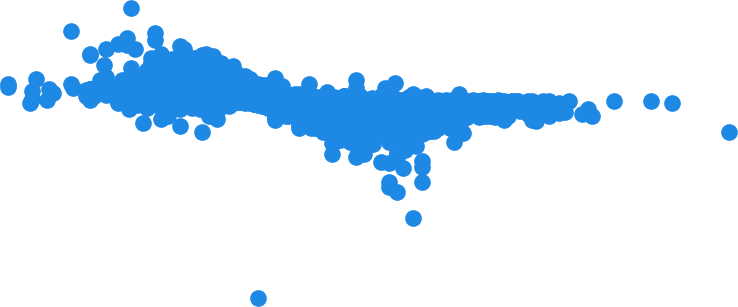}}%
\end{pgfscope}%
\begin{pgfscope}%
\pgfpathrectangle{\pgfqpoint{1.045208in}{0.694630in}}{\pgfqpoint{2.645554in}{1.065370in}}%
\pgfusepath{clip}%
\pgfsetbuttcap%
\pgfsetroundjoin%
\definecolor{currentfill}{rgb}{0.117647,0.533333,0.898039}%
\pgfsetfillcolor{currentfill}%
\pgfsetlinewidth{2.007500pt}%
\definecolor{currentstroke}{rgb}{0.117647,0.533333,0.898039}%
\pgfsetstrokecolor{currentstroke}%
\pgfsetdash{}{0pt}%
\pgfsys@defobject{currentmarker}{\pgfqpoint{0.000000in}{0.000000in}}{\pgfqpoint{0.083333in}{0.000000in}}{%
\pgfpathmoveto{\pgfqpoint{0.000000in}{0.000000in}}%
\pgfpathlineto{\pgfqpoint{0.083333in}{0.000000in}}%
\pgfusepath{stroke,fill}%
}%
\end{pgfscope}%
\begin{pgfscope}%
\pgfsetbuttcap%
\pgfsetroundjoin%
\definecolor{currentfill}{rgb}{0.200000,0.200000,0.200000}%
\pgfsetfillcolor{currentfill}%
\pgfsetlinewidth{0.803000pt}%
\definecolor{currentstroke}{rgb}{0.200000,0.200000,0.200000}%
\pgfsetstrokecolor{currentstroke}%
\pgfsetdash{}{0pt}%
\pgfsys@defobject{currentmarker}{\pgfqpoint{0.000000in}{-0.048611in}}{\pgfqpoint{0.000000in}{0.000000in}}{%
\pgfpathmoveto{\pgfqpoint{0.000000in}{0.000000in}}%
\pgfpathlineto{\pgfqpoint{0.000000in}{-0.048611in}}%
\pgfusepath{stroke,fill}%
}%
\begin{pgfscope}%
\pgfsys@transformshift{1.602751in}{0.694630in}%
\pgfsys@useobject{currentmarker}{}%
\end{pgfscope}%
\end{pgfscope}%
\begin{pgfscope}%
\definecolor{textcolor}{rgb}{0.200000,0.200000,0.200000}%
\pgfsetstrokecolor{textcolor}%
\pgfsetfillcolor{textcolor}%
\pgftext[x=1.602751in,y=0.597407in,,top]{\color{textcolor}\sffamily\fontsize{11.000000}{13.200000}\selectfont \(\displaystyle {50}\)}%
\end{pgfscope}%
\begin{pgfscope}%
\pgfsetbuttcap%
\pgfsetroundjoin%
\definecolor{currentfill}{rgb}{0.200000,0.200000,0.200000}%
\pgfsetfillcolor{currentfill}%
\pgfsetlinewidth{0.803000pt}%
\definecolor{currentstroke}{rgb}{0.200000,0.200000,0.200000}%
\pgfsetstrokecolor{currentstroke}%
\pgfsetdash{}{0pt}%
\pgfsys@defobject{currentmarker}{\pgfqpoint{0.000000in}{-0.048611in}}{\pgfqpoint{0.000000in}{0.000000in}}{%
\pgfpathmoveto{\pgfqpoint{0.000000in}{0.000000in}}%
\pgfpathlineto{\pgfqpoint{0.000000in}{-0.048611in}}%
\pgfusepath{stroke,fill}%
}%
\begin{pgfscope}%
\pgfsys@transformshift{2.286003in}{0.694630in}%
\pgfsys@useobject{currentmarker}{}%
\end{pgfscope}%
\end{pgfscope}%
\begin{pgfscope}%
\definecolor{textcolor}{rgb}{0.200000,0.200000,0.200000}%
\pgfsetstrokecolor{textcolor}%
\pgfsetfillcolor{textcolor}%
\pgftext[x=2.286003in,y=0.597407in,,top]{\color{textcolor}\sffamily\fontsize{11.000000}{13.200000}\selectfont \(\displaystyle {100}\)}%
\end{pgfscope}%
\begin{pgfscope}%
\pgfsetbuttcap%
\pgfsetroundjoin%
\definecolor{currentfill}{rgb}{0.200000,0.200000,0.200000}%
\pgfsetfillcolor{currentfill}%
\pgfsetlinewidth{0.803000pt}%
\definecolor{currentstroke}{rgb}{0.200000,0.200000,0.200000}%
\pgfsetstrokecolor{currentstroke}%
\pgfsetdash{}{0pt}%
\pgfsys@defobject{currentmarker}{\pgfqpoint{0.000000in}{-0.048611in}}{\pgfqpoint{0.000000in}{0.000000in}}{%
\pgfpathmoveto{\pgfqpoint{0.000000in}{0.000000in}}%
\pgfpathlineto{\pgfqpoint{0.000000in}{-0.048611in}}%
\pgfusepath{stroke,fill}%
}%
\begin{pgfscope}%
\pgfsys@transformshift{2.969256in}{0.694630in}%
\pgfsys@useobject{currentmarker}{}%
\end{pgfscope}%
\end{pgfscope}%
\begin{pgfscope}%
\definecolor{textcolor}{rgb}{0.200000,0.200000,0.200000}%
\pgfsetstrokecolor{textcolor}%
\pgfsetfillcolor{textcolor}%
\pgftext[x=2.969256in,y=0.597407in,,top]{\color{textcolor}\sffamily\fontsize{11.000000}{13.200000}\selectfont \(\displaystyle {150}\)}%
\end{pgfscope}%
\begin{pgfscope}%
\pgfsetbuttcap%
\pgfsetroundjoin%
\definecolor{currentfill}{rgb}{0.200000,0.200000,0.200000}%
\pgfsetfillcolor{currentfill}%
\pgfsetlinewidth{0.803000pt}%
\definecolor{currentstroke}{rgb}{0.200000,0.200000,0.200000}%
\pgfsetstrokecolor{currentstroke}%
\pgfsetdash{}{0pt}%
\pgfsys@defobject{currentmarker}{\pgfqpoint{0.000000in}{-0.048611in}}{\pgfqpoint{0.000000in}{0.000000in}}{%
\pgfpathmoveto{\pgfqpoint{0.000000in}{0.000000in}}%
\pgfpathlineto{\pgfqpoint{0.000000in}{-0.048611in}}%
\pgfusepath{stroke,fill}%
}%
\begin{pgfscope}%
\pgfsys@transformshift{3.652509in}{0.694630in}%
\pgfsys@useobject{currentmarker}{}%
\end{pgfscope}%
\end{pgfscope}%
\begin{pgfscope}%
\definecolor{textcolor}{rgb}{0.200000,0.200000,0.200000}%
\pgfsetstrokecolor{textcolor}%
\pgfsetfillcolor{textcolor}%
\pgftext[x=3.652509in,y=0.597407in,,top]{\color{textcolor}\sffamily\fontsize{11.000000}{13.200000}\selectfont \(\displaystyle {200}\)}%
\end{pgfscope}%
\begin{pgfscope}%
\definecolor{textcolor}{rgb}{0.200000,0.200000,0.200000}%
\pgfsetstrokecolor{textcolor}%
\pgfsetfillcolor{textcolor}%
\pgftext[x=2.367985in,y=0.406667in,,top]{\color{textcolor}\sffamily\fontsize{13.000000}{15.600000}\selectfont Mean arterial pressure (raw)}%
\end{pgfscope}%
\begin{pgfscope}%
\pgfsetbuttcap%
\pgfsetroundjoin%
\definecolor{currentfill}{rgb}{0.200000,0.200000,0.200000}%
\pgfsetfillcolor{currentfill}%
\pgfsetlinewidth{0.803000pt}%
\definecolor{currentstroke}{rgb}{0.200000,0.200000,0.200000}%
\pgfsetstrokecolor{currentstroke}%
\pgfsetdash{}{0pt}%
\pgfsys@defobject{currentmarker}{\pgfqpoint{-0.048611in}{0.000000in}}{\pgfqpoint{-0.000000in}{0.000000in}}{%
\pgfpathmoveto{\pgfqpoint{-0.000000in}{0.000000in}}%
\pgfpathlineto{\pgfqpoint{-0.048611in}{0.000000in}}%
\pgfusepath{stroke,fill}%
}%
\begin{pgfscope}%
\pgfsys@transformshift{1.045208in}{0.930405in}%
\pgfsys@useobject{currentmarker}{}%
\end{pgfscope}%
\end{pgfscope}%
\begin{pgfscope}%
\definecolor{textcolor}{rgb}{0.200000,0.200000,0.200000}%
\pgfsetstrokecolor{textcolor}%
\pgfsetfillcolor{textcolor}%
\pgftext[x=0.635370in, y=0.877598in, left, base]{\color{textcolor}\sffamily\fontsize{11.000000}{13.200000}\selectfont \(\displaystyle {\ensuremath{-}0.2}\)}%
\end{pgfscope}%
\begin{pgfscope}%
\pgfsetbuttcap%
\pgfsetroundjoin%
\definecolor{currentfill}{rgb}{0.200000,0.200000,0.200000}%
\pgfsetfillcolor{currentfill}%
\pgfsetlinewidth{0.803000pt}%
\definecolor{currentstroke}{rgb}{0.200000,0.200000,0.200000}%
\pgfsetstrokecolor{currentstroke}%
\pgfsetdash{}{0pt}%
\pgfsys@defobject{currentmarker}{\pgfqpoint{-0.048611in}{0.000000in}}{\pgfqpoint{-0.000000in}{0.000000in}}{%
\pgfpathmoveto{\pgfqpoint{-0.000000in}{0.000000in}}%
\pgfpathlineto{\pgfqpoint{-0.048611in}{0.000000in}}%
\pgfusepath{stroke,fill}%
}%
\begin{pgfscope}%
\pgfsys@transformshift{1.045208in}{1.401370in}%
\pgfsys@useobject{currentmarker}{}%
\end{pgfscope}%
\end{pgfscope}%
\begin{pgfscope}%
\definecolor{textcolor}{rgb}{0.200000,0.200000,0.200000}%
\pgfsetstrokecolor{textcolor}%
\pgfsetfillcolor{textcolor}%
\pgftext[x=0.753657in, y=1.348563in, left, base]{\color{textcolor}\sffamily\fontsize{11.000000}{13.200000}\selectfont \(\displaystyle {0.0}\)}%
\end{pgfscope}%
\begin{pgfscope}%
\definecolor{textcolor}{rgb}{0.200000,0.200000,0.200000}%
\pgfsetstrokecolor{textcolor}%
\pgfsetfillcolor{textcolor}%
\pgftext[x=0.355740in, y=0.699688in, left, base,rotate=90.000000]{\color{textcolor}\sffamily\fontsize{13.000000}{15.600000}\selectfont SHAP value for}%
\end{pgfscope}%
\begin{pgfscope}%
\definecolor{textcolor}{rgb}{0.200000,0.200000,0.200000}%
\pgfsetstrokecolor{textcolor}%
\pgfsetfillcolor{textcolor}%
\pgftext[x=0.538148in, y=0.266861in, left, base,rotate=90.000000]{\color{textcolor}\sffamily\fontsize{13.000000}{15.600000}\selectfont Mean arterial pressure (raw)}%
\end{pgfscope}%
\begin{pgfscope}%
\pgfsetrectcap%
\pgfsetmiterjoin%
\pgfsetlinewidth{0.803000pt}%
\definecolor{currentstroke}{rgb}{0.200000,0.200000,0.200000}%
\pgfsetstrokecolor{currentstroke}%
\pgfsetdash{}{0pt}%
\pgfpathmoveto{\pgfqpoint{1.045208in}{0.694630in}}%
\pgfpathlineto{\pgfqpoint{1.045208in}{1.760000in}}%
\pgfusepath{stroke}%
\end{pgfscope}%
\begin{pgfscope}%
\pgfsetrectcap%
\pgfsetmiterjoin%
\pgfsetlinewidth{0.803000pt}%
\definecolor{currentstroke}{rgb}{0.200000,0.200000,0.200000}%
\pgfsetstrokecolor{currentstroke}%
\pgfsetdash{}{0pt}%
\pgfpathmoveto{\pgfqpoint{1.045208in}{0.694630in}}%
\pgfpathlineto{\pgfqpoint{3.690763in}{0.694630in}}%
\pgfusepath{stroke}%
\end{pgfscope}%
\end{pgfpicture}%
\makeatother%
\endgroup%